# PREDICTING EARLY STAGES OF ALZHEIMER'S DISEASE AND IDENTIFYING KEY BIOMARKERS USING DEEP ARTIFICIAL NEURAL NETWORK AND ENSEMBLE OF MACHINE LEARNING METHODOLOGIES

DEBOPRIYA GHOSH

Master of Science (MSc), Thesis Report

MARCH 2022

# ABSTRACT

Alzheimer's disease (AD) is a neurological disorder that begins slowly with some mild symptoms of memory loss and subsequently affects an individual’s ability to accomplish everyday activities. Alzheimer's disorder affects different areas of the brain that manage thinking, memory, and linguistics. It is the most common form of dementia, the cause of which little is known to scientists. Although it is most common in older people, it often affects younger people as well. The main problem with AD is that the onset, progression, and course of symptoms are highly unpredictable and variable. Along with physical and mental issues, patients with the disease and their families face social issues like isolation, unpredictability, fear, fatigue. There is also a huge economic impact of dementia relating to medical and social care expenditures. Many times, dementia is misdiagnosed as a sign of ageing by physicians, and the severity of the disease becomes more adverse when the person is actually detected with dementia. At present, no drugs are available to prevent this disease, but some symptom-delaying drugs are available that can slow the disease's progression to some extent. Therefore, an automatic early-stage diagnostic system will help physicians control the disease in its primitive phase by taking preventive measures. The purpose of the study is to propose a model which can accurately diagnose the different early stages of Alzheimer's Disease by analysing clinical information, neuropsychological and radiological test measures. This study also helps clinicians to understand the significant biomarkers associated with Alzheimer's disorder. A dataset has been collected from the Alzheimer’s Disease Neuroimaging Initiative study data for this research work. This dataset contains missing values, which will be treated with an iterative imputation technique. Being a healthcare dataset, a class imbalance problem exists in this, which is addressed by the borderline SVM-SMOTE algorithm. Feature selection needs to be performed as fitting a machine learning model with irrelevant features leads to inaccurate detection. This study uses wrapper-based and embedded feature selection techniques to obtain the most significant feature set. A train-test split needs to be carried out on selected predictors, and feature scaling is done to make the features scale free. After completing all data pre-processing tasks, a stacking-based ensemble model, comprised of Logistic Regression, Extra Tree, Bagging KNN and LightGBM classifiers as a base estimator, is designed. A deep learning model (ANN) is also implemented on the same dataset, which may increase the accuracy. Then, a comparative analysis is performed among the above models by evaluating different performance metrics, namely, precision, recall, f1-Score, and AUC-ROC, to determine the best classifier along with the most important biomarkers for early diagnosis of AD.

# TABLE OF CONTENTS

## LIST OF TABLES

## LIST OF FIGURES

## LIST OF ABBREVIATIONS

AD…............................Alzheimer's Disease

AdaBoost…...................Adaptive Boosting

ADAS….........................Alzheimer's Disease Assessment Scale

ADNI….........................Alzheimer's Disease Neuroimaging Initiative

ADNIMERGE…...........Alzheimer's Disease Neuroimaging Initiative Merge table

ANOVA…....................Analysis of Variance

APOE4…......................Apolipoprotein E4

AUC…..........................Area under the Curve

AUCPR….....................Area Under the Curve Precision Recall

AVG…………………...Average

CDR…..........................Clinical Dementia Rating

CDRSB….....................Clinical Dementia Rating Sum of Boxes

CN…............................Cognitive Normal

CSF…...........................Cerebrospinal Fluid

DC…............................Declining Cognition

DT…............................Decision Tree

DX…............................Diagnosis

Ecog……………………Everyday cognitive

EDA….........................Exploratory Data Analysis

ELD….........................Ensemble Linear Discriminant

EMCI….......................Early Mild Cognitive Impairment

FDG….........................Fluorodeoxyglucose

FN…............................False Negative

FP…............................False Positive

GB…............................Gigabyte

GBTClassifier…..............Gradient Boosting Classifier

GDS….........................Geriatric Depression Scale

GTX….........................Giga Texel Shader Xtreme

HIPV…........................Hippocampal Volume

HV…...........................Hard Voting

K-CNN…………………..Keras-Based Convolutional Neural Network

KNN…...............................K-Nearest Neighbour

LASSO…..........................Least Absolute Shrinkage and Selection Operator

LightGBM…......................Light Gradient Boosting Machine

LMCI….............................Late Mild Cognitive Impairment

LR….................................Logistic Regression

MCI…...............................Mild Cognitive Impairment

MidTemp….......................Middle Temporal

ML…................................Machine Learning

MMSE…...........................Mini-Mental State Exam

MOCA…...........................Montreal Cognitive Assessment

MRI…..............................Magnetic Resonance Imaging

MRIv…............................Magnetic Resonance Imaging Volumetric

NaN….............................Not a Number

NB…...............................Naive Bayes

NIA….............................National Institute on Aging

NINCDS-ADRDA…..........Neurological and Communicative Disorders and Stroke-Alzheimer’s Disease and Related Disorders Association

OASIS…..........................Open Access Series of Imaging Studies

PACC……………………Preclinical Alzheimer's Cognitive Composite

PET…..............................Positron Emission Tomography

PSO…..............................Particle Swarm Optimization

PTGENDER…..................Patient's Gender

RAM…............................Random Access Memory

RAVLT….........................Rey's Auditory Verbal Learning Test

RBF…..............................Radial Basis Function

RF…................................Random Forest

RFE…..............................Recursive Feature Elimination

RGF.................................Regularized Greedy Forests

SC…................................Stable Cognition

SMOTE….........................Synthetic Minority Oversampling Technique

SV…................................Soft Voting

SVM…............................Support Vector Machines

TADPOLE…....................The Alzheimer's Disease Prediction of Longitudinal Evolution

TN…................................True Negative

TP…..................................True Positive

US….................................United States

VISCODE….......................Visit Code

XgBoost…........................Extreme Gradient Boosting

# CHAPTER 1

# INTRODUCTION

## 1.1 Background of the Study

According to the World Health Organization, Dementia is a medical condition that affects person's ability to perform daily activities. It includes difficulties with memory, language, problem-solving, and other thinking skills. Alzheimer's disease is the most frequent cause of dementia, contributing for 60 to 70% of all cases (World Health Organization, Dementia, 2021). Alzheimer's disorder is one of the leading fatal illnesses in the United States. It is the 6th deadly illness among American adults. In 2021, approximate 6.2 million Americans aged 65 and above is having Alzheimer's dementia. This number is projected to be nearly 12.7 million people by 2050 (Alzheimer’s Facts and Figures, Alzheimer’s Association, 2021). The synapse in neuron is accountable for transmitting signals in the brain cell and allow neurons to communicate. Along with the signal transmission, neuron releases Amyloid beta, that is a tiny peptide. While the fundamental cause of Alzheimer's disease is still debated, most researchers believe that it grows when Amyloid beta plaques form accumulating Amyloid beta between synapses. Besides that, tau protein forms triangle like structure in memory-related part of the brain, producing swelling and cell injury, leading to loss of memory. A few experts claim that preventing Amyloid Beta plaques from surpassing threshold is the sole way to prevent Alzheimer's disease (Shah et al., 2020). There are three phases for understanding dementia syndrome. Memory lapses, losing sense of time, getting misplaced in known surroundings, etc. symptoms are frequent in the early stages. The disease visibility is more impactful when it progresses into middle stage. Forgetting person’s name, latest incidents, noticing changes in behaviour, such as wandering and repetitive inquiry etc. are symptoms of this phase. The disease becomes more severe at the Late stage. Being oblivious to time and location, having trouble recognising family and friends, trouble in walking, having a growing need for help with self-care etc. are the signs of this advance stage. There is a huge economic impact of dementia relating to medical and social care expenditures. The cost was 1.3 trillion US dollars in 2019, which will expect to exceed 2.8 trillion dollars by 2030. There is no cure for dementia at this time. Anti-dementia drugs and disease-modifying therapies currently on the market have limited efficacy (World Health Organization, Dementia, 2021).

Almost all existing Alzheimer's disease diagnosis procedures entail several neuropathological test results and radiographic examinations, which are both costly and prohibitive for which are both costly and prohibitive for the average person. Therefore, the authors of the (Shah et al., 2020) study examine various machine learning-based algorithms that might be utilised as a cost-effective diagnostic tool for early screening of Alzheimer's disease. Biological markers such as the monumental state examination, clinical dementia rating, and normalized whole brain volume are included in this study. One of the limitations of this study is that it does not take into account medical imaging aspects, which are widely employed in the treatment and identification of neurological illnesses. The future scope of the study is to combine image extracted features along with clinical information. Another study (Lodha et al., 2019) is based on derived dimensions such as RAVLT tests, MOCA, and brain volume. It demonstrates that the RAVLT percentage is a crucial component in detecting Alzheimer's disease. Patients with a high RAVLT forgetting score have a higher risk of dementia, whereas those with a low forgetting score are less likely to be afflicted. In the (Ezzati et al., 2019) study, a normal versus dementia classification algorithm is designed, based on patient demographics and MRI volume measurements, and it is tested against Mild Alzheimer's affected participants to predict disease transformation. It is evident from this study that MRIv, demographics, APOE4 are the best factors for diagnosis of Alzheimer's progression. One drawback of this study is that the dataset used for this study is not a population-based study, and participants are chosen using tight inclusion and exclusion criteria, which may restrict the model's generalisation power. As a result, confirming the outcomes in other population-based research and clinical study findings is a critical next step.

With no proven therapies, finding behavioural activities and biomarkers that may properly evaluate and detect disease development in asymptotic subjects is critical, as medication is most productive at a preliminary phase of Alzheimer's disease (Almubark et al., 2020). Therefore, the purpose of the study is to propose a predictive model on Alzheimer's Disease early detection by analysing clinical information, neuropsychological and radiological test measures to cater a fair decision support system for medical practitioners to diagnose the incident of dementia more accurately and enhance the survivability rate of patients. This study also helps clinicians to understand the significant biomarkers associated with Alzheimer's disorder for effective diagnosis of its stages.

## 1.2 Problem Statement

According to (Department of Health and Human Services, 2020), Alzheimer's disease is the most frequent kind of dementia. Alzheimer's disease wreaks havoc on the parts of the brain that control judgement, memory, and communication. It has the potential to have a serious impact on an individual's ability to do routine chores. Alzheimer's disease is one of the leading causes of death in the United States. In the United States, Alzheimer's ailment is one of the major causes of mortality. In adults aged 65 and up, the sixth leading cause of death. By 2020, 5.8 million Americans aged 65 and over are estimated to have Alzheimer's dementia. The population is predicted to nearly triple by 2060, reaching 14 million people. In 2010, the anticipated cost of treating Alzheimer's disease was between $159 and $215 billion. By 2040, these expenses are predicted to increase from $379 billion to far more over $500 billion yearly. Regardless of the fact that cardiovascular disease and cancer mortality rates are dropping, Alzheimer's disease death rates have gone up. Dementia, including Alzheimer's disease, has been shown to be under-reported in death records, implying that the number of elderly people suffering from Alzheimer's disease is likely to be significantly higher.

Pharmacological intervention can improve the living conditions for people with Alzheimer's disease and their caretakers. Alzheimer's disease presently has no recognized cure. The treatment focuses on various areas: assisting people in maintaining brain health, managing behavioural symptoms, lowering, or postponing illness signs. Therefore, diagnosis of the illness at its early stage is very necessary. Many persons with Alzheimer's disease are being cared for at home by close relatives. Although most people are ready to manage for their near and dear ones, caring for someone with Alzheimer's disease at residence can be a challenging and stressful undertaking at times. Every day presents new problems as the caregiver adjusts to fluctuating levels of competence and new behavioural patterns. People with Alzheimer's disease frequently require increasingly intense care as the condition progresses (Department of Health and Human Services, 2020).

An approach to combine conscious intellectual activity such as reasoning, thinking alongside normal neuropsychological test results to construct an ML-based AD stage detection model is discussed in (Almubark et al., 2020). The aim of the study is to find related biological markers and attitudinal activities that can correctly identify the advancement of the disease in symptomless individuals. The researchers in this study applied the ML solutions on three datasets, each involving 28 elders with AD and 50 elders with the CN category. These datasets contain regular neuropsychological test results with 9 attributes, 260 variables linked to reflex time, and the third is a merged version of previous two datasets with 269 features. Principle Component Analysis and SelectKbest feature selection technique have been applied to finalized top uncorrelated features. Predictive modelling has been implemented on original features, principal components, and top extracted features. It is evident from this study that features selected from neuropsychological test scores are more useful for AD diagnosis than the spatial attention cognitive data. Here the authors' future suggestion is to find informative cognitive factors that can be utilized further and carry out experiment on weighted penalty and neural net-based models. Here, missing value has been imputed by simple mean imputer. But instead of simple imputation, advanced ML based computation technique can result in better performance. These can be identified as one of the problems present in this study. Additionally, embedded tree-based feature selection technique can be performed optimal feature selection.

A stacking-based ensemble approach that can distinguish subjects with mild mental disability, Alzheimer's disease, and healthy people have been proposed in (Khoei et al., 2021) paper. This study also focuses on identification of major biological markers for qualitative measurements. The researchers have used dataset of Alzheimer's Disease Prediction of Longitudinal Evolution which involved cognitive biomarkers like MMSE, RAVLT_immediate, RAVLT_forgetting etc. and quantitative measure biomarkers such as Ventricles, hippocampus, MidTemp, whole brain etc. Imbalance handling, missing value imputation and standardization have been performed as a pre-processing step. Two predictor selection methods namely, Pearson's Correlation Coefficient and Least Absolute Shrinkage and Selection Operator have been used. The heterogeneous stacking-based ML architecture has been trained on 10 biomarkers. The authors have proved that variable RVLT_immediate has become the most significant feature, while variable RVLT_forgetting has become the most insignificant when compared to other biomarkers. Authors have used random oversampling technique for handling imbalance data. This use of random oversampling can introduce redundant information in the data and make the predictive model bias towards that data points. This can be considered as one of the sort-coming presents in this research that can be addressed in future using more improved imbalance handling techniques for achieving better model performance.

The authors in (Thapa et al., 2020), have spoken about feature selection-based procedure that is managed by data for better decision-making on timely detection of AD. This paper proposed how neuropsychological test scores can be integrated with MRI extracted features to control the cases of cognitive impairment. This study has used a filter-based Weka's Bestfirst forward feature selection method to identify best combination of uncorrelated (Wekas's CfsSubsetEval) attribute with superior predictive power. Feature scaling has been applied on the selected predictors before feeding them into models. Machine Learning algorithms have been applied to perform three binary classifications (CN vs AD, MCI vs AD and CN vs MCI). It is observed from this study that attribute atrophy in the hippocampal region has become more relevant for differentiating the stages of AD when combined with neuropsychological test scores like MMSE. This study has used standardization as a scaling technique, instead of this normalization technique can also be tested on this dataset in future. Also, here, SMOTE has been used for generating artificial minority class samples for handling class imbalance. This problem can be addressed in future by utilizing more improved balancing technique such as Borderline-SMOTE or Borderline-SMOTE SVM for creating unambiguous synthetic data points.

The researchers in (Gorji et al., 2020) study, have aimed to determine the most significant features, biological markers and cognitive scores using six conventional variable selection techniques. Filter based methods namely Pearson's Correlation and Chi-Squared, Wrapper based methods Recursive Feature Elimination (RFE) and step forward feature selection and Embedded selection technique Lasso and extra tree classifier algorithms have been implemented on TADPOLE dataset to identify the best set of features which perfectly separate CN, EMCI and AD patients. Here, mean-imputation has been used for replacing NaN values with mean of the specific columns.165 possible feature combinations, after applying feature scaling, have been applied on ML classification model to finalize 8 out of 11 features. It is observed that MidTemp lobe atrophy has become an important feature, although not identified by any selection method. After applying the combination-based selection method, model with {CDRSB and MidTemp} feature combination and {CDRSB, Hippocampus, and MidTemp} combination have gained the highest accuracy. Thus, this study concludes that combination-based feature extraction is the most trusted way for relevant markers selection. This paper has used a very common approach of imputation as it is easy to calculate, but more robust imputation methods can be applied in future. Also, instead of normalization method, standard scaler method can be tested on the dataset. Likewise other tree-based embedded feature selection approach can be performed to check any improvement of prediction.

A multi-level classification architecture constructed with NB at first level, SVM and KNN at the second level, has been introduced in (Kruthika et al., 2019) paper. Here the authors have used FreeSurfer tool to extract features like brain structural volumetric measurements and cortical thickness measurements from MRI scans. To determine most relevant biomarkers for building AD or MCI subject identifying model, Swarm Optimization - PSO algorithm has been implemented. The multi-level classifier proposed in this study, can also be used for image retrieval task based on query image. It is observed that classifier discussed in this study is performing well compared to individual predictive algorithms. However, the authors have noticed some issue at the first level classifier for differentiating AD versus MCI and NC. To overcome the stated issue, they have suggested to choose more effective biological and biochemical information in the future study.

Most of the research publications published on the prediction model for Alzheimer's disease stage detection used a relatively naive missing value imputation strategy, specifically simple mean imputation, which is responsible for model performance decline (Gorji et al., 2020; Khoei et al., 2021). The problem of class imbalance in medical data is pervasive, and many earlier articles have failed to address it (Shah et al., 2020; Eke et al., 2021). In the case of Alzheimer's illness prediction, the majority of previous work has been done on datasets with an extremely uneven distributions of the class labels, such as cognitively normal, early mild cognitive impairment, late mild cognitive impairment, and Alzheimer's disease, and a few papers have used very basic data generation approaches like random Oversampling, SMOTE (Thapa et al., 2020; Khoei et al., 2021).These methods can lead to issues such as random Oversampling, which can introduce selection bias and, as a result, suffer from overfitting. Similarly, SMOTE may produce synthetic data points in an overlapping region of the majority class. As a result, even if the prediction accuracy is great, the results are likely to be biased towards the majority class, which is the non-dementia or cognitively normal group. The selection of key sets of biomarkers is also a crucial step in the ML pipeline. Some articles have only used (Almubark et al., 2020), filter-based correlation selection methods or wrapper-based methods. A combination-based feature selection method considering statistical methods, wrapper-based (like RFE, sequential selection), and tree-based can be examined to achieve a more significant feature sub-set. The development of a prediction model based on class-balanced data, the appropriate feature sub-set, and the suitable imputation method will allow for a more assertive decision-making process during the diagnosis of Alzheimer’s disease.

## 1.3 Aim and Objectives

The main aim of this research is to propose a model to predict early stages of Alzheimer's Disease and understanding the significant biomarkers related to the disease. The identification of Alzheimer's Disease at its initial phase using the well-studied risk factors allows timely and cost-effective diagnosis and the progression of this disease can be controlled by the early phase detection model.

The research objectives are formulated based on the aim of this study, which are as follows:

- To analyse the relationship between the biomarkers via visualization to improve the diagnostic understanding for clinicians and patients.
- To suggest a suitable missing value imputation and balancing technique that can be applied on unbalanced dataset.
- To identify the key risk factors linked with Alzheimer's patients through appropriate feature selection technique.
- To compare among the predictive models for selecting the most accurate model to classify different stages of Alzheimer's disease based on different evaluation metrics.

## 1.4 Research Questions

The following research questions are proposed for each of the research objective as highlighted as follows:

1. Will the combination of identified socio-demographic measures, biomarkers and extracted radiological features associated with Alzheimer's patient improve the understandability of the diagnosis?
2. Which missing value imputation and balancing technique can be implemented on the selected dataset?
3. Which predictive classification model can be proposed to accurately distinguish healthy people from people suffering from Alzheimer’s disorder based on various evaluation metrics?

## 1.5 Scope of the Study

This study is to propose a model to effectively diagnose early stages of Alzheimer's Disease and identify the most significant biomarkers associated with dementia. This current study is limited to only the presented Alzheimer's clinical dataset, the implementation of this model with other domains or datasets will need separate training. Implementation of this model on medical imaging data is outside the scope of this study. Fore missing value imputation, iterative imputer method will be used for replacing null values in numerical columns because this method can take advantage of numerous regressor models and is able to impute missing values more effectively than any other technique.

To overcome the class imbalance problem, an advanced version of SMOTE technique, named as Borderline SVM-SMOTE will be used. SMOTE algorithm only considers minority class samples which result in generation of synthetic data points near the overlapping class region. In contrary, Borderline-SMOTE, generates new synthetic points near the class boundary region. As SMOTE-SVM mitigates the drawbacks of both ordinary SMOTE and borderline SMOTE, it will be used for this study.

For selecting relevant features, RFE, Elastic Net, Random Forest Classifier will be developed. RFE is the most popular semimanual kind of selection approach which is chosen here because of its simple configuration steps and the efficiency of selecting the most suitable features from the training set. Elastic Net embedded technique is selected for this study because it combines both penalties from RIDGE (L2) and LASSO (L1) regularization and effectively eliminates unnecessary features by considering effect of correlated feature groups. Unlike RIDGE, it makes irrelevant feature coefficients zero. Unlike LASSO, it cannot eliminate multicollinear features randomly without averaging the correlated attributes. So, there is no loss of relevant colinear feature. Therefore, the limitations of both L1 and L2 regularization is addressed by Elastic Net and hence, chosen in this research. Random Forest Classifier, which is a tree-based model, will be used as it is faster, more accurate and has an in-build feature importance calculation technique. Based on this feature importance score, attribute selection will be performed.

For modelling and prediction, Logistic Regression, Extra Tree, LightGBM, Bagging KNN Classifier, Stacking Classifier algorithms will be implemented. LR will be selected due to its simplicity and interpretability. Bagging and boosting classifiers will be used to obtain more accurate prediction as they train on many individual estimators and finally generate the outcome based on majority votes for this classification task. Finally, a stacking-blending model will be implemented which includes all the aforementioned algorithms to take advantage of each model and the performance of each individual model will be compared with this stacking-based technique. An ANN model will also be implemented for checking further improvement in the prediction power, as deep learning model usually provides more accurate results.

The usage of other imputation, balancing, feature selection and modelling methodologies are outside the scope of this research.

## 1.6 Significance of the Study

In recent days, Alzheimer's Disease is become one of the leading illness diagnosed among the elderly people. Although it is most common in older ages, it also affects younger people as well. The main problem with AD is the onset, progression, and course of symptoms are highly unpredictable and variable. Along with physical and mental issues the patient with the disease and their families face social issues like Isolation, unpredictability, fear, fatigue. According to (World Health Organization, Dementia, 2021), 60-70% of cognitive impairment cases are due to Alzheimer's disorder. 9% of dementia cases is observed among young people. Women are suffering more in dementia than men. Among total death due to dementia, 65% are women. There is a huge economic impact of dementia relating to medical and social care expenditures. The cost was 1.3 trillion US dollars in 2019, which will expect to exceed 2.8 trillion by 2030. At present, the Alzheimer's Disease is treated by medicines which can only slowdown the disease advancement, but cannot provide a permanent cure of this ailment (Thapa et al., 2020). The main challenge is to diagnose this disease in its early stages, because although the first symptom appears after 60s, it may actually begin decades before. Many times dementia is misdiagnosed as the sign of ageing by the physicians, hence the severity of the disease becomes more adverse when the person is actually detected with dementia.

The aim of the study is to propose a model which can accurately diagnose the different early stages of Alzheimer's Disease by analysing clinical information, neuropsychological and radiological test measures. This study also helps physicians to understand the significant biomarkers associated with Alzheimer's progression. Therefore, utilizing this model, clinician will be able to detect dementia stages and accordingly plan for medications to control the growth of dementia.

Several studies that have been done to create a machine learning model for the Alzheimer's disease detection, but most of them focused on differentiation between physically fit subjcets and Alzheimer's affected subjects. But, differentiation among healthy people, mild to modarate dementia affected people and Alzheimer's patient needs to be performed to restrict the disease growth at early or intermidiate stage. These studies used simple null imputation technique, which is not always suitable for treating missing values as they does not consider all attributes present in the data during calculation. Few of them created models without balancing the dataset. But dataset balancing is an important task in the ML pipeline for getting an accurate and unbiased prediction. Therefore, enhanced imputation and imbalance handling technique needs to be implemented along with advance predictive modelling methods, which would complement the prediction process and can be integrated on top of the existing methodologies.

## 1.7 Structure of the Study

The thesis is organised as follows: Chapter 1 gives the research background, which discusses the effects of dementia caused by Alzheimer's disorder among the aged, the impact of society on the afflicted people and their families, the mortality of the disease, and the problem statements. The study aims and objectives are discussed in section 1.3. Section 1.4 outlines the research questions that the project will attempt to answer. The scope of the study provided in section 1.5. Following that, Section 1.6 talked about the significance of the study.

Chapter 2 provides the necessary theoretical background and underlines the challenges presented in Chapter 1, by methodically evaluating the machine and deep learning models used in dementia prediction, particularly Alzheimer's disease. The benefits of data analytics and machine learning in the health sector have been highlighted in Section 2.2. Section 2.3 explains how data mining is used to diagnose chronic diseases. The effect of Alzheimer's disease, its forms and phases, its cause, and socioeconomic facts etc. are presented in section 2.4. In section 2.5, the importance of data mining in the prediction of dementia disease is explained. In section 2.6, some of the historical relevant research publications on Alzheimer's disease and dementia are discussed. In sections 2.7 and 2.8, the associated research discussion and review summary are discussed and concluded, respectively.

Chapter 3 presents the research design and the recommended framework. Section 3.2 explains the study approach, which includes a data set description, several data pre-processing procedures, data transformation steps, and an imbalance handling technique. Section 3.3 discusses the machine learning as well as deep learning models that will be employed in this project, as well as the precise methods that will be used for predicting early stages of Alzheimer's disease. Section 3.4 provides a summary of the chapter.

The data analysis and design methodology mentioned in Chapter 3 are discussed in Chapter 4. The first section of the chapter will go through the dataset that was used in this dissertation. The subset used to build the machine learning models as well as the number of features and records utilised to perform analysis, details about each attribute, will be mentioned in section 4.2. Section 4.3 will illustrate the pre-processing needed before feeding the data into classifiers. It entails the removal of irrelevant variables and the selection of important features, as well as the transformation of categorical to numerical features and vice versa, the identification of missing values and the selection of an appropriate imputation technique, imbalance analysis and handling, normalization and the preparation of train and test datasets. Section 4.4 will discuss exploratory data analysis, which includes Univariate, bivariate, and correlation/multivariate analysis to identify the relationship between the target variable and other dependent aspects. The six machine learning models, as well as the hyperparameters, will be described in Section 4.5 and applied to both imbalanced and class balanced datasets. Finally, in section 4.6, all the information presented in this chapter will be summarised.

In Chapter 5, the outcomes of the machine learning models employed in this study will be addressed. Section 5.2 will go over the important biomarkers that can be discovered using the visualisation and feature selection techniques. In section 5.3, the best approaches for dealing with class imbalance will be determined, and the assessment results of each modelling algorithm (accuracy, recall, fi-score, AUC, and so on) before and after class imbalance handling will be explained. The selection of the highest performing classification model will be discussed in Section 5.4, based on the comparison of the evaluation metrics calculated from all six models' confusion matrix. Section 5.5 will contain a summary of the entire chapter.

Chapter 6 will cover the conclusion and future recommendation of the study in brief, along with a summary of the entire research. Section 6.2 will discuss how this study will be able to meet all its aim and objectives, as well as whether any gaps or modification scope still exist. If any future recommendations exist, they will be discussed in Section 6.3. This is how the report will be structured and present the overall research.

# CHAPTER 2

# LITERATURE REVIEW

## 2.1 Introduction

Machine learning is a multi-disciplinary field with roots in statistics, mathematics, data processing, knowledge analytics, and other fields, making it challenging to come up with a new definition. Over the last few years, machine learning has become widespread in various industries, such as manufacturing, transportation, and healthcare. Due to the extensive use of Machine Learning/ Deep Learning algorithms in various fields, this technology has become an integral part of our daily lives. This chapter will describe previous research work carried out on visual analytics and data mining in healthcare. This section contains a literature summary that begins with data analytics in health care, progresses to several studies on data mining for diagnosis of chronic diseases and various types of neurological disorders, and ends with studies on Alzheimer's disease.

Data analytics applications in health care have great potential for raising the quality of care, reducing waste and error, and lowering healthcare costs. Data analytics in healthcare will be briefed based on previous literature papers. In healthcare, data mining assesses therapeutic efficacy, saves patients' lives through predictive medicine, detects stage early, detects waste, fraud, and abuse, manages healthcare at different levels, and manages customer relationships. Several studies have been conducted on the data mining techniques on the diagnosis of different chronic diseases such as chronic kidney diseases, diabetes, coronary heart disease, and liver disorders. These will be included in this section.

Apart from that, machine learning approaches used for some of the most common types of dementia, such as Alzheimer's disease, Vascular dementia, Lewy Body Dementia, and Mixed Dementia, will be thoroughly investigated in order to comprehend the current trends used to diagnose and predict the diseases. This chapter describes the development of predictive models using data mining techniques and statistical approaches for diagnosis of mental health issues. The last subsection of this chapter will discuss the overall findings on the recent trends in visual analytics and data mining in healthcare, focusing more on dementia caused by Alzheimer's disease. Any limitations and challenges faced in previous studies will also be highlighted. Data mining improves physician decision making by using analytics in health care instead of relying on only lab test. Today, digitization is changing in the medical system. Big data analysis of medical information enables the diagnosis, treatment, and development of personalised medicines that enable unprecedented treatment. This leads to better patient outcomes while limiting costs.

## 2.2 Machine Learning and Analytics in Healthcare

Healthcare is currently undergoing a comprehensive worldwide overhaul and transformations, making it one of the fastest expanding sectors today. In recent years, great progress has been made in how to use machine learning in the medical domain. (Shailaja et al., 2018) discussed the need of ML based decision support system in healthcare. Decision support systems help doctors investigate illnesses by making them aware of health problems and revealing background knowledge about individual patients. It also helps identify the patient's situation and provides guidance to suggest when to use the right drug at any time (Moreira et al., 2019).

Machine learning is a special method of data analysis that automates model building related to model development. As the machine learns to use certain algorithms, it can extract hidden insights from the data. It's important to note that machine learning doesn't tell the machine where to look. The iterative nature of machine learning allows machines to adjust methods and outputs when exposed to new situations and data (Bhardwaj et al., 2017). (Shailaja et al., 2018) paper has discussed several approaches to machine learning. The first is supervised learning, in which the computer is given a set of inputs and is required to produce a particular number of correct outputs. To identify the error, the empirical results are compared to the proper output during training. The next one is Semi-Supervised Learning, where the given data has a mixture of labelled and unlabelled information. Here the classification performance is achieved by using unlabelled information. The success of this method depends entirely on some basic assumptions. The next category of ML technique is the unsupervised method, where no labels are given to the training data. Clustering, fuzzy clustering, hierarchical clustering, K means clustering etc. are covered by this learning (Simeone, 2018). Another approach to ML is Reinforcement Learning, where computer software grants access to a dynamic environment in order to accomplish a specified goal. Because it navigates its flaws, the software receives feedback in the form of rewards and punishments.

Based on (Bhardwaj et al., 2017), machine learning algorithms are effective for discovering complex patterns in large amounts of data. This facility is ideal for clinical applications, especially for those who rely on advanced genomes and proteomics measures. It is frequently employed in the diagnosis and detection of a variety of illnesses. Machine learning algorithms in medical applications will make better judgements about treatment plans for patients by suggesting how to create a useful health-care system. Prognosis, diagnosis, treatment, and clinical workflow are the four key uses of ML/DL approaches in healthcare. As per (Javed Mehedi Shamrat et al., 2021), prognosis is the process of predicting the expected progression of a disease in clinical practice. It also includes identifying symptoms and signs associated with a particular disease and whether they worsen, improve, or remain stable over time. In a similar fashion to the clinical setting, multimodal patient data is collected that can facilitate the prognosis of the disease by the ML model. (Zheng et al., 2017) discussed the use of ML for the diagnosis of chronic diseases by extracting clinical features from electronic health records (EHRs).

Although ML is primarily used in medicine, there are some challenges associated with data-driven predictive technology. Despite these challenges, ML has the potential to provide better care at a lower cost to both patients and radiologists. Algorithms like KNN, SVM, Neural Network, K-Means, Markov Decision Process, Ensemble Learning Method, etc. are widely used in medical informatics.

Healthcare is one of the industries that may profit greatly from the growing amount of data and its accessibility. Big data refers to huge, complicated datasets that require more than typical data management solutions to store, manage, and process them in a timely and cost-effective manner. Structured, semi-structured, and unstructured data in the petabytes and beyond can be handled using big data solutions (Bhardwaj et al., 2015). Based on a technology survey (Harerimana et al., 2018), big data used in health care is now widespread and is rapidly emerging from a variety of platforms. Due to the urgent need for emergency medical care, analytical systems need to be able to aggregate all data and provide doctors with real-time information. For example, in the past, when a patient came to the hospital with a particular illness, the doctor relied on what he or she generally knew about the disease, or what other doctors in the area knew. With big data technology, physicians can see national trends regarding which treatment strategy is best for their patients in order to prescribe the best medication. Aggregation of individual datasets that would otherwise prove meaningless provides the information physicians need to make better overall medical decisions.

There are numerous and diverse sources of health big data. The sources are determined by the level of technology employed by a health entity. The major contributors to health data are cyber-physical systems, the medical Internet of Things (mIoT), social networks, Electronic Medical Records, and genomic data. The study (Zhang et al., 2018) looked at the big data challenges at each stage of the big data pipeline. The challenge in the data aggregation phase is to combine large amounts of data from different data warehouses with real-time data on a common platform. Another major difficulty in big data analytics is data storage, which is difficult to handle with typical database management systems because data grows at an exponential rate. To tackle the problem, MongoDB, Hadoop Distributed File System (HDFS), and other NoSQL databases can be used. The issue is also present during the data integration and analysis phases. Big data analytics computing platform like Hadoop, Spark, MapReduce, Flink etc. are used to effectively handle both static and real-time data (Yao et al., 2015; Akil et al., 2018). Supporting libraries, such as machine learning libraries and graph libraries, should be included in the platform for relationship analysis.

### 2.3 Data Mining Technique for Diagnosis of Chronic Disease

Data mining is a collection of computational approaches for extracting useful patterns from large amounts of data. Today's healthcare industry generates massive amounts of data about hospitals, resources, disease diagnosis, computerised patient records, and other topics. The enormous amounts of data must be analysed and scrutinised in order to extract knowledge that will aid in comprehending the current state of the health sector. Framing a hypothesis, acquiring data, doing pre-processing, estimating the model, interpreting the model, and drawing conclusions are all part of the data mining process (Umadevi and Marseline, 2018).

Medical data mining refers to a wide range of activities aimed at collecting and maintaining large databases of all types of medical information and finding patterns in the information found. Most commonly, such information includes electronic health records, drug research results, a list of medicines available on the market, and other data. Data mining is closely linked to machine learning, artificial intelligence, and big data technology, including the processing, analysis, storage, and protection of large amounts of collected information. Because of the numerous benefits it offers, data mining is becoming increasingly popular in the medical field. The benefits include improved diagnosis accuracy, more effective treatment, improved fraud detection, access to predictive analytics, drug quality assessment etc. (Bahri et al., 2019).

Healthcare is one industry that generates a huge amount of data that is characterised by high speed and variety, such as lab data, medical prescriptions, appointments, machine-generated data, insurance patient data, and administrative data. Indeed, in the United States, only clinical data reached 150 exabytes in 2011. In the near future, clinical data is expected to grow to zettabytes, or even yottabytes (Shafqat et al., 2020). According to (Wang et al., 2018) analysing this Big Data is necessary to improve patient satisfaction and quality of care as well as reduce expenditures, which reached 17.9% of the gross domestic product in 2010.

In (Snegha et al., 2020), the authors have employed two data mining techniques, the Random Forest algorithm, and the Back Propagation Neural Network, to diagnose chronic kidney illness and evaluate the data in order to determine the optimum algorithm for forecasting chronic kidney disease. Chronic kidney disease is a rapidly onset human disease. In today's world, many people suffer from chronic kidney disease (CKD). To identify the symptoms, attributes such as blood pressure, sodium, haemoglobin, potassium, white blood cells, hypertension, diabetes, stirrup edema, etc. need to be analysed. They are verified using robust diagnostic and data mining methods. The symptoms are less noticeable in the early stages of the disease, while kidney loss is more noticeable in the later stages. The aim of this study is to demonstrate that the application of data mining techniques aids in the prevention and control of disease to some extent. This study clearly shows that the Back Propagation neural network method is more effective than the Random Forest Classifier technique in detecting chronic kidney disease.

The authors of (Woldemichael and Menaria, 2018) have used data mining techniques to assist doctors in making accurate diagnoses and treating diseases, reducing the workload of specialists. The goal of this study was to use data mining techniques to predict diabetes. Diabetes is a chronic disease represented by elevated blood glucose levels in the human body. Diabetes, over time, causes a slew of complications in the kidneys, eyes, and heart, among other organs. According to the International Diabetes Federation, 382 million people worldwide have diabetes. By 2035, this figure will have more than doubled to 592 million. In this research, the back propagation technique is used to determine whether a person is diabetic or not. To predict diabetes, J48, naive bayes, and support vector machine were also utilised. These neural networks had an input layer with eight parameters, a hidden layer with six neurons, and an output layer with six neurons. The model's performance was improved using a 5-fold cross-validation technique and a large value learning rate. This study aids in the support of medical decisions as well as the better medical treatment and diagnosis of patients. According to the findings of this study, the accuracy of Back propagation in diabetes prediction outperforms SVM, J48, and Naive Bayes algorithms.

Another study on coronary heart disease by (Krishnani et al., 2019) explained the use of Data mining for retrieving information from a large set of medical data. In this paper, researchers proposed a comprehensive pre-processing technique to predict Coronary Heart Diseases (CHD). Replacing null values, resampling, scaling, classifying, and prediction are all part of the method. The objective of this paper is to predict the risk of coronary artery disease using data mining algorithms such as Random Forest, Decision Trees, and K-Nearest Neighbours. In addition, a comparison study of these algorithms based on prediction accuracy is carried out. K-fold Cross Validation is also used to generate randomness in the data. These models can accurately predict the risk of heart disease, allowing researchers to determine whether a person is likely to develop CHD within the next ten years. The results show that Random Forest is the most compatible competitor for prediction model and provides superior performance measure among K-Nearest Neighbour and Decision Tree with the suggested data pre-processing extended research.

In (Kumar and Katyal, 2019), data mining methods were applied to a dataset of liver disorder patients in the classification mode in attempt to predict the right liver diagnosis. The liver is one of the largest organs in the human body. Every liver ailment can cause acute or chronic pain and swelling, liver weakness, and even damage to other organs within the body. The authors applied five classification approaches such as Naive-Bayes, RF, K-means, C5.0 and KNN to a specific dataset as performance was measured using precision, recall, and accuracy. Investigations were conducted to increase the performance of categorization models. Finally, the findings indicated that the method could be used to diagnose liver disease in its early stages. When all algorithms are implemented without any adjustments, the Random Forest Algorithm provides the maximum accuracy. However, following adaptive boosting, the C5.0 method produces a more accurate result. The implementation of the K-Means Algorithm yielded the highest precision, whereas the implementation of the KNN Algorithm yielded the highest recall.

### 2.4 Fundamental of Alzheimer and Dementia Disease

Dementia is a phrase used to describe a loss of capacity to recall, reason, or make judgments that disrupts with daily tasks. The most frequent form of dementia is Alzheimer's disease. Although dementia primarily affects the elderly, it is not a natural part of aging. Dementia is a broad word with a wide range of symptoms. Memory loss, attention, communication, judgement, and problem solving are among the issues that dementia patients face. Dementia manifests itself in a variety of ways, including becoming lost in a nearby neighbourhood, using weird phrases to refer to familiar items, forgetting the name of a closest relative or friend, erasing previous memories, and being unable to do tasks independently. Age, family history, poor heart health, traumatic brain injury, etc. are factors that increase the risk of dementia (Alzheimer's Disease, National Institute on Health, NIA, 2021).

As follows, there are five common types of dementia (Ahmed et al., 2019):

- Alzheimer's disease: Alzheimer's disease is the most common neurodegenerative disease, responsible for 60 to 80% of cases. It is brought on by changes in the brain. The most significant risk factor is family history. Having an immediate family who has Alzheimer's disorder raises the chances of getting it by 10% to 30%.
- Vascular dementia: Stroke and other disorders of blood flow to the brain are linked to about 10% of instances of dementia. Diabetes, hypertension, and a high cholesterol level are all risk factors. The size and location of the afflicted brain determine the symptoms. The condition worsens over time. As the number of strokes and mild strokes rises, the symptoms become more severe.
- Lewy body dementia: Along with the more common symptoms of dementia, such as forgetfulness, patients with this type of dementia may experience stiffness and tremors. Arousal abnormalities, such as daytime tiredness, anxiety, and staring, are also common. Individuals may also have difficulty sleeping or have night-time hallucinations.
- Fronto-temporal dementia: This type of dementia often results in personality and behavioural changes due to the part of the brain it affects. Person with this disorder may be ashamed of themselves or misbehave. Language abilities, such as speaking and understanding, may also be an issue.
- Mixed dementia: This type of dementia happens when multiple types of dementia may be present in the brain at the same time, especially in people over the age of 80. The disease can progress faster than any type of dementia.

According to (World Health Organization, Dementia, 2021) dementia signs and symptoms can be classified into three categories:

1. Early stage: A person with Alzheimer's disease can operate independently in the initial stages. The individual is still able to drive, work, and engage in social activities. Even yet, the individual may experience memory problems, such as recalling common words or having trouble finding common objects.

2. Middle stage: In the mid stages of Alzheimer's disease, the dementia symptoms become increasingly noticeable. Words may be misread, causing the individual to become agitated or enraged, and causing them to respond in unpredictable ways, such as rejecting to take bathe. Injury to cerebral nerve cells can make it hard for a person to express their thoughts and do everyday chores without assistance.
3. Late stage: The clinical manifestations of dementia in its latter stages are devastating. Individuals lose their ability to react to their surroundings, communicate, and, finally, control their movement. They may still speak in phrases or words, but conveying their pain becomes more challenging. As memory and concentration abilities deteriorate, individual personality changes may occur, demanding a great deal of attention.

Scientists are working to learn more about Alzheimer's disease plaques, tangles, and other biological aspects. Through advancements in brain imaging, researchers can now monitor the growth and distribution of aberrant amyloid and tau proteins in the living brain, as well as alterations in brain structure and function. Scientists are also looking into the early stages of Alzheimer's disease by looking at changes in the brain and fluids that can be noticed years before symptoms appear. These studies' findings will aid in our understanding of Alzheimer's disease's causes and diagnosis (Alzheimer’s Disease, National Institute on Health, NIA, 2021).

Alzheimer's disease is more than just a loss of memory. Alzheimer's disease is lethal. According to (World Health Organization, Dementia, 2021) dementia currently affects over 55 million people globally, with approximately 10 million new cases diagnosed each year. Alzheimer's disease, which accounts for 60–70% of all dementia cases, is the most frequent.

## 2.5 Data Mining Technique for Diagnosis of Dementia

Data mining in medicine is an emerging field of great importance in providing prognosis and insight into disease classification, especially in the mental health fields. A high level of impairment, such as affective illness, which causes depression and various anxiety disorders, is used to assess mental health. In both rich and developing countries, 25% of people suffer from mental health issues. Data is growing to terabytes and petabytes, with 80% of it unstructured, rendering traditional database management tools and procedures ineffective. The global cost of mental health care is estimated to be around $2.3 trillion. This cost can be minimised by enhancing the quality of therapies, and this quality can be increased by using data mining tools and techniques in mental health.

The (Shohieb and El-Rashidy, 2019) study applies intelligent strategies to meet the individual demands of patients with cognitive deficiencies and to analyse interaction changes depending on patient behaviour at different phases of disease. The goal of this research is to create a data mining framework that is both usable and effective for the elderly. The evaluation phase, data mining phase, and assistance phase are the three primary phases of the suggested framework. The patient progresses through the three levels of behaviour analysis and feedback to receive interactive evaluation and help. The evaluation phase will concentrate on developing neurocognitive expertise through the use of games and tests that assess selective memory, attention, and concentration. The proposed test will concentrate on the assessment of memory functions related to current events and discussions. During the data mining stage, the system will attempt to collect data from patient sensors and analyse it on a real-time basis with a clinical and a middle data mining layer. The analysis of physiological and vital data will be performed, and essential features will be identified using a subset of features approaches, followed by training and on-time classification. This step is critical for understanding and building a useable and accessible human-computer interaction (HCI) interface that will help the patient improve his or her cognitive capacities. The most crucial phase of our proposed system is this one. It offers features such as using data mining from the previous stage to predict patient needs, providing mental tasks to build awareness and recognition, assessing the patient on a regular basis, providing feedback, and automatically adapting the system based on the patient's situation.

With the growing ageing population, dementia in the elderly has become a major health concern. It is critical to take efforts to avoid or delay the onset of dementia symptoms. Using a large nationally representative cross-sectional data of elderly adults from the English Longitudinal Project of Ageing (ELSA), the (Yang and Bath, 2020) study evaluated a huge and diverse range of characteristics from several domains related to dementia. The ELSA is a nationally representative study of community-dwelling adults aged 50 and over, representing the general population of Great Britain, United Kingdom. This study identifies dementia in two ways: (a) Participants recorded a physician's diagnosis of dementia or Alzheimer's disease. (b) An adapted short-form IQCODE assessment to check an individual’s ability to perform various functions like memorizing details about relatives. The study includes predictors such as demographic and socio - economic factors, social interaction and social networking factors, physical health and impairment factors, mental health - related factors, lifestyle factors, neurocognitive factors, and more. Different sub-set of feature selection was done by combining the top-ranking features from XgBoost, LGBM, CatBoost ML algorithms. Seven machine learning models, CatBoost, LGBM, GBM, RF, LR, Keras-Based Convolutional Neural Network, Regularized Greedy Forests and Ensemble model by combining different base models have been implemented. SMOTE, ADASYN and class weighted technique were used to handle highly imbalanced data. This research uses Normalised Gini Coefficient for performance measurement. The best outcome, a Gini score of 0.9333, was obtained by combining the scores of four different models: K-CNN, RF, RGF, and XgBoost.

The proportion of people living with dementia is rising in tandem with the global trend of increased life expectancy. The most serious problem is that clinician evaluations are lengthy, subjective, and prone to error because they are dependent on clinical records for individuals and questionnaires from their guardians. An entire data mining modelling can be used to overcome this problem, allowing physicians to make sensible decisions. As a result, the (Bang et al., 2017) paper proposed a four-module quad-phased data mining approach for the CREDOS project, which includes clinical information obtained from 37 university-affiliated institutions in the Republic of Korea between 2005 and 2013. Potential clinical factors that are effective for investigations were selected in the Proposer Module. Then, using a machine learning algorithm, a model was built to predict and identify dementia in the Predictor Module. This module implements machine learning algorithms such as SVM, ANN, and DT. A threshold-independent index AUC (area under a receiver operating characteristic (ROC) curve) was used to assess the expected outcome for these three models. In the Descriptor Module, diagnostic causes were evaluated by profiling patient groups to enable clinical practitioners to better comprehend the predictive model's outcomes. In this phase, the input value and predicted value of SVM were used to create DT as an interpretable model. Therefore, 20,836 health records are divided into 31 leaf node groups. Finally, visualisation was provided in the visualisation module to effectively study the characteristics of patient groupings. SVM was set up as a final decision-making model for a physician with a significant AUC of 0.96.

The (Moreira and Namen, 2018) work proposes a hybrid data mining paradigm that combines text mining with structured data mining. The goal of this model is to aid doctors in the diagnosis of individuals who have clinical suspicions of dementia. The investigations were based on a set of 605 health records containing 19 various features regarding individuals who had reported cognitive deterioration. To begin, a text mining approach was used to construct a new structured attribute. This was achieved by grouping the patient's diagnostic history information into an unstructured textual feature. Various classification techniques, including nave bayes, bayesian belief networks, and decision tree, were used to develop Alzheimer's disease and mild cognitive impairment prediction model. To enhance the performance of the created models, ensemble approaches like Bagging, Boosting, and RF were applied. These techniques were tested on two data sources: one with only the original structured data, and the other with the original structured data plus the derived attribute obtained from text mining (hybrid model). Finally, for the dementia class of interest, Bayesian classification algorithms and decision trees employing ensemble methods yielded the best results for FPR and FNR. With AdaBoost and Bagging, the J48 algorithm gave the best rates in terms of the AUC measure. A significant improvement in the accuracy rate and, more importantly, the number of false positives was seen with the incorporation of the new structural attribute Cluster PPH, a result of the text mining process. A similar trend was identified in the false negative rate and AUC measures, but to a lesser degree. The optimum model was developed using the AdaBoost ensemble strategy, which employed the J48 technique as the basic classifier and applied the K-means segmentation method to examine the grouping of medical storylines.

## 2.6 Related Research Publications

The dataset used in this study is frequently updated by Alzheimer's Disease Cooperative Study data system, hence it contains recent study data. Thus, limited papers are available on the exact same dataset. Therefore, this section provides related study that was done on the same or similar fields.

At present, more than 55 million individuals worldwide are suffering from dementia, with roughly 10 million new cases diagnosed each year. Dementia ranks as the seventh major cause of death of all the diseases and one of the key reasons of disability among the elderly around the world. Alzheimer's disease is the most common type of dementia, which may account for 60 – 70% of cases. However, there is limited treatment available for Alzheimer's disorder (World Health Organization, Dementia, 2021). If the diagnose is done at primitive stage, the progression of the disease can be controlled.

In (Divya and Shantha Selva Kumari, 2021) paper, various attribute selection methods and classifiers have been used to classify the chronic mental condition as normal control, cognitive decline, and Alzheimer's disorder using MRI scans from the ADNI dataset. The majority of aged persons suffer from dementia, primarily as a result of Alzheimer's disease. AD advances gradually from minor phase, moderate to severe dementia. It is caused primarily by an unusual accumulation of neurochemicals, such as amyloid plaques and tau tangles. The main aim of the study is to perform efficient feature elimination to build a good classifier. When there are lesser data with high dimensions, dimensionality reduction performs a significant impact on improving performance of the classifier. The authors have dropped the columns with missing values without any imputation. To choose the highly important variables from a huge feature set, the wrapper-based strategy combining RFE and Genetic Algorithm algorithms with LR and linear SVM classifiers (RFE-LR, RFE-SVM, GA-LR, GA-SVM) have been used. Also, data range has been selected between interquartile range (IQR) to make the model stable against outliers. After all the stated pre-processing steps, LR, RF, SVM, XgBoost algorithms have been implemented and performance are evaluated using Accuracy, Sensitivity, Specificity metrics. The GA-LR attribute selection algorithm outperforms the others. The addition of the MMSE score to the structural MRI compressed feature set improves the classification performance in recognizing MCI from AD. For the Random Forest classifier in MCI verses AD diagnosis, the model constructed from the characteristics chosen using GA-LR paired with the MMSE score increased accuracy, sensitivity, and specificity. For binary classification of NC verses AD and NC verses MCI, SVM with RBF kernel offers better outcomes. The gap identified in this paper, i.e., researchers should have used some efficient ML based imputation technique like KNN imputer or iterative imputer instead of removing the columns with missing values. Also, feature scaling needs to be performed using either standardization or min-max scaler method to get better result.

The primary objective in (Eke et al., 2021) study is to create an ML-based approach for identifying blood related biochemical markers of premature Alzheimer's disease on the basis of non-amyloid proteins, with the ability to detect the ailment before amyloid-beta deposition in the brain matter. The ADNI website was utilized to obtain the blood metagenomic data used in this research project to distinguish healthy group from affected group. The pre-processed samples were split into two non-intersecting sets, sets 1 and 2, to make optimal use of existing data while decreasing our approach's vulnerability to overfitting concerns. Baseline data from Alzheimer's and Healthy control make up Dataset 1. Dataset 1 was employed in our technique to conduct a powerful feature selection and model construction. Dataset 2 was used to assess the algorithms that were produced. Month-12 data from MCIs and HCs is included in Dataset 2. During model building, primarily set 1 was used to train models. Other than that, feature scaling and feature selection were performed. This study built a correlation-based feature selection technique on dataset 1 to identify most significant and necessary features for categorization of AD and HC groups, reducing the dimension of the research data before model building. SVM with polynomial kernel of degree 2 and 3 and RBF kernel has been applied on the training data. A2M, ApoE, RAGE, Eot3 etc. were shown to be critical protein profiles that contributed significantly to the classification results. From the above study it is observed that the author has not applied any class imbalance strategy on the dataset. Also, more advanced feature selection technique like tree-based, embedded method or statistical techniques can be developed. Moreover, other effective non-linear algorithms such as bagging, boosting, ensemble methods can be implemented to make the classifier more robust. These gaps would be addressed in this specific research project.

Alzheimer's disorder is the most prevalent forms of neurocognitive condition that causes brain cell damage over time. There is no specific medicine available to slow or stop the progression of this disease. However, if detected early, the disease can be controlled. The (Khan and Zubair, 2020b) paper proposed a system for a definitive task of input to output transformation which can effectively and accurately classify a demented and non-demented population of individuals. The suggested framework was developed using MRI imaging data from the Open Access Series of Imaging Studies database. This study has used chi-squared based feature selection technique to identify the most related categorical feature having strong relation with target column. Also, correlation matrix, feature importance etc. were analysed to finalize most relevant feature set. Rows corresponding to 8 missing values were eliminated, leaving just 142 participants in the sample for training, validation, and testing activities. Median based simple imputation strategy has been used for imputation purpose. Besides that, for each sample of the specified column, a standardisation procedure is used to conform it to a specified scale. AdaBoost, Extra tree, RF, GBM, Gaussian Process classifier, LRCV, Passive Aggressive classifier, Ridge ClassifierCV, SGD, Perceptron classifier, NB, KNN, DT, SVM algorithms were implemented on the pre-processed data. The researchers have selected the Random Forest algorithm with imputation by a median at the ROC curve's peak point based on the results of the experiment. Therefore, it has the best accuracy of 86.84 % among all classification techniques. The gap identified in this research is that the researchers used median-based imputation, which is a simple imputation technique. Rather, any other advanced method like KNN imputation or iterative imputation can be tried. Also, a combination of different feature selection techniques like RFE, tree-methods, elastic nets, etc., can be applied along with the chi-squared statistical method to obtain a more concise feature set. This improvement may lead to enhanced model performance.

In another study (Stamate et al., 2019), authors have discussed that in the absence of a remedy or a structured diagnostic test, using a machine learning approach to determine patients at high risk of developing Alzheimer's disease could be a new milestone toward proactive intrusion.

The purpose of this project is to conduct a preliminary investigation that will lead to the development of different classifiers with better capabilities for discriminating indications of Cognitively Normal, Mild Cognitive Impairment, and Dementia. The sample applied in this research is a portion of the ADNIMERGE data which only includes baseline (the first visit for every subject) component measurements with a diagnosis. The study used random forest imputation to fill in missing variables. Relevant features have been identified using information gain and ReliefF technique combines variable selection with statistical permutation tests on the basis of 500 randomised label combinations. Also, for handling class imbalance SMOTE strategy has been implemented here. RF, SVM with linear and RBF kernel, Gaussian Processes, SGD, XgBoost algorithms have been applied on the pre-processed dataset. An XgBoost model yielded the best results, with attribute selection achieved via the ReliefF approach. Only variables having an estimated Relief score equivalent to a p-value of less than 0.1 in the statistical permutation test were considered in this feature selection technique. The gap observed in this research are as follows: the use of SMOTE may introduce overlapping minority class samples, which may not be the correct representative of actual samples. Authors should have use other advanced synthetic data generation methods like SMOTE-Tomek, Borderline SMOTE or class weighted technique. Improvement may also be done for feature selection approaches used in this study.

The authors of (You et al., 2020) study looked at whether paralinguistic elements obtained from recordings of an episodic memory screening test (LOGOS) may distinguish between individuals at high and low risk of dementia based on their sensitivity to low and high cognitive reserve stimuli, respectively. This study used audio samples assembled from volunteers in the Australian National Health and Medical Research Council. This research indicates that paralinguistic traits such as stuttering, stops, and tone can be utilised to diagnose not only dementia but also moderate memory deficits. Here the authors used variable selection on the normalised attribute set, the Mann-Whitney U-test, Neighbourhood Component Analysis (NCA), and ReliefF to determine the best selective set of paralinguistic features and minimise the size of the attribute set. Standardization was performed to set the column values centred around the mean with a unit standard deviation. After performing data pre-processing and feature selection, KNN Classifier and SVM algorithm have been built on the paralinguistic attributes. To make greater use of both algorithms' capabilities, a parallel ensembled classifier has been created using the k-NN and SVM models' respective bipolar optimism ratings. The proposed method had an accuracy of 97.2% when both paralinguistic and episodic cognitive test variables were utilised and 94.7% when just paralinguistic variables were used in classifying older individuals at high or low risk of dementia. When both paralinguistic and episodic cognitive test features were employed, the parallel classification scheme improved by 5.6% compared to a k-NN model alone and 12% compared to a SVM classifier, and 10.6% and 13.9% correspondingly when just paralinguistic information were used.

The timely identification of several dementia phases is critical in developing efficient Alzheimer disease therapeutic interventions. High-resolution MRI images can gradually be used in the categorization of distinct phases of dementia, assisting in the creation of effective therapy stratagems. Given the difficulty of analysing large amounts of data, authors in (Khan and Zubair, 2020a) investigated the ability of machine learning based solutions to identify onset of dementia in subjects. The Open Access Series of Imaging Studies database was utilised to compile the information for this publication. The sample was gathered from medically apparent dementia patients who showed signs of extremely mild to moderate impairment. The dementia category was determined by scoring characteristics such as CDR, SES, and MMSE, as well as other variables. The 75th percentile and highest values of age, CDR, and eTIV varied significantly. This indicates the presence of anomalies, or skewed data, in the collection. This study has used PCA, Factor Analysis (FA), LDA, Neighbourhood Components Analysis, and Fast Independent Component Analysis methods as dimensionality reduction approaches. Other pre-processing steps like sampling without replacement, imputing missing values, categorical variable encoding, outlier capping, variable scaling etc. were also implemented on the dataset before building ML models. KNN, SVM, GaussianNB, DT, RF, XgBoost, SGD, and AdaBoost algorithms were developed, with the ensemble-based RF predictor for LDA dimensionality reduction approach demonstrating the superior prediction performance.

In (Ryu et al., 2020) paper, the researchers have proposed an Alzheimer risk detection model based on XgBoost employing derived parameter extraction from numericalized demented facts and hyper-parameters refinement. This study was performed on OASIS dataset. The suggested method used gradient boosting to extract feature importance from typical predictor variables and then creates derived components. The resultant derived elements are then used to perform variable significance analysis, resulting in the formation of a Top-N group. Because missing values can have a detrimental impact on Alzheimer assessment, the data row with missing values should be eliminated. A data volume is lowered in this method. As a result, it is possible to protect additional data in a range of situations by merging the OASIS-1 and 2 data sets with the same format. ID, Hand, Delay, and MRI ID are examples of insignificant variables. Because it has no effect on the response variable CDR, the variable has been eliminated from the predictor factors. After that, normalisation is done using Min-Max Scaling approach. To identify optimum parameters more quickly, the Grid Search approach and parallel processing were used. This study implemented three types of XgBoost models. One without hyper-parameter tuning and without using derived features. One with parameter tuning but without using derived variables. One with parameter optimization and use of the top-20 group through the retrieval of derived features. The model with hyper-parameter optimization results of the Top-20 group showed the best outcome.

Another study (Raut and Dalal, 2018) proposed the use of a Machine Learning methodology to predict Alzheimer's Disease from Magnetic resonance imaging. Using the Gray-Level Co-Occurrence Matrix method and the seven moment invariants technique, the suggested method recovers texture and shape properties of the Hippocampus area from MRI images. Finally, for the detection of different phases of the disease, a Neural Network is leveraged as a Multi-Class Classification model on the extracted features. This research work has used scans from Open Access Series of Imaging Studies along with that subject's age, gender, socio-economic status, Mini Mental State Examination score etc. factors were also considered. The input images were pre-processed through noise removal and normalization steps before extracting features from Region of Interest. The authors can use other feature extraction technique like convolution neural network for extracting feature maps from MRI images.

The authors in (Shah et al., 2020), have discussed that all current Alzheimer's detection methods involve different neuropathological test outcomes, albeitno concrete test, radiological examination, which are quite expensive and unaffordable for common people. Therefore, they have performed analysis on different machine learning based techniques that can be used as a cost-effective diagnostic tool for detection of Alzheimer's disease at initial phase. Researchers have collected data for 437 elderly patients between ages of 60 and 96. 72 of them are in non-demented category and 64 are in demented class. As the brain wave activity differs between right- and left-handed individuals, all the selected participants in this study are right-handed. Support Vector Machine, Random Forest, Hard and Soft Voting Classifier, XGBoost and Decision Tree Classifier have been applied on training data containing several biomarkers such as Education, Monumental State Examination, Clinical Dementia Rating etc. and different performance metrics have been evaluated to get the best suited model. It is clear from the above study that HV and SV classifiers have outperformed all other models on the selected biomarkers. Finally, the researchers have recommended to combine textual data with medical image extracted meaningful features, to further improve the model's efficiency.

In another study (Alroobaea et al., 2021), authors have introduced a computerized examination mechanism based on machine learning, which can eliminate the need of extensive and expensive manual Labour and magnetic scans. The purpose of this study is to timely diagnose dementia because Alzheimer is one of the fatal illnesses among senior citizens and timely detection helps healthcare professionals minimizing the threat of fatality. In this study authors have used datasets from two popular databases namely Alzheimer's disease Neuroimaging Initiative and Open Access Series of Imaging Studies brain dataset. The ADNI dataset includes 502 features for 1737 North American male and female personals. Although the dataset contains CN, LMCI, EMCI, AD groups, this study has focused on only CN and AD classes. ML methodologies such as LR, DT, SVM, K-nearest Neighbours, RF, Naïve Bayes, Linear Discriminant Analysis have been designed on these datasets. The authors have performed a relative study on the accuracy their models have achieved with the state-of-the-art models they have identified from past studies. Finally, on ADNI dataset, LR and SVM models have performed best, whereas on OASIS dataset, LR and RF models have shown more accurate result. Here the authors' future suggestion is to be confined on MRI data without any handcrafted attributes.

A proposal for using derived dimensions like whole brain volume from MRI scans, cognitive factors like RAVLT tests, MOCA and FDG score has been brought up in a research study (Lodha et al., 2019). This study applied neuroimaging technologies to process scanned images, collected from ADNI2 study, for generating numeric data, which the researchers have further utilized for proposing ML based solutions. The ADNI study data comprises individuals of ages between 55 and 90. Logistic Regression, SVM, Gradient boosting, Neural Network, K-Nearest Neighbour, Random Forest ML algorithms have been implemented on this data. The study result has shown that RF and Neural Network have given more accurate prediction in terms of accuracy than other methods. It is observed from this research that participants whose RAVLT percentage forgetting score is very high has Alzheimer's disease and whose RAVLT percentage forgetting score is not high don't have Alzheimer disease.

In (Neelaveni and Devasana, 2020) study, authors have proposed a learning-based algorithm utilizing education, age, number of visits, MMSE etc. psychological parameters. SVM and DT algorithms have been used in this study and trained on 70% of data. The SVM model has performed well with 85% accuracy to diagnose the disease. Again (Ali et al., 2021) has demonstrated a custom ML model called Modified Random Forest to distinguish between healthy people and people who likely to have AD. The aim of this study is to build a model that achieves higher prediction accuracy, including the patients with very less symptoms and that can handle large dataset. MRI images from OASIS-2 database have been selected for training the m-RF model. Pre-processing steps like NaN value imputation with median value, categorical to numerical conversion using LabelEncoder method etc. have been applied on the dataset before fitting it to the model. This paper shows that the proposed model has obtained an accuracy of 96.43% which is higher than the accuracy achieved by the pre-existing algorithms mentioned in related studies.

In another research (Uspenskaya-Cadoz et al., 2019), the authors have aimed to design a predictive model that can recognize preliminary AD subjects in broader community, so that they can register for further medical investigation. This study has used data from proprietary IQVIA US data assets - LRx and Dx database, which are prescription and claims databases of 72,670,283 US inhabitants. Furthermore, for more granular analysis, the positive cohort is subdivided into four different age groups. LR, DT, RF and GBT classifiers have been implemented on the imbalanced training dataset. As the dataset is unbalanced with higher negative class, the precision recall curve has been chosen by authors over other evaluation metrics. The study result shows that AUCPR achieved by GBTClassifier has exceeded all other models. This study helps to identify 76% asymptomatic prodromal AD patients who can be timely referred for further assessments. The current study has few limitations like the methods chosen by authors for generating positive and negative class, results in high similarities between the target groups, also, the past 2 years medical data has been utilized to train models, which leads to imperfection in model's prediction. As a future scope, the authors have proposed to include family health history, genetic data, information from sensor–based wearable digital devices to strengthen prediction accuracy.

In another research (Memon et al., 2020), authors have mentioned that Alzheimer's Disease identification at preliminary phase is very crucial to apply anti-dementia drugs in a planned way. This study has been conducted on ADNI dataset for developing machine learning-based solution. 70% of data has been used for training of LR, DT and SVM models. Among the three classifiers Logistic regression has gained 95% specificity and 90% sensitivity, which means LR model can correctly diagnosis 95% non-AD personnel in the dataset. Likewise, it can effectively predict 90% AD positive subjects. In future the researchers have a plan to apply effective feature extraction and model optimization techniques to acquire more perfect prediction.

A CN versus AD classification model has been implemented in (Ezzati et al., 2019) study, leveraging patient's demographics and MRIv measures, which has been tested against MCI subjects to predict MCI to AD transformation. The aim of the study is to find out best possible classifier comparing various linear and non-linear models' performance, compare the effective increase in accuracy after using multiple brain structures as opposed to a single brain region such as hippocampus. Four different sub-set of features 1) Hippocampal volume, II) HipV along with age, sex, education, APOE4, III) All MRI volumetric measures, IV) All MRIv in addition to age, sex, education, APOE4 have been selected on which ensemble linear discriminant, boosted trees, SVM, DT, KNN and RF algorithms have been applied. ELD model using feature set IV (MRIv, Demographics, APOE4) has been selected as best predictive model by McNemar test.

Clinical trial is very crucial to slowdown Alzheimer's Disease severity in mild cognitive impaired people, which has been discussed in (Ezzati and Lipton, 2020) paper. The primary goal of this study is to use a learning-based techniques to predict disease progression for increasing the efficiency of clinical test. The training and test data involve ADNI and Semagacestat trials' study data which adhere the NINCDS-ADRDA inclusion criteria for considering a potential AD subject. Based on ADAS-cog scores variation, between 12-months follow-up and baseline visit, patients have been categorized as Stable cognition and Declining cognition groups, which is the target labels for this study. MRIv measures and other continuous features such as MMSE, GDS etc. have been scaled before passing to learning models. Authors have implemented KNN, DT and LR algorithms on ADNI study data and validated the model on LFAN test sample. McNemar test has been conducted which confirmed that KNN is the more accurate model than DT to differentiate DC versus SC function among subjects.

In a recent study (Akter and Ferdib-Al-Islam, 2021), researchers have implemented a clinically translatable machine learning algorithm based on Extreme Gradient Boosting, to categorise different types of dementia. A dataset with 1229 items of 7 variables such as CDR, MMSE, ageAtEntry, memory, DX1 etc. has been used for this paper. Data pre-processing like categorical to numerical conversion using label encoder, feature normalization has been implemented in the machine learning pipeline. The data visualization performed here shows that AD dementia is more common in men. It is evident from this study that XGBoost model has performed well for classifying AD Dementia, No Dementia and Uncertain Dementia patients compared to pre-existing work proposed in related study. According to the current study, variable ageAtEntry has become a significant contributor for classifying dementia. More detailed analysis using oversampling techniques can be conducted to improve the model performance in future.

### 2.7 Discussion

In this section, previous dementia research, especially Alzheimer's disease research, will be summarized in order to provide information on data mining approaches that have been investigated for dementia detection and prediction. Because this research is focused on a specific cause of dementia, only data mining techniques used in Alzheimer's disease is described. The summaries of prior dementia diagnosis studies will be found in table 2.1. SVM, Decision Tree, Bagging classifier, Random Forest, Logistic Regression and k-NN were the most widely utilized data mining techniques in both tables. Apart from dementia diagnosis, these classification systems are well-known in other fields of study. Most of the previous Alzheimer's research has concentrated on building prediction models for survivorship at various time intervals, dementia stage or type categorization, and a computerized system for dementia early detection.

*Table 2.1 Summary of past studies on Dementia*

| Disease | Author & Year | Best Data Mining Technique | Performance Measure | Scope of Study |
|---|---|---|---|---|
| Alzheimer's Dementia | (Shah et al., 2020) | Soft Voting Classifier | Accuracy = 86% | Early Detection of Alzheimer's Disease with various psychological tests |
| | (Alroobaea et al., 2021) | Logistic Regression | Accuracy = 99.43%<br>Precision = 99.30%<br>Recall = 99.70% | Alzheimer's disorder verses Cognitively Normal prediction model |
| | (Lodha et al., 2019) | Neural Network | Accuracy = 98.36% | Detection model for Alzheimer's Disease in its primitive phase |
| | (Neelaveni and Devasana, 2020) | SVM | Accuracy = 85% | Prediction model for Alzheimer's Disease with psychological parameters |
| | (Uspenskaya-Cadoz et al., 2019) | GBT classifier | Precision = 80%<br>AUCPR = 44% | Identify non-diagnosed prodromal Alzheimer’s subjects |
| | (Memon et al., 2020) | Logistic Regression | Accuracy = 98.12%<br>Specificity = 95%<br>Sensitivity = 90% | Prediction model for early-stage Alzheimer’s Disease diagnosis |
| Alzheimer's Dementia | (Ezzati et al., 2019) | Ensemble linear discriminant | Accuracy = 92.8%<br>Sensitivity = 95.8%<br>Specificity = 88.3% | Predictive model for Alzheimer’s Disease |
| | (Akter and Ferdib-Al-Islam, 2021) | XgBoost | Accuracy = 81%<br>Precision = 85% | Prediction model for AD Dementia verses No Dementia verses Uncertain Dementia classification |
| Other form of Dementia | (Ryu et al., 2020) | XgBoost | Accuracy = 85.61%<br>F1-Score = 79.28% | Prediction model for Dementia risk diagnosis |
| | (Moreira and Namen, 2018) | Bagging classifier with NB | AUC = 87.4%<br>Accuracy = 84.9% | Data mining model for diagnosis of Dementia |
| | (Bang et al., 2017) | SVM | AUC = 96% | Data mining modelling for Dementia risk detection |
| | (Khan and Zubair, 2020a) | RF | Accuracy = 87% | Prediction model of Dementia progression using Cross-Sectional MRI data |
| | (You et al., 2020) | Ensembled classifier (KNN & SVM) | Accuracy = 97.2% | Dementia risk predictor using Paralinguistic and Memory Test Features |

Looking at the numerous Dementia Disease studies listed above, each one employs a data mining strategy that is widely utilized. Most researchers have claimed successful Alzheimer's prediction utilizing Logistic Regression, Random Forest, SVM, and XgBoost approaches. Meanwhile, in other investigations of dementia illness prediction, the Bagging classifier, SVM etc. appear to have created the most accurate detection model. One of the reasons for the differences in classifier selection could be related to the varied features identified in each of the datasets used in these experiments. The structure of the prediction model will differ depending on the importance of the factors in Alzheimer's disease prediction. This could be due to the scope of the investigation, as different pre-processing techniques will be used in different studies. Variation in the final prediction model can be caused by feature selection, class balance, and hybrid data mining techniques. As a result of these previous research, it is critical to use the same methodology on the same dataset when comparing alternative data mining techniques.

## 2.8 Summary

The use of big data analytics and machine learning in healthcare were examined in terms of disease prediction and diagnosis. The use of data mining techniques in the treatment of several chronic diseases was discussed in depth. The majority of earlier data mining studies focused on kidney illness, coronary heart disease, diabetes mellitus, mental health etc. chronic disease diagnosis. Different types of dementia, its stages and cause of the mental health issues were discussed in brief. Several sorts of dementia prediction utilizing data mining approaches were also incorporated, and prediction models were constructed to assess the survivability rate and to detect Alzheimer's early using risk criteria. Because different datasets produce different performance measures with varying percentages of accuracies, the classifiers used to create predictive models will be more data-dependent and dependent on the study's scope. The discussion revealed that previous studies differed in terms of data pre-processing procedures and dataset types used, making it difficult to compare the effectiveness of data mining approaches fairly.

# CHAPTER 3

# RESEARCH METHODOLOGY

## 3.1 Introduction

This chapter covers the research technique as well as the theoretical framework that underpins it. It explains each stage in the approach. Its purpose is to explain the technical jargon that will be utilized throughout the research. Each step of the investigation will be explained in depth in the research approach. Data set definition, exploratory data analysis, data preparation, imbalance handling, feature selection, data mining techniques, and evaluation of results will be discussed in these parts. Oversampling, under sampling, SMOTE, SVM-SMOTE, and combinations of these are among the class balancing approaches discussed. Various feature selection approaches, such as recursive feature selection, elastic net, and random forest classifier, as well as their combinations, are discussed. The data mining section will discuss the supervised learning method, which is a classification methodology, and how it can be used to detect Alzheimer's disease. The confusion matrix and performance evaluation measures will be explained in detail in the evaluation of results section, particularly regarding the technical jargon that will be employed. Following that, the proposed classification methods will be discussed, including the underlying theories and principles of six methods namely, Extra tree classifier, ANN, Logistic regression, LGBM, Bagging with KNN, and Stacking blending classifiers.

## 3.2 Research Methodology

Alzheimer's disease is a neurological disorder that begins slowly with some mild symptom of memory loss and subsequently affect a person’s ability to carry out daily activities. Alzheimer's disease affects parts of the brain that control thinking, memory, and language. It is the most common form of dementia, the cause of which is very little known to scientists. Current research says, a genetic mutation may be the cause of initial Alzheimer's stage, and a complex series of brain changes may be the cause of disease advancement. A combination of genetic, environmental, and behavioural factors may be the possible cause of the disease. Alzheimer is more prominent among older adults. Researchers are studying how ageing in brain can influence other brain cells and damage connectivity among neurons which leads to Alzheimer's disease. These ageing in brain cell involves atrophy of certain parts of the brain, soreness, nerves damage, free radicals' production, and mitochondrial malfunction (Alzheimer's Disease, National Institute on Health, NIA, 2021).

Alzheimer's disease is one of the leading fatal illnesses in the United States. It is the $6^{th}$ deadly illness among American adults. In 2021, approximate 6.2 million Americans aged 65 and above is having Alzheimer's dementia. This number is projected to be nearly 12.7 million people by 2050 (Alzheimer's Facts and Figures, Alzheimer's Association, 2021).

There are several research that has been conducted to predict the Alzheimer's disease and its stages. In (Almubark et al., 2020; Gorji et al., 2020), authors used the mean imputation technique for handling missing values, which is not a good technique for substituting null values because it does not consider the correlation among the columns while imputing with mean. Balancing the datasets before feeding them into the model is a vital step in ML pipeline. Some studies (Khoei et al., 2021) implemented an over-sampling strategy to deal with class imbalance problem, and the drawbacks of the technique are that it does increase the sample size of the minority class, and thus the possibility of overfitting, and does not add new information to the dataset. There are some other related studies (Gorji et al., 2020; Khoei et al., 2021), which used L1 regularization for feature selection, and the limitation of this method is that it does not consider the collinearity among the variables and randomly makes one of the features' coefficients zero. To address these issues, advanced predictive modelling algorithms along with efficient missing value imputation, class imbalance handling, and feature selection techniques will be implemented in this specific project.

The entire workflow of research methodology for this study is given in below figure 3.1:

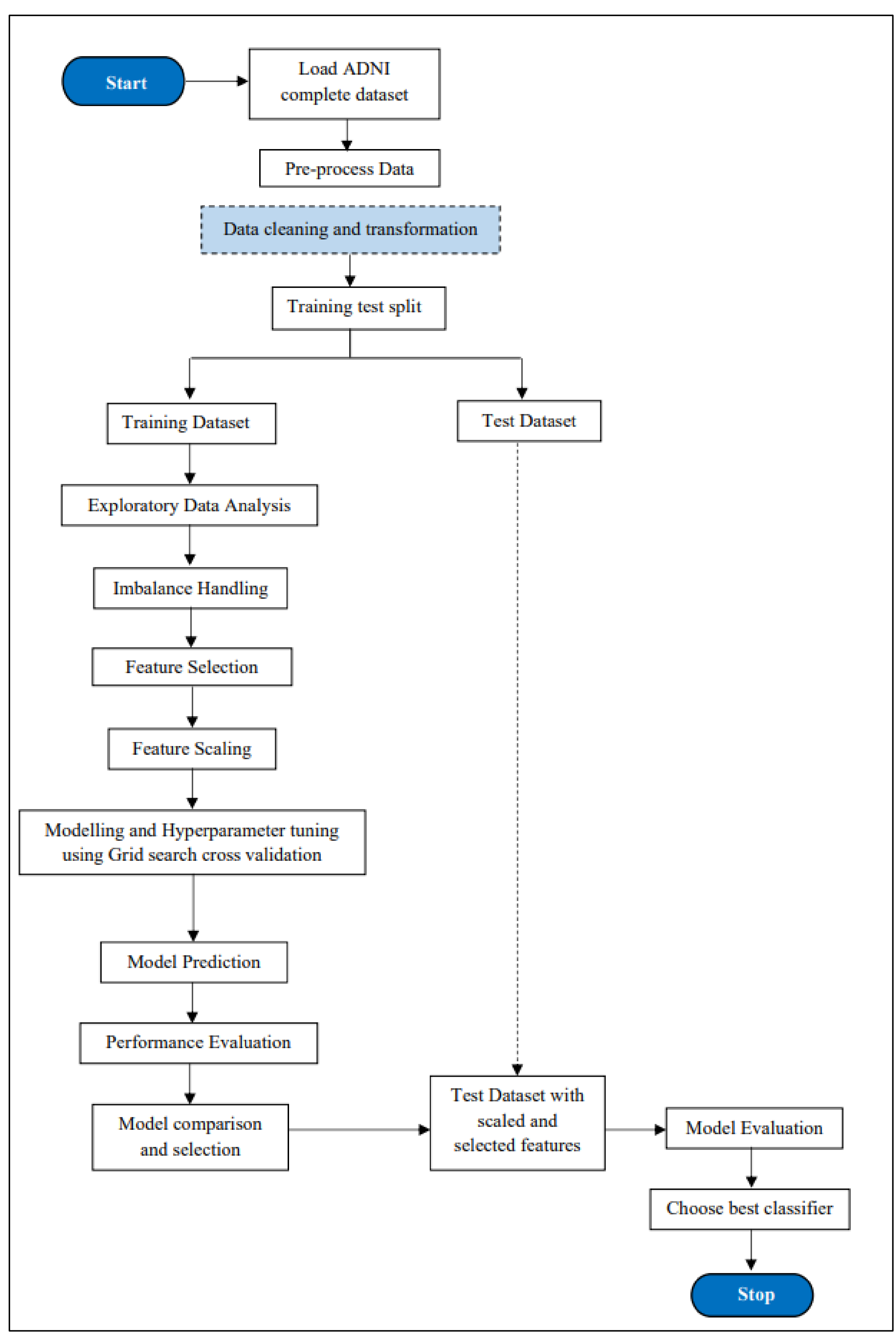

*Figure 3.1 Research Methodology Workflow*

### 3.2.1 Dataset Description

In disease-related analysis, the desired data selection is critical since it impacts the prediction model's performance. Large volumes of health data are created at a rapid rate in a variety of forms, necessitating the use of data mining tools to comprehend the data and extract useful information. The dataset used for this research has been collected from Alzheimer's Disease Neuroimaging Initiative database (ADNI, Alzheimer's Disease Neuroimaging Initiative Database, 2020). The ADNI was launched in 2004 as a public-private partnership and continues their study on improve diagnostic methods for early detection of AD and improve clinical trial design. The ADNI database has enrolled cognitively normal, MCI, and AD participants aged between 55 and 90 who are recruited at 57 sites in the United States and Canada. For this study, a subset of the ADNI dataset called ADNIMERGE will be used, which is updated daily directly from the Alzheimer's Disease Cooperative Study data system. This datafile contains 116 attributes from four ADNI study phases (ADNI-1, ADNI-GO, ADNI-2 and ADNI-3) which includes demographics, neuropsychological testing scores, MRI and PET summaries, CSF measures of both a baseline diagnosis and follow-up assessments performed every 6 months. The datafile is imbalance and comprises of numerous missing values, hence pre-processing of the raw data needs to be done, before passing it to the ML pipelines.

Below table 3.1 describes some of the important variables and its description extracted from the original dataset for better understanding of the predictor variables for Alzheimer's Disease prediction.

*Table 3.1 Alzheimer's Risk Parameters and their Definitions*

| Alzheimer's Risk Parameters | Attributes | Definition |
|---|---|---|
| Baselines Demographics | AGE | As predictors, age, gender, ethnicity, race, marital status, and level of education were all used. Demographic data was converted to numeric format. |
| | PTGENDER | |
| | PTEDUCAT | |
| | PTETHCAT | |
| | PTRACCAT | |
| | PTMARRY | |
| Functional Activities Questionnaire | FAQ | This can be used to examine a subject's dependency on some other people to carry out typical daily chores. FAQ is |

| | | made up of questionnaires and multiple-choice questions that result in an overall score ranging from 0 to 30. |
|---|---|---|
| Mini-Mental State Exam | MMSE | It is used to determine the level and development of cognitive impairment, as well as to track an individual's cognitive changes over time. |
| PET measurements (Participant's Brain function measurements) | FDG | Average FDG-PET of angular, temporal, and posterior cingulate |
| | PIB | Average PIB SUVR of frontal cortex, anterior cingulate, precuneus cortex, and parietal cortex |
| | AV45 | Average AV45 SUVR of frontal, anterior cingulate, precuneus, and parietal cortex relative to the cerebellum |
| MRI measurements (Structural measurements) | Hippocampus | The volume of hippocampus |
| | ICV | The Intra Cranial Volume |
| | MidTemp | The volume of the middle temporal gyrus |
| | Fusiform | The volume of the fusiform gyrus |
| | Ventricles | The Volume of ventricles |
| | Entorhinal | The volume of the entorhinal cortex |
| | WholeBrain | The volume of Brain |
| Apolipoprotein E4 | APOE4 | An integer value indicating the presence of epsilon 4 allele of the APOE gene. |
| CSF Biomarkers | ABETA | These are cerebrospinal fluid (CSF) biomarker measurements. |
| | TAU | |
| | PTAU | |
| RAVLT | RAVLT_Immediate | The Rey's Auditory Verbal Learning Test (RAVLT) is a set of neurophysiological assessments that measures a patient's episodic memory. |
| | RAVLT_Learning | |
| | RAVLT_Forgetting | |
| | RAVLT_Perc_Forgetting | |
| Everyday cognitive evaluations | EcogPtMem, EcogPtLang, | Everyday cognitive (Ecog) assessments are questionnaires that demonstrate a |
| | EcogPtVisspat, EcogPtPlan etc. | person's ability to execute day-to-day chores. |
| Neuropsychological tests to assess cognitive abilities | LDELTOTAL | Logical Memory - Delayed Recall Total - A neuropsychological test that examines a people's ability to remember information after a timeframe is the number of storey units recalled. |
| | ADAS11 | This test has been demonstrated to be effective in distinguishing between dementia patients and healthy controls, as well as in determining the degree of cognitive decline. Sum of 11 elements of the ADAS-cog |
| | ADAS13 | Sum of 13 elements of the ADAS-cog |
| | MOCA | Montreal Cognitive Assessment evaluates an individual's cognitive ability (e.g., memory, visuospatial, etc.). |
| PACC test | mPACCdigit | ADNI modified Preclinical Alzheimer's Cognitive Composite (PACC) with Digit Symbol Substitution |
| | mPACCtrailsB | ADNI modified Preclinical Alzheimer's Cognitive Composite (PACC) with Trails B |
| Unique primary keys | RID | Participant roster ID |
| | VISCODE | Visit code |

### 3.2.2 Exploratory Data Analysis

Exploratory Data Analysis is an approach for performing initial investigation on data to discover underlying patterns, detect outliers, test hypotheses, identify missing values, and to check correlations with the help of summary statistics (quantitative) and graphical representations. To solve any data science problem, EDA is very essential to first understand the problem statement and to identify the relationships between data features. As this study involves data from the health sectors, it has several missing values, which need to be analysed during the data analysis phase. Further analysis on the data distribution on different attributes in this dataset during EDA can be useful during feature engineering.

### 3.2.3 Data Preprocessing

A technique of transforming raw data into an understandable format that better represents the underlying problem statement to the predictive models, which improves efficiency and accuracy of the model on unseen data is known as Data Pre-processing. It is an important step in ML pipeline as machine cannot interpret the raw data format. Data cleaning, feature scaling, data formatting, feature encoding, data integration etc. are some of the major tasks of Data Pre-processing step. Also, analysis of duplicate entries is part of this step which is done based on the key columns present in the dataset.

#### 3.2.3.1 Missing Value Imputation

The data used for this research is medical data and contains missing values because not all attributes apply to all patients participated in this study and some data are missing during data collection. When imputing missing values, if any feature contains larger missing values (more than a specific threshold), their imputation will alter the entire data set. Therefore, it is good practice to remove these columns instead of imputing the same. There are few columns in this study data file, that contains greater than 45% missing data, which needs to be handled specifically.

Attributes with less than 45% missing values will be treated before feeding them into machine learning models because ML algorithms does not perform well with missing data. In (Almubark et al., 2020; Gorji et al., 2020; Khoei et al., 2021) studies, authors used mean imputation technique for handling missing values. In mean imputation, the mean of a particular column (which contains NaN values) is calculated, and the NaN values of that column are replaced by the mean value. But this is not a good and robust technique for substituting null values because this simple mean imputation process does not consider the correlation among the columns while imputing with mean.

There are two powerful techniques compared to simple imputation that yield a better predictive model when used for imputation. They are KNN Imputation and Iterative Imputation technique. Based on the problem statement, KNN Imputer sometimes works better than Iterative Imputer or vice versa. None of the study have used these robust methos yet. For this study, Iterative Imputation method will be used over KNN Imputation for handling null values of continuous variables, because KNN Imputer returns most similar neighbours based on distant metrics and replaces the missing values with the average of its neighbours. Therefore, it does not always give the most appropriate results for some problems as mean is involved in the computation.

For handling categorical columns such as VISCODE, PTGENDER etc. simple mode Imputation technique will be used. The subsections below describe mode imputation and Iterative imputation in details.

#### 3.2.3.1.1 Simple Imputation Technique

A common and simple procedure for entering missing data includes statistical methods to compute a value in a column from all its known values, then impute all missing values of that column using the calculated statistics. It is common as many times this technique proves very effective based on problem statement and it is simple because statistics are computationally efficient irrespective of data volume.

The following statistics are calculated as part of simple Imputation strategy.

- Mean Imputation – Here the NaN values are replaced by the mean value of the column. This imputation strategy works only for numerical data columns. However, it is unsuitable to use if the column comprises outliers/anomalies.
- Median Imputation – Here the null values are getting replaced by median of the column. This is also only applicable for numeric data columns.
- Mode Imputation – This technique uses the most occurring value of a particular column to fill the missing value. It can be applied on both numerical as well as categorical column.
- A constant value – The missing value of both the numerical and categorical column can be substituted by a constant value.

In this study, the null values in the categorical columns will be replaced by mode imputation strategy using SimpleImputer function from sklearn library.

#### 3.2.3.1.2 Iterative Imputation Technique

Iterative Imputation technique is a missing value handling approach where each feature is modelled as a function of other features. It is called iterative because the procedure of missing value imputation is repeated over and over, allowing for the calculation of continuously improved estimates of missing values because missing values across all features are estimated. This strategy is useful while dealing with multivariate data. It is inappropriate for treating categorical columns, only columns with numerical data are imputed by this method. Various regression algorithms can be used to estimate the missing values for each continuous feature, although linear methods are preferable for the ease of simplicity. The steps below define the more granular description of this approach (Zhu et al., 2011).

1. A regressor algorithm needs to be passed as an estimator of the round-robin imputation.
2. A feature column having missing data is selected as an output column y.
3. Other feature columns are chosen as input columns X.
4. The dataset is now split into train and test set, where the training set contains all the rows with known data of the feature to be imputed, and the test set is containing the rows with missing samples.
5. A regressor is fit on (X, y) for known y.
6. After that the regressor is used to predict the missing values of output column y.
7. This process continues till all the columns with missing data are imputed.
8. The above steps 1 to 7 comprise a single iteration. These steps are carried out again and again to allow ever improved estimates of the missing values.

#### 3.2.3.2 Feature Scaling

Feature Scaling is one of the most critical data pre-processing steps, to align the independent features present in data in a fixed range. In real-world data, different features are represented in different units. Columns with highly varying magnitude needs to be scaled, otherwise some machine learning algorithms will give more weightage to them comparing to the low valued columns, which is not correct way of modelling. However, there are some algorithms which are virtually insensible to it.

Algorithms like linear regression, logistic regression, neural network etc. which use gradient descent as an optimization technique needs to be scaled, because the steps for gradient descent are updated at the same rate for all the features to reach the global minima smoothly. It can help the gradient descent converges faster towards minima. Some distance-based algorithms like KNN, SVM, K-Means also require scaling as they utilized the distance between data points to find neighbouring datapoints. Models will become biased towards the larger valued columns, which will have impact on the performance of the model.

There are two kind of scaling technique namely,

1. Normalization
2. Standardization

Normalization, which is also known as Min-Max scaling, is a scaling approach in which the features values are scaled and shifted between 0 and 1. That means, all the data values of the columns lie between 0 and 1. The following equation 3.1 is for normalization,

$$x_{ne\omega} = \frac{x - x_{min}}{x_{max} - x_{min}} \tag{3.1}$$

Standardization, often known as Z-score, is another feature scaling technique, where feature gets transformed by subtracting from mean and dividing by standard deviation. Here the data points are not having a bounding range, instead they are centred around the mean with a unit standard deviation. This scaling strategy is not affected by outliers as the transformed features are not ranged between predefined value like normalization. The equation 3.2 for standardization is as follows,

$$x_{\eta e\omega} = \frac{x - \mu}{\sigma} \tag{3.2}$$

Researchers in (Thapa et al., 2020; Khoei et al., 2021) papers used standard scaler method for transforming the unscaled data. While (Gorji et al., 2020) paper rescaled all datapoints between 0 and 1 by applying min-max scaler. Depending on the underlying data distribution during the implementation phase of this project, the scaling approach will be decided. If the data follows a Gaussian distribution, standardization can be helpful. If the distribution is not Gaussian, normalization is useful.

### 3.2.4 Imbalance Handling

Data imbalance means that in a classification problem, the number of observations in one class differs from the number of observations in the other class. This is a major problem in both binary and multi class classification problem, and if not treated then results in a misleading predicting model. If there is majority in one class in training dataset, the classifier trained on that data, is biased towards that same class.

Unbalanced datasets are very common problem in health data, as there are usually fewer people with certain medical conditions than healthy people. For example, the data set proposed for this study contains information on 1256 people at baseline visits, of which 409 were diagnosed with dementia and 848 were cognitively normal. Hence, for this research problem, dementia has become a minority group over majority of normal people. The presence of majority class is almost two times higher than the minority class.

For diagnosing medical domain problem, prediction accuracy is very much crucial for any machine learning model. If the model is trained with imbalance data, it will get a higher accuracy just by predicting all data points as majority class. All the data points with minority class will be misclassified and hence, the model will fail to diagnose the healthcare problem efficiently. Therefore, checking and treating imbalanced data is a vital step before fitting the learning-based model.

The main objective of imbalance handling technique is to create equal number of instances for both the majority class and the minority class by increasing the number of instances of the minority class or decreasing the number of instances of the majority class.

There are three commonly used resampling techniques as follows:

- Random Under-Sampling
- Random Over-Sampling
- Synthetic Minority Over-sampling Technique (SMOTE)

Random under sampling method randomly eliminates data points from majority class to balance class distribution. This is done until instances of the majority class and the minority class are balanced. This method is inefficient when the dataset contains a smaller number of data points. Also, removing random samples may discard potentially useful information, which might be important for model training. Therefore, provides inaccurate outcome for the actual test dataset.

Random over sampling technique increases data points from minority class by randomly repeating them in the dataset. There is a related study (Khoei et al., 2021) which implemented this over sampling strategy to replicate samples of the Alzheimer's class (minority class). Although this does not result in loss of information like under sampling, it does increase the sample size of the minority class, and thus the possibility of overfitting. It also does not add new information to the dataset by replicating the minority class samples.

In Synthetic Minority Over-sampling technique, new synthetic samples are generated by duplicating the samples of minority class and the created synthetic instances are added to the original dataset. To improve the performance of machine learning models, the researchers in (Thapa et al., 2020) paper used SMOTE method to solve the problem of unbalanced data by generating synthetic instances of minority AD classes. This resampling technique is more efficient and robust than random over sampling and under sampling technique. This is because creating synthetic points not only eliminates loss of information, but also eliminates overfitting problem. Despite being advantageous than simple over sampling or under sampling methods, SMOTE can introduce some ambiguity in dataset by increasing overlapping classes as the neighbouring majority class samples are not considered by this method.

In this research, an advance version of SMOTE technique, called SVM-SMOTE will be used, which is more powerful than SMOTE. This method uses SVM algorithm to select minority class samples to use for generating new synthetic data points.

#### 3.2.4.1 Imbalance Handling using SVM-SMOTE

SVM-SMOTE is an enhanced variation of SMOTE. This technique uses line segments to connect minority class points and then places artificial points on those lines. The step-by-step working of SMOTE algorithm are as follows:

1. First, select a minority class set for generating new points.
2. For each data sample x from the above chosen set, find the k nearest neighbours by calculating the Euclidean distance between x and every other sample in the set.
3. Next, choose one of the neighbouring points and place a synthetic point on the line joining the point x and its chosen neighbour.
4. For each neighbour xk (k=1, 2, 3…N), the new sample point is generated utilizing the formula, $x' = x + rand(0, 1) * |x - xk|$ in which rand(0, 1) represents the random number between 0 and 1.
5. Repeat the step 1 - 4 until the data is balanced for all the classes.

As discussed in previous segment that SMOTE is having few drawbacks due to considering only the minority class samples which result in overlapping class samples, an advance variant of this namely, Borderline-SMOTE has been introduced. Unlike SMOTE, which created synthetic points randomly between two data points, Borderline-SMOTE generates new datapoints along the decision boundary between two classes.

As Borderline-SMOTE generates synthetic data points near the overlapping class boundary regions, it is more suited to use when the misclassification happens near the boundary decision. This limitation of Borderline-SMOTE can be mitigated by SVM-SMOTE technique. The main difference between SVM-SMOTE and Borderline SMOTE technique is unlike Borderline-SMOTE, which uses nearest neighbours to identify the misclassification, SVM-SMOTE utilizes SVM algorithm to determine the same. Here, the borderline points are determined by the support vector points, where the data points are well separated, which makes SVM-SMOTE special by synthesizing new samples away from the class overlapping region (Tang et al., 2009).

### 3.2.5 Feature Selection Method

When designing machine learning solutions for real life problems, feature selection is an important part of eliminating insignificant features because real-world dataset usually contains a lot of features. During model training, unimportant features, not only degrade the accuracy of the model, but also increase the complexity of the mode, reduce the generalizability of the model, and introduce bias into model predictions. Therefore, it is very vital to reduce dimensionality to mitigate overfitting, reduce training time and increase decision-making power of model.

Various manual, semi-manual and automated techniques are available for choosing best set of relevant features. Following are the three broad categories of feature selection methods:

- Filter Based Method
- Wrapper Based Method
- Embedded Method

Filter based method selects significant features from dataset using mainly statistical methods, which are faster and computationally inexpensive. It does not involve any machine learning algorithm, instead it uses Pearson's, Chi-Squared, ANOVA etc. statistics. Although this method is good for eliminating correlated and redundant features, it is unable to remove multicollinearity as it looks for pair-wise correlation.

Wrapper based selection method finds feature combination based on specific machine learning algorithm and the algorithm stops when predefined number of top features are selected, or the performance of the model starts to decrease. Multiple models are evaluated against a specific evaluation metrics that could be p-value, Adjusted R-squared in case of regression problem and precision, recall in case of classification problem that add or remove independent features to find the optimum group of predictors that maximizes model efficiency. Although this method is computationally expensive than filter method, it provides more appropriate set of predictors for training the model. Recursive feature elimination, Forward and Backward selection etc. are few examples of Wrapper method.

Embedded selection method combines the advantages of both filter and wrapper-based feature selection methods but also maintains acceptable computational cost. This technique is having its own built-in feature selection or feature importance calculation procedure, which make it more beneficial than other methods. L1, L2 regularization, Elastic nets, Tree-based methods like Random Forest, Gradient Boosting etc. are few examples of Embedded selection technique. In regularization embedding process the model penalizes the coefficients such that it controls overfitting and the feature with high penalty (least or zero coefficient value) can be eliminated from dataset. On the other side, tree-based techniques calculated feature importance to find the optimal set of features which are having maximum impact on target variable.

In (Almubark et al., 2020) study, the authors used both wrapper-based feature selection techniques such as Sequential Forward Selection, Sequential Backward Selection, RFE, and filter-based technique called SelectKBest. Another study (Khoei et al., 2021), computed Pearson's Correlation as a filter-based selection method and applied Lasso Regularization, Extra tree-based technique as embedded methods for determining optimum input variables. A similar study (Gorji et al., 2020), which was aimed to detect related biomarkers, utilizes all three categories of feature selection method. This includes Pearson's Correlation and Chi-Squared statistical methods from filter-based, RFE and step forward feature selection from wrapper-based and Lasso and Tree-based algorithms from embedded method. From these above studies, it is observed that tree-based selection technique has performed better in most of the cases.

Of the three feature selection categories discussed above, this study will implement three selection methods in two categories: RFE from wrapper-based method and Elastic net, Random Forest tree from embedded based method and evaluates model performance for each set of features. Next few sub-sections illustrate the details of these three selection methods and their advantages.

#### 3.2.5.1 Recursive Feature Elimination Technique (RFE)

For both the classification and regression-based machine learning problems, RFE is the most widely used method to identify optimal set of features that are more relevant in prediction. It is a very popular and useful method for feature selection in many machine learning solutions because of its simple configuration steps and the efficiency of selecting the most suitable features (feature ranking) from the training set based on coefficient values or feature importance. For these stated advantages RFE will be used in this project in conjunction with other selection techniques. RFE is considered as a wrapper-based method because different machine learning algorithm is introduced and used in the core of the method, which is wrapped by RFE. It takes a supervised model as an estimator parameter and fits the model on the entire set of predictors. Then it computes the importance score for every predictor by the model's coefficient or feature importance attributes and ranks them according to the importance score. The predictor with least important score is removed and the model is retrained again with the remaining predictors. This recursive elimination and retraining process continues per loop until the specified number of features is reached (Ustebay et al., 2019). The main configuration options involve selection of number of features and the estimator, which are also known as hyperparameters. Hyperparameters needs to be tuned such a way the model's performance increases.

#### 3.2.5.2 Feature Selection using Elastic Net

The two major problems in any machine learning models are Overfitting and Underfitting, which occur due to improper data pre-processing, incorrect tunning of hyperparameters etc. Any ML model overfits when it starts learning every detail including the noise present in training data and performs very well on seen data, but poorly perform on unseen test data. On the other hand, Underfitting happens when model unable to perform well on both training and test data.

Regularization is a technique which helps to address the model overfitting problem by shrinking the coefficient estimates towards zero. The commonly used regularization techniques are as follows:

- LASSO Regularization (L1)
- RIDGE Regularization (L2)
- Elastic Net Regularization

Some of the related study (Gorji et al., 2020; Khoei et al., 2021), used Lasso (L1) regularization, although it has some limitations. If there are two or more collinear variables exist in dataset, Lasso randomly selects one of them, which may lose the interpretability of data. Also, if the number of predictors (p) is more than the number of observations (n), Lasso will choose at most n non-zero predictors, even if all the predictors are important for the training. In contrast, Ridge (L2) regularization is not an efficient technique for feature selection because it only minimizes the coefficient values, but not make it zero, which makes it unsuitable for feature reduction. To create an interpretable ML model, it is very important that the relevant meaningful variables be selected using feature selection technique by removing irrelevant predictors. None of the Ridge or Lasso method individually satisfies both the said criteria, therefore they are not applicable to this specific research. Instead, a hybrid technique that utilizes the strengths of both ridge and lasso will be used for this case study.

Elastic Net is a regularized method that linearly combines the L1 and L2 penalties of the lasso and ridge methods. The L1 part of penalty creates a sparse model, whereas the L2 part of penalty stabilizes the regularization path L1, average out the coefficients of correlated features, which enforces a grouping effect and removes the limit on the number of selected features. Thus, Elastic Net has learned from the weaknesses of ridge and lasso and combines both to achieve most relevant feature set. Therefore, Elastic Net is a two-stage procedure, where the first stage is to calculate Ridge coefficients and then apply the lasso on them for shrinking the coefficients. This way Elastic Net performs feature reduction without easily eliminating high collinearity coefficient (Challita et al., 2015).

Below equation 3.3 is the regression function of Elastic Net with penalty term, where $\sum \beta_j^2$ is the ridge penalty term and $\sum|\beta_j|$ is the lasso penalty term.

$$L_{enet}(\hat{\beta}) = \frac{\Sigma_{i=1}^{n}\left(y_i - x_i^T\hat{\beta}\right)^2}{2n} + \lambda\left(\frac{1-\alpha}{2}\sum_{j=1}^{m}\hat{\beta}_j^2 + \alpha\sum_{j=1}^{m}\left|\hat{B}_j\right|\right) \quad (3.3)$$

#### 3.2.5.3 Feature Importance using Random Forest

Random Forest is an ensemble learning technique, consisting of several decision trees, where each tree is trained on a random extraction of observations from dataset and random extraction of the features. The final decision making by the forests is done by averaging the outcome from each single tree in forest in case of regression problem or by taking a majority vote in case of classification problem. It is also known as Bagging algorithm because each tree in forest builds on a bootstrap sample from dataset, which maintains the diversity among each learner in the ensemble. RF is having very good predictive power, low overfitting, high generalizability.

Each decision tree in the forest is build following below steps:

1. Random samples are created by sampling the training data with replacement.
2. For any node, the best split is calculated by searching in a subset of randomly selected features of size square root of number of predictors.
3. The feature for which the normalized reduction in Gini Index is maximum, is considered for the split.

4. The size of each decision tree in the forest is decided by the max_length hyper parameter of the model.

Random Forest performs feature selection using its own build-in importance calculation technique. RF uses tree-based strategies to naturally rank the features by how well they improve purity (Gini impurity) of a node by decreasing the overall impurity. The nodes which are at the start of the trees are having highest decrease in impurity, therefore features at those top nodes are most important. While the nodes which are at the end of the tree are having least decrease in impurity, therefore those bottom nodes contain least important features. Feature importance of each decision tree is calculated by averaging the Gini index for each feature in a single tree. Therefore, the average Gini over all trees in the forest is the measure of the ensemble's feature importance (Peng et al., 2019).

(Gorji et al., 2020; Khoei et al., 2021) used Extra tree-based embedded method for computing feature importance, which works almost similar like Random Forest, but there is slight difference in the selection of samples and splitting the nodes. Random Forest based feature selection will be used in this project to see if the model's performance improves or declines, as it has great generalisation ability, is simple to apply, and provides explicit ranking results for all features.

## 3.3 Proposed Classification Methods

Selecting the appropriate classification model for properly segregating the early stages of Alzheimer's Disease is very important for obtaining better predictive power of the system. Before training the model on the dataset, several data pre-processing tasks such as data cleaning, imbalance handling, data transformation, encoding, feature selection etc. needs to be performed. Therefore, dataset is divided into training and test set, where the classifier learns from training set and the learned model parameter is used to discriminate the new observation in test set. In the following section different classification algorithms along with their advantages and why these methods should be for this specific research work, will be discussed in brief.

### 3.3.1 Logistic Regression

Logistic regression is a supervisor learning algorithm which is used for solving classification problems, where the dependant (target) variable is categorical in nature. This predictive analysis algorithm works on the concept of Probability. Hence, the outcome is always between 0 and 1 as probability value cannot exceed 1. Logistic Regression is a parametric modelling technique as it works on assumption that target variable is categorical or discreate is nature and zero collinearity exists among the predictors. Instead of fitting a best fit regression line, logistic regression fits a sigmoid (logistic) function which is a “S” shaped curve, ranges between 0 and 1. Sigmoid function maps the predicted values to probabilities. The sigmoid curve is shown in below figure 3.2:

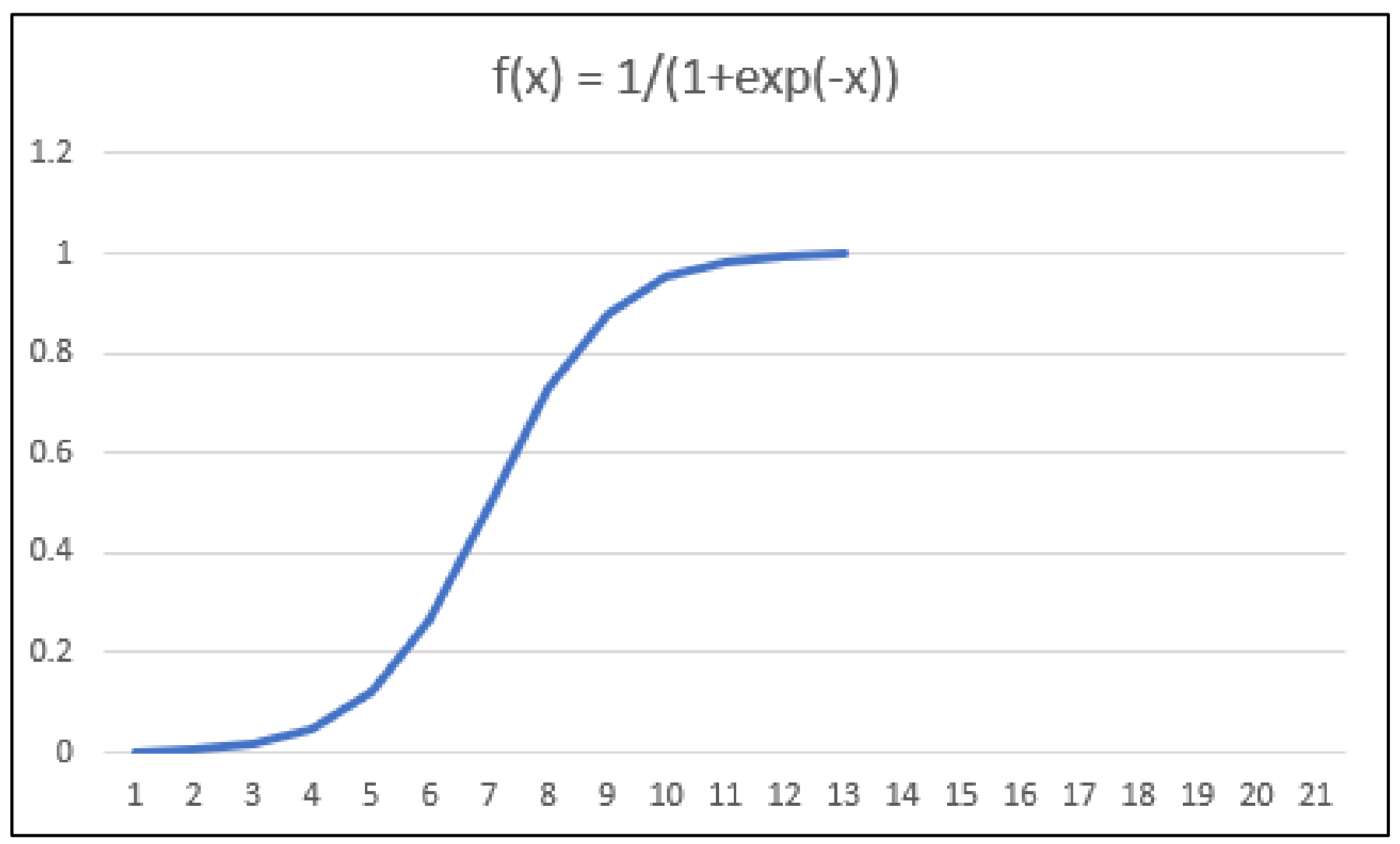

*Figure 3.2 Logistic Function Curve*

The equation 3.4 for the sigmoid (logistic) function is as follows:

$$f(x) = \frac{1}{1 + e^{-x}} \tag{3.4}$$

This curve indicates the likelihood of something such as whether the patient is suffering from a medical condition or not based of its weights (Dingen et al., 2019).

In (Lodha et al., 2019; Memon et al., 2020), Logistic Regression was used for detecting AD dementia. This algorithm is preferrable in many research studies because of its simplicity, ease of implementation, less prediction time, and interpretability. The model coefficients are the measure of the variable importance. As a medical field problem, building an interpretable model is very essential, which helps to understand the cause of the disease by selecting the important biomarkers. For these advantages, this algorithm will be used for this project.

### 3.3.2 Extremely Randomized Trees Classifier (Extra Tree)

Extra Tree, also known as Extremely Randomized Trees Classifier, is an ensemble learning technique, which is built on many decision trees. This is very similar to bagging algorithm, where each decision tree is built in parallel on a part of training data and combines the results of multiple decorrelated trees collected in the forest, to produce an outcome. Unlike bagging that develop each decision tree in the forest from a bootstrap sample, extra tree uses whole training sample. Also, like Random Forest, Extra Tree selects a random sub-set of features at each cut point of decision tree to split nodes. But Extra Tree chooses the split point at random, while random forest selects the optimal split. Ensemble's outcome is calculated by averaging the predictions from each decision trees in case of regression or using majority wins principle in case of classification.

The step-by-step procedure of how Extra Tree works is as follows:

1. Each decision tree in the extra tree forest is built with all the data available in the dataset.
2. A Random sub-set of features are selected to split nodes. This subset is selected by the equation sqrt (number of features).
3. Split points are selected randomly.
4. Best features to split is selected by calculating Gini Impurity.
5. Built each decision tree with the selected features on whole original sample.
6. Choose the number of decision tree that needs to be build.
7. Repeat the steps from 1 and 5.
8. For new datapoints, find the prediction from each of the decision tree and assign the class to the new datapoint that wins the majority votes.

Therefore, by randomly selecting the cut points, Extra Tree adds randomization, although it is maintaining optimization. The generated model is having low variance and sometimes it outperforms Random Forest based on training data (Camana Acosta et al., 2020).

In (Ezzati and Lipton, 2020), decision tree algorithm was used which is a high variance model, as they are very prone to overfitting if hyperparameters are not tuned correctly. This shortcoming of decision tree is overcome by ensemble algorithm Extra Tree as it consists of numerous decision trees, mistake of one tree can be mitigated by another tree in the ensemble. Thus, outcome is not affected by underperformance of any individual tree.

Ensemble method Random Forest was implemented in (Uspenskaya-Cadoz et al., 2019; Thapa et al., 2020). Although RF and Extra Tree is very similar ensemble method, Extra Tree algorithm is faster than RF, which minimizes the computational cost as well. For these above discussed advantages, Extra Tree classifier will be used in this research.

### 3.3.3 Light Gradient Boosting Machine (LightGBM)

LightGBM is a gradient boosting framework that uses decision tree algorithm, which can be used for solving both regression and classification problem. This lies in the boosting family of ensemble learning algorithm, which consists of n numbers of decision trees (weak learner), where each weak learner learns from the mistake of its previous weak learners. After multiple iterations weak learners are combined and form a strong learner that can generate more accurate outcome. Models are fit with a differentiable loss function and gradient descent optimization algorithm is performed on it, hence the name “Gradient Boosting”.

LightGBM in many ways special from ordinary gradient boost trees. It uses Gradient-based One-Side Sampling, which considered the training instances with larger gradients to compute the information gains. The way LightGBM builds the tree is also makes this algorithm beneficial than other tree-based models. It grows the tree vertically leaf-wise, while most other algorithms grow tree horizontally level-wise. Level-wise tree growth can result in unnecessary leaves and nodes as the tree building does not stop until the maximum depth reached. In contrary, trees with leaf-wise growth choses the leaf with max delta loss to grow, which reduces more loss than level-wise algorithm and hence results in much higher accuracy (Su and Zhao, 2020).

The advantages of LightGBM over other boosting methods are as follows:

1. It is computationally faster and offers high efficiency by binning the continuous features into discrete buckets which reduces number of splits computation and expedites the training procedure.
2. Performing binning on continuous values result in lower memory uses.
3. Leaf-wise tree growing approach of LightGBM helps it to achieve higher accuracy than any other boosting algorithm.
4. It can work well with large data sets with significantly reduced training time compared to XgBoost.
5. LightGBM can support parallel learning approach by utilizing GPU learning.

XgBoost algorithm was used in (Shah et al., 2020; Thapa et al., 2020) for classifying stages of Alzheimer. But the XgBoost tree is having few drawbacks over LightGBM, such as it is slower than LightGBM, cannot automatically handle categorical variable like LightGBM, not perform well on sparse data.

Due to the above-mentioned advantages of LightGBM over XgBoost, in this study, LightGBM will be used for classifying different early stages of dementia.

### 3.3.4 Bagging Ensemble Classifier with KNN Base Estimator

Bootstrap Aggregation, or Bagging, is an ensemble machine learning algorithm, consists of many base estimators. It is called “Bootstrap” because each estimator in ensemble uses bootstrap samples of dataset with replacement. As the samples selected from a dataset is replaced, this allowing the sample to be reselected again in a new sample set. It is called “Aggregator” because the outcome from meta-estimator is the aggregation (voting for classifiers or averaging for regressors) of individual estimator’s prediction. In bagging each base estimators are trained in parallel in randomly drawing independent training samples and feature sets. Therefore, it can be able to reduce the variance present in individual base estimator and minimize overfitting by averaging or voting (Wang et al., 2021).

To identify the dementia progression among mild Alzheimer's patients, (Ezzati and Lipton, 2020) study used K-nearest neighbour algorithm, which was proven best over other ML algorithms with sensitivity of 80.1%. KNN is very simple to implement, no explicit training time is needed, hence can evolve with new data constantly, can handle non-linear data very well, having a single hyperparameter to tune (number of neighbours, k), it does not consider any assumption about underlying data distribution. Because of these advantages KNN is a popular choice in many research problems. To avoid the risk of overfitting (high variance problem), this study uses KNN as a base estimator inside the Bagging ensemble classifier. The working mechanism of K-nearest neighbour algorithm will be discussed in detail in next paragraph.

The architecture of Bagging classifier using KNN estimators is shown in below figure 3.3:

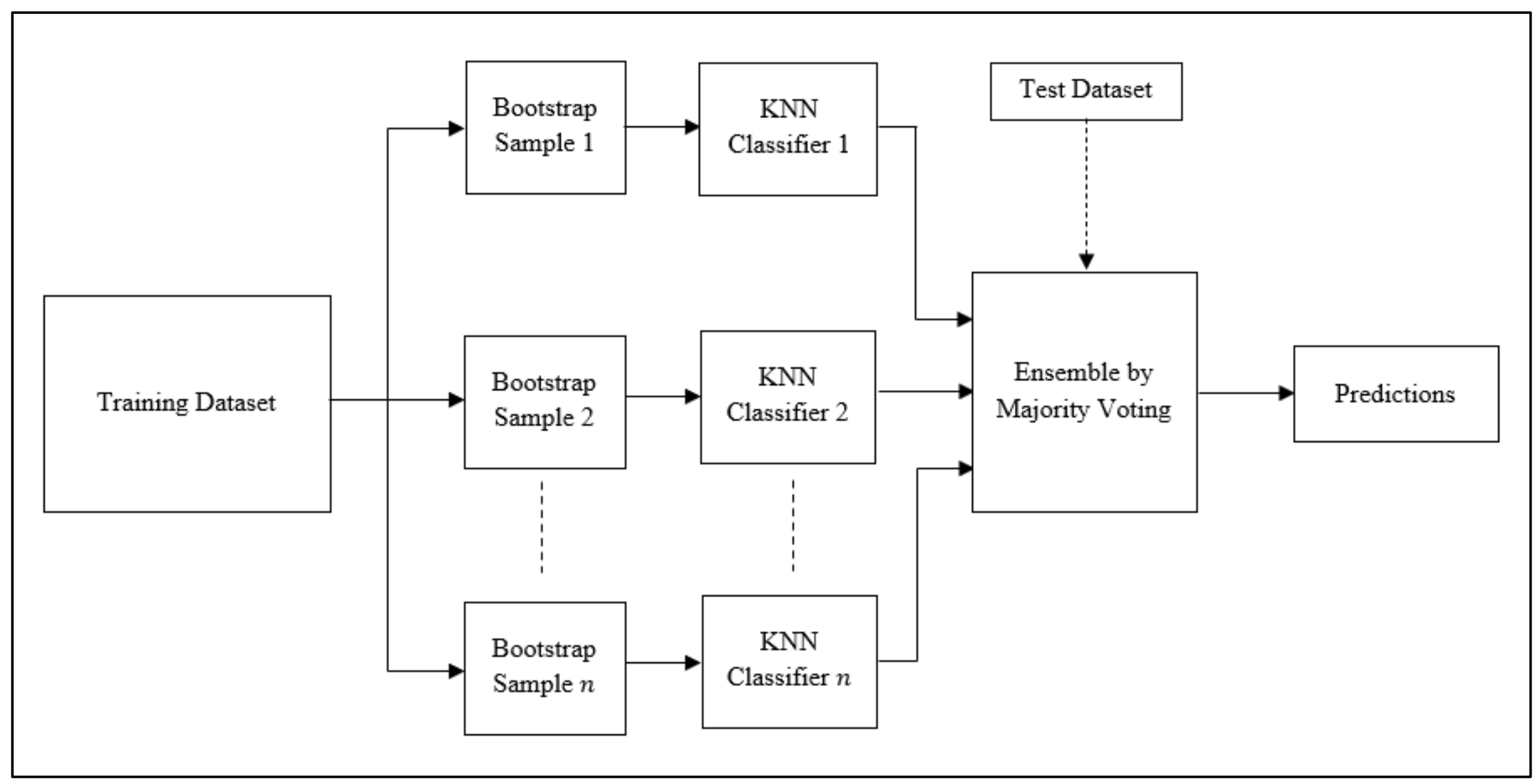

*Figure 3.3 Bagging-KNN Classifier Architecture*

KNN measures the similarity between unseen data and available training data points and groups the unseen data into categories similar to the available categories. KNN algorithm can be referred as lazy learner algorithm, as it does not involve much in training, instead stores the training dataset during training phase. When any new data point comes, it classifies the data from the stored dataset by calculating some distance metrics, into a similar category.

Choosing an appropriate value for k (number of nearest neighbour) is a key hyperparameter for tuning, which helps to achieve good accuracy. There is no specific rule to determine the value of k, so different values of k need to be passed to model and prediction error made by the model needs to be validated before finalizing the k value. If the value of k is low, there is overfitting of data (high variance) and error in test data is high. With increasing value of k, the test error starts decreasing. But if the k value is become too high again the test error goes high. Hence, the optimum k value is needs to be selected from where test error goes low and stabilizes and with further increase in K value, it again increases (due to bias) (Vieira et al., 2019).

The step-by-step procedure of how KNN algorithm works are as follows:

1. Loading the dataset.
2. Choose the value of number of nearest neighbour points (k) by hyperparameter tuning. K can be any integer.
3. Next, following steps needs to be performed for each unseen (test) data points.
    - Using any distance metrics (Euclidean, Manhattan etc.), calculate the distance between test data and each instance of training data. The most used method to

calculate distance is Euclidean.

- Then, sort the training data points in ascending order based on the distance value.
- Next, choose top k rows from the sorted collection.
- Now, among the k rows which is the most frequent occurring class, assign that class to the unseen test point.

4. End of prediction.

Therefore, because the KNN algorithm depends on the distance between data points, feature scaling and outlier processing before feeding data to the mode are important pre-processing steps.

### 3.3.5 Stacking-Based Ensemble Learning

Stacking, also known as Stacked Generalization, is an ensemble learning algorithm, consists of several heterogeneous base-models and combined the outcomes of these base-models to output the final prediction by training a meta-model. Here, base-models, which are capable to learn some part of the problem, are trained individually on same dataset. These multiple different learners can build intermediate predictions. Therefore, a new model is built to learn how to combine the intermediate predictions and predict the same target. This is called stacking because the final meta-model is stacked on top of the others. Thus, often the stacked model improves the overall performance than individual intermediate models.

The working mechanism of stacking ensemble model is as follows:

1. Split the training data in two groups.
2. choose n base-models (n >= 2) and fit them to data of the first group.
3. For each of the n base-models, make predictions for observations in the second group.
4. Fit the meta-model on predictions made by the base-models as inputs.
5. Next, meta-model is used to make a prediction on the second fold.

The more dissimilar the base-models, the more appropriate result will obtain. it is often a good idea to use a variety of models that make very different assumptions about how to solve the predictive modelling task. Thus, the bae-models are usually complex and diverse. In contrary, the meta-model is relatively simple and interpretable. Although the performance of the model is largely depending on the complexity of the problem, the representation of dataset, choice of base-model, in most cases stacking outperforms the individual models (Zian et al., 2021).

The architecture of stacking-based ensemble classifier is shown in below figure 3.4:

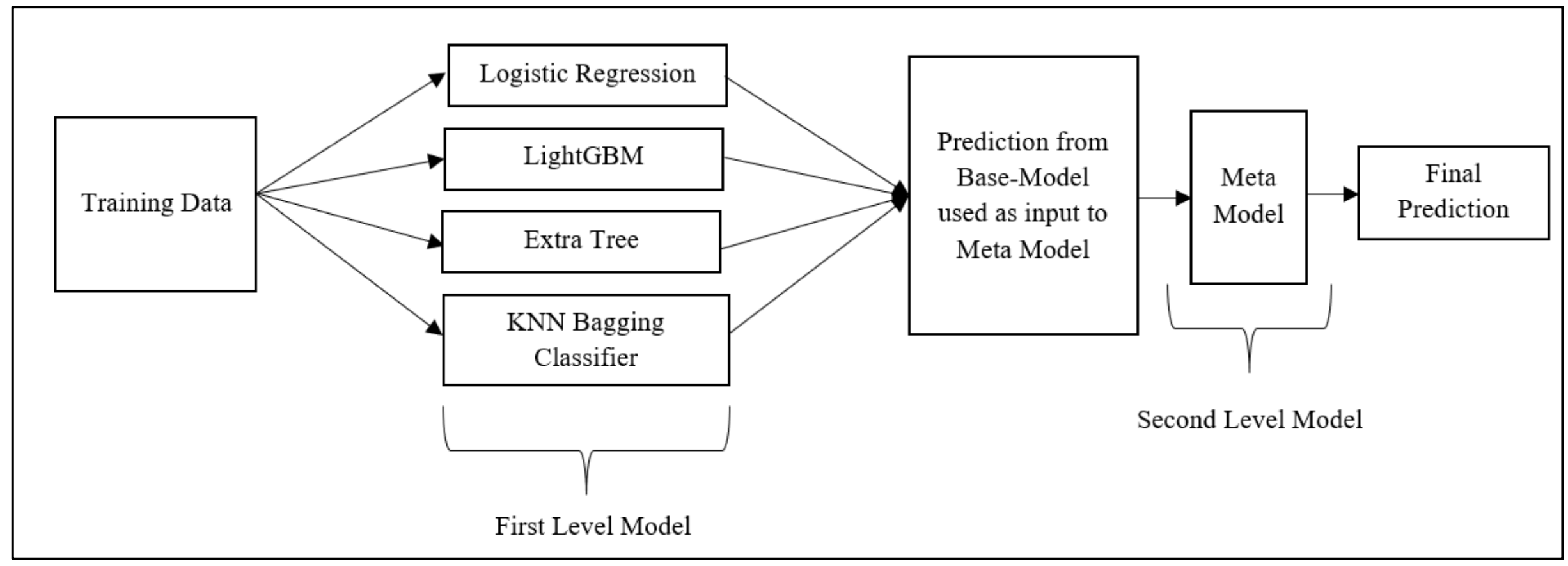

*Figure 3.4 Stacking Ensemble Architecture*

In this research, Stacking-based Ensemble methodology will be implemented using the modelling algorithms discussed in previous sub-sections. Where these individual models will act as base-model or first-level classifiers and therefore a meta-model will be stacked on top of the base-models to exploit the ability of each individual models to its very extent. Therefore, this kind of stacking ensemble model can provide better prediction than any individual model and improves model performance. The reason to include the same classifiers discussed in above sub-sections as the first level models is to demonstrate the power of stacking and to compare the evaluation results of the classifiers when working independently versus when functioning as part of a stacking ensemble technique.

### 3.3.6 Artificial Neural Network (ANN)

Artificial Neural Network is a branch of artificial intelligence that mimic the work of the human brain to process combinations of stimuli into output signals. Neurons are the important part of neural networks. Like human brains which consists of many interlinked brain cells, ANN is the collection of neurons which are interconnected to one another in various layers of the network. ANN is very popular because it can solve complex problem, understand things, and make decisions in a human-like manner. As the neurons or nodes in each layer works in parallel, ANN is often called Parallel distributed processing systems.

The network is constructed with a large number of artificial neurons arranged in sequence of layers. The main three layers of the network are as follows:

- Input Layer: This layer is responsible for receiving the inputs. There is always only one input layer in any neural network. The number of neurons in input layer is equal to number of attributes in dataset.

- Hidden Layer: Hidden layer is placed between input and output layers and perform calculations to find hidden features and patterns. Multiple hidden layers with different number of neurons can be present between input and output layer according to the complexity of the problem.
- Output Layer: The output from the last hidden layer, which goes through series of transformation, is the input of the output layer. Finally, the results generated from output layer is conveyed as the network prediction. The number of neurons in output layer is always equal to one for regression problem or the number of class levels in case of classification problem.

After receiving input from input layer, the network computes the weighted sum and adds a bias to it. Then, the weighted sum including bias is passed to the hidden layer for processing. If the weight associated with interconnection is $\omega^T$, input to the network is $x$ and bias is $b$, then the transfer function can be written as follows in equation 3.5:

$$z = \sum_{i=1}^{n} \omega^T \cdot x + b \tag{3.5}$$

The output from the activation ($\sigma$) layer, $a$ can be written as follows in equation 3.6:

$$a = \sigma(z) \tag{3.6}$$

The output from hidden layer (weighted total) is passed through activation function, which either dampen the signal or amplify the signal, thus keeping the neuron fired or turned off. The purpose of the activation function is to introduce non-linearity into the output of a neuron. The activation function performs a nonlinear transformation of the input, making it capable of learning and performing more complex tasks. SoftMax, sigmoid, RELU, ELU, Leaky-RELU etc. are commonly used activation functions. Among these SoftMax (multi-level classification) and sigmoid function (binary classification) are usually used at only output layer. The architecture of Artificial Neural Network is shown in below figure 3.5:

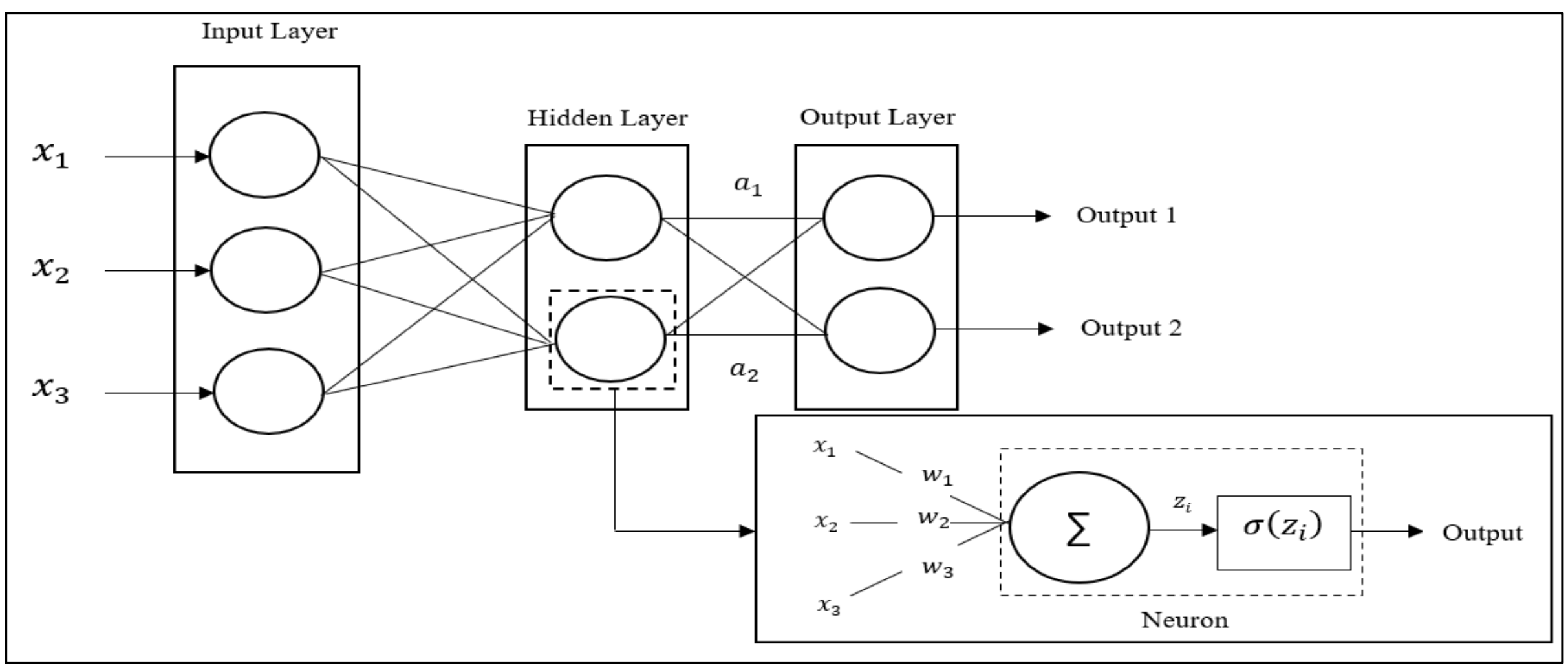

*Figure 3.5 Artificial Neural Network Architecture*

Step by step working of Artificial Neural Network is as follows:

1. Information is passed into input layer, which transfer the data into hidden layer.
2. The interconnections between the two layers assign weights to each input randomly.
3. A bias is added to the weighted sum, after the weights are multiplied with each input individually and summed up.
4. Next, the weighted sun including the bias is fed to the activation layer.
5. The activation function determines which input to pass on and which ones to supress. Accordingly, it fires the neurons for extracting hidden information.
6. The model applies an SoftMax or sigmoid activation function at output layer to deliver probabilistic outcome for classification problem.
7. Then loss is calculated (cross entropy loss for classification) based on predicted and actual output and weights are updated during feedback loop propagation to minimize the classification error.
8. This entire feed forward and feedback loan ops repeats to improve accuracy.

ANN can learn non-linear, complex relationship, have good fault tolerance, supports parallel processing, and can often generates more accurate model than any other ML models (Carkli Yavuz et al., 2018). Although ANN is difficult to interpret, keeping the aforementioned advantages in mind, Artificial Neural Network modelling will be implemented in this project to generate more accurate results for diagnosing and controlling the dementia progression. As the prediction accuracy is very crucial for any lethal disease prediction task, therefore, neural network will be implemented in this study.

### 3.3.7 Evaluation Metrics

An evaluation metric is a measure of how accurate a machine learning model is. Therefore, the choice of the appropriate metric is very crucial for both assessing the classification performance and guiding the classifier modelling. For quantifying the model predictive power following steps are usually performed in any machine learning solution.

- Training the model on a dataset, which is exposed to the model during training task.
- Using the model making predictions on holdout dataset, kept aside during training. Hence, unseen to model.
- Then comparing the predicted values to the expected values in the holdout dataset. In case of classification problem, expected class label is compared to the predicted class label.
- When the predicted label belongs to the actual expected class, it is called True Positive (TP). When the predicted label does not belong to a class and the actual outcome also the same, then it is called as True Negative (TN). Similarly, when the predicted class belongs to a class, but actual class does not belong to the same, then it is called as False Positive (FP) and when the predicted class does not belong to a class, but the actual is belonging to, it is called False Negative (FN).
- Base on the four types of outcomes the performance of any model can be quantified.

An evaluation metric must be chosen based on the type of problem the model tries to address. For some classification problems, stakeholders want to minimize false positives, while for some problem statements, it is more sensible to reduce false negatives. Therefore, the real problem statement, right evaluation metric needs to be finalized for achieving more again.
According to past related works on disease-related datasets (Woldemichael and Menaria, 2018; Kumar and Katyal, 2019; Shanthakumari and Jayakarthik, 2021) and some on Alzheimer's Disease datasets (Almubark et al., 2020; Shah et al., 2020), the most frequently used measures are primarily derived from the confusion matrix, which illustrates the four outcomes; the measures include Accuracy, Precision, Recall, Specificity, and F-measure. Apart from these, graphical display-based measurements such as the receiver operating characteristic curve and precision-recall curve are also used to verify the efficiency of classifiers (Uspenskaya-Cadoz et al., 2019; Khan and Zubair, 2020b; Thapa et al., 2020). The performance measures used to evaluate the classifier's performance include terminology that will be detailed below.

1. Accuracy: It is the proportion of the true outcomes among the total number of instances examined. The accuracy metric is shown in the below equation 3.7.

$$Accuracy = \frac{TP + TN}{TP + TN + FP + FN} \tag{3.7}$$

2. Precision: It explains how many of the correctly predicted instances actually turn out to be positive. It is the ratio of number of true positive divided by number of predicted positive cases. The precision metric is shown in the below equation 3.8.

$$Precision = \frac{TP}{TP + FP} \tag{3.8}$$

3. Recall: It explains how many of the actual positive cases predicted correctly by the model. It is the ratio of number of true positive divided by number of actually positive cases. The recall metric is shown in the below equation 3.9.

$$Recall = \frac{TP}{TP + FN} \tag{3.9}$$

4. Specificity: It explains how many of the actual negative cases predicted correctly by the model. It is the ratio of number of true negative divided by number of actually negative cases. The specificity metric is shown in the below equation 3.10.

$$Specificity = \frac{TN}{TN + FP} \tag{3.10}$$

5. F1 Score: It gives a combined idea about both precision and recall score. This metric is evaluated when the balance between precision and recall needs to be maintained. The f1-score metric is shown in the below equation 3.11.

$$F1 = 2 . \frac{Precision * Recall}{Precision + Recall} \tag{3.11}$$

6. AUC-ROC: The Receiver Operator Characteristic (ROC) is a probability curve that plots the Sensitivity (True Positive Rate) against the (1 - Specificity) (False Positive Rate) at various threshold values. It indicates how well the probabilities from the positive classes are separated from the negative classes. The higher the value of Area Under the Curve the better the performance of the classification model.

Sensitivity, specificity, Accuracy, F1-Score, AUC etc. evaluation metrices were evaluated in (Ezzati and Lipton, 2020; Memon et al., 2020; Thapa et al., 2020). Accuracy is not a good metric to evaluate when there is sufficiency of one class than other class, as it biases the model towards the majority class. Being a health care problem, the dataset use for this specific study is unbalanced, hence, accuracy is not a good metric to evaluate. Here, the model should concentrate more on reducing false negative diagnosis, because patients with false negative diagnosis (despite having the disease, model will predict them healthy or no disease) will not undergoes any treatment which can be life threatening. Therefore, all the positive classes need to be identified correctly. To do the same, Recall, also known as Sensitivity and AUC-ROC evaluation metric needs to be evaluated accurately along with the other classification metrices.

Finally a comparison study will be performed among the above discussed classifiers based on the evaluation metrics and model with the best performance will be selected for predicting various stages of Alzheimer's Disease.

### 3.3.8 Hardware Requirements

The hardware requirements to execute this project are as follows:

1. RAM - 16 GB
2. Hard Disk - 500 GB
3. Graphic Card - NVIDIA GeForce GTX 1650 (4 GB) or above
4. Processor - Intel Core i5 or above

### 3.3.9 Software Requirements

The software requirements to execute this project are as follows:

1. Jupyter Notebook
2. Python version 3.8.5

Python libraries required to execute this project are as follows:

1. NumPy

2. Seaborn
3. Matplotlib
4. Pandas
5. Scikit-learn
6. Statsmodels
7. Imblearn
8. Keras
9. TensorFlow
10. SciPy

### 3.4 Summary

The detailed research approach and dataset that will be used for the various early phases of Alzheimer's disease prediction tasks are briefly described in this chapter. Some of the most widely utilized evaluation measures for assessing the proposed classification model's performance have also been discussed in depth.

The following part began with a detailed overview of the dataset and the risk parameters associated with Alzheimer's disease. Following that, there was a brief discussion of Exploratory Data Analysis, which will be used to find several underlying patterns in the dataset utilized for this study. Following that, various data pre-processing techniques that will be done on raw data collected from the ADNI dataset were discussed. Feature selection is one of the crucial tasks as all the attributes present in the dataset may not be equally contribute towards the outcome. It is proven from past studies that the use of appropriate feature set will increase model performance. Therefore, different combination of filter-based, wrapper-based, and embedded approach of feature selection has been mentioned in above section to choose the best subset of features, which will be implemented in this research. Handling class imbalance is also necessary for improved model performance, and this has been accomplished using target similarity.

In the subsequent section, different machine learning based algorithms has been discussed which will be implemented in this research. Various methods of ensemble learning have been employed by researchers in previous studies. Individual models such as Logistic Regression, Extra tree, LightGBM, Bagging with KNN, etc. will be implemented in this study, as well as their ensembled framework. The preceding section also includes a discussion of neural networks, which is also part of the implementation plan. The use of ensemble modelling is being examined because it has been shown to produce superior outcomes in the past. As a result, the stacking ensemble of all these strategies will be applied in this thesis to increase overall performance. Apart from that different evaluation metrics like accuracy, precision, recall, AUC etc., required for model's performance analysis has been discussed in detail.

This research necessitates a number of hardware and software requirements, all of which have been detailed. This study will look into cloud infrastructure options because the deep learning model uses GPU, which is readily available on the cloud and does not require a local setup.

# CHAPTER 4

# ANALYSIS AND DESIGN

## 4.1 Introduction

The model building process is covered in this chapter, and the sub-chapters will provide a full explanation of the technique used in this study. The chapter will begin with a discussion of the dataset used in this dissertation, as well as the various parameters that are included in it. All procedures involved in data preparation for further analysis will be provided. These processes involve removing unrelated variables and converting the data to a categorical format, as well as eliminating variables and transforming them into categorical variables. The distribution of the variables in the dataset will be determined through the identification of missing values and univariate analysis. Following that, missing values will be treated to make subsequent analysis easier and to avoid bias or error during analysis. The Alzheimer's dataset was subjected to outlier analysis, detection, and treatment. For training and evaluation purposes, the dataset will be split into training and validation sets. On the entire dataset, bi-variate and multi-variate exploratory data analysis was performed using visualisation libraries such as matplotlib, seaborn, and others to establish the significance of the relationship between the independent factors and the target variables.

After that, the class imbalance problem in the dataset will be addressed, with synthetic data generation class imbalance techniques being used on the dataset prior to the use of data mining techniques. The procedure for performing the oversampling (SMOTE-SVM) procedures on the training dataset will be described in detail. The model validation approach will be discussed, with the process flow of the classifiers being illustrated to explain the configuration and any tuning in the parameters. The steps taken during the training phase, as well as the outcomes generated by each classifier, will be detailed. The measures used to evaluate the performance of the classifiers based on the confusion matrix, classification reports, and area under the curve measurements will also be included in this section.

## 4.2 Dataset Description

The desired data selection is crucial in disease-related analysis since it affects the prediction effectiveness of the classification algorithm. Large amounts of health data are generated at a quick rate in a variety of formats, necessitating the employment of data mining technologies to decipher the data and extract meaningful information. The Alzheimer's Disease Neuroimaging Initiative database (ADNI, Alzheimer's Disease Neuroimaging Initiative Database, 2020) provided the data for this study. The ADNI, which began as a public-private partnership in 2004, is still working to enhance diagnostic procedures for early identification of Alzheimer's disease and improve clinical trial design. Participants in the ADNI database range in age from 55 to 90 years old and are recruited from 57 sites across the United States and Canada. The ADNIMERGE dataset will be used in this investigation, which is a subset of the ADNI dataset that is updated daily straight from the Alzheimer's Disease Cooperative Study data system. This datafile comprises 15762 rows and 116 attributes from four ADNI study phases (ADNI-1, ADNI-GO, ADNI-2, and ADNI-3), including demographics, neuropsychological testing scores, MRI and PET summaries, and CSF measurements from both a baseline diagnosis and six-month follow-up examinations. Because the datafile is unbalanced and has multiple missing values, it must be pre-processed before being fed into the machine learning pipelines. This study's ethical difficulties and ramifications were all taken into account. Because the dataset was acquired from an open data source available for research on the ADNI website, there were no substantial ethical considerations in this work. As a result, there are no ethical forms attached to this report.

Since the main focus of this study is early detection of Alzheimer's illness, analysis will be carried out on the baseline assessments. To filter out baseline diagnoses, selection has been done using the VISCODE column, which represents the visit number for each patient. The rows with a VISCODE value of "bl" have been picked. After filtering, 2,379 patient records remained in the dataset, which will be subjected to further study. Out of 116 columns, 65 parameters have been refined, which corresponds to the patient's baseline visit. Following the use of feature selection methods, approximately 43 relevant features out of 65 elements were identified. Table 4.1 presents the key parameters, their datatypes, range of values, and their descriptions for a better understanding of the predictor factors for Alzheimer's disease diagnosis.

*Table 4.1 Selected Attributes and its Definitions from presented ADNI dataset*

| Attributes | Type | Definition and Permissible values (MIN-MAX Range) |
|---|---|---|
| ADAS11 | Numerical | Alzheimer's Disease Assessment Scale (11 items): a thorough examination to assess cognitive and non-cognitive signs of Alzheimer's disease. Scores range from 0 (no impairment) to 70 (severe impairment) |
| ADAS13 | Numerical | Alzheimer's Disease Assessment Scale (13 items): a thorough examination to assess cognitive and non-cognitive signs of Alzheimer's disease. Scores range from 0 (no impairment) to 70 (severe impairment) |
| ADASQ4 | Numerical | Alzheimer's Disease Assessment Scale. Value ranges from 0 to 10 |
| AGE | Numerical | Participant's age at time of visit. Value ranges from 50.40 to 91.40 |
| APOE4 | Numerical | Presence of the APOE gene, which produces the ApoE4 protein, is linked to delayed Alzheimer's illness. Scores range from 0 to 2 |
| CDRSB | Numerical | Clinical Dementia Rating - Sum of Boxes: tracks the course of dementia, particularly in patients with mild to moderate cognitive impairment. Scores range from 0 to 10 |
| EcogPtDivatt | Numerical | Measurement of Everyday Cognition, a short questionnaire completed out by the patient (Pt) and spouse (Sp) in which they rate the individual's performance in various cognitive domains (memory, language, visuo-spatial ability, thinking, organisation, divided concentration) in comparison to his or her performance 10 years ago. Scores |
| EcogPtLang | Numerical | |
| EcogPtMem | Numerical | |
| EcogPtPlan | Numerical | |
| EcogPtVisspat | Numerical | |
| EcogSPDivatt | Numerical | |
| EcogSPLang | Numerical | |
| EcogSPMem | Numerical | |

| EcogSPOrgan | Numerical | range from 1 (no change) to 4 (much worse), or 5 (unknown) |
|---|---|---|
| EcogSPPlan | Numerical | |
| EcogSPTotal | Numerical | |
| EcogSPVisspat | Numerical | |
| Entorhinal | Numerical | Entorhinal cortex volume: value ranges from 1,106 to 6,265 |
| FAQ | Numerical | Functional Activities Questionnaire: measures a patient's ability to conduct daily tasks unassisted. Scores range from 0 (normal) to 30 (highly dependent) |
| FDG | Numerical | Measured uptake of Fluorodeoxyglucose: reported to be a powerful differentiator between Alzheimer's disease and other kinds of dementia, as well as a predictor of cognitive impairment development. Value ranges from 0.693671 to 1.701130 |
| FLDSTRENG | Categorical | MRI Field Strength. Values: 0 - 1.5 Tesla MRI; 1 - 3 Tesla MRI |
| Fusiform | Numerical | The volume of the fusiform gyrus. Value ranges from 8,991 to 29,950 |
| Hippocampus | Numerical | Hippocampal volume: Value ranges from 2,991 to 10,769 |
| LDELTOTAL | Numerical | Logical Memory - Delayed Recall. Score ranges from 0 to 23 |
| MidTemp | Numerical | Middle temporal gyrus volume: value ranges from 9,375 to 32,324 |
| MMSE | Numerical | Mini Mental State Examination: a simple and extensively used method of assessing cognitive function. Scores of 25-30 are considered normal, 21-24 mild, 10-20 moderate, and <10 severe |
| MOCA | Numerical | Montreal Cognitive Assessment: for mild cognitive impairment, a 10-minute exam is recommended. |

| | | Scores range from 0 (severe impairment) to 30 (normal) |
|---|---|---|
| mPACCdigit | Numerical | ADNI modified Preclinical Alzheimer's Cognitive Composite with Digit Symbol Substitution. Score ranges from -23.61 to 6.30 |
| mPACCtrailsB | Numerical | ADNI modified Preclinical Alzheimer's Cognitive Composite with Trails B. Score ranges from -23.61 to 7.41 |
| PTEDUCAT | Numerical | The patient’s education. Score ranges from 4 to 20 |
| PTETHCAT | Categorical | The patient’s ethnicity. Having Not Hisp/Latino, Hisp/Latino, Unknown labels |
| PTGENDER | Categorical | The patient’s gender. Values: 0 – Male; 1 – Female |
| PTMARRY | Categorical | The patient’s marital status. Value contains Married, Widowed, Divorced, Never married, Unknown |
| PTRACCAT | Categorical | The patient’s race information. Contains White, Black, Asian, More than one, Unknown, Am Indian/Alaskan, Hawaiian/Other PI labels |
| RAVLT_forgetting | Numerical | Rey Auditory Verbal Learning Test for forgetting: assesses auditory learning and memory. Scores range from 0 to 15 |
| RAVLT_immediate | Numerical | Rey Auditory Verbal Learning Test for immediate response: assesses auditory learning and memory. Scores range from 0 to 75 |
| RAVLT_perc_forgetting | Numerical | Rey Auditory Verbal Learning Test for percentage of forgetting: assesses auditory learning and memory. Scores ranges in percentage 0 to 100. |
| SITE | Numerical | Site, Score ranges from 2 to 941 |
| TRABSCOR | Numerical | Trails B. Score ranges from 0 to 379 |
| Ventricles | Numerical | The Volume of ventricles. Score ranges from 6,823 to 1,57,713 |
| WholeBrain | Numerical | Entire brain volume: value ranges from 5,21,287 to 14,86,040 |

There are two outcome variables in the dataset: DX and DX_bl. In ADNI dataset, there are two "initial" visits – the screening visit, which is a subset of information collected to ensure the individual meets all the inclusion/exclusion criteria and then a baseline visit which includes a full neuropsychological battery along with additional information. DX_bl is the diagnosis at the screening visit (note, this diagnosis also dictated the visitation schedule) while DX contains the diagnosis at each visit. For most participants, DX_bl will be the same as DX when VISCODE is "bl". The Clinical Core has generally recommended using the diagnosis at the baseline visit (rather than the screening visit), since it is based on a more complete picture rather than just the screening instruments. Therefore, this study uses DX column as the target variable. Below table 4.2 describes the classes present in the target column.

*Table 4.2 Selected Target Attribute from presented ADNI dataset*

| Attributes | Type | Definition and Permissible values (MIN-MAX Range) |
|---|---|---|
| DX | Categorical | Diagnosis: Output labels<br>0 – Cognitive Normal (CN)<br>1 – Mild Cognitive Impairment (MCI)<br>2 – Dementia (AD) |

### 4.3 Data Preparation

Although the original data is in a structured format, it is not clean and requires a number of pre-processing processes before any operations can be conducted to achieve the study's objectives. Each pre-processing stage will be described in depth, including the objective of each step as well as the results of these procedures.

#### 4.3.1 Elimination of Variables

The "RID" (Participant roster ID) and "PTID" (Participant ID) variables indicate whether each observation in the training set or validation set can uniquely identify a participant. But this variable was removed from the original dataset as it is not a parameter of disease diagnosis. The "COLPROT" variable is holding the study protocol of data collection and the "ORIGPROT" column has protocol values from which the subject originated (e.g., ADNI1, ADNI2, ADNIGO, ADNI3), which do not have any impact on the final outcome variable as all the records from each protocol are being considered in this study. Hence, it was also removed. The "VISCODE" column was used to filter out the baseline assessments. Therefore, after selecting the necessary sub-samples of records, this column contains only one value, i.e., "bl" for all the datapoints. Hence, not useful for final diagnosis and removed from the dataset. Columns "EXAMDATE", "Month" (Months from baseline to nearest 6 months, based on EXAMDATEs), "M" (Months from baseline based on VISCODE) are having no impact as those attributes contain date and month related information of disease progression, and not having direct relation with early detection of dementia. Column "FSVERSION" stores FreeSurfer Software Version, which is also unrelated to the target variable and was discarded from dataset. The variable "update stamp" is removed from the dataset because it contains the time when the ADNI datafile was last updated with each observation. Above identified unrelated features were eliminated using Pandas library with Python version 3.8.5. As a result of this pre-processing stage, the data now contains 55 variables, all of which are related to dementia diagnosis, with 2,379 observations.

### 4.3.2 Transformation of Variables

Out of these selected variables some variables are represented in numerical format, whereas some are marked as categorical variables. This identification is done using python libraries. For the better data analysis and visualization purpose some of the numerical features are required to be represented into categorical format. On the other hand, there are 5 such categorical variables that are needed to be converted into numerical features before feeding into classification model. The below sub-section describes the details of these variable transformation before and after conversions.

#### 4.3.2.1 Transformation into Numerical Variables

Every value for each of the remaining 55 variables in the dataset needs to be represented in numerical format. Out of these 55 variables, 5 variables are identified as categorical type using python Pandas library. But for the modelling purpose, these values are required to be decoded into numerical format. Below table 4.3, is showing the categorical columns and its converted format with the allowable values. Following these conversions, the total number of traits becomes 64 in the complete dataset.

*Table 4.3 Categorical to Numerical Transformed Attributes*

| **Categorical Feature** | **Transformation by** | **Dummy Feature** | **Permissible Values** |
|---|---|---|---|
| PTGENDER | Mapping | NA | 0: Male; 1: Female |
| FLDSTRENG | Mapping | NA | 0: 1.5 Tesla MRI; 1: 3 Tesla MRI |
| PTETHCAT | Dummification | PTETHCAT_ Not Hisp/Latino | 1: Not Hisp/Latino; 0: Other Value |
| | | PTETHCAT_ Unknown | 1: Unknown; 0: Other Value |
| PTRACCAT | Dummification | PTRACCAT_Asian | 1: Asian; 0: Other value |
| | | PTRACCAT_Black | 1: Black; 0: Other value |
| | | PTRACCAT_ Hawaiian/Other PI | 1: Hawaiian/Other PI; 0: Other value |
| | | PTRACCAT_More than one | 1: More than one; 0: Other value |
| | | PTRACCAT_Unknown | 1: Unknown; 0: Other value |
| | | PTRACCAT_White | 1: White; 0: Other value |
| PTMARRY | Dummification | PTMARRY_Married | 1: Married; 0: Other Value |
| | | PTMARRY_Never married | 1: Never married; 0: Other Value |
| | | PTMARRY_Unknown | 1: Unknown; 0: Other Value |
| | | PTMARRY_Widowed | 1: Widowed; 0: Other Value |

#### 4.3.2.2 Transformation into Categorical Variables

There are 4 numerical features which were converted into categorical column using binning methodology of Pandas library. This transformation is useful to visualize which range of value is a suitable indicator with respect to the target variable. These transformed variables were solely used for visualisation and were not taken into account when developing the model. Below table 4.4, describes the binning values for the converted categorical features.

*Table 4.4 Numerical to Categorical Transformed Attributes*

| Numerical Attribute | Converted Column | Permissible Values |
|---|---|---|
| AGE<br>(Participant's Age) | AGE_BIN | 10 to 20: Age between 10 years to 20 years<br>20 to 30: Age between 20 years to 30 years<br>30 to 40: Age between 30 years to 40 years<br>40 to 50: Age between 40 years to 50 years<br>50 to 60: Age between 50 years to 60 years<br>60 to 70: Age between 60 years to 70 years<br>70 to 80: Age between 70 years to 80 years<br>80 to 90: Age between 80 years to 90 years<br>90 to Above: Age between 90 years and above |
| MMSE<br>(Mini Mental State Examination) | MMSE_BIN | 0 to 5: MMSE score between 0 to 5<br>5 to 10: MMSE score between 5 to 10<br>10 to 15: MMSE score between 10 to 15<br>15 to 20: MMSE score between 15 to 20<br>20 to 25: MMSE score between 20 to 25<br>25 to 30: MMSE score between 25 to 30 |
| FAQ<br>(Functional Activities Questionnaire) | FAQ_BIN | 0 to 5: FAQ score between 0 to 5<br>5 to 10: FAQ score between 5 to 10<br>10 to 15: FAQ score between 10 to 15<br>15 to 20: FAQ score between 15 to 20<br>20 to 25: FAQ score between 20 to 25<br>25 to 30: FAQ score between 25 to 30 |
| MOCA<br>(Montreal Cognitive Assessment) | MOCA_BIN | 0 to 5: MOCA score between 0 to 5<br>5 to 10: MOCA score between 5 to 10<br>10 to 15: MOCA score between 10 to 15<br>15 to 20: MOCA score between 15 to 20<br>20 to 25: MOCA score between 20 to 25<br>25 to 30: MOCA score between 25 to 30 |

### 4.3.3 Identification of Missing Values

Upon examination of the dataset, it was discovered that the majority of the Cognitive Tests, PET and MRI scan and CSF biomarkers related variables have missing values. The inclusion of missing values reduces the efficiency of any study and introduces complications during pre-processing and model construction. To avoid this issue, the missing value must be handled. The missing or unknown value is represented by NaN in the original dataset. This missing value analysis has been performed using Missingno and Pandas libraries with Python version 3.8.5. Below table 4.5 listed down all the features along with respective missing values percentage.

*Table 4.5 Attributes with their Missing Value Percentage from ADNI Dataset*

| Features Name | Missing Value Percentage |
|---|---|
| PIB | 99.159 |
| FBB | 86.885 |
| DIGITSCOR | 65.784 |
| AV45 | 53.047 |
| ABETA | 48.928 |
| TAU | 48.928 |
| PTAU | 48.928 |
| FLDSTRENG | 39.975 |
| EcogSPOrgan | 38.714 |
| FDG | 37.495 |
| EcogSPDivatt | 37.327 |
| EcogSPVisspat | 37.116 |
| EcogSPPlan | 36.612 |
| EcogPtOrgan | 36.402 |
| EcogSPMem | 36.15 |
| EcogSPTotal | 36.15 |
| EcogSPLang | 36.066 |
| MOCA | 35.982 |
| EcogPtVisspat | 35.435 |
| EcogPtDivatt | 35.393 |
| EcogPtMem | 35.099 |

| EcogPtLang | 35.099 |
|---|---|
| EcogPtPlan | 35.057 |
| EcogPtTotal | 35.015 |
| Entorhinal | 16.982 |
| Fusiform | 16.982 |
| MidTemp | 16.982 |
| Hippocampus | 15.805 |
| Ventricles | 8.281 |
| APOE4 | 6.936 |
| WholeBrain | 6.852 |
| ICV | 5.254 |
| IMAGEUID | 5.212 |
| TRABSCOR | 2.564 |
| FAQ | 1.513 |
| ADAS13 | 1.051 |
| RAVLT_perc_forgetting | 0.757 |
| ADAS11 | 0.546 |
| RAVLT_forgetting | 0.504 |
| RAVLT_immediate | 0.462 |
| RAVLT_learning | 0.462 |
| ADASQ4 | 0.294 |
| AGE | 0.168 |
| LDELTOTAL | 0.168 |
| mPACCdigit | 0.126 |
| mPACCtrailsB | 0.126 |
| MMSE | 0.042 |
| DX | 1.597 |

Below figure 4.1 and 4.2 provides the visualization of missing values and its percentage with respect to all cognitive test and brain-related features present in the ADNIMERGE dataset.

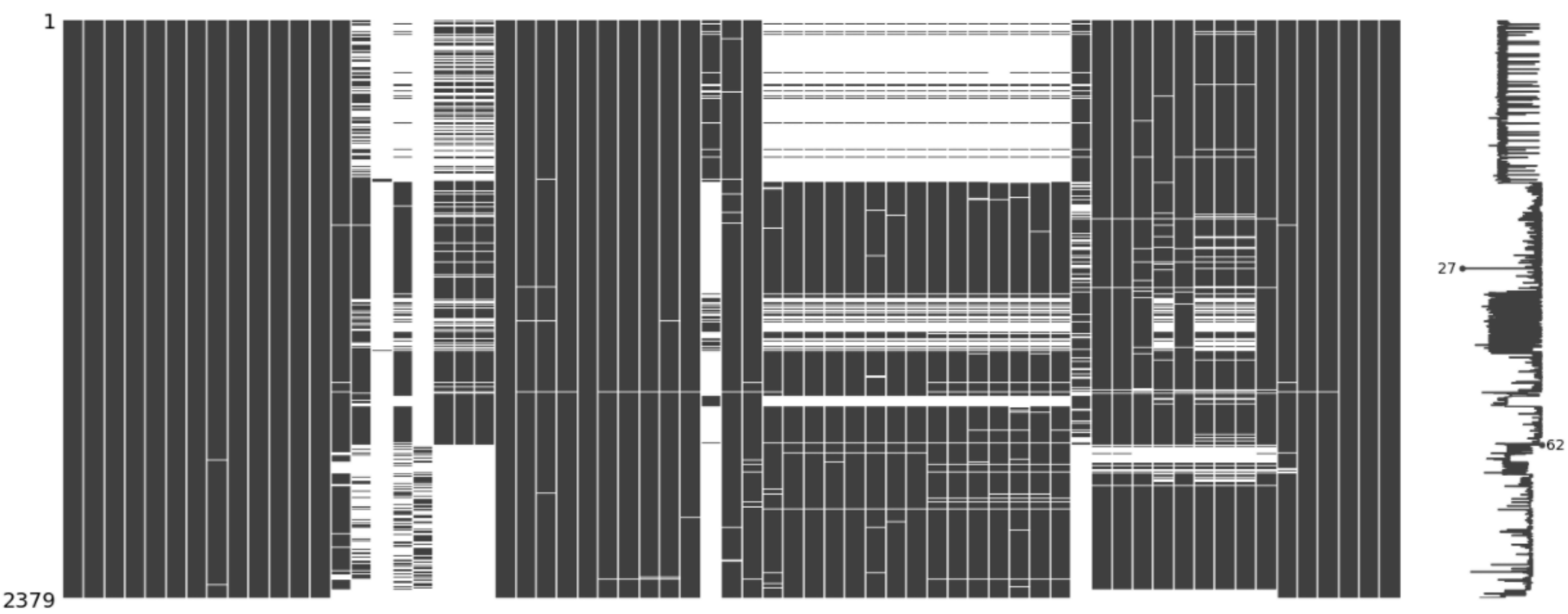

*Figure 4.1 Matrix Plot to Visualize the Missing Values in the Presented Dataset*

The white space in above plot signifies absent of data in the data frame columns.

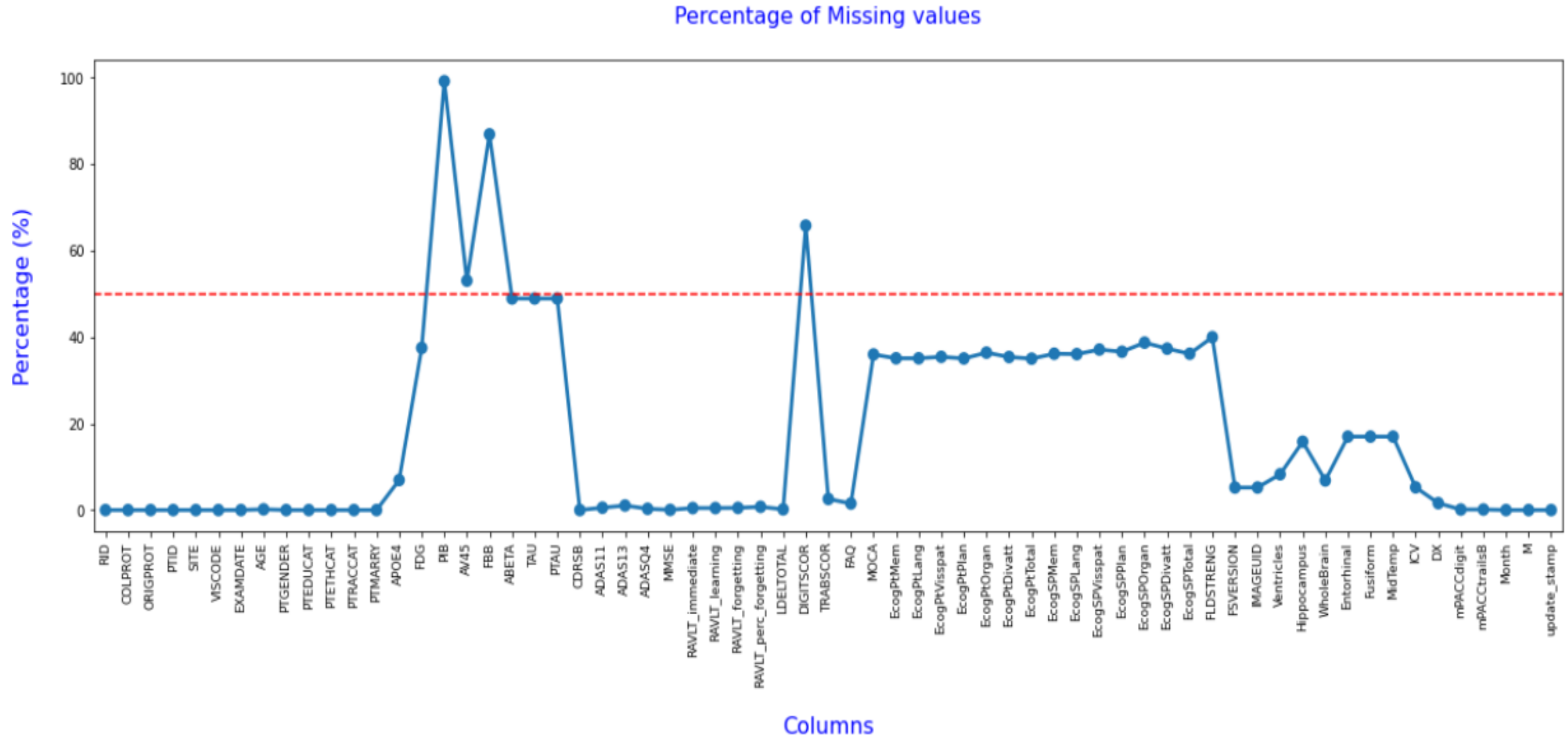

*Figure 4.2 Point Plot to Represent Missing Value % in the Presented Dataset*

The percentage of missing values are depicted along y-axis with respect to each feature across x-axis in figure 4.2. The red line highlights a threshold of 45% beyond which the missing percentage is unacceptable. Following sub-section describes the methodologies of treating the missing values in detail.

### 4.3.4 Treatment of Missing Values

It has been determined from the above missing value analysis that around 48 features have missing values, and it is critical to treat the missing value in order to achieve a fair outcome and prevent bias or errors during analysis. It has been observed that a few numbers of attributes have a large number of missing values. Imputation of those high missing values introduces bias into the dataset and has an impact on model performance. As a result, a threshold of 45 percent was used in this investigation, after which the features were dropped without treatment. Table 4.5 shows that the properties PIB, FBB, DIGITSCOR, AV45, ABETA, TAU, and PTAU have more than 45 percent missing values. As a result, the seven CSF biomarkers and PET scan-related traits were removed from the dataset without imputation.
Above table 4.5 shows that the target column DX has 1.597% of missing data as well. Performing imputation on the outcome variable is not a good practice. Hence, this study excluded 38 rows from the dataset where DX is holding null values and keep those rows separated so that they could be utilized during testing phase.

From the remaining 40 features, there are 39 numerical features that consist of missing values and the same has been treated using sklearn's iterative imputation library with Python version 3.8.5. Iterative imputation is a refined approach involves defining a model for imputing missing values by simulating each feature that predicts each missing values in an attribute as a function of all other feature recursively by evaluating the function values. Succeeding repetitions of predicting missing values using other features provide a refined estimate, which improves the quality of the null data being imputed.

It can be seen in table 4.5 that there is one nominal categorical feature FLDSTRENG consisting of missing values, which has been handled using the mode value of that column utilizing simple imputation method with strategy 'most frequent'. Because the variable is a nominal categorical type, the missing values have been imputed using modes, which are the values with the highest frequency in the relevant columns. To impute the missing values for nominal categorical features, the Pandas package with Python version 3.8.5 has been utilized.

The dataset with imputed missing values has been considered for further analysis. The dataset used after this pre-processing step contained no missing values and all the observations are identified as valid. Following this pre-processing stage, the data now contains 57 variables, including the target column, and 2,143 rows, on top of which further pre-processing operations will be carried out. Below figure 4.3 is a visualization of entire dataset, after imputing the missing values related to the dementia related biomarkers in the presented dataset.

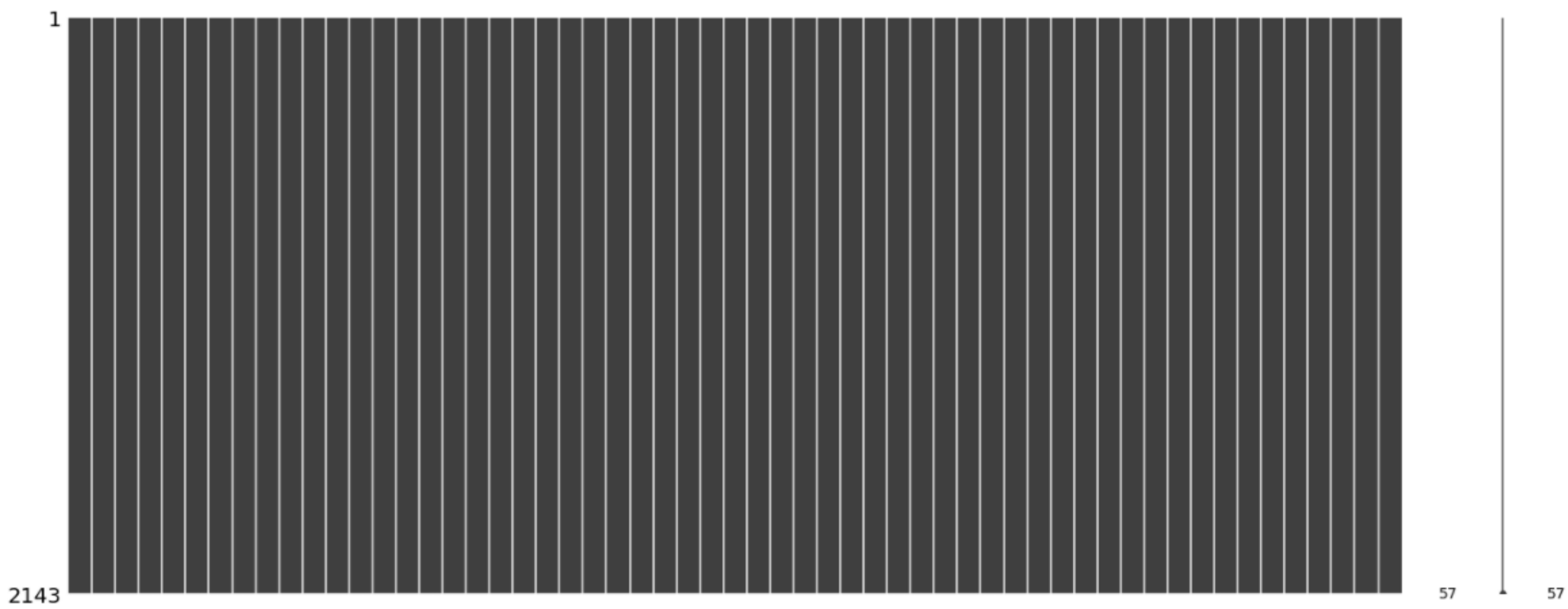

*Figure 4.3 Matrix Plot to indicate zero Missing values in Imputed Dataset*

Above matrix plot clearly depicts that no data sparsity exists across all data frame columns after applying imputation technique.

### 4.3.5 Identification of Data Imbalance

In a classification problem, data imbalance indicates that the number of observations in one class differs from the number of observations in the other. This is a significant issue in both binary and multiclass classification problems, and if left unaddressed, it leads to a deceptive predictive model. If one class in the training dataset has a majority, the classifier trained on that data is biased towards that class. Prediction accuracy is critical for any machine learning model when diagnosing medical domain problems. If the model is trained on unbalanced data, just predicting all data points as majority class will result in a greater accuracy. Because all data points in the minority class have been misclassified, the algorithm is unable to efficiently diagnose health conditions. As a result, prior to implementing a learning-based model, it is critical to check and process imbalanced data.

The presented ADNI dataset is having class imbalance problem, where the instances of "Mild Cognitive Impairment (MCI)" class is higher than all other classes. Below table 4.6 shows the target class labels, label description and respective instance count under each target class.

*Table 4.6 Target Class Labels with Number of Observations in Each Class*

| Target Class Label | Label Description | Number of Observation |
|---|---|---|
| 0 | Cognitive Normal (CN) | 758 |
| 1 | Mild Cognitive Impairment (MCI) | 985 |
| 2 | Dementia (AD) | 400 |

The class instance count in percentage along with target class labels are visualized in below figure 4.4. This pie chart clearly depicts the presence of class imbalance problem in the presented dataset.

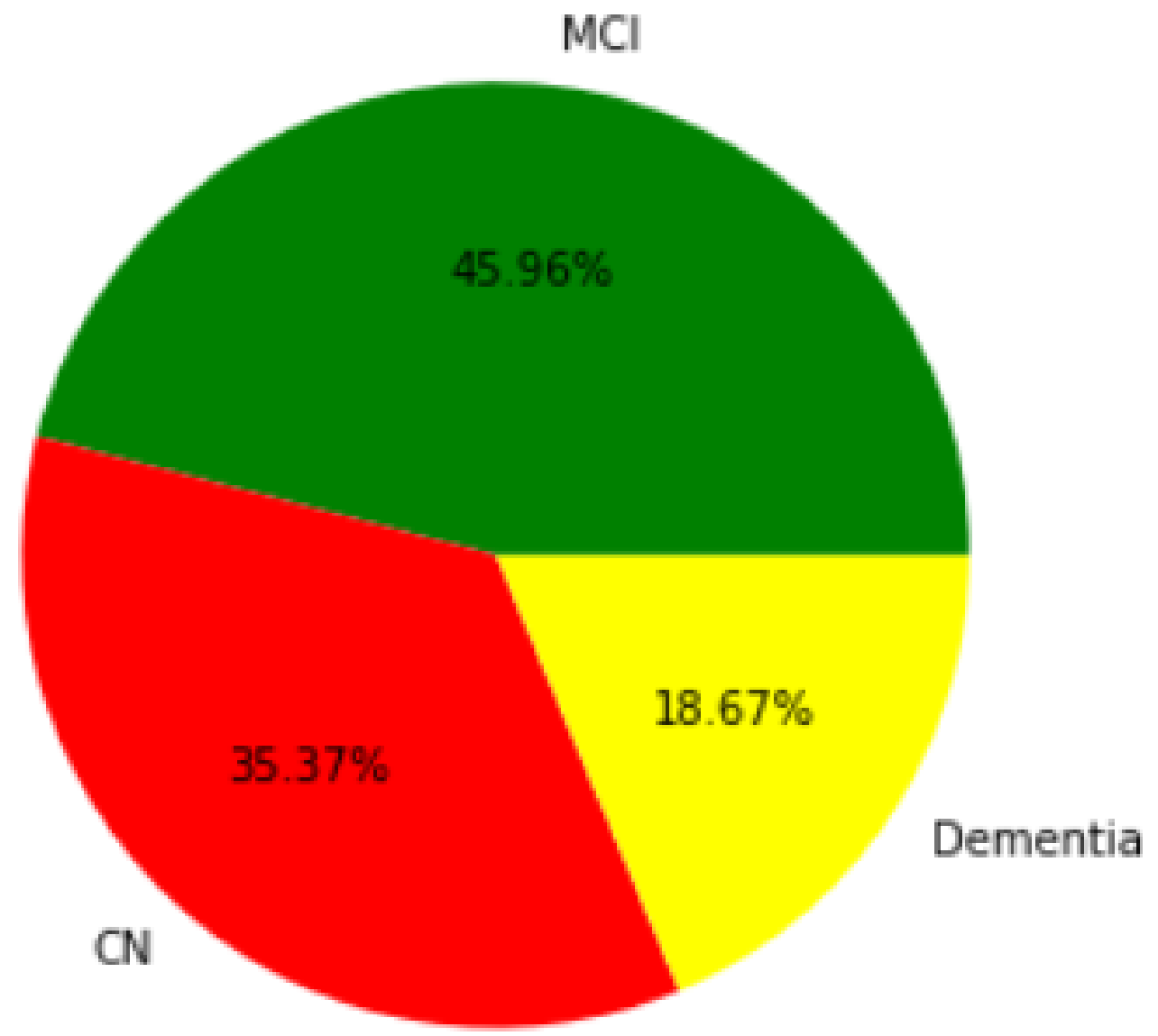

*Figure 4.4 Pie Chart representing Class Imbalance in Presented Dataset*

### 4.3.6 Splitting of Original Dataset

The training and testing of the dataset are a key step in the construction of a supervised learning model. Without training the algorithm, the model's parameters do not settle on appropriate values, rendering the model useless. There is no way to know how good the model is unless the methodology is validated after training the model parameters on a test or validation set, which is, once again, pointless.

Apart from the original dataset with the imputed missing values, two other datasets were constructed from the entire dataset, one including only the training set and the other containing only the test set. The data was partitioned in a 70:30 ratio, with 70% of the data from the entire dataset going into the training set and the remaining 30% going into the test set. To allow other researchers to compare their work on the same dataset, the default 70-30 was used. With Python 3.8.5, the train test split function of the Scikit-learn module was used to split the data. There are now three datasets, each with all 57 variables with no missing values. There are 56 input variables and one target variable which is DX (Diagnosis).

The three datasets that will be used for further analysis in this study are listed below, along with the total number of observations in each dataset. The training set is used to train the classifiers and develop a predictive model while the test set is used to examine and assess the model's performance.

i. Complete Dataset (includes both training and validation) – 2,143 records
ii. Training Dataset – 1,500 records (70 %, for model training)
iii. Testing Dataset – 643 records (30 %, for model validation)

### 4.3.7 Treatment of Imbalance Dataset

From sub-section 4.3.5, it is seen that the dataset is not balanced among the classes. The number of observations with MCI class is more than the other two classes. To handle this data imbalance, class balancing approach has been applied in this study. The main goal of class balancing is to either increase the minority class's frequency or decrease the dominant class's frequency. This is done to ensure that each class have roughly the same number of instances. When dealing with unbalanced datasets, balancing classes technique needs to be applied in the training data during data preparation stage before the data is fed into the machine learning algorithm.

Below table 4.7 shows the class imbalance among the training datasets after partitioning the dataset into training and test sets in sub-section 4.3.6. Because the imbalance needs to be treated on training data, no modification was done in the test dataset.

*Table 4.7 Target Class Imbalance with Instance Count in Training Dataset*

| Target Class Label | Label Description | Number of Observation |
|---|---|---|
| 0 | Cognitive Normal (CN) | 525 |
| 1 | Mild Cognitive Impairment (MCI) | 691 |
| 2 | Dementia (AD) | 284 |

The pie chart in figure 4.5 depicts the class imbalance ratio in percentage among the training datasets.

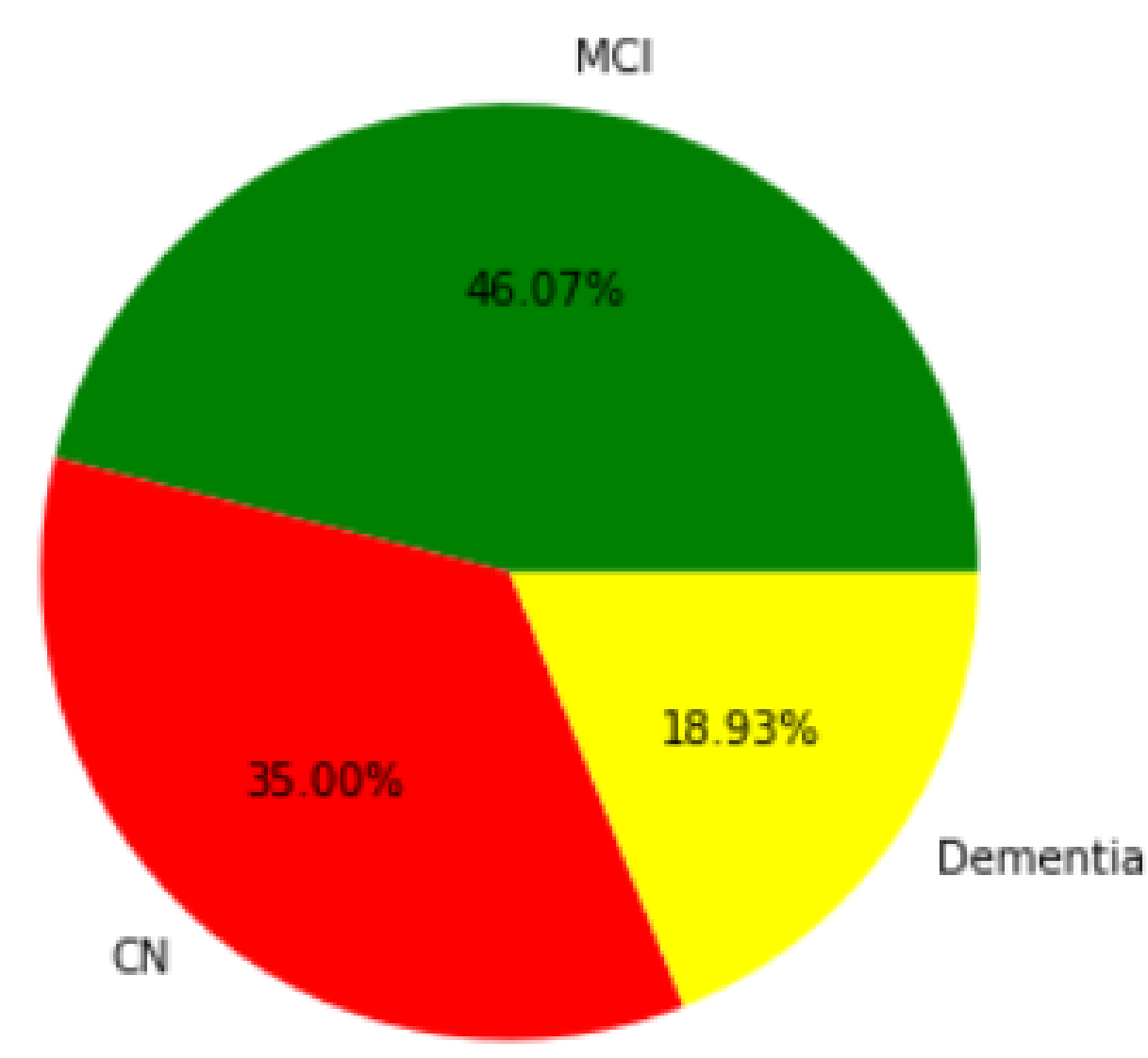

*Figure 4.5 Pie Chat to Visualize Class Imbalance in Training Dataset*

This research used borderline-SMOTE SVM minority class over sampling strategy to address the class balancing problem. In this technique the support vectors acquired after training a typical SVMs classifier on the original training set are used to estimate the borderline area. Using the interpolation method, new instances will be created at random along the lines connecting each minority class support vector with a number of its nearest neighbours.

After adjusting for class imbalances in the training dataset, each instance of the CN (Class 0), MCI (Class 1), and Dementia (Class 2) class labels contained 691 records. The class balanced training dataset after SMOTE SVM is depicted in figure 4.6 using a Pie Chart.

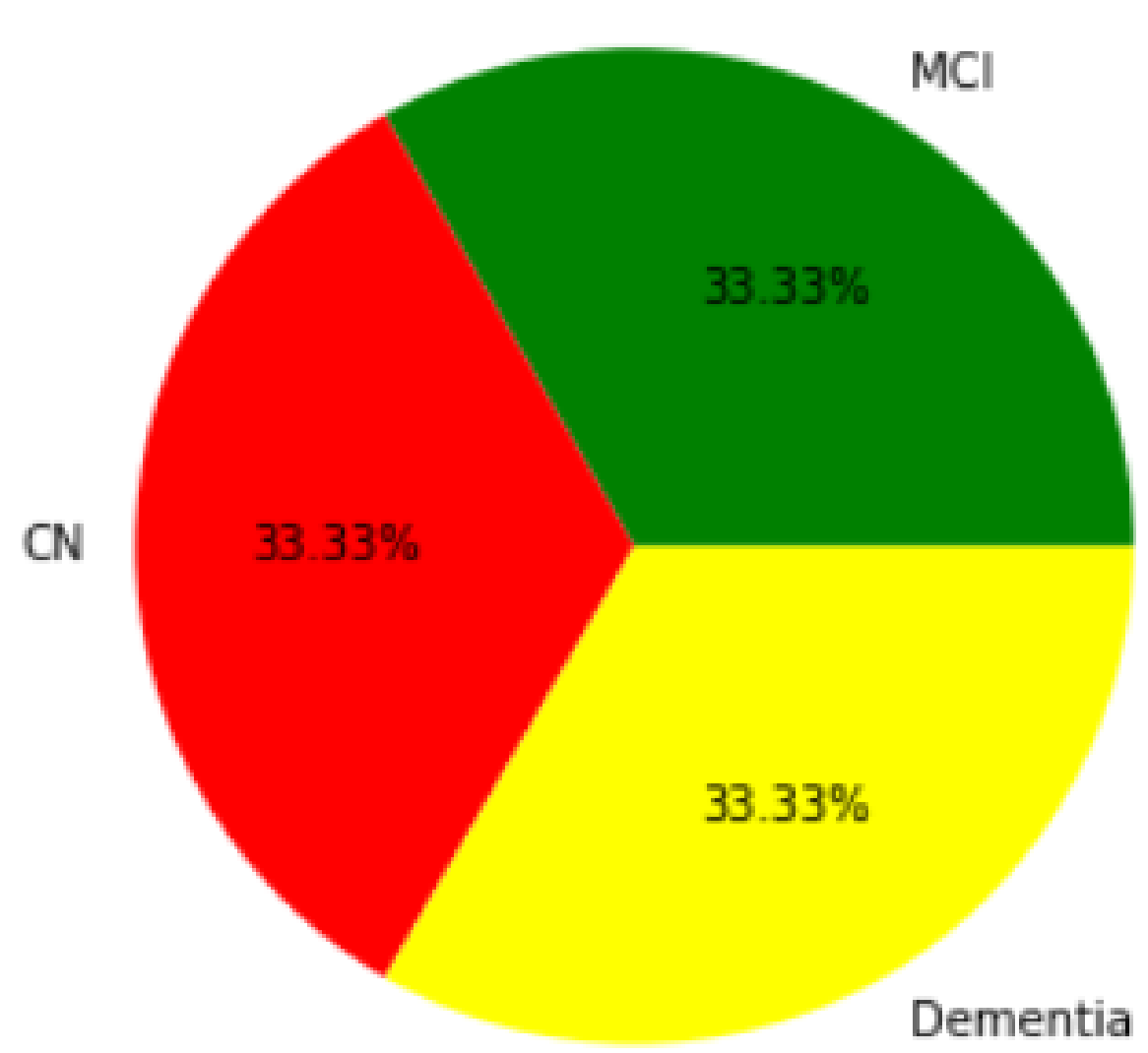

*Figure 4.6 Pie Chart representing the Class Balanced Training Dataset after SMOTE SVM*

Above chart clearly depicts that after applying borderline SMOTE SVM minority class oversampling strategy the class imbalance problem of original presented dataset has been solved.

After implementing this pre-processing step, a new class balanced resampled training dataset has been generated with 2,073 rows and 57 columns including target column. Both the training dataset (before imbalance and after imbalance handling) will be used to train the classification model. Test dataset will be used (without imbalance treatment) to evaluate the classifier performance.

i. Training Dataset without Imbalance Handling – 1,500 records (for model Training)
ii. Training Dataset with Imbalance Handling – 2,073 records (for model Training)
iii. Testing Dataset – 643 records (for model Testing)

#### 4.3.8 Feature Scaling

Feature scaling is one of the most significant pre-processing phases in machine learning algorithms. This guarantees that all independent features are scale free, scale independent, or scale equal, allowing the machine learning algorithm to take each feature into account equally. For gradient descent-based algorithms like Linear regression, Logistic regression, and neural networks, scaling ensures that all features are updated with the same step size during gradient descent and converges more quickly towards minima.

Standardization was chosen over Normalization in this analysis since most of the features in this study are gaussian distributed in nature and some of the features contain real outliers. The StandardScaler function from the sklearn package is used with Python 3.8.5 for feature scaling.

### 4.3.9 Feature Selection

As real-world datasets typically contain a large number of features, feature selection is an important aspect of building machine learning solutions for real-world problems. Insignificant characteristics not only lower model effectiveness during training, but they also raise the mode's complexity, reduce the model's universal applicability, and bring bias into simulated results. Therefore, reducing dimensionality is critical for preventing overfitting, reducing training time, and increasing the model's decision-making power. The three feature selection techniques used in this study are listed below along with the findings.

#### 4.3.9.1 Feature Selection using Recursive Feature Elimination

RFE is a common and beneficial method for feature selection in many machine learning solutions because to its easy configuration steps and reliability in selecting the most relevant features from the training set based on coefficient values or feature importance. This wrapper-based selection approach was employed in this project, with the core of the method being the supervised learning algorithm Logistic Regression, which is wrapped by RFE. RFE chose the top 45 relevant features, which are ranked as 1 (using ref.ranking_ property), based on the value of the feature coefficients determined using logistic regression. With Python version 3.8.5, the RFE function was imported from Scikit-Learn package's feature_selection class.

The original dataset used in this study was uneven in terms of classes. In this study, model training will be done on two different training sets: one on the class imbalanced dataset and other on the class balanced training set. RFE is used twice to perform feature selection from those two different sets. RFE was used to choose top 45 features from training set of 1,500 rows and 57 features before dealing with the class balancing problem. Following the resolution of the imbalance issue, the RFE feature selection approach was used again to select the top 45 features from a balanced training set of 2,073 rows and 57 features. Table 4.8 listed down 45 first-ranked features selected by RFE, before and after imbalance handling.

*Table 4.8 List of Features Selected via RFE Pre and Post Imbalance Handling*

| **RFE Selected Features Before Balance** | **RFE Selected Features After Balance** |
|---|---|
| AGE | SITE |
| PTGENDER | AGE |
| PTEDUCAT | PTGENDER |
| APOE4 | PTEDUCAT |
| FDG | FDG |
| CDRSB | CDRSB |
| ADAS11 | ADAS11 |
| ADAS13 | ADAS13 |
| MMSE | ADASQ4 |
| RAVLT_immediate | MMSE |
| RAVLT_forgetting | RAVLT_immediate |
| RAVLT_perc_forgetting | RAVLT_forgetting |
| LDELTOTAL | RAVLT_perc_forgetting |
| TRABSCOR | LDELTOTAL |
| FAQ | TRABSCOR |
| MOCA | FAQ |
| EcogPtMem | MOCA |
| EcogPtVisspat | EcogPtMem |
| EcogPtPlan | EcogPtLang |
| EcogPtDivatt | EcogPtVisspat |
| EcogSPMem | EcogPtDivatt |
| EcogSPLang | EcogSPMem |
| EcogSPVisspat | EcogSPLang |
| EcogSPPlan | EcogSPVisspat |
| EcogSPOrgan | EcogSPPlan |
| EcogSPDivatt | EcogSPOrgan |
| EcogSPTotal | EcogSPTotal |
| FLDSTRENG | FLDSTRENG |
| Hippocampus | Ventricles |

| WholeBrain | Hippocampus |
|---|---|
| Entorhinal | WholeBrain |
| Fusiform | Entorhinal |
| MidTemp | Fusiform |
| mPACCdigit | MidTemp |
| mPACCtrailsB | mPACCdigit |
| PTETHCAT_Not Hisp/Latino | mPACCtrailsB |
| PTETHCAT_Unknown | PTETHCAT_Unknown |
| PTRACCAT_Asian | PTRACCAT_Asian |
| PTRACCAT_Black | PTRACCAT_Black |
| PTRACCAT_Hawaiian/Other PI | PTRACCAT_Hawaiian/Other PI |
| PTRACCAT_More than one | PTRACCAT_More than one |
| PTRACCAT_White | PTRACCAT_White |
| PTMARRY_Married | PTMARRY_Married |
| PTMARRY_Never married | PTMARRY_Never married |
| PTMARRY_Widowed | PTMARRY_Widowed |

#### 4.3.9.2 Feature Selection using Random Forest Classifier

Random Forest uses its own built-in relevance calculation technique to choose features. RF ranks features based on how well they increase purity (Gini impurity) of a node by lowering overall impurity using tree-based techniques. Because the impurity decreases the greatest near the beginning of the tree, the features at those top nodes are the most relevant. Because the nodes at the bottom of the tree have the smallest drop in impurity, they have the smallest number of essential features.

The RandomForestClassifier method from sklearn's ensemble package was used to create the Random Forest Classifier for feature selection, with Python version 3.8.5 and a number of base estimators (Decision Tree) of 500 utilised as a hyperparameter. The top 45 features chosen by RFE are now run through a further tree classifier to refine the top 40 features, which are subsequently used in model construction.

Because the reported dataset is unbalanced, a feature selection using the Random Forest classifier was performed once on the train dataset with 1,500 rows and 45 features (chosen through RFE) before the imbalance treatment. Finally, from a total of 45 independent variables, RF chose the top 40 features based on Gini significance.

Following the adjustment of the class imbalances in the presented dataset, the train dataset with 2,073 rows and 45 features was subjected to feature selection using RF once again (selected via RFE). Finally, a random forest classifier chose the top 40 features from a total of 45 independent characteristics based on Gini significance.

Below figure 4.7 represents the top 40 features ordered by its importance after applying this tree-based feature selection strategy on imbalanced training dataset.

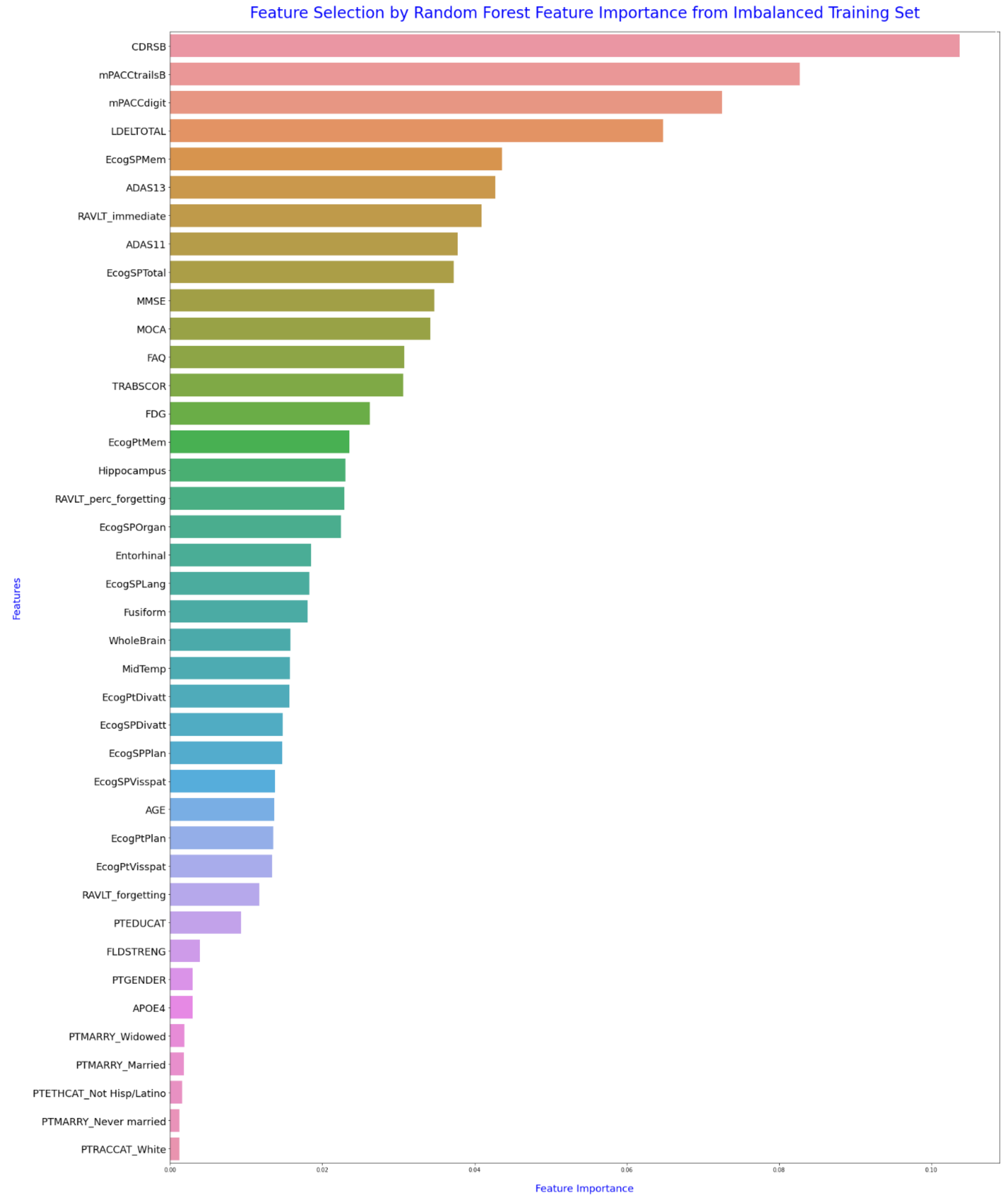

*Figure 4.7 Top 40 Features selected using Feature Importance before Imbalanced handling*

Below figure 4.8 represents the top 40 features ordered by its importance after applying this tree-based feature selection strategy on class balanced training dataset.

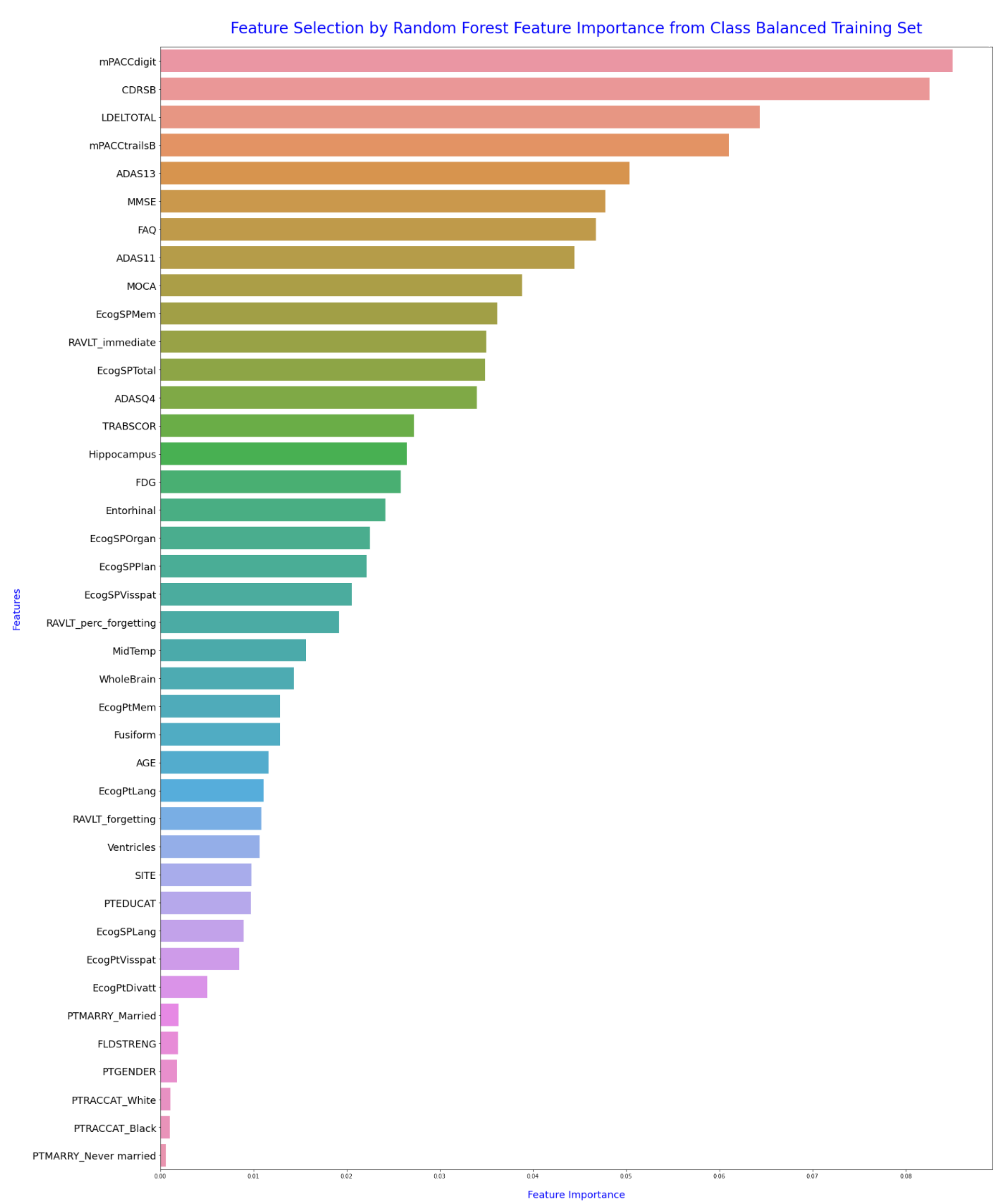

*Figure 4.8 Top 40 Features selected using Feature Importance after Imbalanced handling*

#### 4.3.9.3 Feature Selection using Elastic Net

The Elastic Net technique is a regularised approach that linearly integrates the lasso and ridge methods' L1 and L2 penalties. The L1 half of the penalty provides a sparse model, while the L2 part stabilises the regularisation route L1. The L2 element of the penalty averages out the coefficients of associated features, enforcing a grouping effect and removing the constraint on the number of selected features. As a result, Elastic Net has taken the flaws of both ridge and lasso and combined them to obtain the most relevant feature set. If any of the feature coefficients becomes zero using elastic net regularisation, it can be assumed that the feature (with the coefficient value of zero) has the least importance in identifying the target class and can be deleted during feature selection. In case of multi-class classification problem, the coefficient matrix shape will be (number of classes, number of features). That is, n number of coefficients will be calculated for each class, where n is the number of features. If the coefficient of any common feature becomes 0 for all classes, that feature can be removed before feeding the data to the model.

Elastic Net Regularization was implemented using Logistic Regression in this project. With Python 3.8.5, the Elastic Net model for feature selection was created using the LogisticRegression function with penalty 'elasticnet' from sklearn's linear model package. The top 40 features, as determined by Random Forest significance in previous sub-section, are now ran through this classifier again to refine the remaining features, which will be used in model creation.

Because the reported dataset is unbalanced, the train dataset with 1,500 rows and 40 features (selected through RF) was subjected to a feature selection using the Elastic Net before the imbalance treatment. Finally, Elastic Net unable to regularize any feature coefficient value as zero from a total of 40 independent variables, which is common among all three target classes. For class 0, it sets 1 feature coefficient as zero. Similarly, for class 1, the elastic net penalises 1 feature coefficient's value to zero. But those two features are not common between the two classes. So, those features cannot be dropped, although they are less important for one class's prediction but have significance for the other class's prediction. Therefore, it has been determined that all 40 attributes chosen by Random Forest are dependent to classify all three target classes. As a result, all 40 features will be carried forward for building the classifiers.

Following the adjustment of the class imbalances in the presented dataset, the train dataset with 2,073 rows and 40 features was subjected to feature selection using Elastic Net once again (selected via RF). For all three classes, Elastic Net was unable to regularise any of the feature coefficients to zero. Therefore, all the elements of coefficient matrix of shape (3,40) become non-zero, which means all 40 RF-selected features are important in classifying the target labels. As a result, the top 40 features selected by the random forest classifier have been finalised and applied to model construction.

Below Venn diagram 4.9 visualizes the intersection of all features among the three classes with coefficient value calculated as zero.

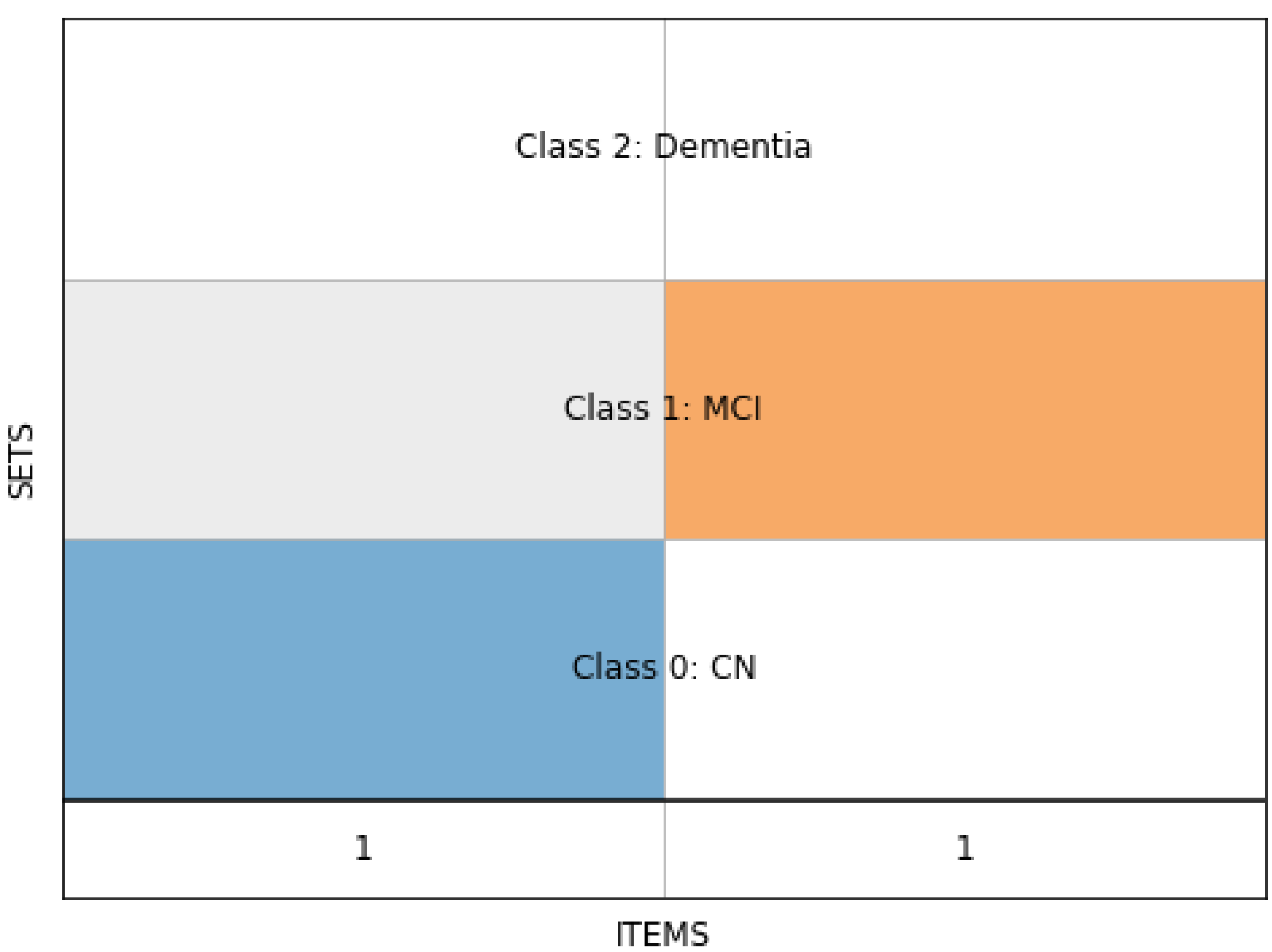

*Figure 4.9 Intersection of Feature sets with zero valued coefficients among 3 Classes Selected via Elastic Net Regularization*

As shown above, there are no coefficients that are common in all the 3 classes which are calculated as zero, hence all the coefficients for selected 40 features are significant in this multiclass classification problem and should be consider for model development.

### 4.4 Exploratory Data Analysis

Exploratory Data Analysis is a method for using descriptive statistic (quantifiable) and visualisations to investigate data in order to find underlying patterns, locate outliers, testing hypotheses, uncover missing values, and evaluate relationships. To answer any data science problem, EDA is required to first comprehend the problem formulation and find data feature correlations. EDA allows users to investigate data distribution for each predictor using plots such as histograms and distribution plots (univariate analysis), as well as find relationships between dependent and independent variables using bivariate and multivariate plots such as bar plots, scatter plots, and heatmaps etc. Below sub-sections describe all the exploratory data analysis performed in this study.

#### 4.4.1 Univariate Analysis

The simplest method of data analysis is univariate analysis. Data has only one variable because "uni" signifies "one." It does not concern with causes or relationships (unlike regression), and its primary goal is to describe; It takes information, summaries it, and looks for patterns. In univariate analysis, a variable simply refers to a condition or subset of data. It does not look at more than one variable at a time. It examines each variable individually in a dataset. There are two types of variables that can be used: categorical and numerical. Central Tendency (mean, mode, and median), Dispersion (range, variance), Quartiles (interquartile range), Frequency and Standard deviation are some of the patterns that may be easily discovered using univariate analysis. In this study the univariate analysis has been carried out using three important Python libraries with python version 3.8.5. For visualization Matplotlib and Seaborn packages are utilized, while the Pandas library is used for descriptive statistics.

##### 4.4.1.1 Visualization and Summary Statistics

This section explains how the statistical data and visualisation graphs are used to analyse the variables. Few of the important biomarkers among all the predictive factors are discusses here. This study looks at both numerical and category features. For numerical variables distribution plot, Central Tendency, Quartiles etc. statistics are observed. On the other hand, frequency distribution table, mode value, count plot etc. were inspected for categorical attributes. The details of each of the variables are explained below.

**a) CDRSB (Clinical Dementia Rating Sum of Boxes)**

The table 4.9 shows that for the patients who participated in ADNI trial, their clinical dementia rating spans from 0 to 10 (MIN-MAX range). The study had a total of 2,143 individuals, with an average CDRSB score of 1.54993. The 0.00 number in the first quartile (Q1) indicates that 25% of participants have a CDRSB score of 0.00. The second quantile number (Q2) denotes that the Dementia Rating of 50% of the population is less than 1.00. Similarly, third quartile or Q3 shows, 75% individuals are having clinical dementia rating less than 2.50. This reported analysis is as per the number of observations present in the dataset.

*Table 4.9 Descriptive Statistics for variable CDRSB*

| CDRSB | | | | | | | |
|---|---|---|---|---|---|---|---|
| **Count** | **Mean Value** | **Standard Deviation** | **Minimum value** | **25% (Q1)** | **50% (Q2)** | **75% (Q3)** | **Maximum Value** |
| 2143 | 1.54993 | 1.81454 | 0.00 | 0.00 | 1.00 | 2.50 | 10.00 |

Below figure 4.10 shows that the distribution of CDRSB variable is right skewed. It is observed from the plot that for most of the participants the dementia rating sum of boxes score ranges between 0 to 3. There are very few subjects in this ADNIMERGE study data whose score goes beyond 4.

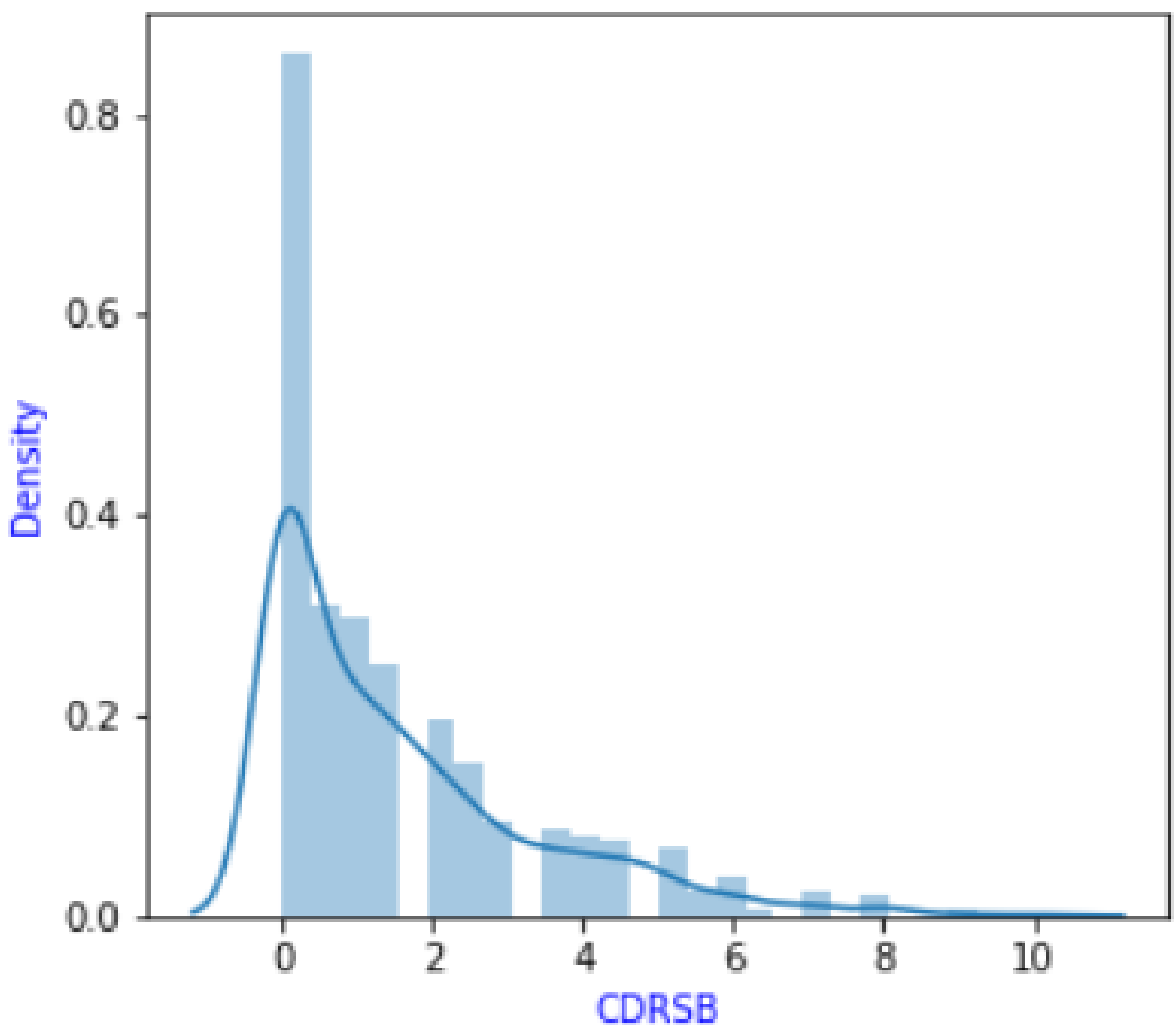

*Figure 4.10 Distribution Plot for variable CDRSB*

**b) mPACCdigit (modified Preclinical Alzheimer's Cognitive Composite with Digit)**

The mPACCdigit cognitive test assesses working memory by asking the patient to repeat a series of digits until they are unable to do so any longer. The table 4.10 shows that for the patients who participated in ADNI trial, their PACCdigit test results span from -23.6077 to 6.3004 (MIN-MAX range). The study had a total of 2,143 individuals, with an average test score of -5.7128. The -10.3572 number in the first quartile (Q1) indicates that 25% of participants have a test score of below -10.3572. The second quantile number (Q2) denotes that the cognitive test result of 50% of the population is less than -4.5510. Similarly, third quartile or Q3 shows, 75% individuals are having test outcome less than -0.6857. This reported analysis is as per the number of observations present in the dataset.

*Table 4.10 Descriptive Statistics for variable mPACCdigit*

| **mPACCdigit** | | | | | | | |
|---|---|---|---|---|---|---|---|
| **Count** | **Mean Value** | **Standard Deviation** | **Minimum value** | **25% (Q1)** | **50% (Q2)** | **75% (Q3)** | **Maximum Value** |
| 2143 | -5.7128 | 6.3454 | -23.6077 | -10.3572 | -4.5510 | -0.6857 | 6.3004 |

Below figure 4.11 shows that the distribution of mPACCdigit variable is almost normal. Hence the reported sample distribution of patient’s PACC digit test score is a good representation of the population’s test score those who are suffering from dementia. Also, it can be observed from the plot that most of the patients those who are reported in ADNI trials having cognitive test score between -4 to 3.

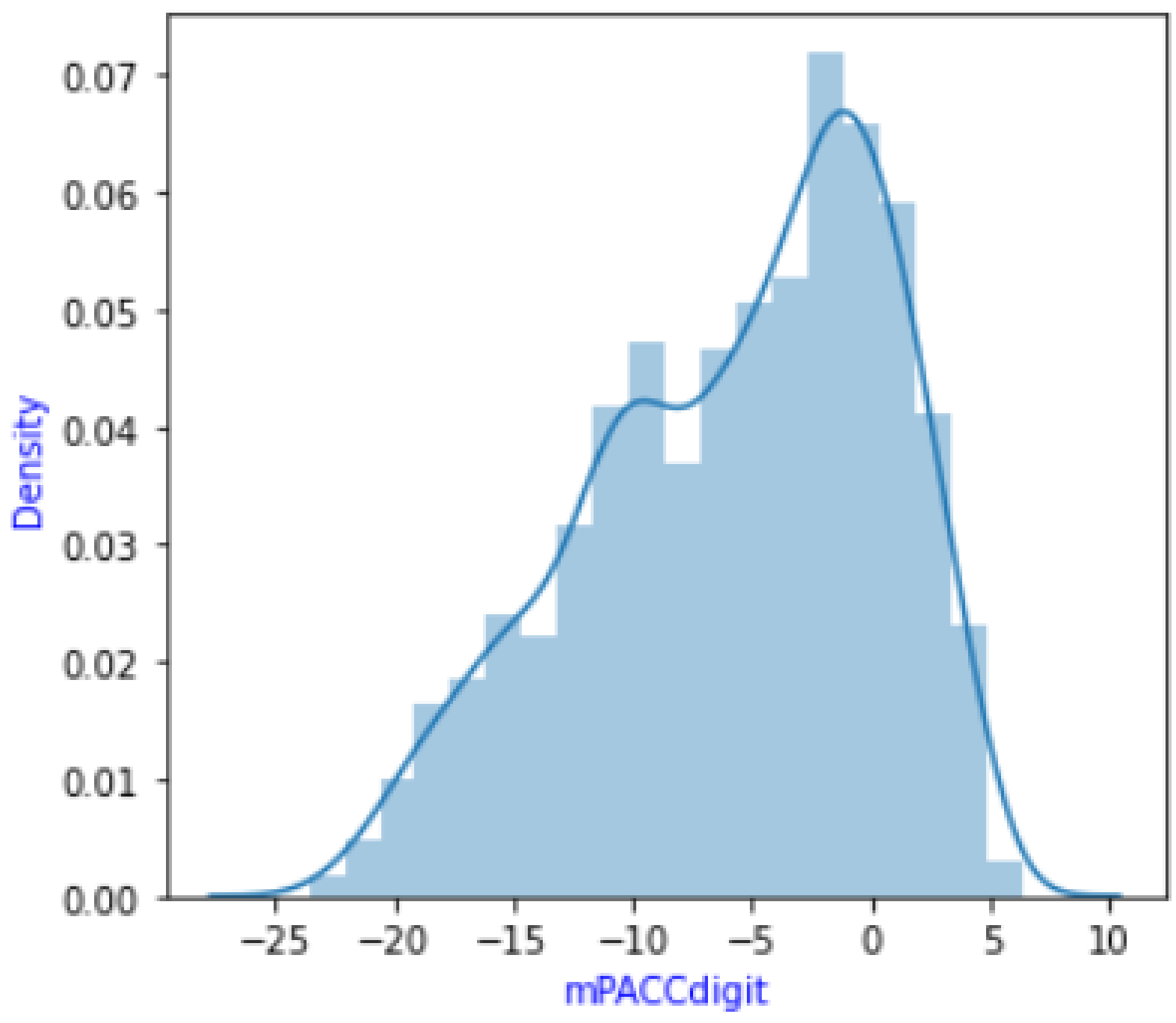

*Figure 4.11 Distribution Plot for variable mPACCdigit*

**c) mPACCdigitB (modified Preclinical Alzheimer's Cognitive Composite with Trails B)**

The mPACCtrailsB test measures processing speed, with a lower score suggesting more severe impairment. The table 4.11 shows that for the patients who participated in ADNI trial, their PACCdigitB test results span from -23.6077 to 7.4086 (MIN-MAX range). The study had a total of 2,143 individuals, with an average test score of -5.3552. The -9.8969 number in the first quartile (Q1) indicates that 25% of participants have a test score of below -9.8969. The second quantile number (Q2) denotes that the cognitive test result of 50% of the population is less than -4.1136. Similarly, third quartile or Q3 shows, 75% individuals are having test outcome less than -0.5510. This reported analysis is as per the number of observations present in the dataset.

*Table 4.11 Descriptive Statistics for variable mPACCdigitB*

| mPACCdigitB | | | | | | | |
|---|---|---|---|---|---|---|---|
| **Count** | **Mean Value** | **Standard Deviation** | **Minimum value** | **25% (Q1)** | **50% (Q2)** | **75% (Q3)** | **Maximum Value** |
| 2143 | -5.3552 | 6.0159 | -23.6077 | -9.8969 | -4.1136 | -0.5510 | 7.4086 |

Below figure 4.12 shows that the distribution of mPACCdigitB variable is almost normal. Hence the reported sample distribution of patient's PACC digit Trails B test score is a good representation of the population's test score those who are suffering from dementia. Also, it can be observed from the plot that most of the patients those who are reported in ADNI trials having cognitive test score between -4 to 3.

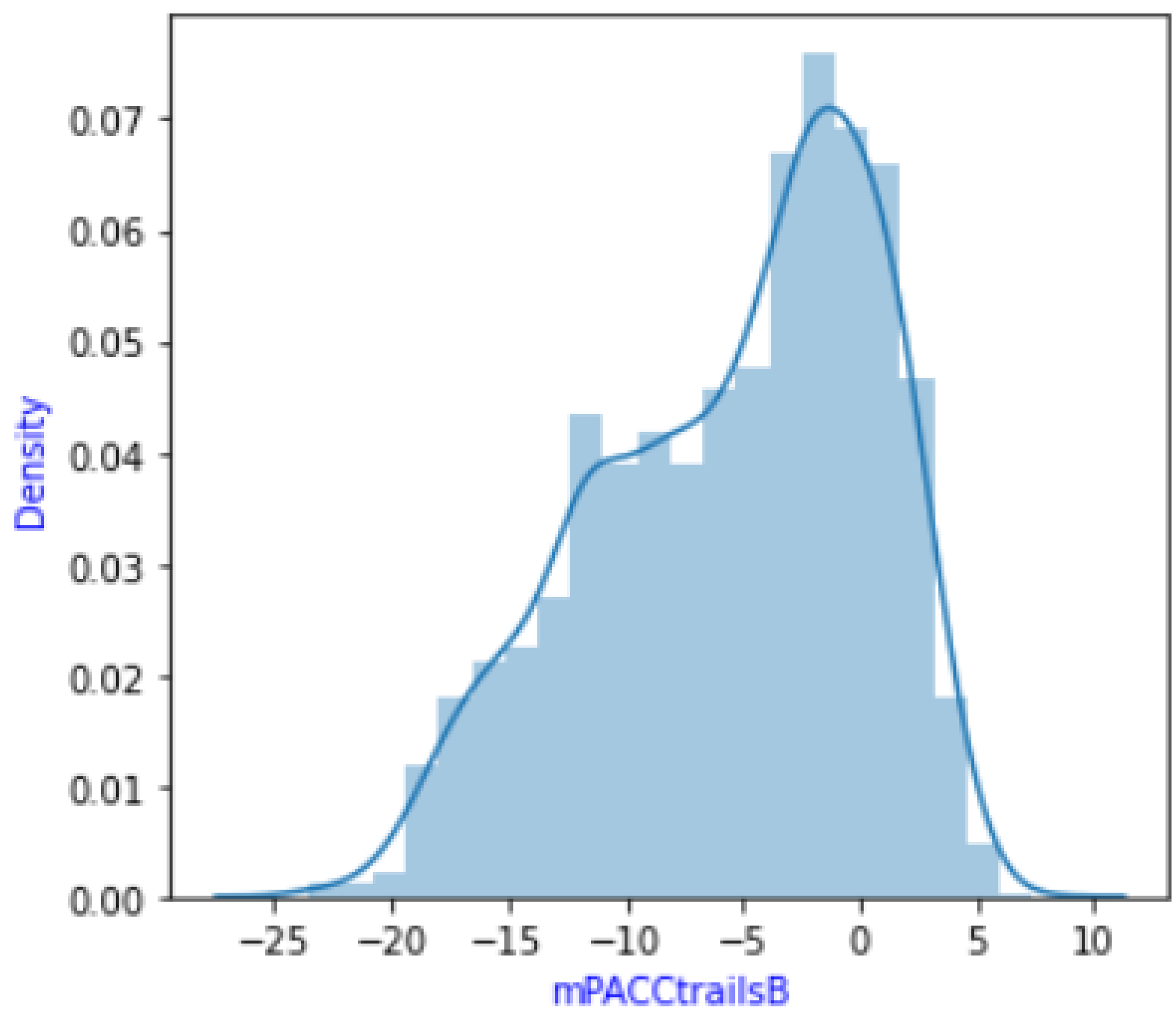

*Figure 4.12 Distribution Plot for variable mPACCdigitB*

**d) MMSE (Mini-Mental State Examination)**

The MMSE test is widely used for assessing the cognitive function. The table 4.12 shows that for the patients who participated in ADNI trial, their MMSE test results span from 16.00 to 30.00 (MIN-MAX range). The study had a total of 2,143 individuals, with an average test score of 27.2734. The 26.00 number in the first quartile (Q1) indicates that 25% of participants have a test score of below 26.00. The second quantile number (Q2) denotes that the cognitive test result of 50% of the population (median value) is less than 28.00. Similarly, third quartile or Q3 shows, 75% individuals are having test outcome less than 29.00. This reported analysis is as per the number of observations present in the dataset.

*Table 4.12 Descriptive Statistics for variable MMSE*

| MMSE | | | | | | | |
|---|---|---|---|---|---|---|---|
| **Count** | **Mean Value** | **Standard Deviation** | **Minimum value** | **25% (Q1)** | **50% (Q2)** | **75% (Q3)** | **Maximum Value** |
| 2143 | 27.273448 | 2.714019 | 16.00 | 26.00 | 28.00 | 29.00 | 30.00 |

Below figure 4.13 shows that the distribution of MMSE variable is left skewed. It is observed from the plot that for most of the participants the MMSE rating ranges between 28 to 30. There are very few subjects in this ADNI study data whose score goes below 22.5.

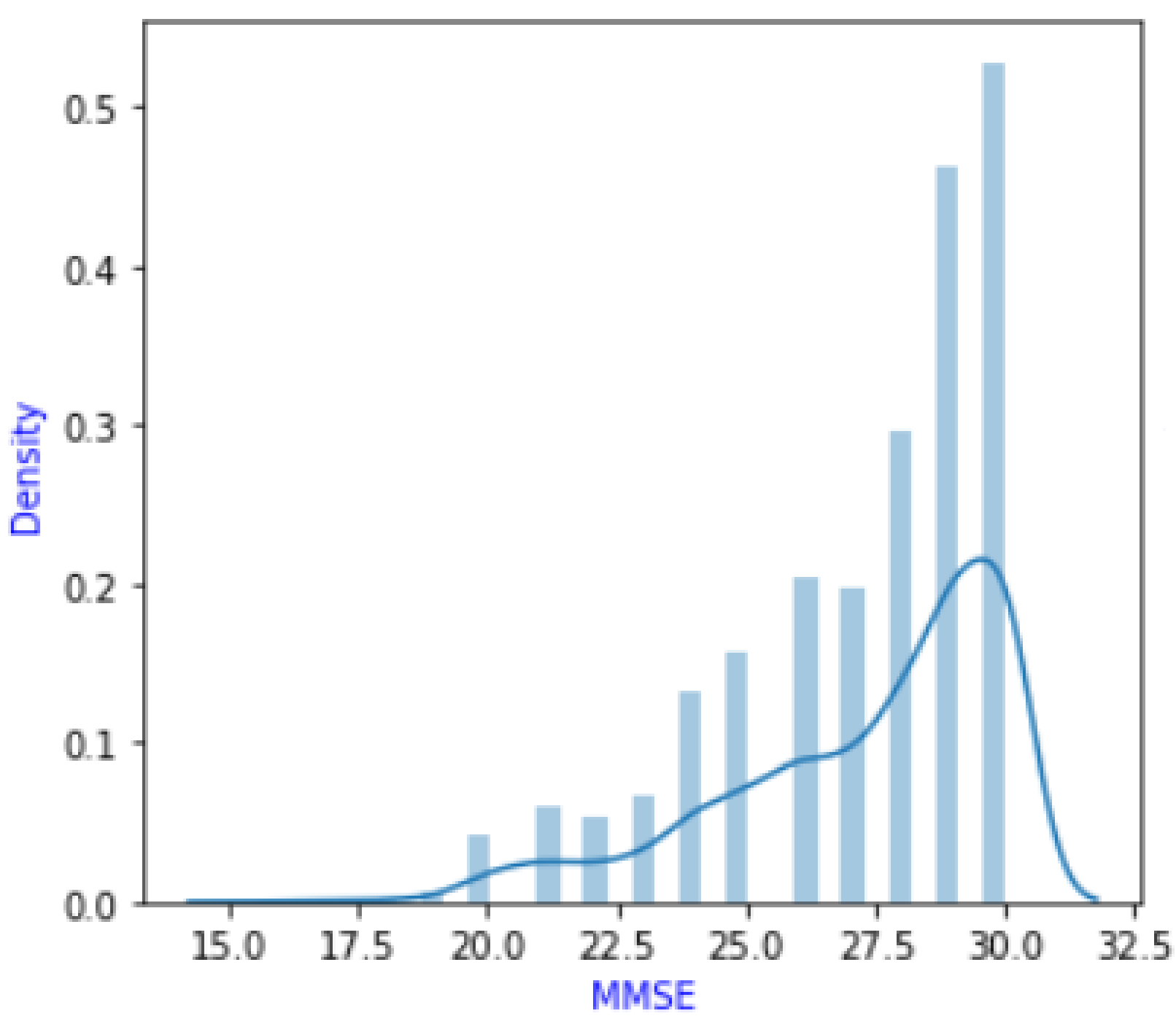

*Figure 4.13 Distribution Plot for variable MMSE*

**e) FAQ (Functional Activities Questionnaire)**

The FAQ test is widely used to measure individuals' ability to carry out daily work. The table 4.13 shows that for the patients who participated in ADNI trial, their FAQ score span from -1.00 to 30.00 (MIN-MAX range). The study had a total of 2,143 individuals, with an average test score of 4.0401. The 0.00 number in the first quartile (Q1) indicates that 25% of participants have a test score of below 0.00. The second quantile number (Q2) denotes that the Questionnaire test result of 50% of the population (median value) is less than 1.00. Similarly, third quartile or Q3 shows, 75% individuals are having test result less than 6.00. This reported analysis is as per the number of observations present in the dataset.

*Table 4.13 Descriptive Statistics for variable FAQ*

| FAQ | | | | | | | |
|---|---|---|---|---|---|---|---|
| **Count** | **Mean Value** | **Standard Deviation** | **Minimum value** | **25% (Q1)** | **50% (Q2)** | **75% (Q3)** | **Maximum Value** |
| 2143 | 4.040131 | 6.180092 | -1.00 | 0.00 | 1.00 | 6.00 | 30.00 |

Below figure 4.14 shows that the distribution of FAQ variable is right skewed. It is observed from the plot that for most of the participants the FAQ rating ranges between 0 to 1. There are very few subjects in this ADNI study data whose FAQ score goes above 10.

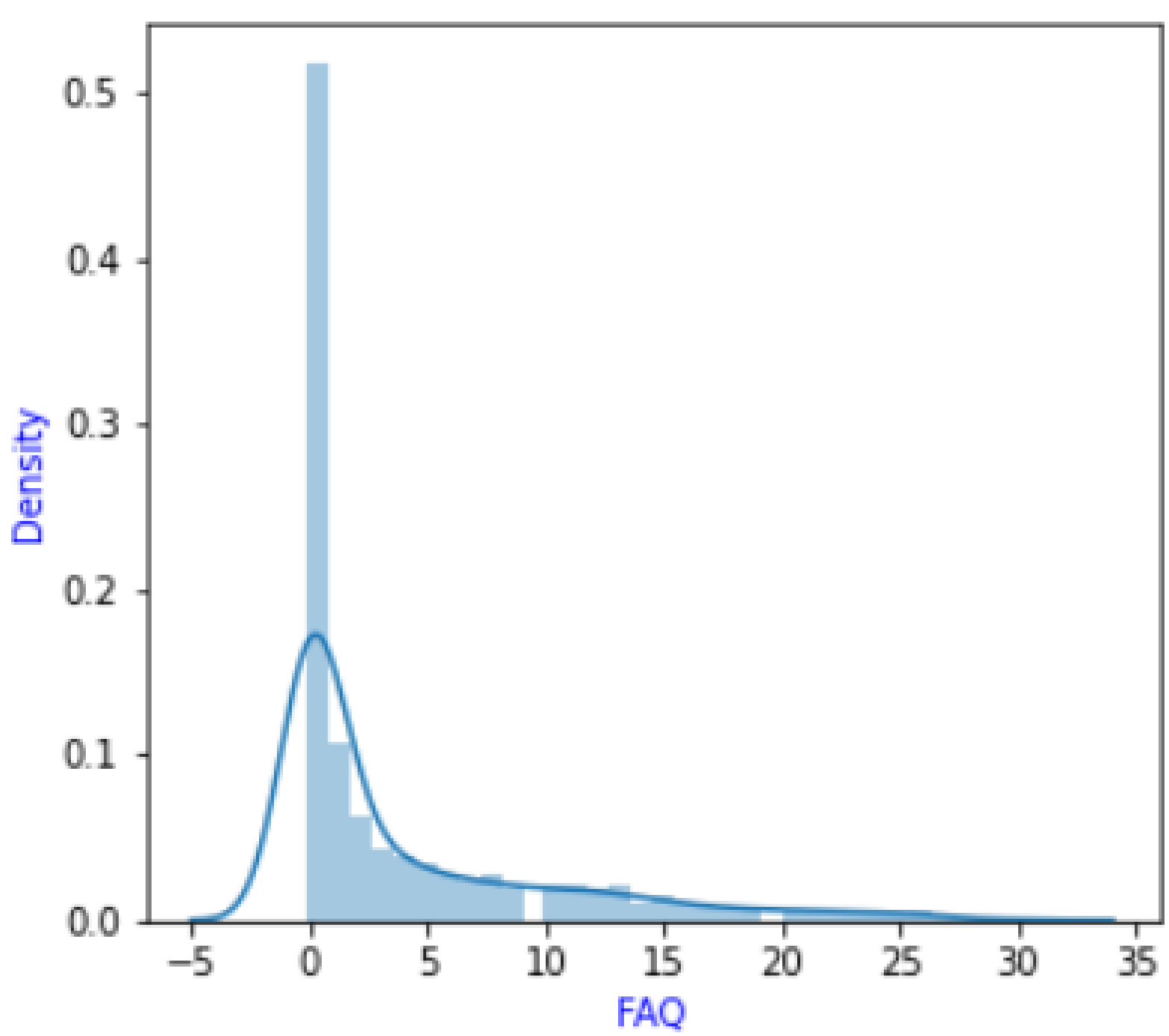

*Figure 4.14 Distribution Plot for variable FAQ*

**f) LDELTOTEL (Logical Memory - Delayed Recall)**

The LDELTOTEL (Delayed recall) is frequently utilised in neuropsychological tests to compare the rate of loss of previously delivered knowledge to established standards. The table 4.14 shows that for the patients who participated in ADNI trial, their LDELTOTEL score span from 0.00 to 23.00 (MIN-MAX range). The study had a total of 2,143 individuals, with an average test score of 7.550163. The 3.00 number in the first quartile (Q1) indicates that 25% of participants have a test score of below 3.00. The second quantile number (Q2) denotes that the test result of 50% of the population (median value) is less than 8.00. Similarly, third quartile or Q3 shows, 75% individuals are having test result less than 11.00. This reported analysis is as per the number of observations present in the dataset.

*Table 4.14 Descriptive Statistics for variable LDELTOTEL*

| LDELTOTEL | | | | | | | |
|---|---|---|---|---|---|---|---|
| **Count** | **Mean Value** | **Standard Deviation** | **Minimum value** | **25% (Q1)** | **50% (Q2)** | **75% (Q3)** | **Maximum Value** |
| 2143 | 7.550163 | 5.466234 | 0.00 | 3.00 | 8.00 | 11.00 | 23.00 |

Below figure 4.15 shows that the distribution of LDELTOTEL variable is right skewed. It is observed from the plot that there are two modes in the distribution. For most of the participants the LDELTOTEL rating ranges between 0 to 1 and between 9 to 10. There are very few subjects in this ADNI study data whose LDELTOTEL score goes above 15.

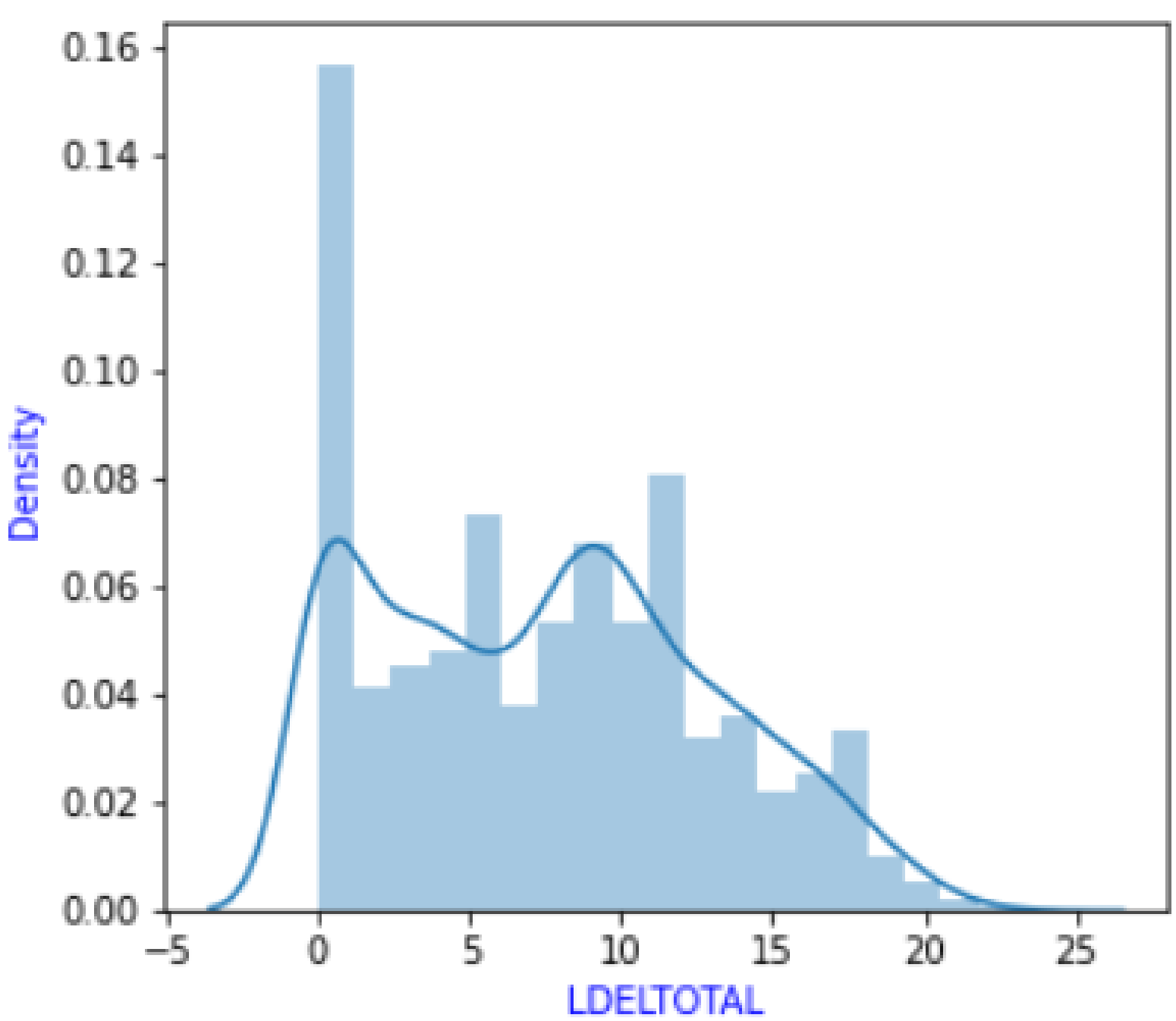

*Figure 4.15 Distribution Plot for variable LDELTOTEL*

**g) AGE (The Patient's Age)**

As per the table 4.15, for the patients who participated in ADNI trial, their AGE lies between 50.40 to 91.40 (MIN-MAX range). The study had a total of 2,143 individuals, with an average age of 73.16. The 68.20 number in the first quartile (Q1) indicates that 25% of participants have age below 68.20. The second quantile number (Q2) denotes that the age limit of 50% of the population (median value) is less than 73.30, who enrolled for the trials. Similarly, third quartile or Q3 shows, 75% individuals are having age limit less than 78.30. This reported analysis is as per the number of observations present in the dataset.

*Table 4.15 Descriptive Statistics for variable AGE*

| AGE | | | | | | | |
|---|---|---|---|---|---|---|---|
| **Count** | **Mean Value** | **Standard Deviation** | **Minimum value** | **25% (Q1)** | **50% (Q2)** | **75% (Q3)** | **Maximum Value** |
| 2143 | 73.16640 | 7.3042 | 50.40 | 68.20 | 73.20 | 78.30 | 91.40 |

Below figure 4.16 shows that the AGE variable almost follows a normal distribution. Hence the reported sample distribution of patient's Age is a good representation of the population's age those who are suffering from dementia. Also, it can be observed that most of the patients those who are suffering from this disease having age between 67 to 80.

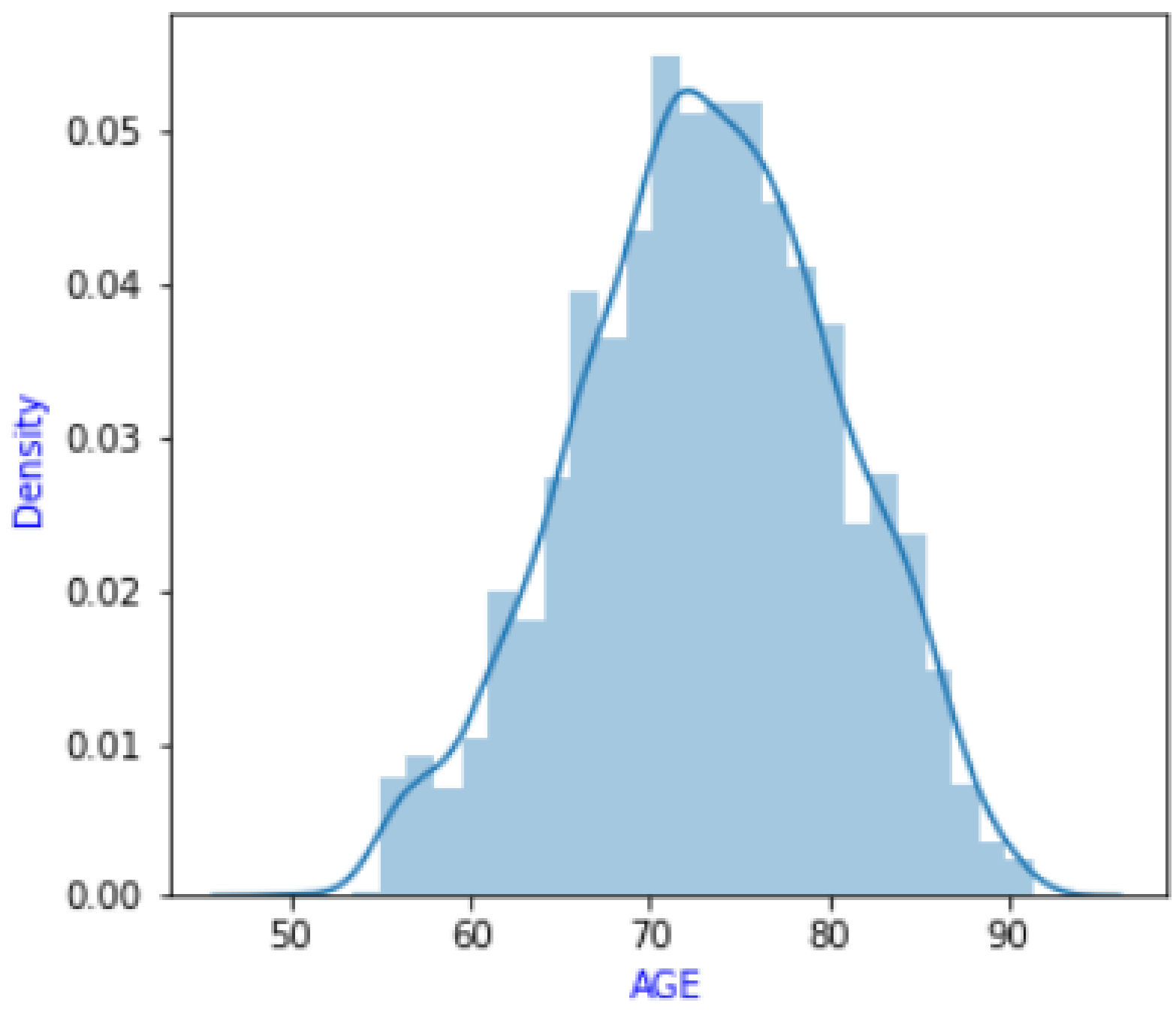

*Figure 4.16 Distribution Plot for variable AGE*

**h) EcogSPMem (Measurement of Everyday Cognition completed out by Study Partner)**

Measurement of Everyday Cognition is a short questionnaire filled out by a study partner (Sp) in which they score the individual's performance in numerous cognitive areas in contrast to 10 years earlier. As per the table 4.16, for the patients who participated in ADNI trial, their EcogSPMem score spans from 0.931257 to 4.00 (MIN-MAX range). The study had a total of 2,143 individuals, with an average score of 2.058. The 1.50 number in the first quartile (Q1) indicates that 25% of participants have score below 1.50. The second quantile number (Q2) denotes that the score of 50% of the population (median value) is less than 2.0382, who enrolled for the trials. Similarly, third quartile or Q3 shows, 75% individuals are having measurement score less than 2.2768. This reported analysis is as per the number of observations present in the dataset.

*Table 4.16 Descriptive Statistics for variable EcogSPMem*

| EcogSPMem | | | | | | | |
|---|---|---|---|---|---|---|---|
| **Count** | **Mean Value** | **Standard Deviation** | **Minimum value** | **25% (Q1)** | **50% (Q2)** | **75% (Q3)** | **Maximum Value** |
| 2143 | 2.058112 | 0.747287 | 0.931257 | 1.50 | 2.038294 | 2.276830 | 4.00 |

Below figure 4.17 shows that the distribution of EcogSPMem variable is right skewed. It is observed from the plot that there are two modes in the distribution. For most of the participants the EcogSPMem rating ranges between 0.9 to 1.5 and between 2.0 to 2.4. There are very few subjects in this ADNI study data whose EcogSPMem score goes above 2.5.

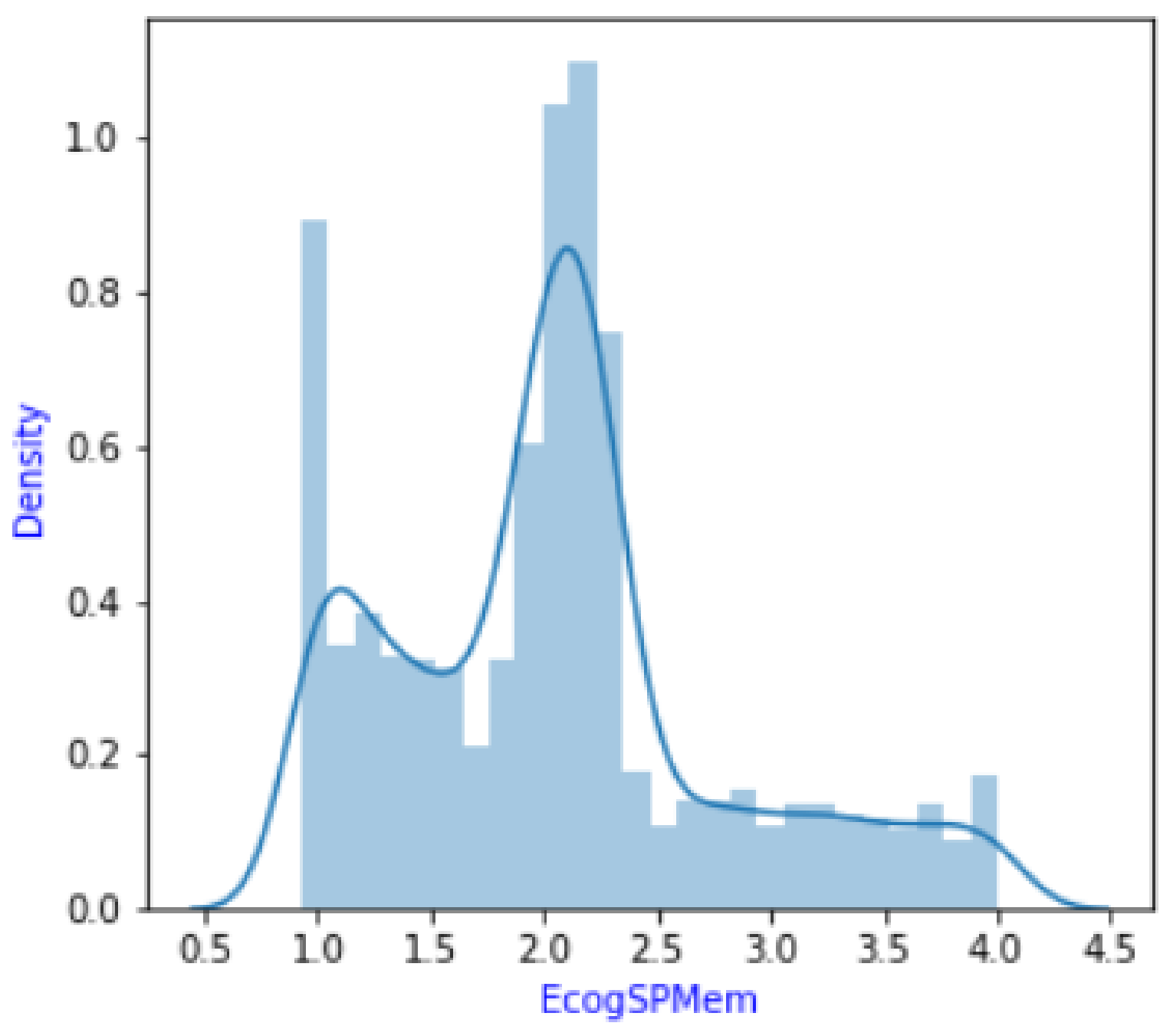

*Figure 4.17 Distribution Plot for variable EcogSPMem*

**i) RAVLT_immediate (Rey Auditory Verbal Learning Test for immediate response)**

The Rey's Auditory Verbal Learning Test is a series of neurophysiological tests that assesses a patient's memory performance. As per the table 4.17, for the patients who participated in ADNI trial, their RAVLT_immediate score spans from 0.00 to 71.00 (MIN-MAX range). The study had a total of 2,143 individuals, with an average score of 36.0279. The 26.00 number in the first quartile (Q1) indicates that 25% of participants have score below 26.00. The second quantile number (Q2) denotes that the score of 50% of the population (median value) is less than 35.00, who enrolled for the trials. Similarly, third quartile or Q3 shows, 75% individuals are having test score less than 45.00. This reported analysis is as per the number of observations present in the dataset.

*Table 4.17 Descriptive Statistics for variable RAVLT_immediate*

| RAVLT_immediate | | | | | | | |
|---|---|---|---|---|---|---|---|
| **Count** | **Mean Value** | **Standard Deviation** | **Minimum value** | **25% (Q1)** | **50% (Q2)** | **75% (Q3)** | **Maximum Value** |
| 2143 | 36.0279 | 12.7563 | 0.00 | 26.00 | 35.00 | 45.00 | 71.00 |

Below figure 4.18 shows that the RAVLT_immediate variable almost follows a normal distribution. Hence the reported sample distribution of patient's test score is a good representation of the population's score those who are suffering from dementia. Also, it can be observed that most of the patients those who are suffering from this disease having this clinical test score between 30 to 40.

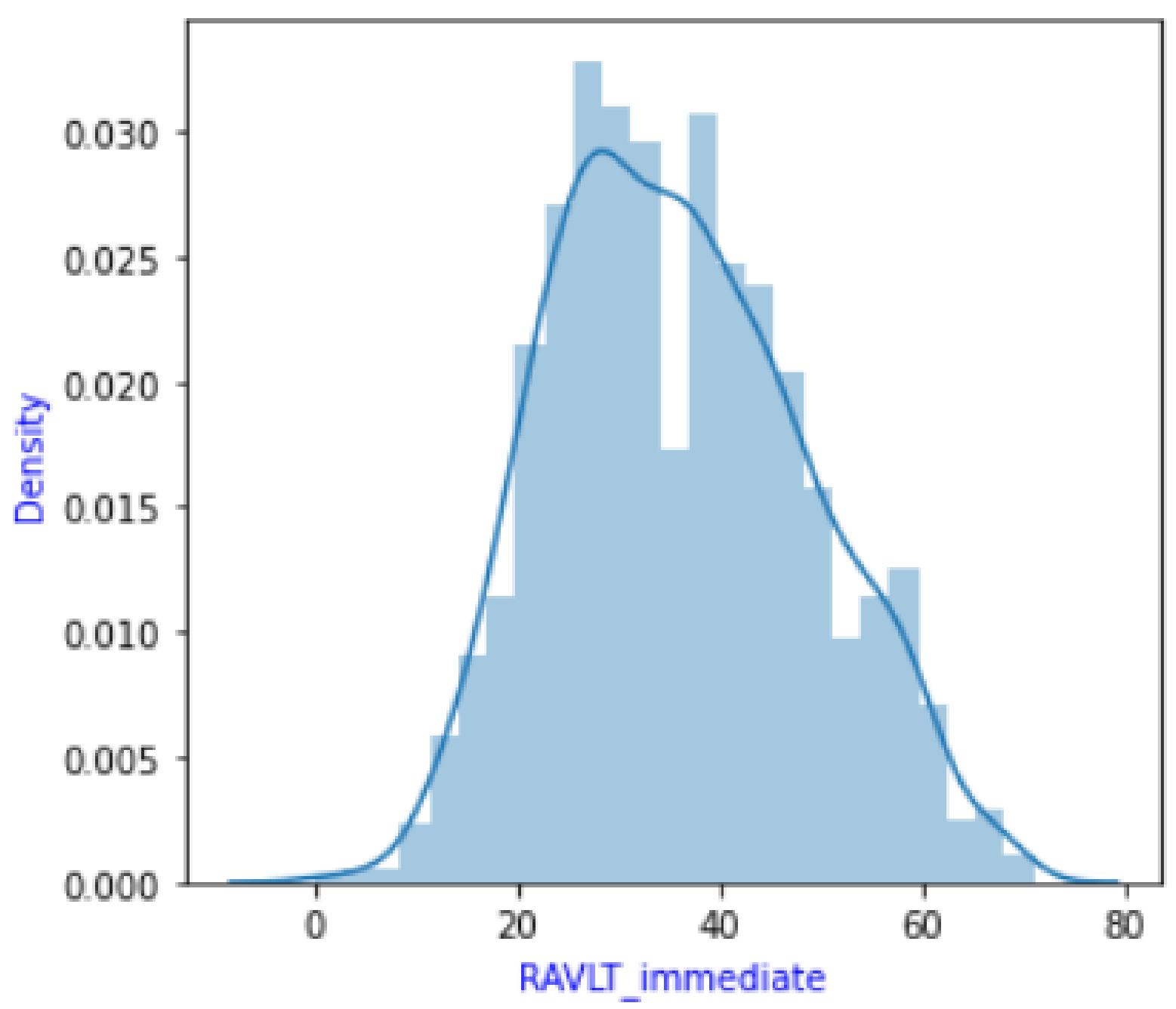

*Figure 4.18 Distribution Plot for variable RAVLT_immediate*

**j) Fusiform (Fusiform Gyrus)**

The fusiform gyrus is a vast gyrus in the cerebral hemispheres that spans the basal surface of the temporal and occipital lobes and plays a key role in object and face identification. As per the table 4.18, for the patients who participated in ADNI trial, their fusiform gyrus size spans from 8,991.00 to 29,950.00 (MIN-MAX range). The study had a total of 2,143 individuals, with an average fusiform size of 17,415.4484. The 15,641.00 number in the first quartile (Q1) indicates that 25% of participants have fusiform size below 15,641.00. The second quantile number (Q2) denotes that the fusiform gyrus size of 50% of the population (median value) is less than 17,412.00, who enrolled for the trials. Similarly, third quartile or Q3 shows, 75% individuals are having fusiform structure of size less than 19,055.00. This reported analysis is as per the number of observations present in the dataset.

*Table 4.18 Descriptive Statistics for variable Fusiform*

| Fusiform | | | | | | | |
|---|---|---|---|---|---|---|---|
| **Count** | **Mean Value** | **Standard Deviation** | **Minimum value** | **25% (Q1)** | **50% (Q2)** | **75% (Q3)** | **Maximum Value** |
| 2143 | 17415.4484 | 2628.4928 | 8991.00 | 15641.00 | 17412.00 | 19055.00 | 29950.00 |

Below figure 4.19 shows that the Fusiform variable almost follows a normal distribution. Hence the reported sample distribution of patient's size of fusiform gyrus structure is a good representation of the population's fusiform size those who are suffering from dementia. Also, it can be observed that most of the patients those who are suffering from this disease having this fusiform gyrus size of between 17,000 to 20,000.

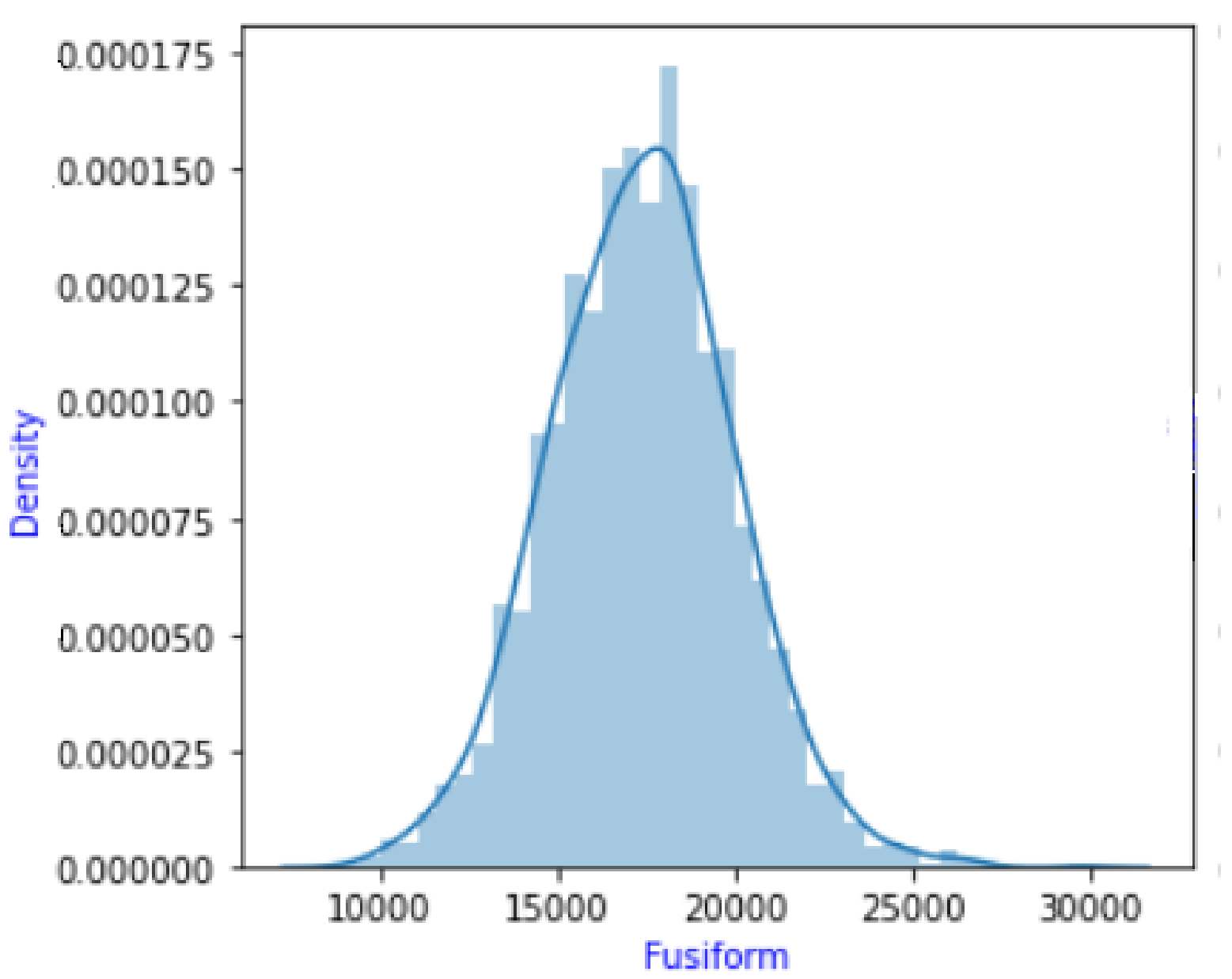

*Figure 4.19 Distribution Plot for variable Fusiform*

**k) WholeBrain (Entire brain volume)**

As per the table 4.19, for the patients who participated in ADNI trial, their whole brain volume rangers from 5,21,287.00 to 14,86,040.00 (MIN-MAX range). The study had a total of 2,143 individuals, with an average brain volume of 10,25,643.00. The 9,47,287.50 number in the first quartile (Q1) indicates that 25% of participants have brain volume below 9,47,287.50. The second quantile number (Q2) denotes that the volume of entire brain of 50% of the population (median value) is less than 10,24,400.00, who enrolled for the trials. Similarly, third quartile or Q3 shows, 75% individuals are having volume of whole brain less than 10,98,765.00. This reported analysis is as per the number of observations present in the dataset.

*Table 4.19 Descriptive Statistics for variable WholeBrain*

| **WholeBrain** | | | | | | | |
|---|---|---|---|---|---|---|---|
| **Count** | **Mean Value** | **Standard Deviation** | **Minimum value** | **25% (Q1)** | **50% (Q2)** | **75% (Q3)** | **Maximum Value** |
| 2143 | 1025643 | 109913.60 | 521287 | 947287.50 | 1024400 | 1098765 | 1486040 |

Below figure 4.20 shows that the WholeBrain variable almost follows a normal distribution. Hence the reported sample distribution of patient's volume of entire brain structure is a good representation of the population's brain volume those who are suffering from dementia. Also, it can be observed that most of the patients those who are suffering from this disease having volume of whole brain between 1 to 1.1 (in log scale).

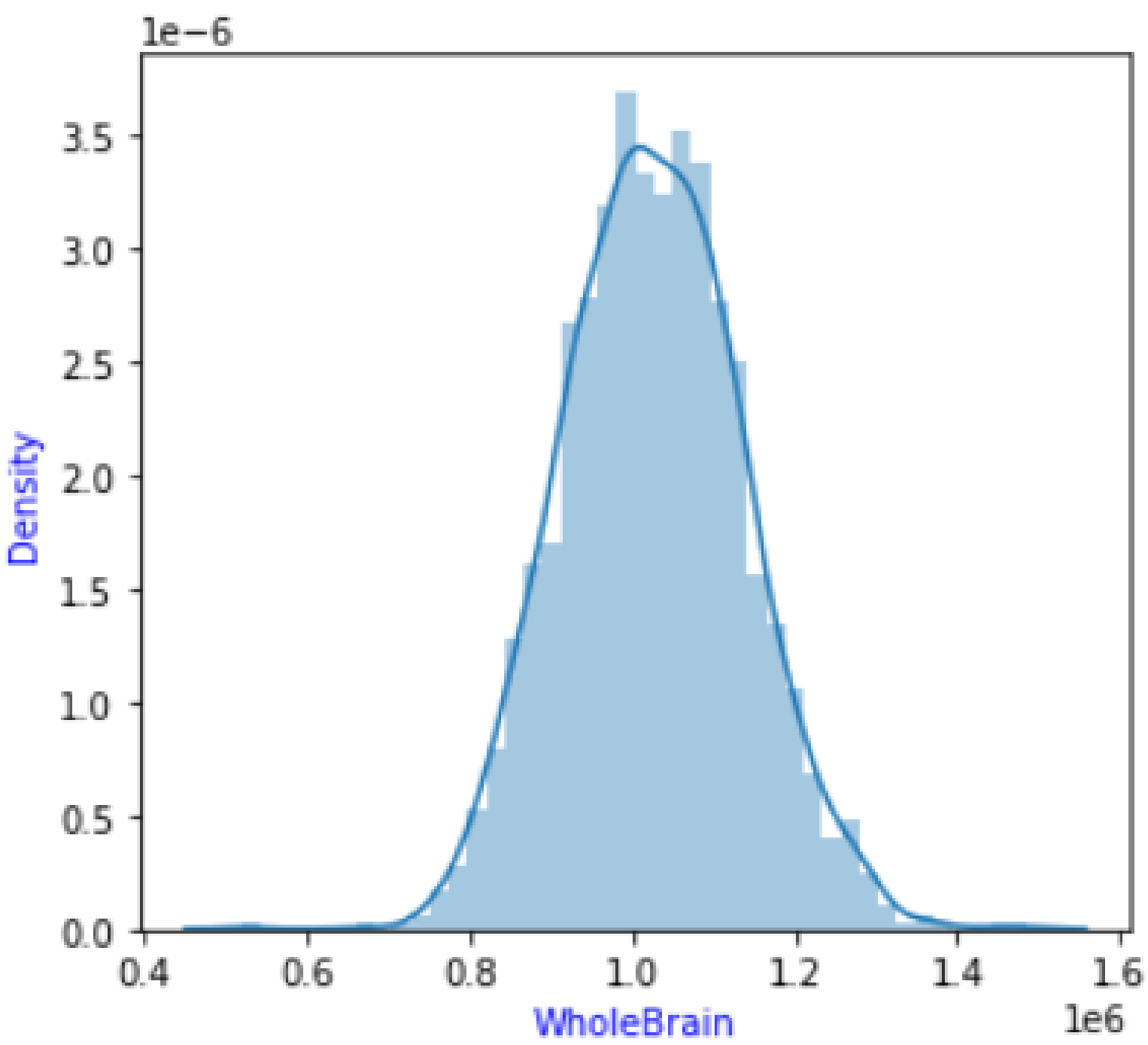

*Figure 4.20 Distribution Plot for variable WholeBrain*

### l) ADAS13 (Alzheimer's Disease Assessment Scale - 13 items)

Alzheimer's Disease Assessment Scale is relatively detailed test to assess cognitive and non-cognitive symptoms of AD. As per the table 4.20, for the patients who participated in ADNI trial, their ADAS13 score spans from 0.00 to 52.00 (MIN-MAX range). The study had a total of 2,143 individuals, with an average score of 16.4201. The 9.00 number in the first quartile (Q1) indicates that 25% of participants have score below 9.00. The second quantile number (Q2) denotes that the score of 50% of the population (median value) is less than 14.33, who enrolled for the trials. Similarly, third quartile or Q3 shows, 75% individuals are having exam score less than 22.67. This reported analysis is as per the number of observations present in the dataset.

*Table 4.20 Descriptive Statistics for variable ADAS13*

| ADAS13 | | | | | | | |
|---|---|---|---|---|---|---|---|
| **Count** | **Mean Value** | **Standard Deviation** | **Minimum value** | **25% (Q1)** | **50% (Q2)** | **75% (Q3)** | **Maximum Value** |
| 2143 | 16.4201 | 9.6961 | 0.00 | 9.00 | 14.33 | 22.67 | 52.00 |

Below figure 4.21 shows that the distribution of ADAS13 variable is right skewed. For most of the participants the ADAS13 rating ranges between 10.00 to 15.00. There are very few subjects in this ADNI study data whose ADAS13 score goes above 30.

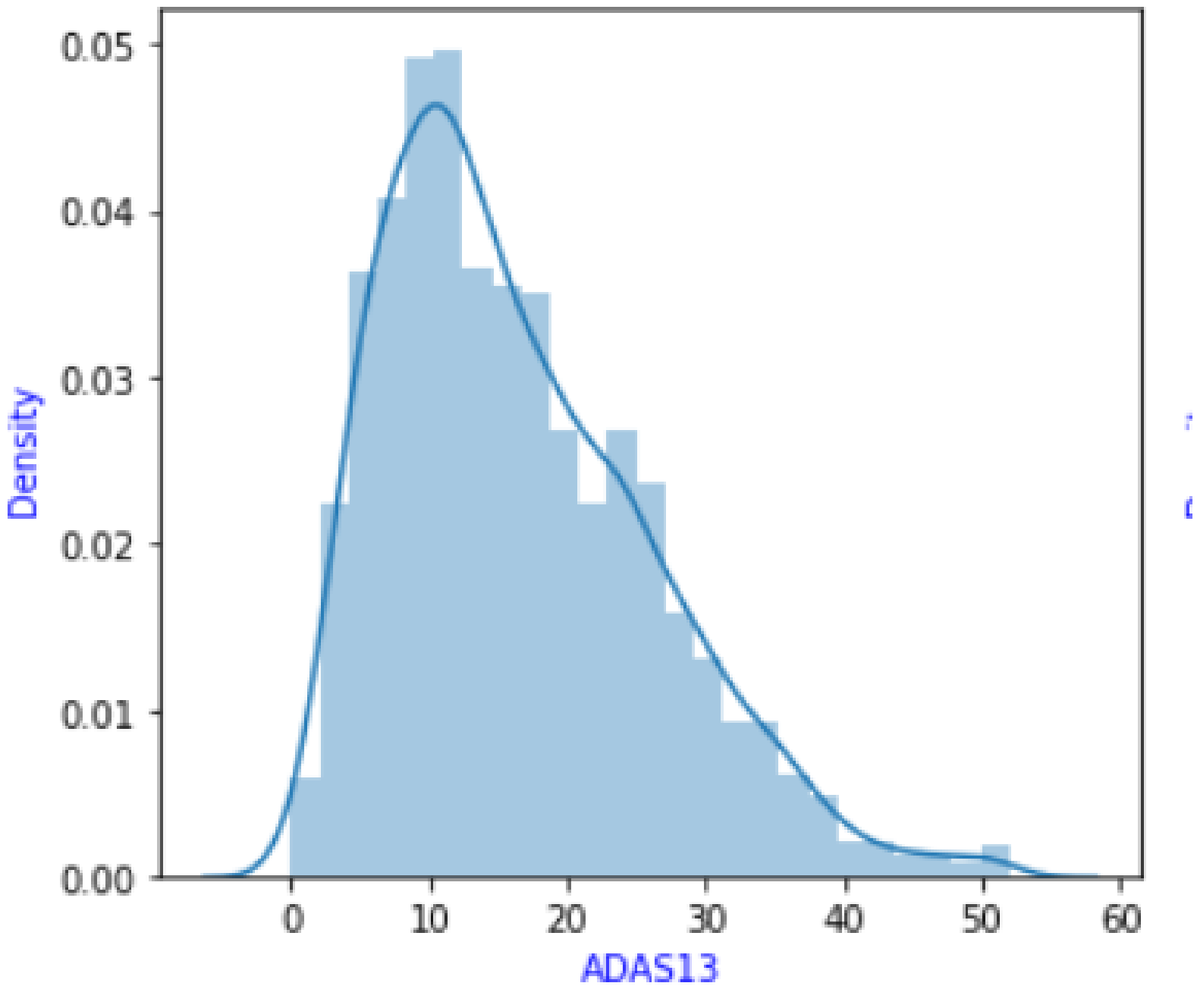

*Figure 4.21 Distribution Plot for variable ADAS13*

**m) EcogPtMem (Measurement of Everyday Cognition completed out by Participant)**

Measurement of Everyday Cognition is a short questionnaire filled out by a Participant (Pt) in which they score the individual's performance in numerous cognitive areas in contrast to 10 years earlier. As per the table 4.21, for the patients who participated in ADNI trial, their EcogPtMem score spans from 1.00 to 4.00 (MIN-MAX range). The study had a total of 2,143 individuals, with an average score of 2.0715. The 1.75 number in the first quartile (Q1) indicates that 25% of participants have score below 1.75. The second quantile number (Q2) denotes that the score of 50% of the population (median value) is less than 2.0719, who enrolled for the trials. Similarly, third quartile or Q3 shows, 75% individuals are having measurement score less than 2.1612. This reported analysis is as per the number of observations present in the dataset.

*Table 4.21 Descriptive Statistics for variable EcogPtMem*

| **EcogPtMem** | | | | | | | |
|---|---|---|---|---|---|---|---|
| **Count** | **Mean Value** | **Standard Deviation** | **Minimum value** | **25% (Q1)** | **50% (Q2)** | **75% (Q3)** | **Maximum Value** |
| 2143 | 2.0715 | 0.5799 | 1.00 | 1.75 | 2.0719 | 2.1612 | 4.00 |

Below figure 4.22 shows that the distribution of EcogPtMem variable is right skewed. For most of the participants the EcogPtMem rating ranges between 2.0 to 2.2. There are very few subjects in this ADNI study data whose EcogPtMem score goes above 2.5.

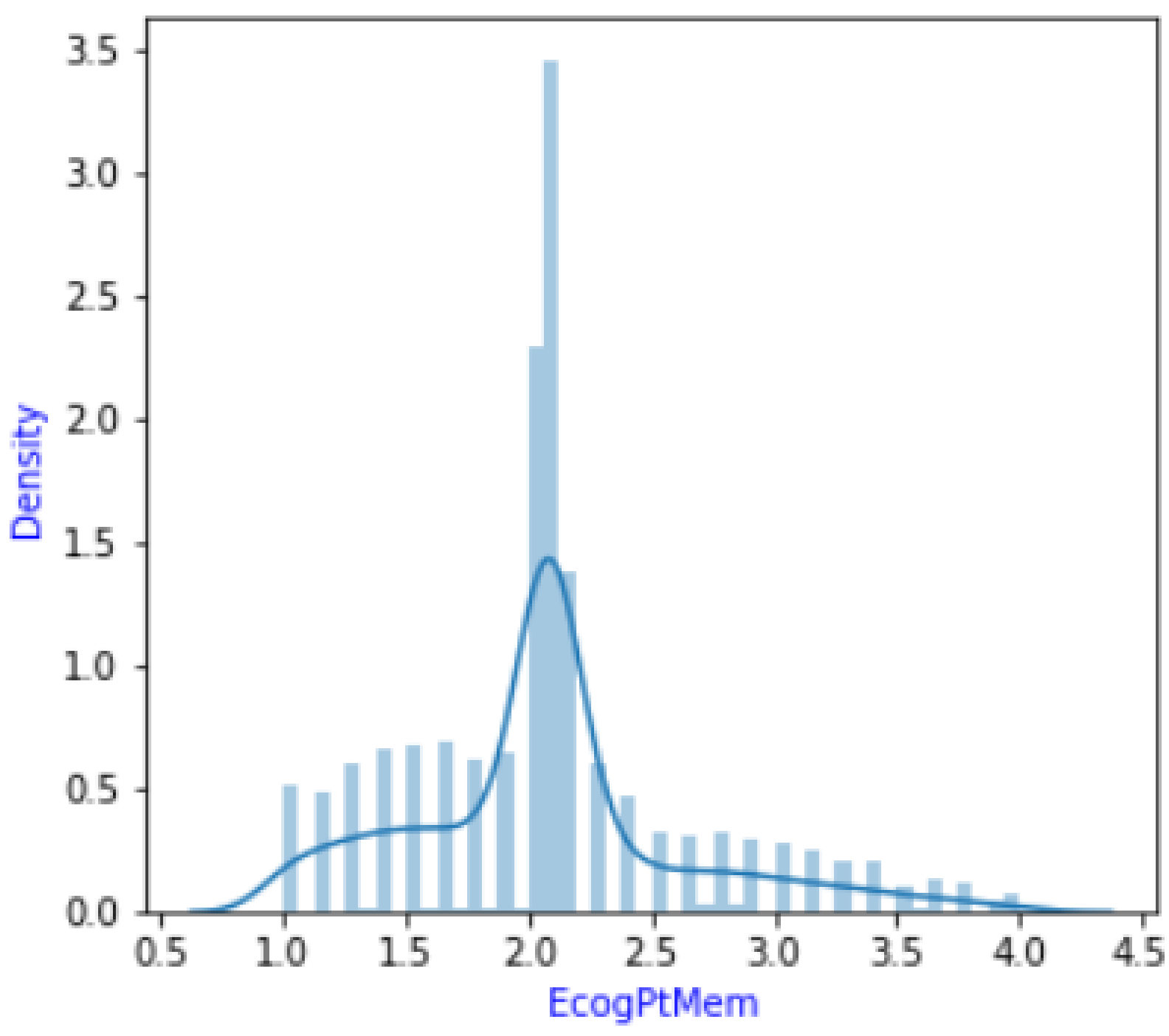

*Figure 4.22 Distribution Plot for variable EcogPtMem*

**n) FDG (Fluorodeoxyglucose)**

As per the table 4.22, for the patients who participated in ADNI trial, their FDG score rangers from 0.6936 to 1.7011 (MIN-MAX range). The study had a total of 2,143 individuals, with an average score of 1.2243. The 1.1410 number in the first quartile (Q1) indicates that 25% of participants have FDG scan score below 1.1410. The second quantile number (Q2) denotes that the FDG score of 50% of the population (median value) is less than 1.2456, who enrolled for the trials. Similarly, third quartile or Q3 shows, 75% individuals are having scan score less than 1.3177. This reported analysis is as per the number of observations present in the dataset.

*Table 4.22 Descriptive Statistics for variable FDG*

| FDG | | | | | | | |
|---|---|---|---|---|---|---|---|
| **Count** | **Mean Value** | **Standard Deviation** | **Minimum value** | **25% (Q1)** | **50% (Q2)** | **75% (Q3)** | **Maximum Value** |
| 2143 | 1.2243 | 0.1357 | 0.6936 | 1.1410 | 1.2456 | 1.3177 | 1.7011 |

Below figure 4.23 shows that the FDG variable almost follows a normal distribution. Hence the reported sample distribution of patient's FDG scan measure is a good representation of the population's score those who are suffering from dementia. Also, it can be observed that most of the patients those who are suffering from this disease having scan measurement ranges between 1.2 to 1.4.

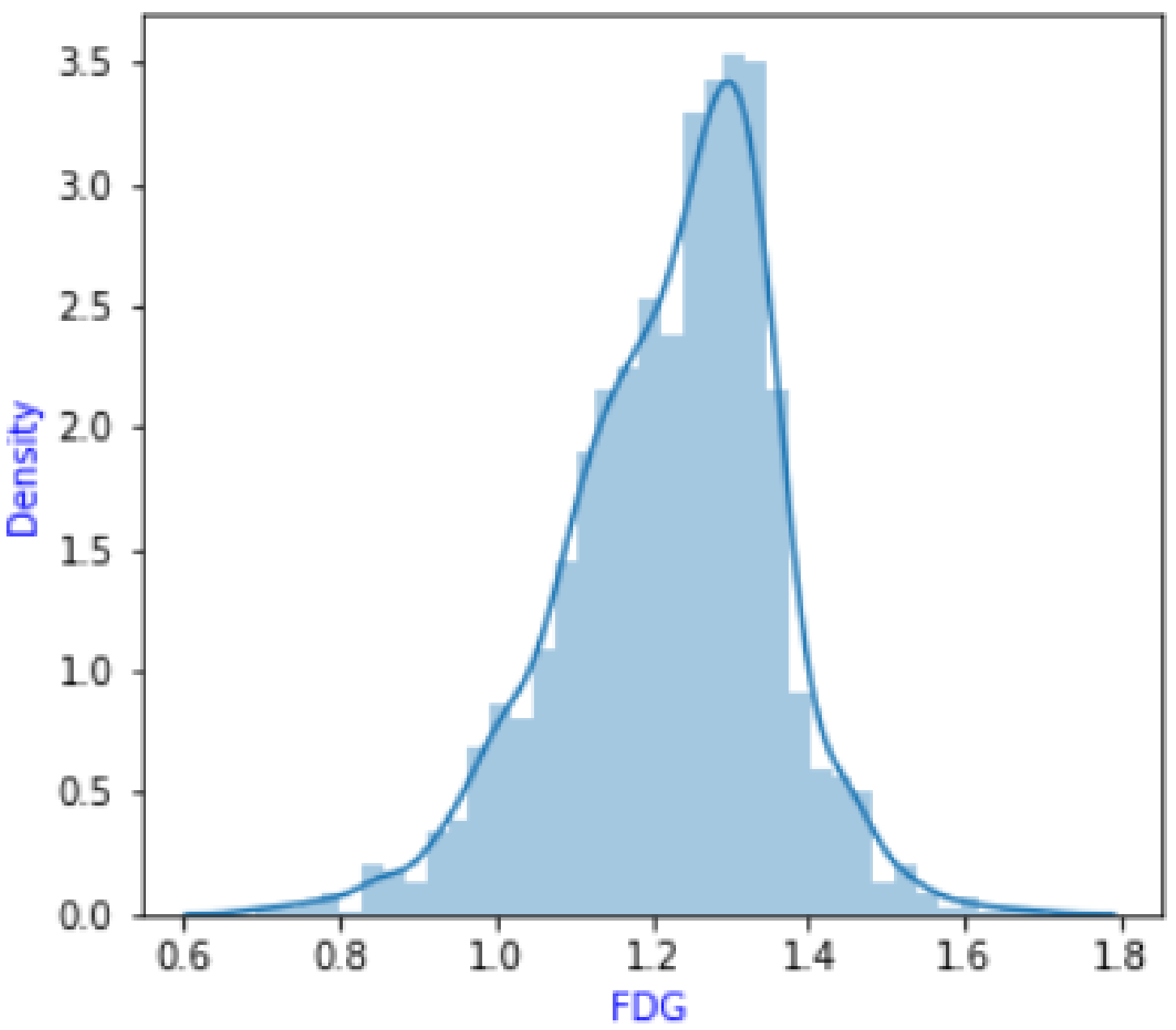

*Figure 4.23 Distribution Plot for variable FDG*

**o) Ventricles (Brain Ventricles)**

The ventricles of the brain are a network of cavities inside the brain parenchyma that communicate and are filled with cerebrospinal fluid. As per the table 4.23, for the patients who participated in ADNI trial, their ventricles size spans from 6,823.00 to 1,57,713.00 (MIN-MAX range). The study had a total of 2,143 individuals, with an average ventricles size of 40,503.5248. The 24,939.35 number in the first quartile (Q1) indicates that 25% of participants have ventricles size below 24,939.35. The second quantile number (Q2) denotes that the size of 50% of the population (median value) is less than 36,469.00, who enrolled for the trials. Similarly, third quartile or Q3 shows, 75% individuals are having ventricles size less than 50,735.90. This reported analysis is as per the number of observations present in the dataset.

*Table 4.23 Descriptive Statistics for variable Ventricles*

| Ventricles | | | | | | | |
|---|---|---|---|---|---|---|---|
| **Count** | **Mean Value** | **Standard Deviation** | **Minimum value** | **25% (Q1)** | **50% (Q2)** | **75% (Q3)** | **Maximum Value** |
| 2143 | 40503.5248 | 21654.4777 | 6823.00 | 24939.35 | 36469.00 | 50735.90 | 157713.00 |

Below figure 4.24 shows that the distribution of Ventricles variable is right skewed. For most of the participants the Ventricles size ranges between 25,000 to 50,000 (in log scale). There are very few subjects in this ADNI study data whose brain's ventricles size goes above 90,000 (in log scale).

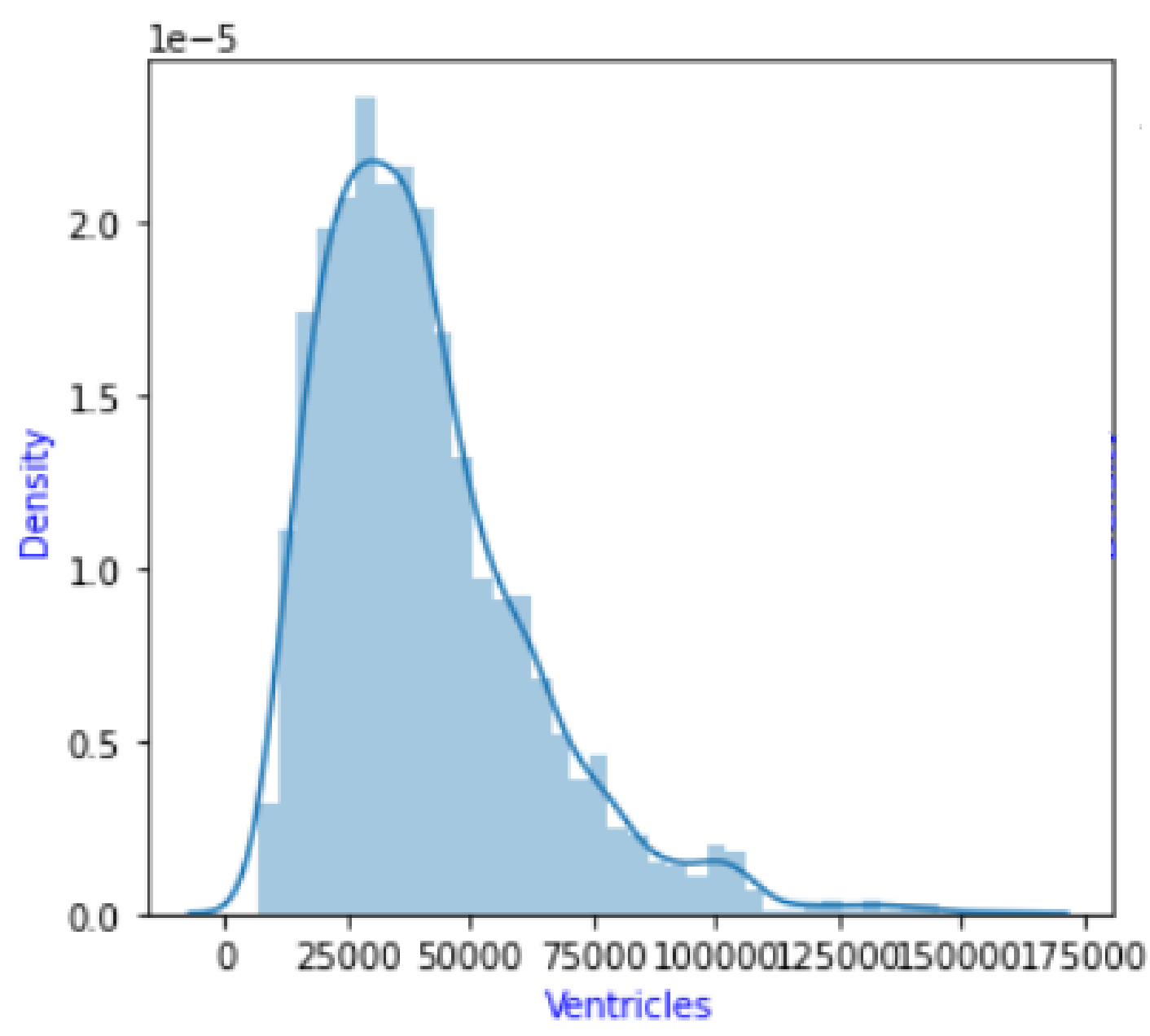

*Figure 4.24 Distribution Plot for variable Ventricles*

**p) EcogSPTotal (Everyday Cognition Study Partner - Total)**

Measurement of Everyday Cognition is a short questionnaire filled out total by a study Partner (SP) in which they score the individual's performance in numerous cognitive areas in contrast to 10 years earlier. As per the table 4.24, for the patients who participated in ADNI trial, their test score spans from 1.00 to 3.9487 (MIN-MAX range). The study had a total of 2,143 individuals, with an average score of 1.6918. The 1.2564 number in the first quartile (Q1) indicates that 25% of participants have test result below 1.2564. The second quantile number (Q2) denotes that the score of 50% of the population (median value) is less than 1.6787, who enrolled for the trials. Similarly, third quartile or Q3 shows, 75% individuals are having test less than 1.7283. This reported analysis is as per the number of observations present in the dataset.

*Table 4.24 Descriptive Statistics for variable EcogSPTotal*

| EcogSPTotal | | | | | | | |
|---|---|---|---|---|---|---|---|
| **Count** | **Mean Value** | **Standard Deviation** | **Minimum value** | **25% (Q1)** | **50% (Q2)** | **75% (Q3)** | **Maximum Value** |
| 2143 | 1.6918 | 0.5884 | 1.00 | 1.2564 | 1.6787 | 1.7283 | 3.9487 |

Below figure 4.25 shows that the distribution of EcogSPTotal variable is right skewed. For most of the participants the total Ecog test score ranges between 1.7 to 1.9. There are very few subjects in this ADNI study data whose score goes above 2.0.

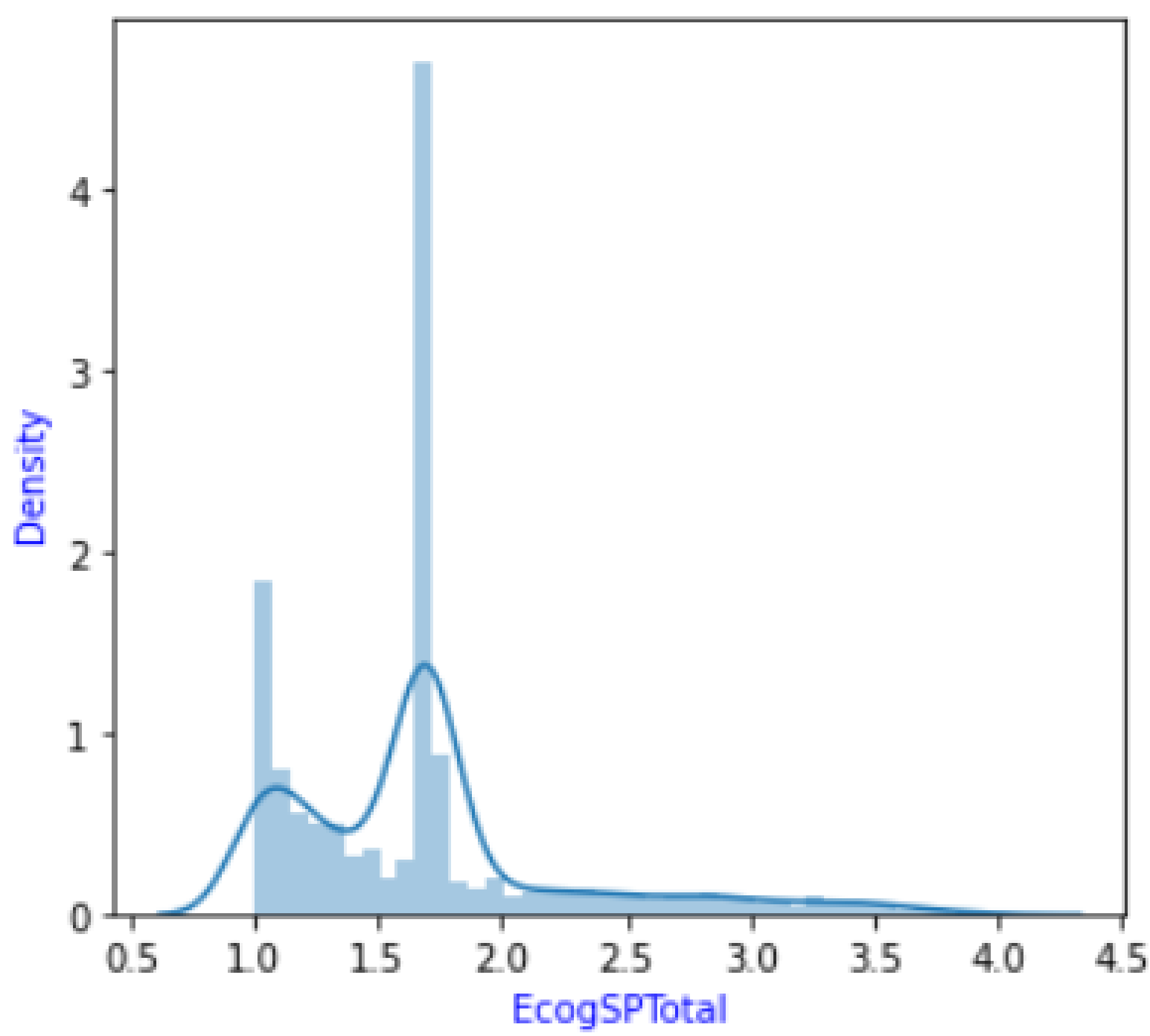

*Figure 4.25 Distribution Plot for variable EcogSPTotal*

### q) PTGENDER (Participant's Gender)

As it can be observed from below table 4.25, a total of 2,143 individuals have been participated for Alzheimer's Disease Neuroimaging Initiative trials. Among 2,143 patients, 1,144 patients or 53.38% of them are male patients, while 999 participants or 46.62% of them are female.

*Table 4.25 Frequency and Percentage analysis for variable PTGENDER*

| PTGENDER | | |
|---|---|---|
| | **Frequency/Count** | **Percentage** |
| Male | 1144 | 53.38 |
| Female | 999 | 46.62 |
| **Total** | 2143 | 100.00 |

Figure 4.26 displays the percentage of male and female patients have taken part in the ADNI study and were suffering from dementia. It is noted that in the overall number of observations, there are more male patients than female patients.

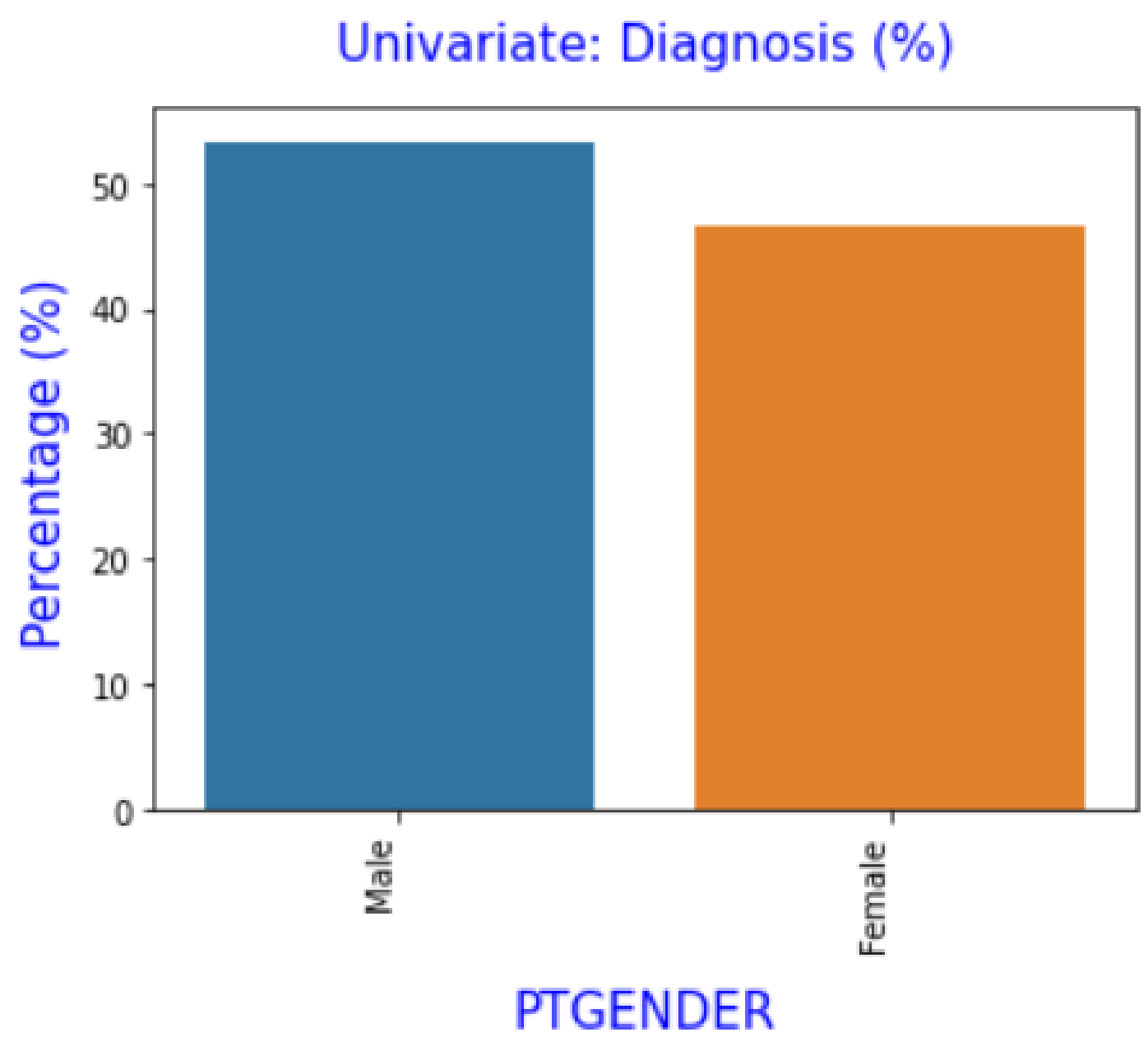

*Figure 4.26 Percentage Plot for variable PTGENDER*

**r) PTETHCAT (Participant's Ethnicity)**

As it can be observed from below table 4.26, a total of 2,143 individuals have been participated for Alzheimer's Disease Neuroimaging Initiative trials. Among 2,143 patients, 2,031 patients or 94.77% of them are belonged to Not Hisp/Latino, while 100 participants or 4.67% of them are member of Hisp/Latino. On the other hand, 12 participants or 0.56% are from unknown ethnicity.

*Table 4.26 Frequency and Percentage analysis for variable PTETHCAT*

| PTETHCAT | | |
|---|---|---|
| | **Frequency/Count** | **Percentage** |
| Not Hisp/Latino | 2031 | 94.77 |
| Hisp/Latino | 100 | 4.67 |
| Unknown | 12 | 0.56 |
| **Total** | 2143 | 100.00 |

Figure 4.27 displays the percentage of patients with different ethnicity have taken part in the ADNI study and were suffering from dementia. It is noted that in the overall number of observations, there are more patients who are part of Not Hisp/Latino ethnic group than other categories.

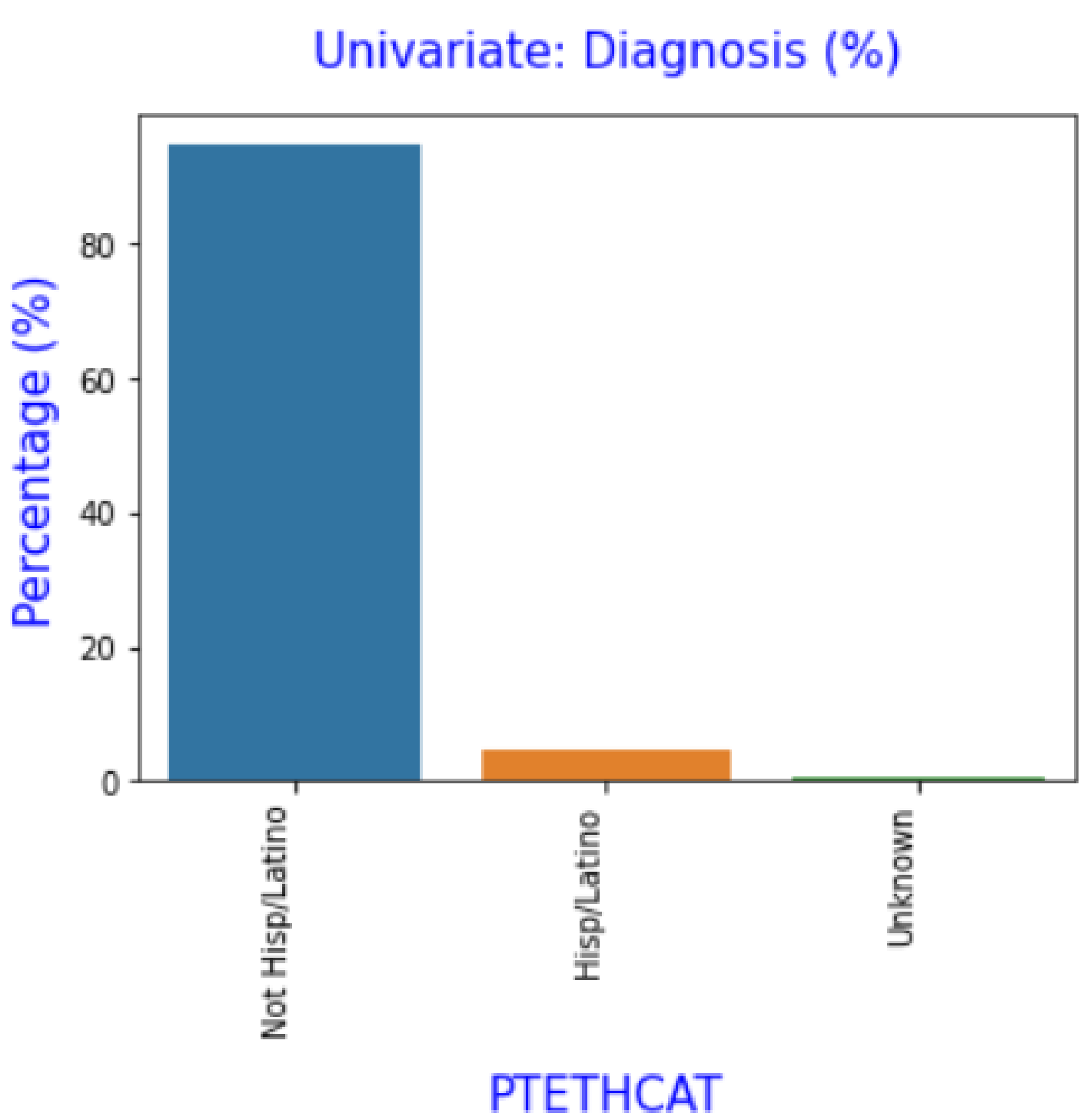

*Figure 4.27 Percentage Plot for variable PTETHCAT*

**s) PTRACCAT (Participant's Race)**

As it can be observed from below table 4.27, a total of 2,143 individuals have been participated for Alzheimer's Disease Neuroimaging Initiative trials. Among 2,143 patients, 1,922 patients or 89.69% of them are belonged to White race, while 132 participants or 6.16% of them are member of Black race. On the other hand, 47 participants or 2.19% are from Asian race, another 24 subjects or 1.12% are belonged to more than one race. 11 people or 0.51% people's race are unknown. There are 5 person who are part of Am Indian/Alaskan race group, rest 2 participants or 0.09% of them are of Hawaiian/Other PI race.

*Table 4.27 Frequency and Percentage analysis for variable PTRACCAT*

| PTRACCAT | | |
|---|---|---|
| | **Frequency/Count** | **Percentage** |
| White | 1922 | 89.69 |
| Black | 132 | 6.16 |
| Asian | 47 | 2.19 |
| More than one | 24 | 1.12 |
| Unknown | 11 | 0.51 |
| Am Indian/Alaskan | 5 | 0.23 |
| Hawaiian/Other PI | 2 | 0.09 |
| **Total** | 2143 | 100.00 |

Figure 4.28 displays the percentage of patients with different race have taken part in the ADNI study and were suffering from dementia. It is noticed that a majority of population is white.

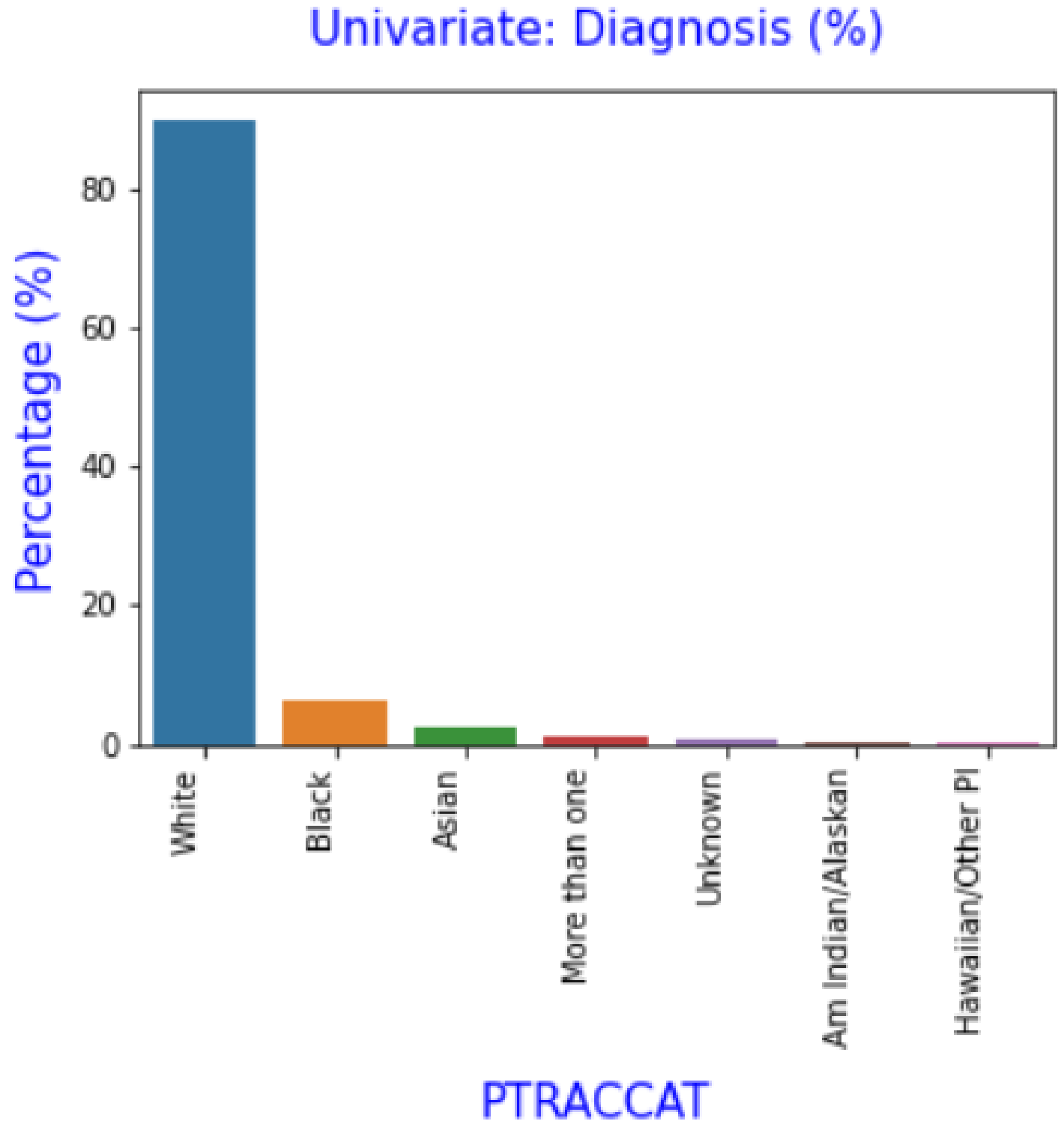

*Figure 4.28 Percentage Plot for variable PTRACCAT*

### t) PTMARRY (Participant's Marital Status)

As it can be observed from below table 4.28, a total of 2,143 individuals have been participated for Alzheimer's Disease Neuroimaging Initiative trials. Among 2,143 patients, 1,640 patients or 76.53% of them are married, while 233 participants or 10.87% of them are widowed. On the other hand, 185 participants or 8.63% are of divorced status, another 79 subjects or 3.69% are never married. Rest 6 participants or 0.28% peoples' marital status is unknown.

*Table 4.28 Frequency and Percentage analysis for variable PTMARRY*

| PTMARRY | | |
|---|---|---|
| | **Frequency/Count** | **Percentage** |
| Married | 1640 | 76.53 |
| Widowed | 233 | 10.87 |
| Divorced | 185 | 8.63 |
| Never married | 79 | 3.69 |
| Unknown | 6 | 0.28 |
| **Total** | 2143 | 100.00 |

Figure 4.29 displays the percentage of patients with different race have taken part in the ADNI study and were suffering from dementia. It is noticed that a majority of population is Married who are enrolled in this study.

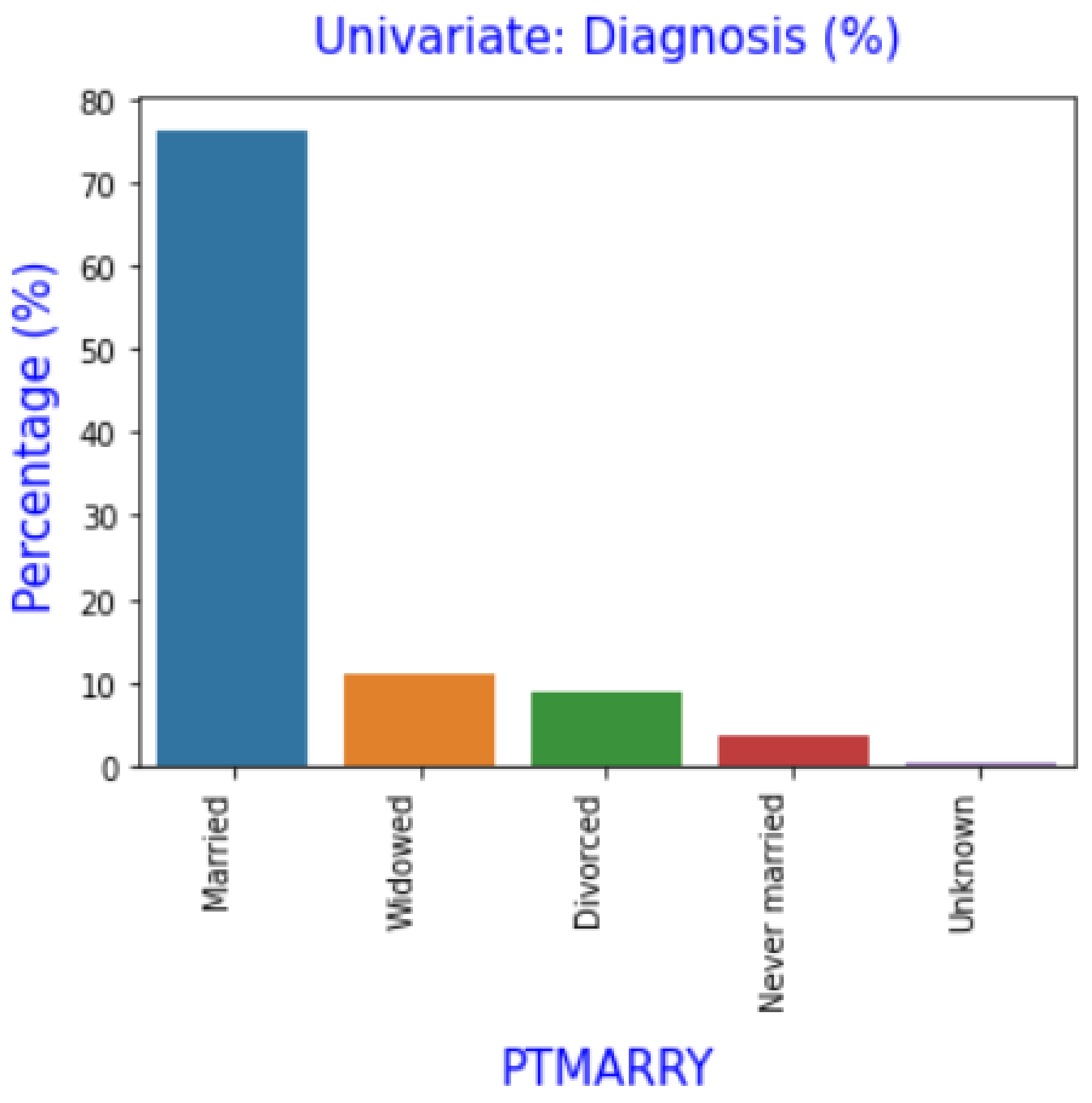

*Figure 4.29 Percentage Plot for variable PTMARRY*

**u) AGE_BIN (Participant's Age Group)**

As it can be observed from below table 4.29, a total of 2,143 individuals have been participated for Alzheimer's Disease Neuroimaging Initiative trials. Among 2,143 patients, 1,040 patients or 48.62% of them are belonged to age group 70 to 80, while 618 participants or 28.89% of them are in age group 60 to 70. On the other hand, 384 participants or 17.95% are between 80 to 90years of age, another 91 subjects or 4.25% are part of 50 to 60 age bin. Rest 6 participants or 0.28% peoples are having age more than or equal to 90. No reported participants are between age more than 10 or less than 50, which shows mostly more than 50 aged peoples are suffering from dementia.

*Table 4.29 Frequency and Percentage analysis for variable AGE_BIN*

| AGE_BIN | | |
|---|---|---|
| | **Frequency/Count** | **Percentage** |
| 70 to 80 | 1040 | 48.62 |
| 60 to 70 | 618 | 28.89 |
| 80 to 90 | 384 | 17.95 |
| 50 to 60 | 91 | 4.25 |
| 90 to Above | 6 | 0.28 |
| 40 to 50 | 0 | 0 |
| 30 to 40 | 0 | 0 |
| 20 to 30 | 0 | 0 |
| 10 to 20 | 0 | 0 |
| **Total** | 2143 | 100.00 |

Figure 4.30 displays the percentage of patients with different age group have taken part in the ADNI study and were suffering from dementia. The majority of the people that are participating in this study are between the ages of 70 and 80. The second and third most populous age groups are 60 to 70 and 80 to 90 years old, respectively. There are only a few persons in this study that are between the ages of 50 and 60.

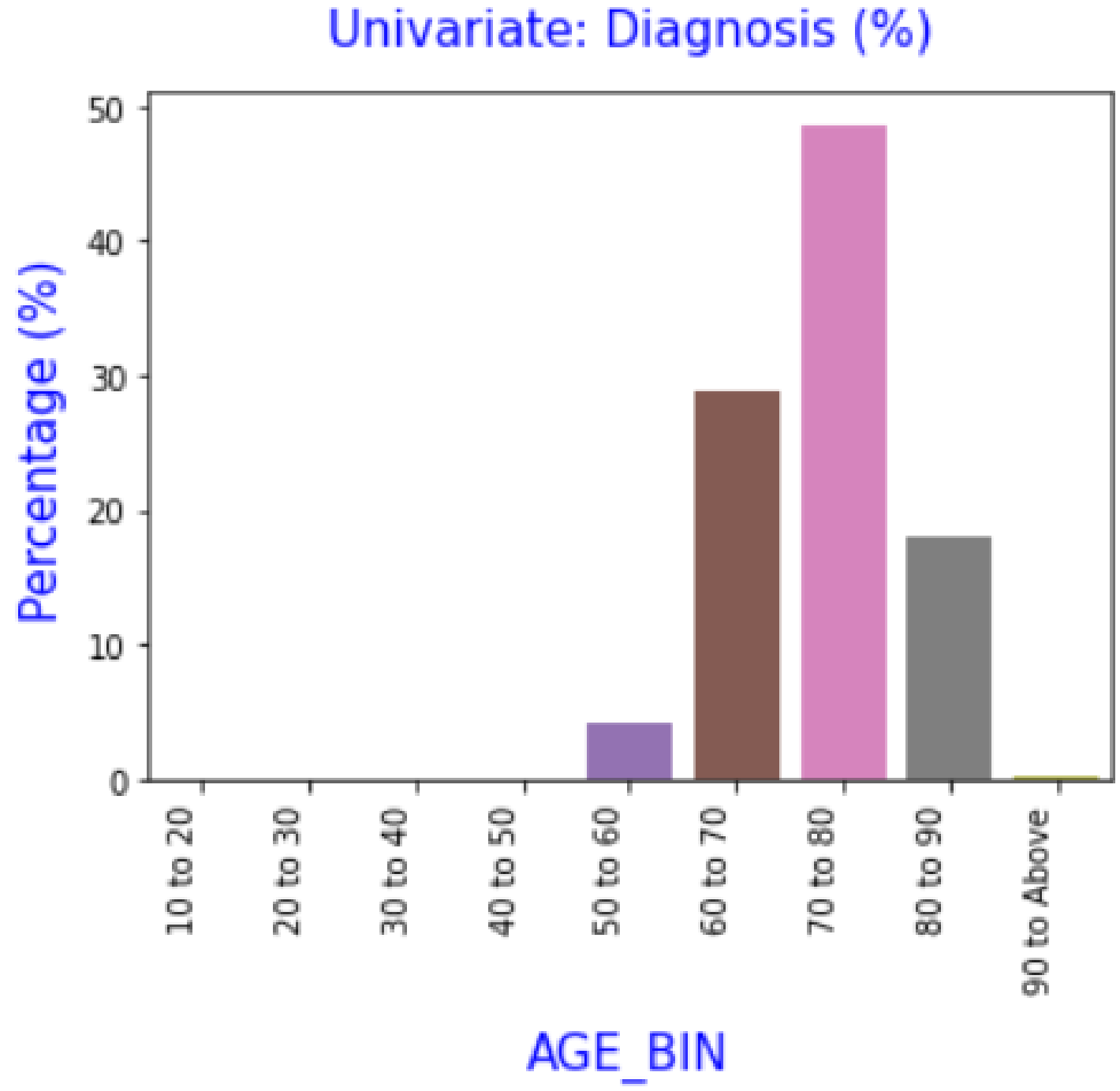

*Figure 4.30 Percentage Plot for variable AGE_BIN*

**v) MMSE_BIN (Mini Mental State Examination Score Group)**

As it can be observed from below table 4.30, a total of 2,143 individuals have been participated for Alzheimer's Disease Neuroimaging Initiative trials. Among 2,143 patients, 1,636 patients or 76.38% of them are scored between 25 to 30, while 453 participants or 21.15% of them are in range between 20 to 25. Rest 53 participants or 2.47% people are having MMSE score of intervals 15 to 20. No reported participants are scored between 0 to 15, which shows mostly more than 50 aged peoples are suffering from dementia.

*Table 4.30 Frequency and Percentage analysis for variable MMSE_BIN*

| MMSE_BIN | | |
|---|---|---|
| | **Frequency/Count** | **Percentage** |
| 25 to 30 | 1636 | 76.38 |
| 20 to 25 | 453 | 21.15 |
| 15 to 20 | 53 | 2.47 |
| 10 to 15 | 0 | 0 |
| 5 to 10 | 0 | 0 |
| 0 to 5 | 0 | 0 |
| **Total** | 2143 | 100.00 |

Figure 4.31 displays the percentage of patients with different age group have taken part in the ADNI study and were suffering from dementia. The majority of the people that are participating in this study are scored between 25 to 30. The second most populous score groups are 20 to 25. There are only a few persons in this study whose MMSE score obtained 15 to 20.

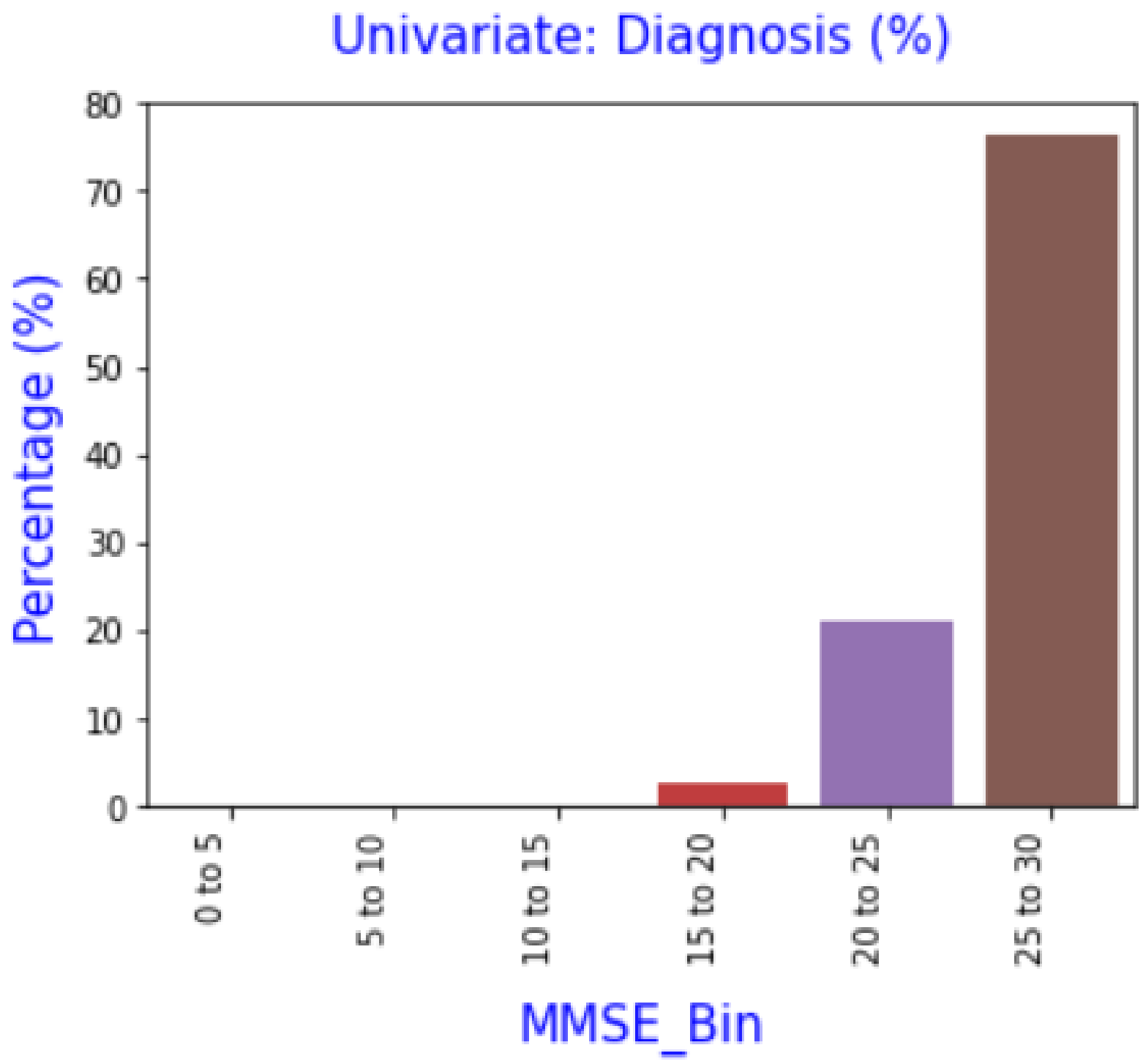

*Figure 4.31 Percentage Plot for variable MMSE_BIN*

#### 4.4.1.2 Outlier Analysis and Treatment

Outliers are extreme results that differ from other data observations; they may represent measurement variability, experimental errors, or uniqueness. To put it another way, an outlier is an observation that deviates from a sample's main pattern. Outlier analysis was performed on the numerical features in this study, and treatment was applied to a few of them after determining whether the outliers were valid or not. Feature-wise outlier inspection are discussed below in details.

**a) AGE (The Patient's Age)**

As per figure 4.32, there is one data point which lies below the lower whisker level. From initial analysis one can consider this data point as an outlier or anomaly point. But from descriptive static in table 4.15, it can be seen that the data point value is 50.40. As most of the participant's age is ranges between 70 to 80 years (shown in figure 4.30), age value 50.40 looks like an outlier. Although this value is rare in dataset, but 50.40 years is a valid age, hence should not be treated. Therefore, in this study the AGE variable is processed without handling any outlier.

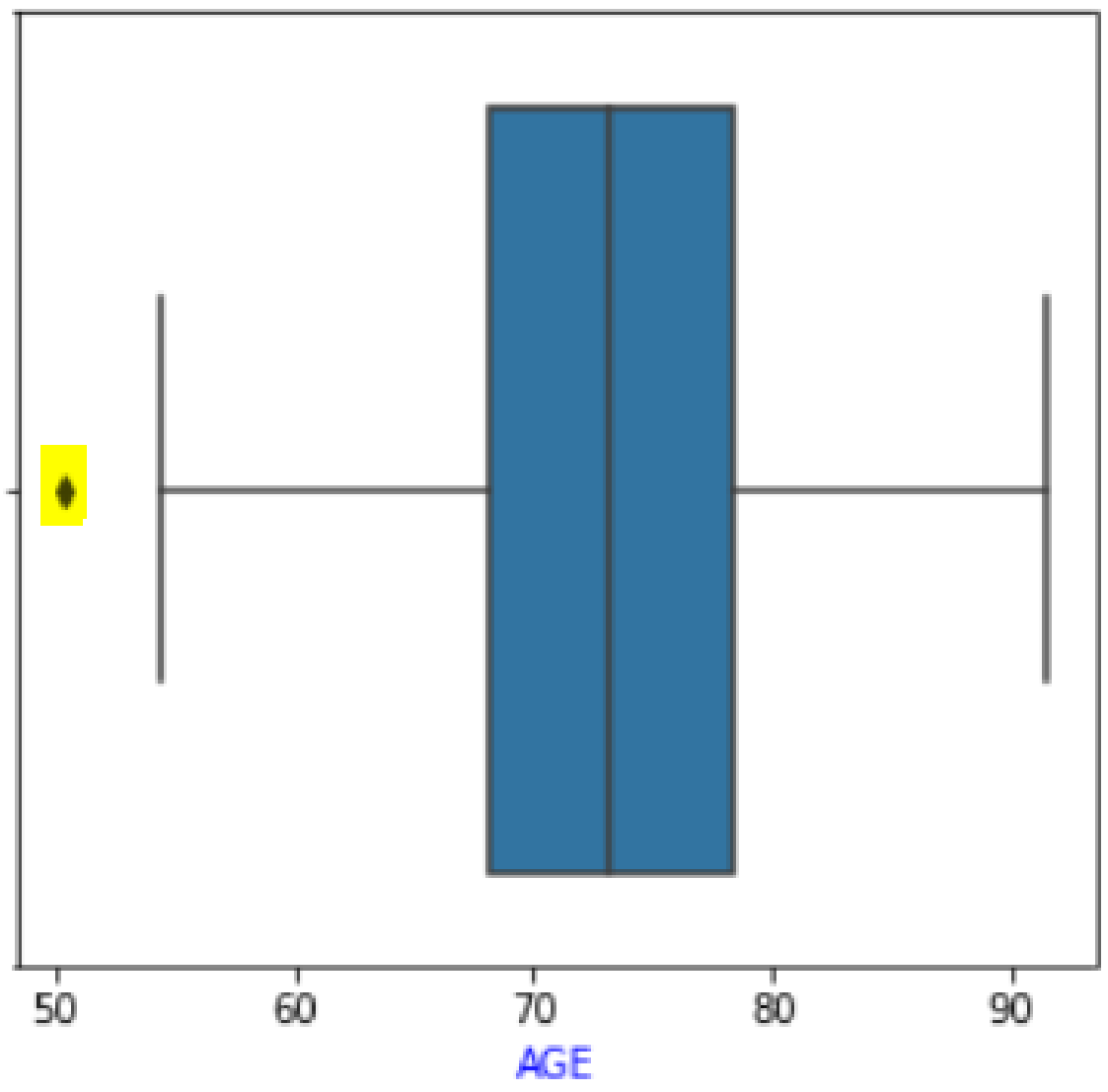

*Figure 4.32 Box Plot for variable AGE*

**b) CDRSB (Clinical Dementia Rating Sum of Boxes)**

From figure 4.33, one can see that after the upper whisker level there are few data points which are showing as outliers. As in the dataset most of the patient's CDRSB score lies between 0 to 4, shown in figure 4.10, data points beyond 4 looks like anomaly points. But according to (O'Bryant et al., 2008) Clinical Dementia Rating Sum of Boxes ranges from 0 to 18, where 0 to 0.5 rating means cognitive normal, 0.5 to 7.0 rating signifies mild cognitive impairment and 7.0 to 18.0 signifies dementia. Hence, these outliers are valid points and need not be treated.

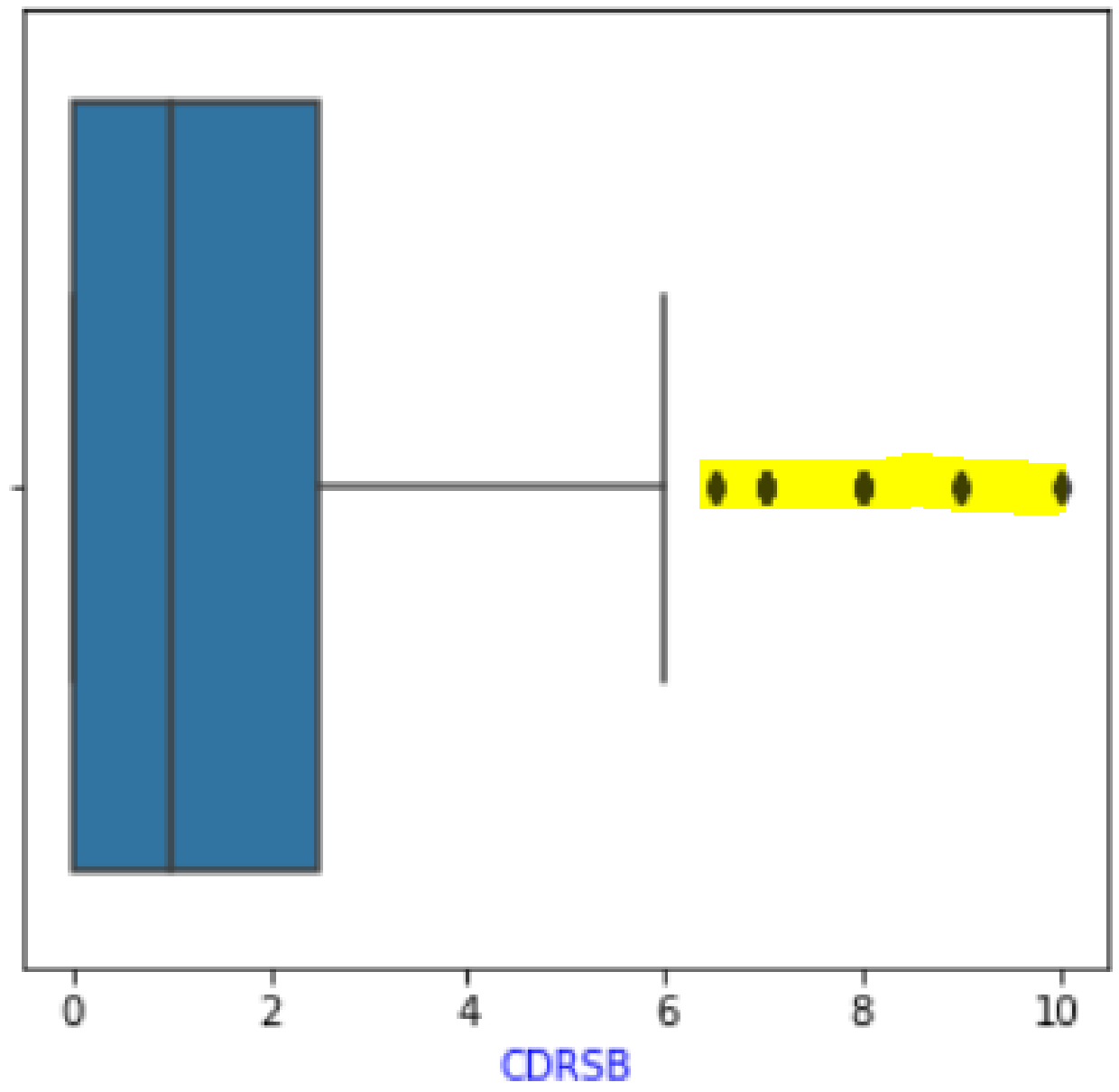

*Figure 4.33 Box Plot for variable CDRSB*

**c) ADAS13 (Alzheimer's Disease Assessment Scale - 13 items)**

From figure 4.34, one can see that after the upper whisker level there are few data points which are showing as outliers. As in the dataset most of the patient's ADAS13 score lies between 0 to 40, shown in figure 4.21, data points beyond 40 looks like anomaly points. But ADAS13 score can ranges from 0 to 85. Hence, these outliers are valid points and need not be treated.

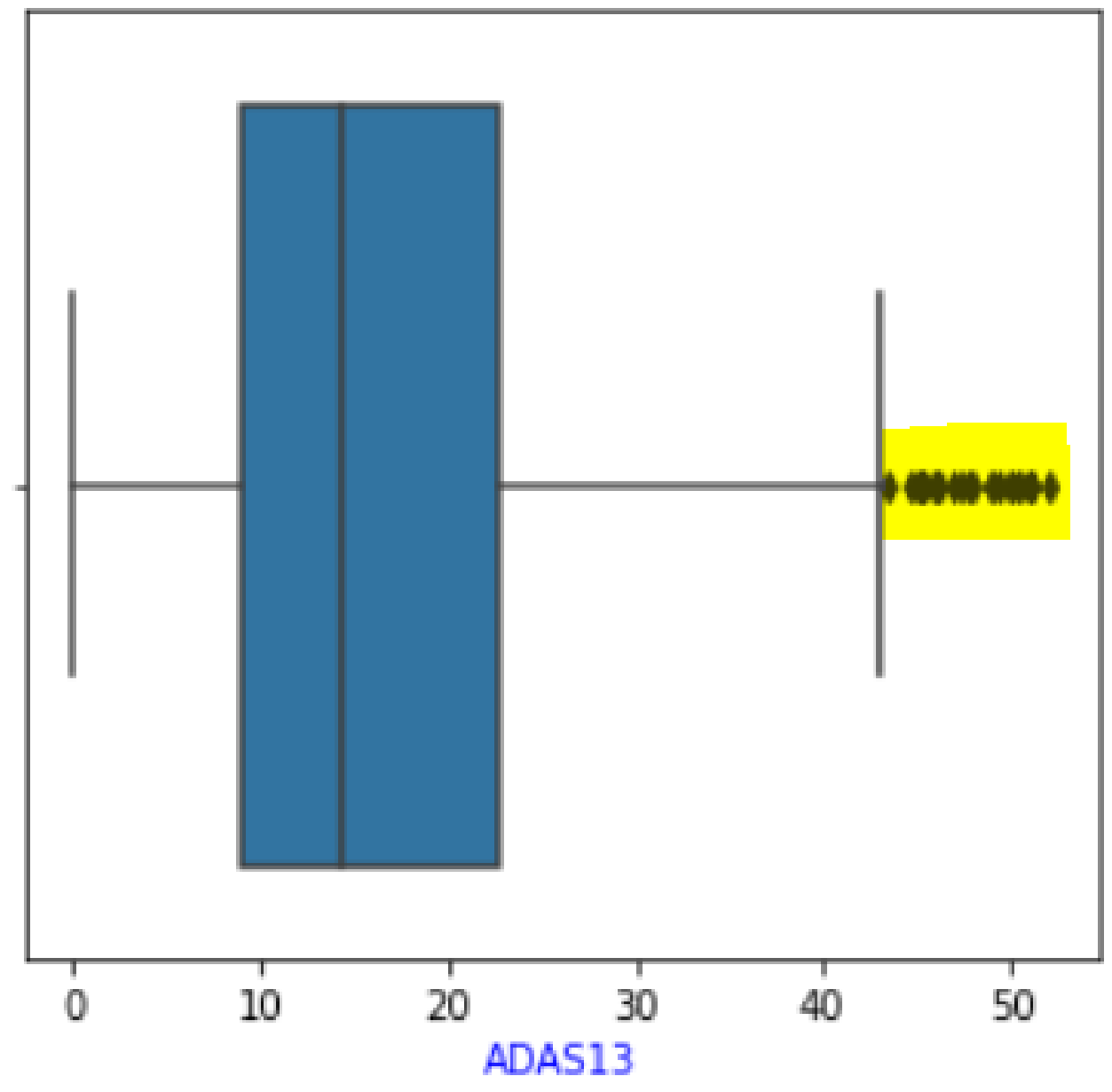

*Figure 4.34 Box Plot for variable ADAS13*

**d) MMSE (Mini-Mental State Examination)**

From figure 4.35, one can see that below the lower whisker level there are few data points which are showing as outliers. As in the dataset the patient's minimum MMSE score lies at 16.00 as shown in descriptive statistics table 4.12, and 25% of participants score resides below 26.00. Data points in between that range looks like anomaly points. But MMSE score can ranges from 0 to 30. Hence, these outliers are valid points and need not be treated.

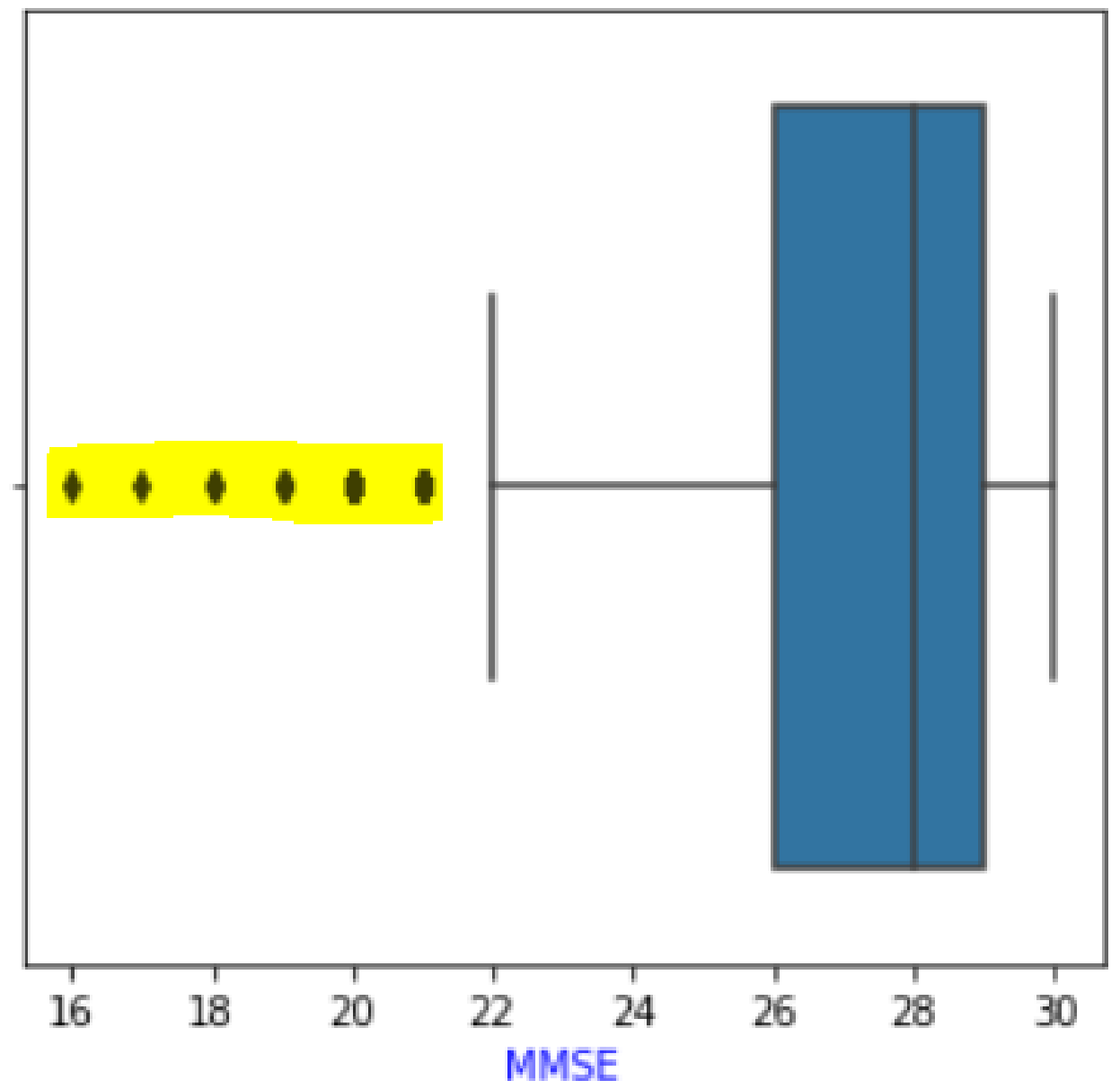

*Figure 4.35 Box Plot for variable MMSE*

**e) RAVLT_forgetting (Rey Auditory Verbal Learning Test for forgetting)**

From figure 4.36, one can see that both below the lower whisker level and above the upper whisker, there are few data points which are showing as outliers. As in the dataset the patient's minimum RAVLT_forgetting score is negative -28.00, and 5% of participants score resides below 0.00. Data points in between that range looks like anomaly points. The data points which are above the upper fence, should be considered as valid points as this test score can go up to 75. Hence, here only the data points, below the lower fence has been treated.

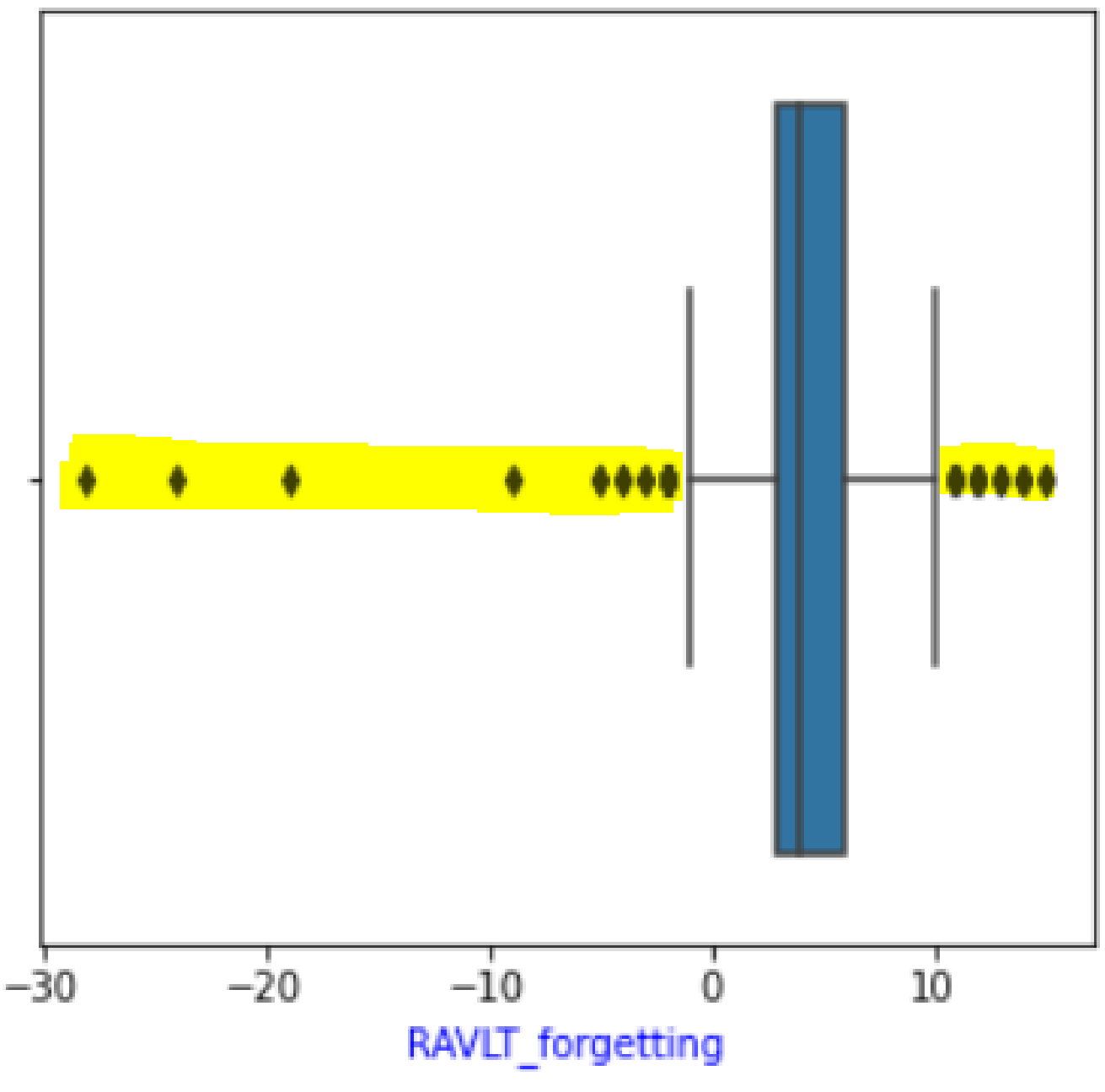

*Figure 4.36 Box Plot for variable RAVLT_forgetting*

For the outliers below the lower fence for feature RAVLT_forgetting was treated by hard capping, means all the value below 5 percentiles for this feature was capped with the value of 5 percentile itself. Figure 4.37 shows the boxplot of the feature RAVLT_forgetting after the outlier treatment below the lower fence.

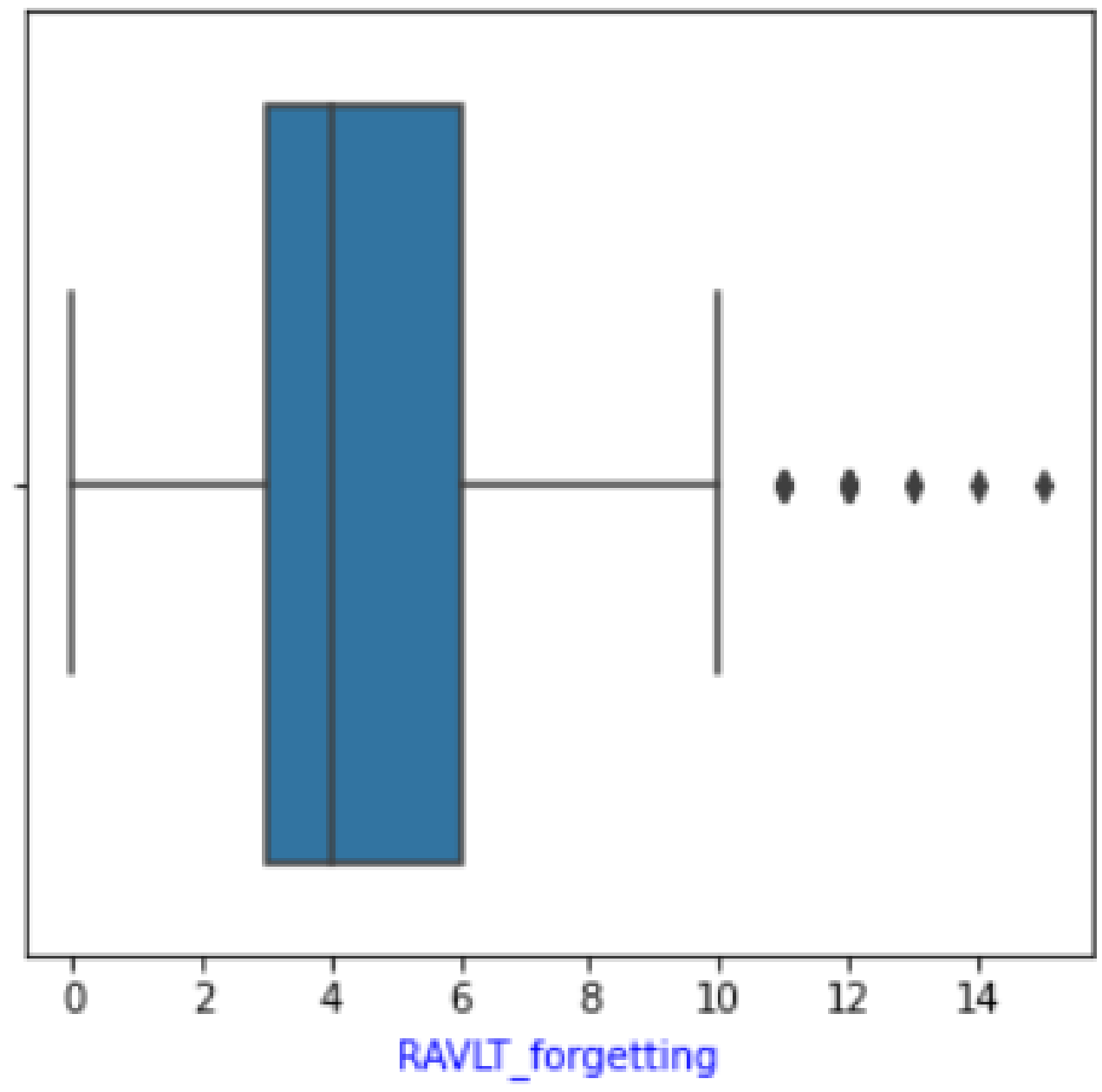

*Figure 4.37 Box Plot for variable RAVLT_forgetting post Outlier treatment*

**f) RAVLT_learning (Rey Auditory Verbal Learning Test)**

From figure 4.38, one can see that below the lower whisker level, there is one data points which are showing as outliers. As in the dataset the patient's minimum RAVLT_learning score is negative -8.00, and 25% of participants score resides below 2.00. Data points in between that range looks like anomaly points as this score can range between 0 to 75. Hence, here the negative data points, below the lower fence has been treated.

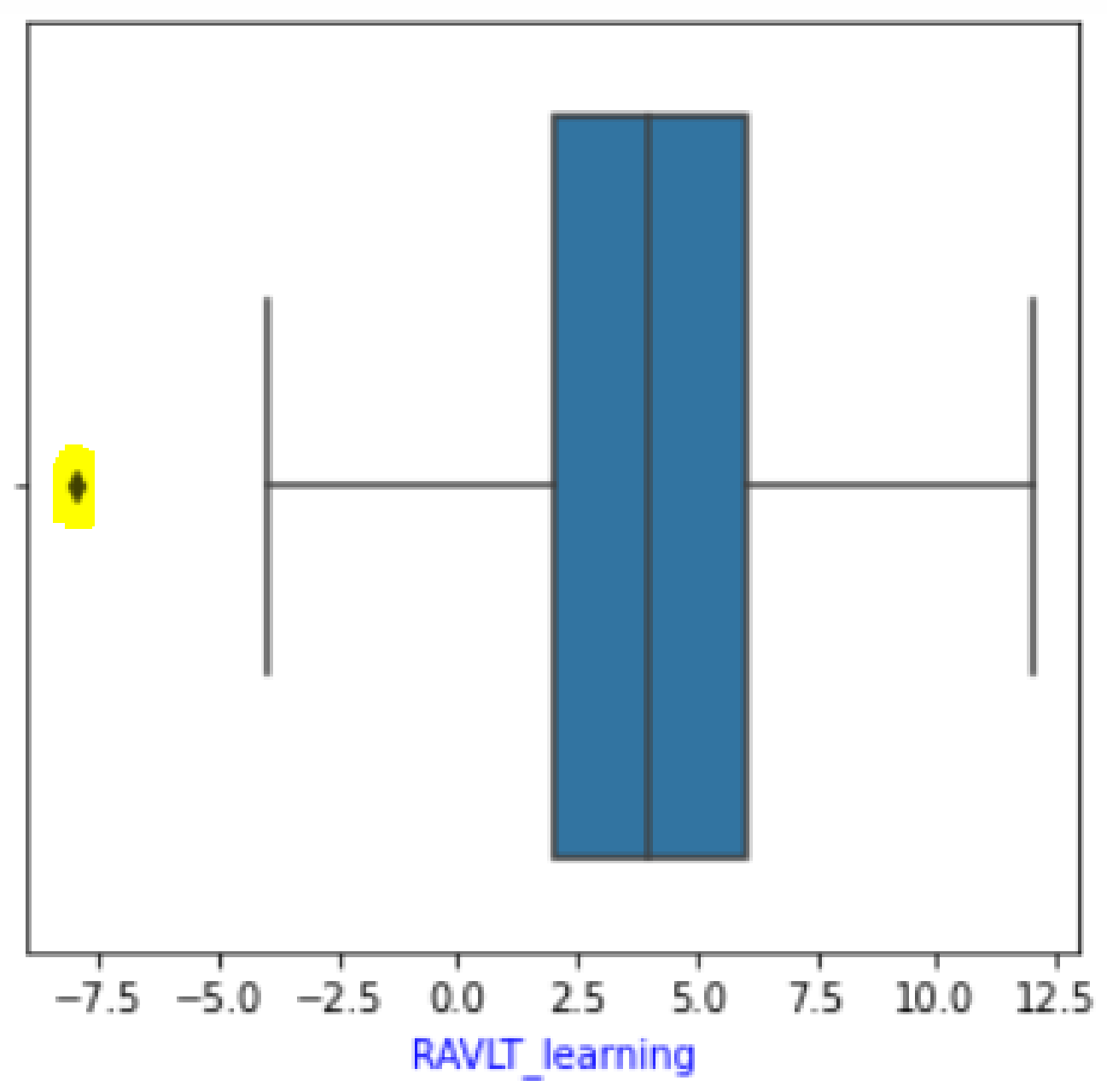

*Figure 4.38 Box Plot for variable RAVLT_learning*

For the outliers below the lower fence for feature RAVLT_learning was treated by hard capping, means all the value below 5 percentiles for this feature was capped with the value of 5 percentile itself. Figure 4.39 shows the boxplot of the feature RAVLT_learning after the outlier treatment below the lower fence.

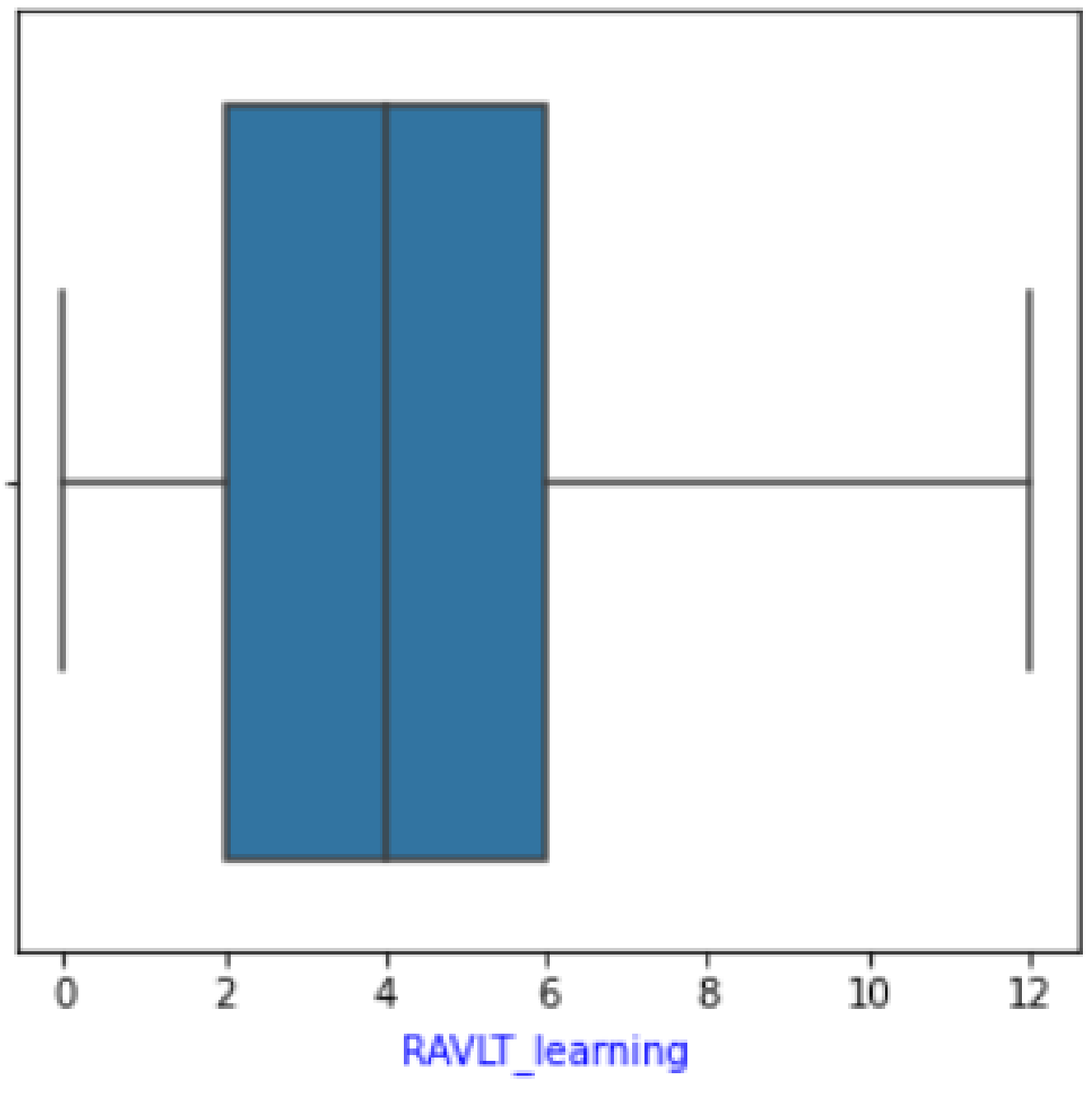

*Figure 4.39 Box Plot for variable RAVLT_learning post Outlier treatment*

### 4.4.2 Bivariate Analysis

One of the most basic types of quantitative analysis is bivariate analysis. It entails the examination of two variables in order to establish an empirical relationship between them. It is a study of the relationship between two variables. This analysis can be categorized into three sub-sections. Details of each are described below.

#### 4.4.2.1 Categorical vs Numerical Analysis

This section will go over various bivariate visualisations, such as bar plots and hist plots, for some significant independent numerical and categorical attributes. Among all the predictive indicators, this section focuses on few most important biomarkers. This analysis focuses on the target variable (DX) as a categorical feature versus some vital numerical features.

**a) CDRSB (Clinical Dementia Rating Sum of Boxes)**

As per below figure 4.40, one can see that the average Clinical Dementia Rating Sum of Boxes test score is highest (more than 4) for the people suffering from Dementia or Alzheimer's illness (Target Class 2). In contrary, among the control group (Target Class 0), average value of this test score is close to zero. This cognitive test score is showing a good separability among three classes at base line diagnosis. This variable might become an important distinguishing factor of Alzheimer's prediction.

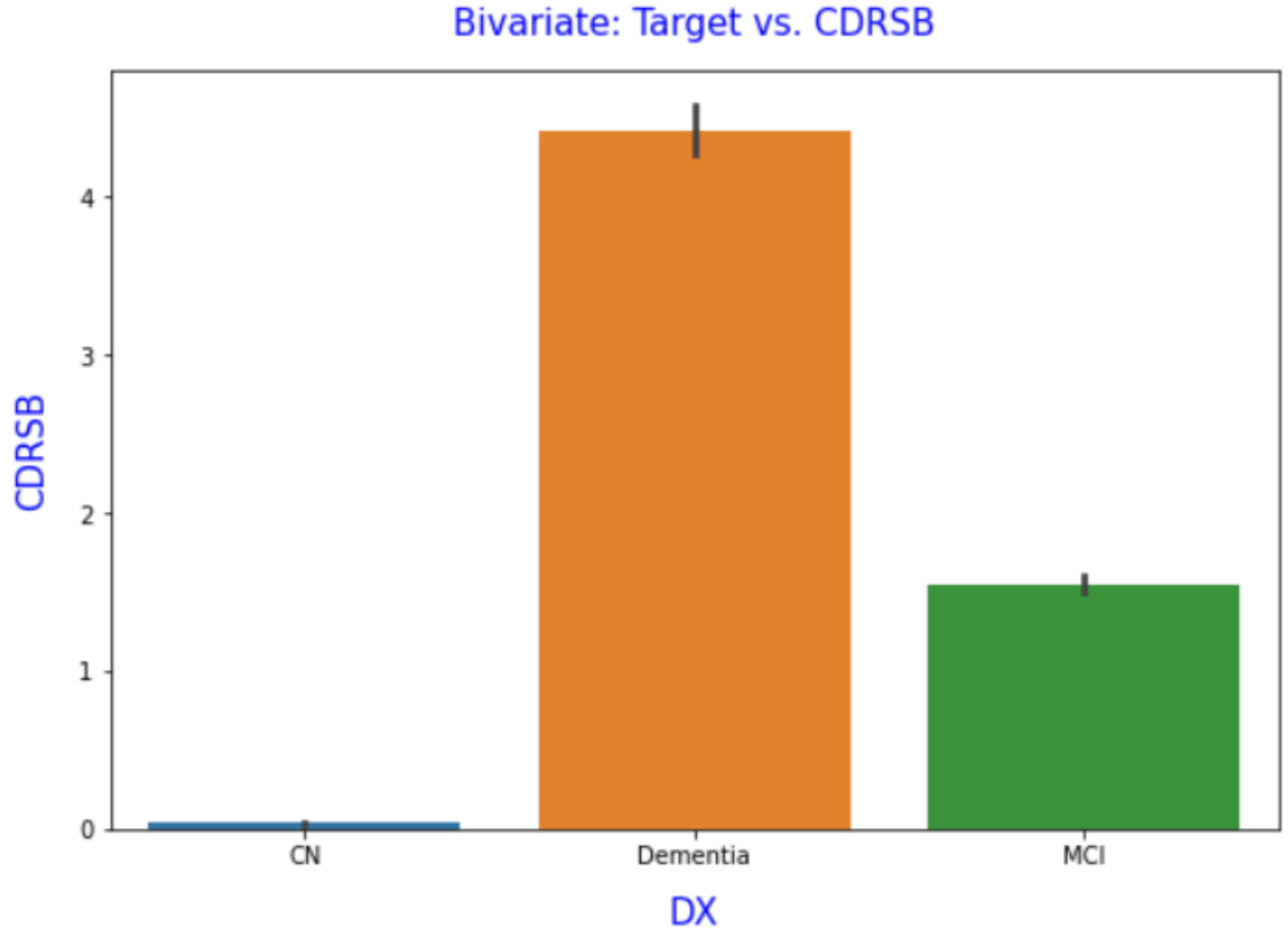

*Figure 4.40 Bar Plot between DX (Target) vs CDRSB*

From below figure 4.41, it is clearly observed that CDRSB rating for Dementia affected people are mostly ranges above 4. While rating zero is a perfect indicator of cognitive normal people. For mild affected people this score mostly spans between 1 to 3. Hence, this variable can give a clear identification of target groups.

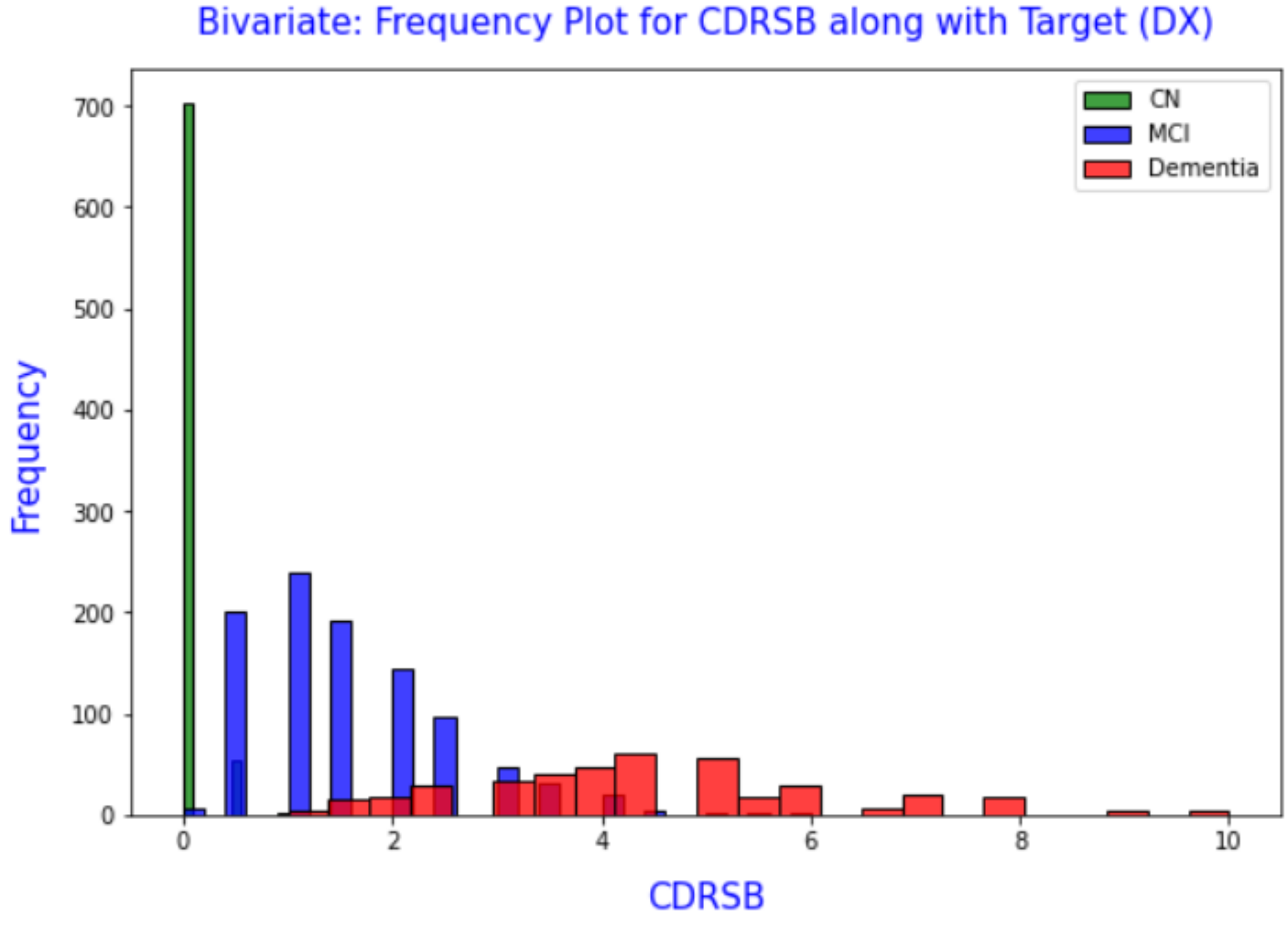

*Figure 4.41 Comparative Hist Plot between DX (Target) vs CDRSB*

**b) mPACCdigit (modified Preclinical Alzheimer's Cognitive Composite with Digit)**

As shown in figure 4.42, the average PACC digit test score is more negative (greater than -14) for participants affected by Alzheimer's disease or Dementia (Target Class 2). On the contrary, people with mild dementia or mild cognitive impairment (Target Class 1) having lesser negative score than other classes. While Cognitive Normal people are scored approximately zero. This cognitive test score is showing a good separability among three classes at base line diagnosis. This variable might become an important distinguishing factor of Alzheimer's prediction.

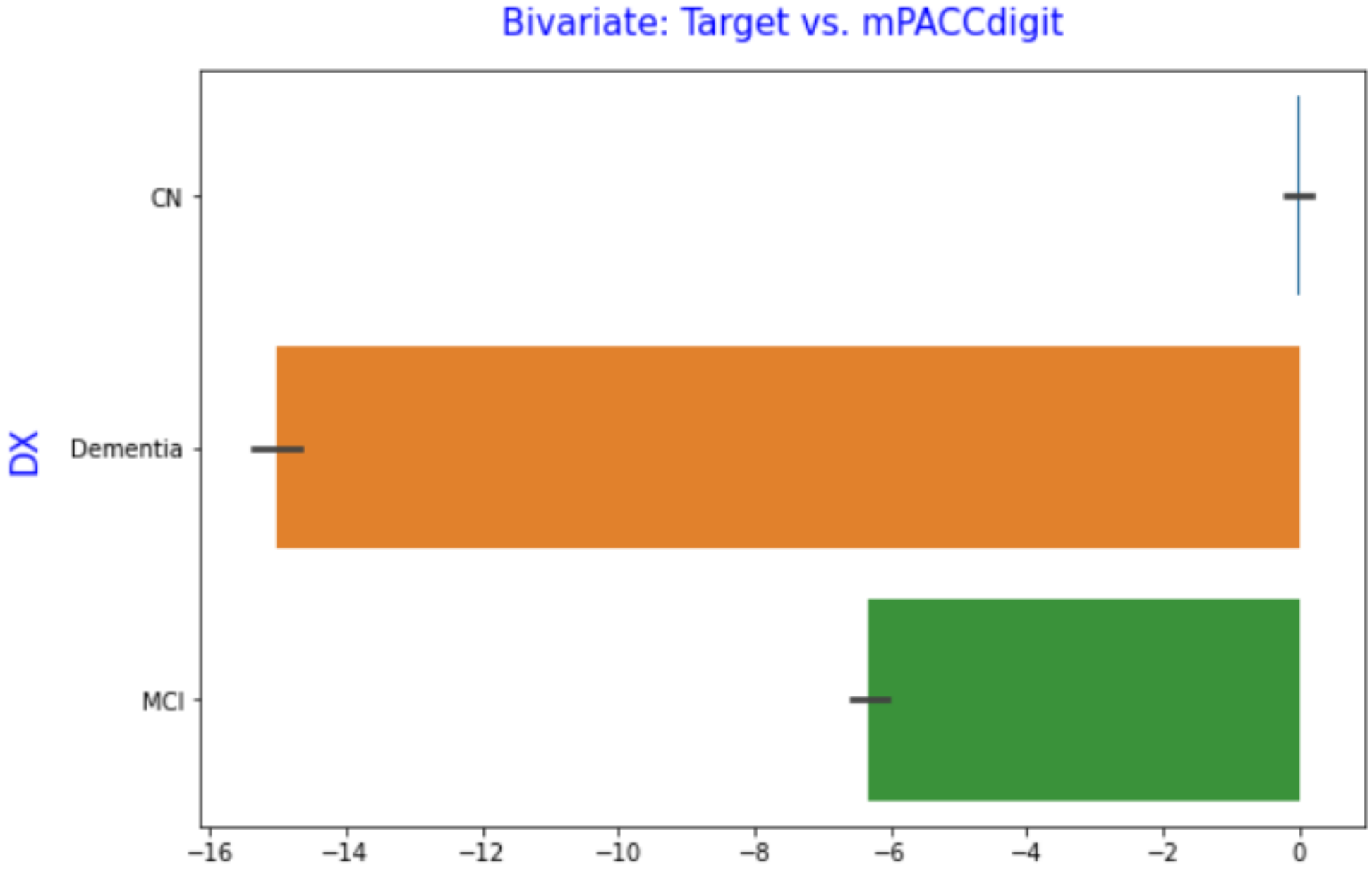

*Figure 4.42 Bar Plot between DX (Target) vs mPACCdigit*

From below figure 4.43, it is clearly observed that PACC digit test rating for Dementia affected people are mostly ranges between -25 to -5. While -15 to -5 is an overlapping between AD and MCI. For healthy people this score mostly spans above 1. Hence, this variable can give a clear identification of target groups.

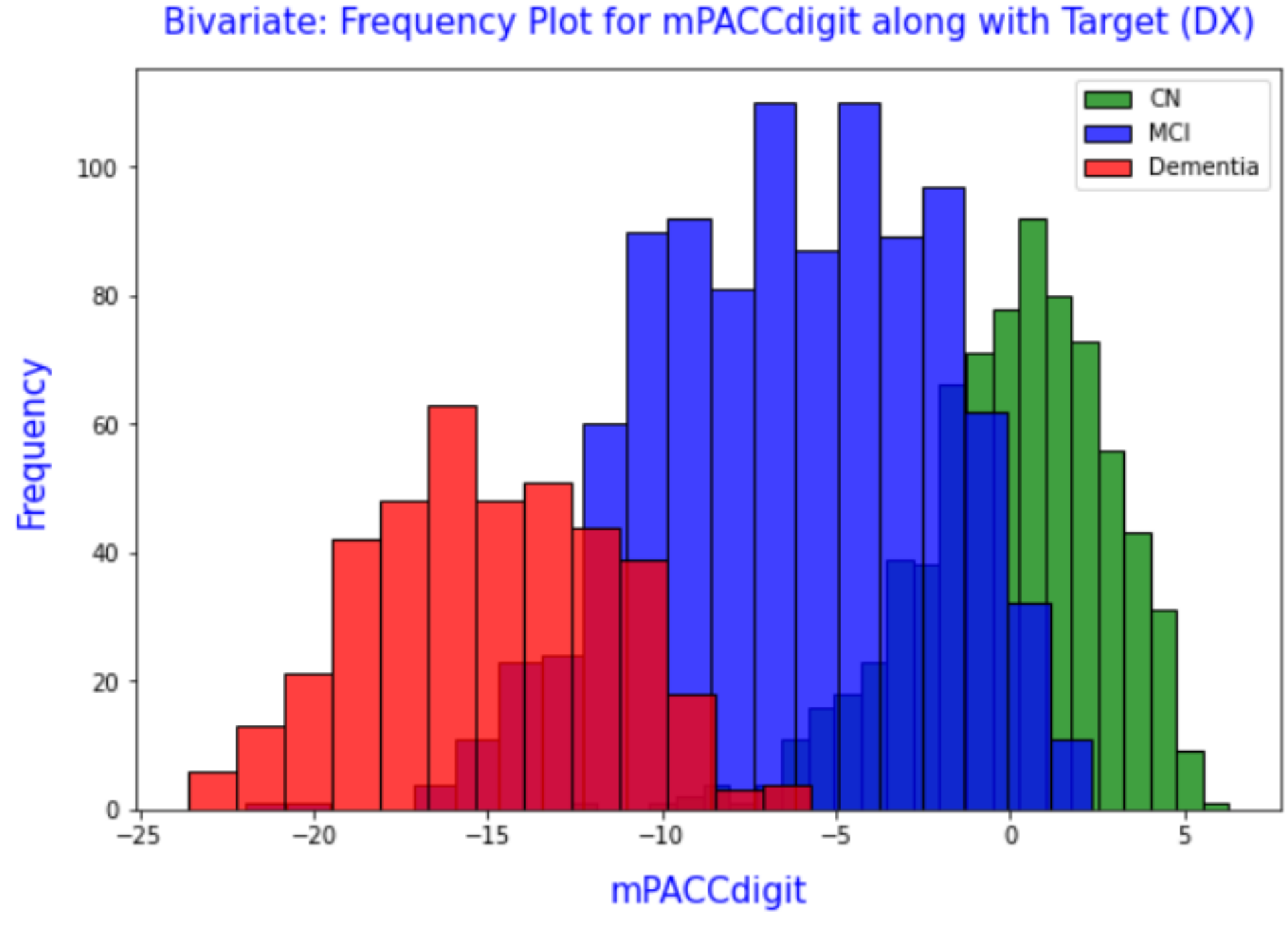

*Figure 4.43 Comparative Hist Plot between DX (Target) vs mPACCdigit*

**c) AGE (Participant's Age)**

As shown in figure 4.44, the average age of the patients is almost same among all three classes. While some people with average age more than 70 are suffering from cognitive impairment or dementia, others in the same age group are part of the healthy population. This age variable is not showing a good separability among three classes at base line diagnosis. Therefore, this variable might not be an important distinguishing factor of Alzheimer's prediction.

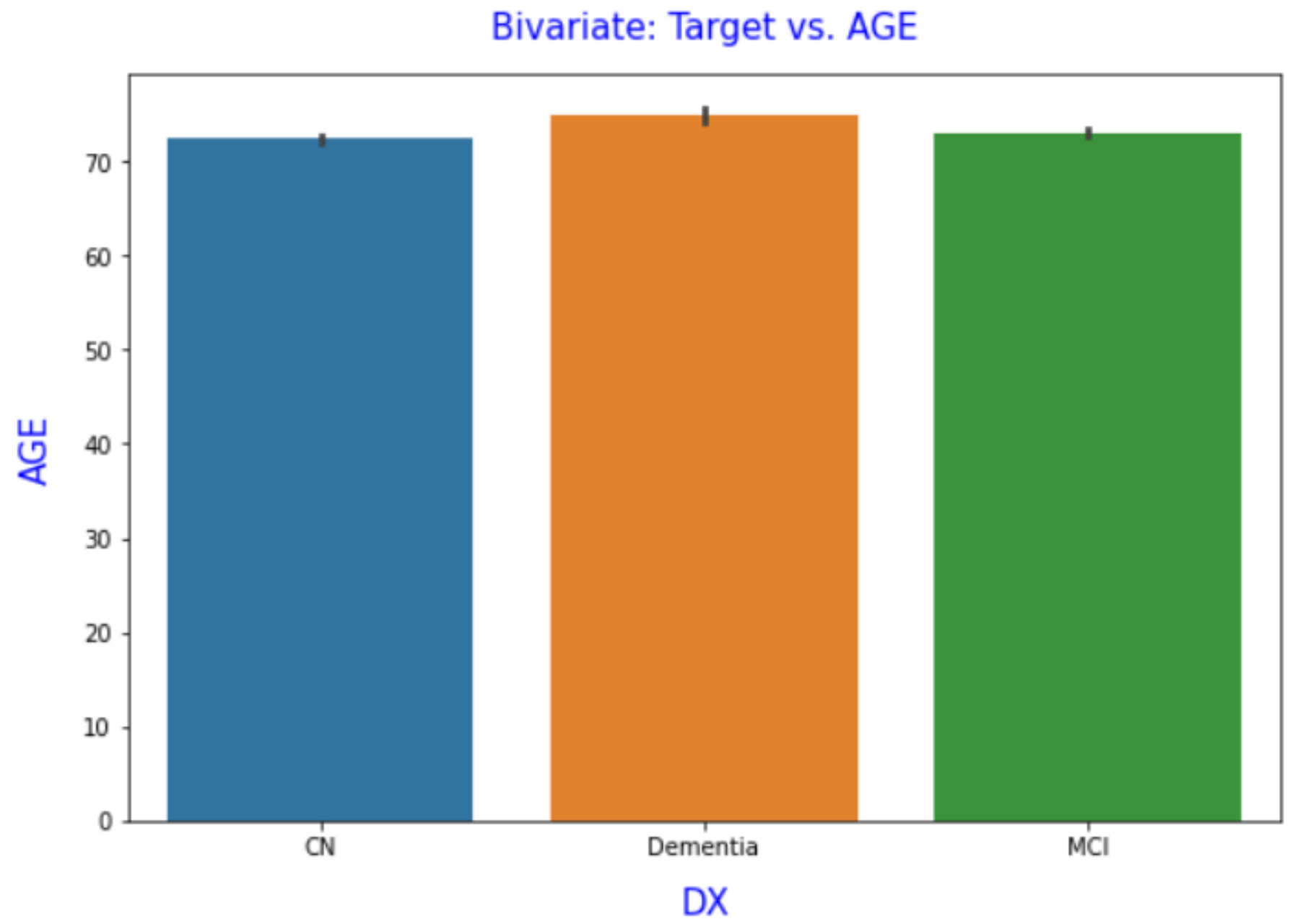

*Figure 4.44 Bar Plot between DX (Target) vs AGE*

**d) RAVLT_immediate (Rey Auditory Verbal Learning Test for immediate response)**

As shown in figure 4.45, the average Rey Auditory Verbal Learning test score is more (greater than 40) for participants who are cognitive normal (Target Class 0). On the contrary, for people with Alzheimer's disease or Dementia (Target Class 2), the test score is lowest among all other classes. This cognitive test score is showing a moderate separability among three classes at base line diagnosis. This variable might become an important distinguishing factor of Alzheimer's prediction.

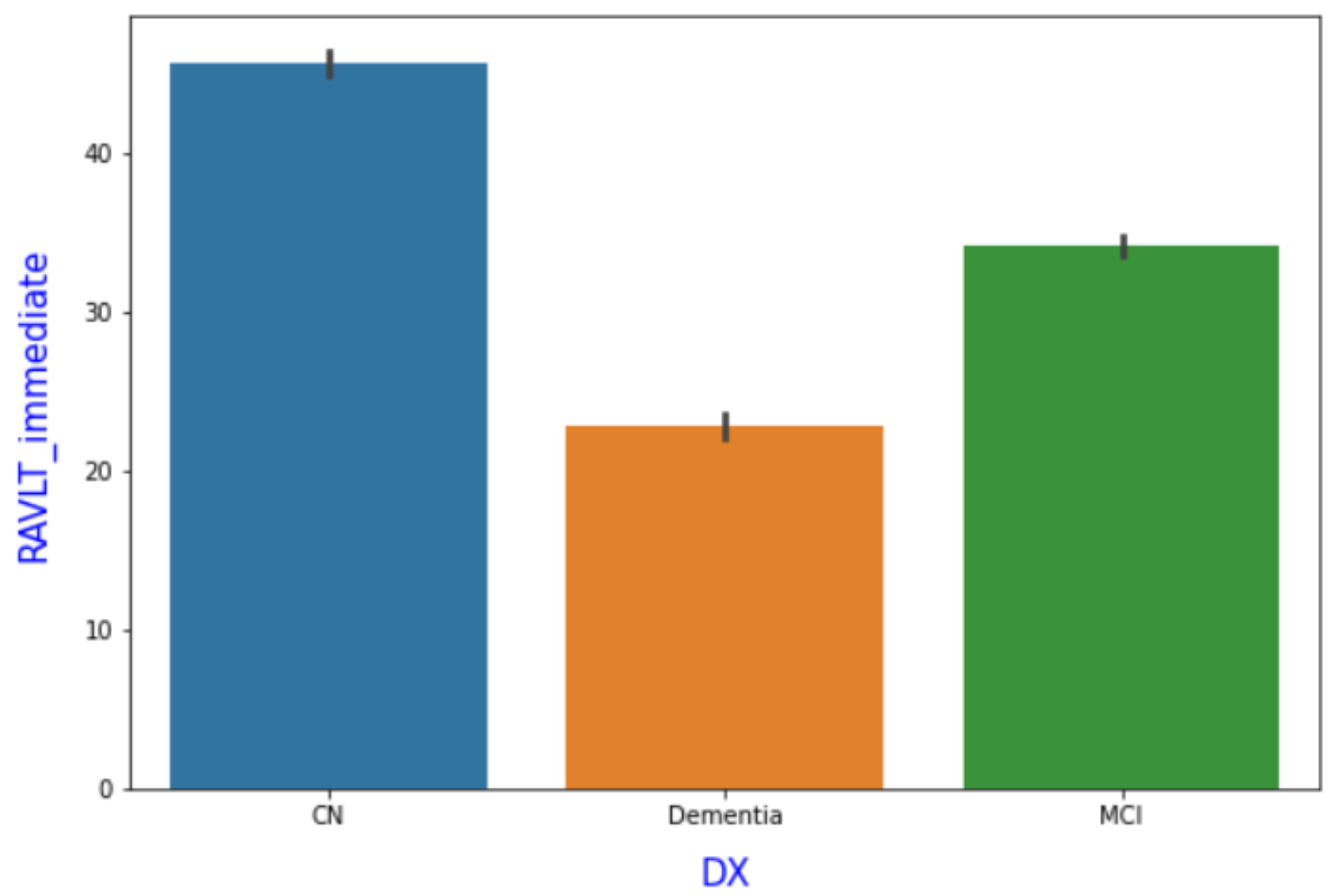

*Figure 4.45 Bar Plot between DX (Target) vs RAVLT_immediate*

From below figure 4.46, it is clearly observed that RAVLT_immediate test rating for Dementia affected people are mostly ranges between 0 to 40. While 15 to 40 is an overlapping between AD and MCI. For healthy people this score mostly spans above 60. Hence, this variable can give a clear identification of target groups.

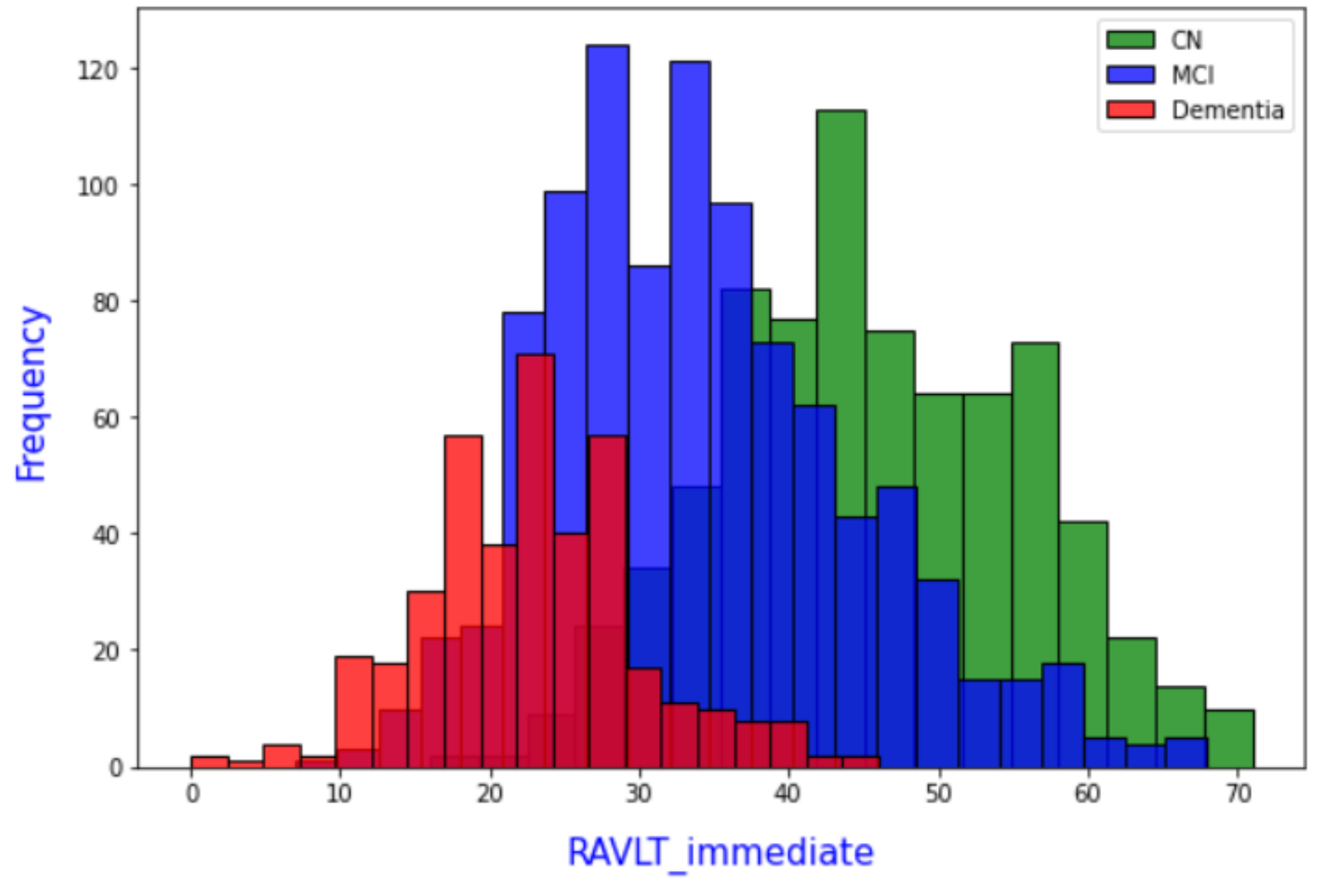

*Figure 4.46 Comparative Hist Plot between DX (Target) vs RAVLT_immediate*

**e) WholeBrain (Entire brain volume)**

As shown in figure 4.47, the average brain volume is more (greater than $1\times10^6$) for participants who are cognitive normal (Target Class 0) as well as who are suffering from mild cognitive impairment (Target Class 1). On the contrary, for people suffering from Alzheimer's disease or Dementia (Target Class 2), the whole brain volume is less than $1\times10^6$. This variable might not become an important distinguishing factor of Alzheimer's prediction as the values are slightly different among the three classes.

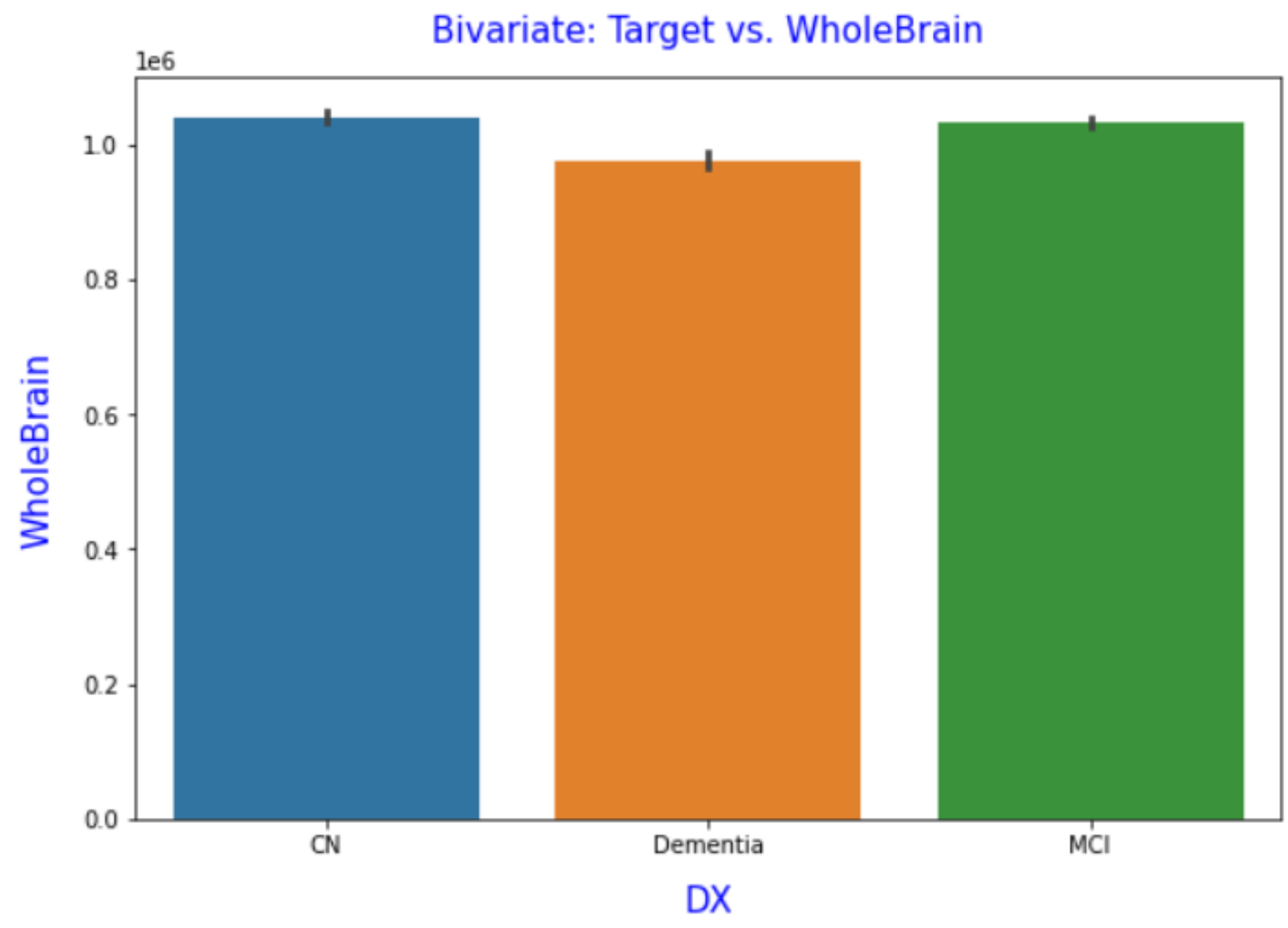

*Figure 4.47 Bar Plot between DX (Target) vs WholeBrain*

**f) ADAS13 (Alzheimer's Disease Assessment Scale - 13 items)**

As shown in figure 4.48, the average score of this cognitive and non-cognitive symptoms assessment test is high (between 25 to 30) for participants who are suffering from dementia (Target Class 2). On the contrary, people who are part of healthy population are having less score for this cognitive test (below 10). This variable might become an important distinguishing factor of Alzheimer's prediction as the values lie in different range among the three classes.

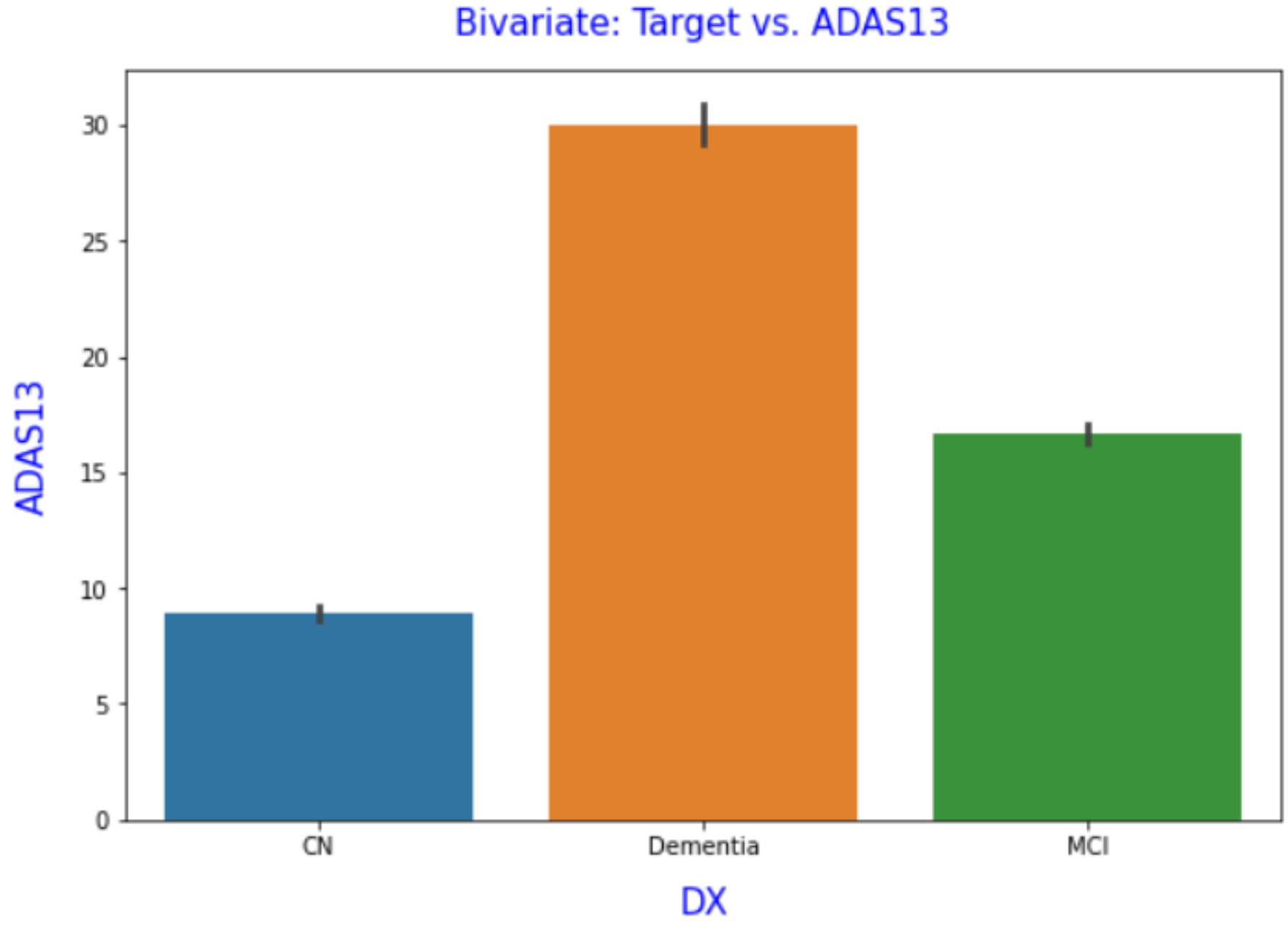

*Figure 4.48 Bar Plot between DX (Target) vs ADAS13*

From below figure 4.49, it is clearly observed that ADAS test rating for Dementia affected people are mostly ranges between 15 to 50. While 15 to 40 is an overlapping between AD and MCI. For healthy people this score mostly spans between 0 to 20. Hence, this variable can give a clear identification of target groups CN vs AD.

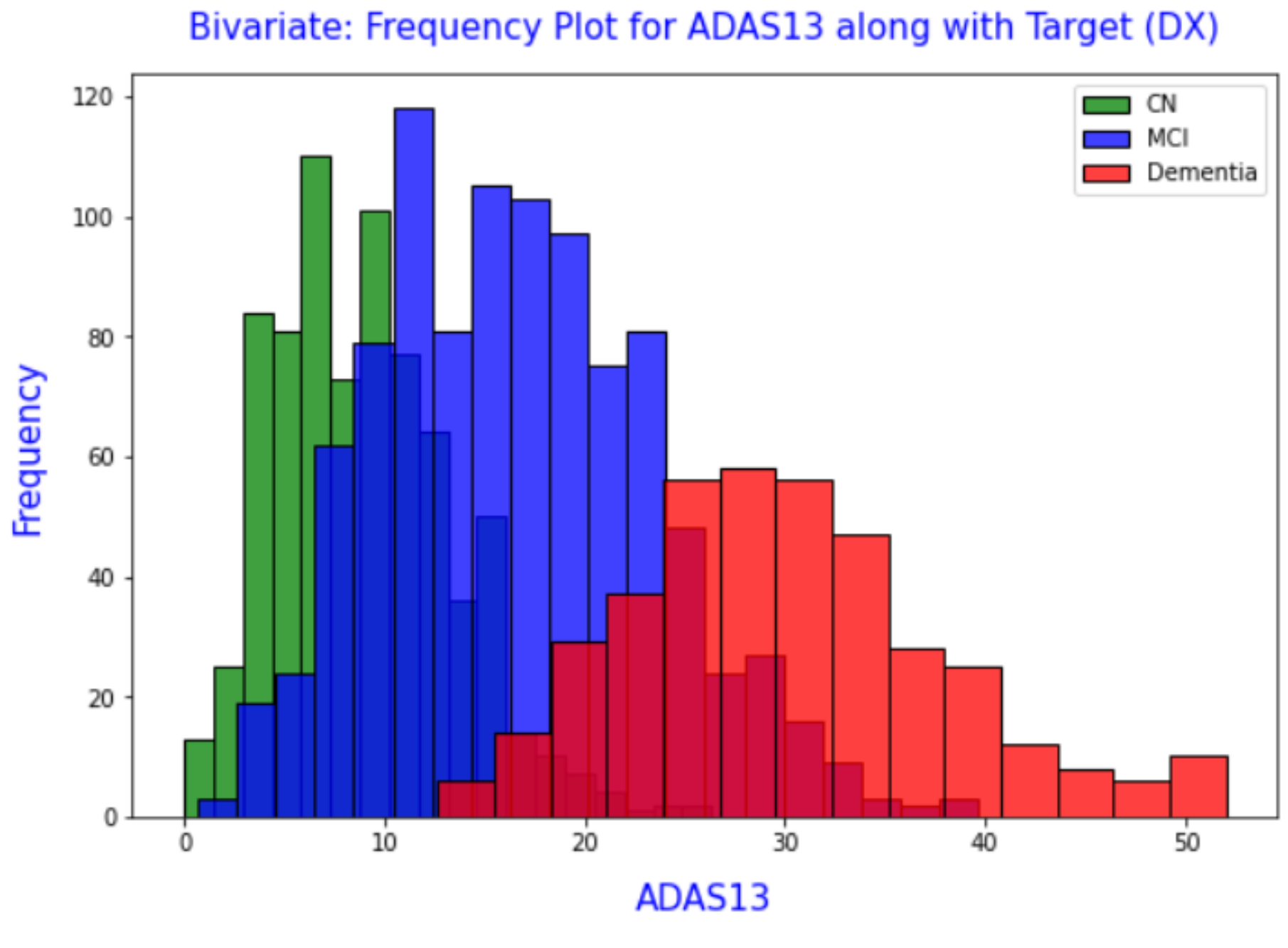

*Figure 4.49 Comparative Hist Plot between DX (Target) vs ADAS13*

**g) EcogPtMem (Measurement of Everyday Cognition completed out by Participant)**

As shown in figure 4.50, the average score of this cognitive assessment is high (around 2.3 to 2.5) for participants who are suffering from dementia (Target Class 2). On the contrary, people who are part of healthy population are having less score for this cognitive test (around 1.7). The average test score for the severe dementia affected and mild affected people are almost similar. This variable might not be able to separate class 1 and class 2 distinctly.

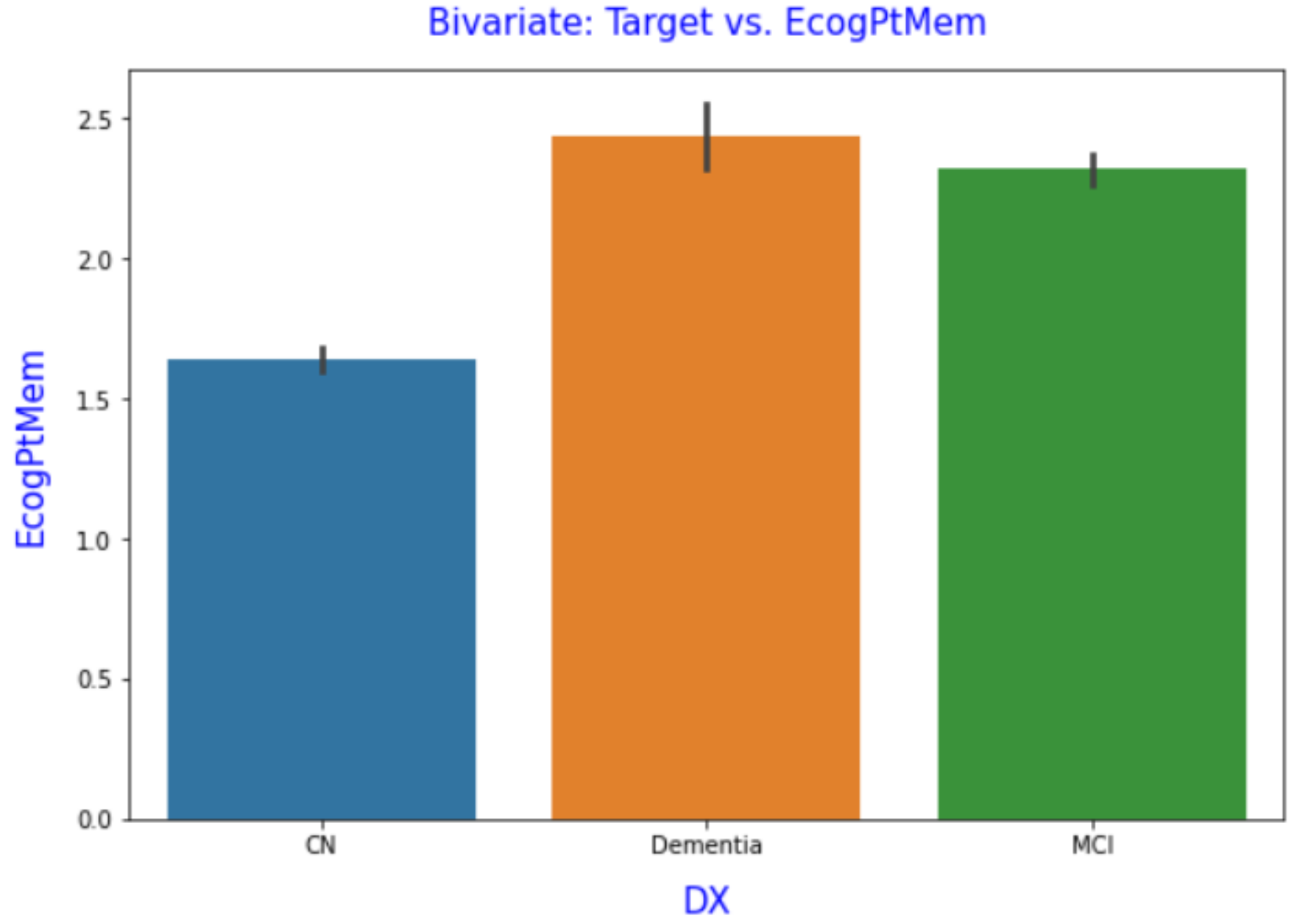

*Figure 4.50 Bar Plot between DX (Target) vs EcogPtMem*

**h) EcogSPTotal (Everyday Cognition Study Partner - Total)**

As shown in figure 4.51, the average score of this cognitive symptom assessment is high (above 2.5) for participants who are suffering from dementia (Target Class 2). On the contrary, people who are part of healthy population are having less score for this cognitive test (around 1.3). This variable might become an important distinguishing factor of Alzheimer's prediction as the values lie in different range among the three classes.

*Figure 4.51 Bar Plot between DX (Target) vs EcogSPTotal*

From below figure 4.52, it is clearly observed that Ecog test rating for Dementia affected people are mostly ranges between 1.5 to 4. While 1.5 to 3.5 is an overlapping between AD and MCI. For healthy people this score mostly spans between 1 to 1.5. For majority of normal people this score is 1.0

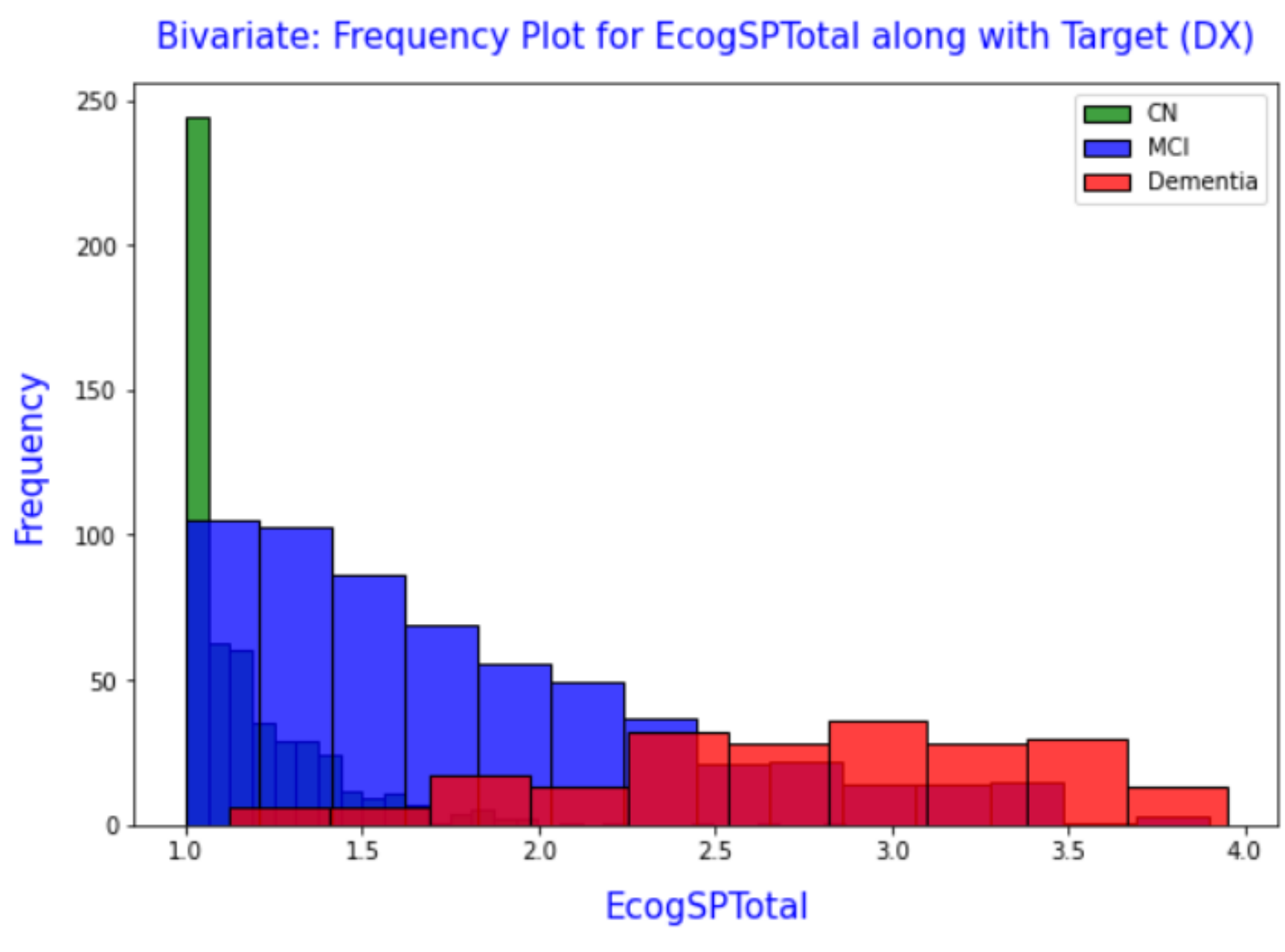

*Figure 4.52 Comparative Hist Plot between DX (Target) vs EcogSPTotal*

### i) MMSE (Mini-Mental State Examination)

As shown in figure 4.53, the average score of this cognitive function test is low (below 25) for participants who are suffering from dementia (Target Class 2). On the contrary, people who are part of healthy population are having high score for this cognitive test (around 29). This variable might become an important distinguishing factor of Alzheimer's prediction as the values lie in different range among the three classes.

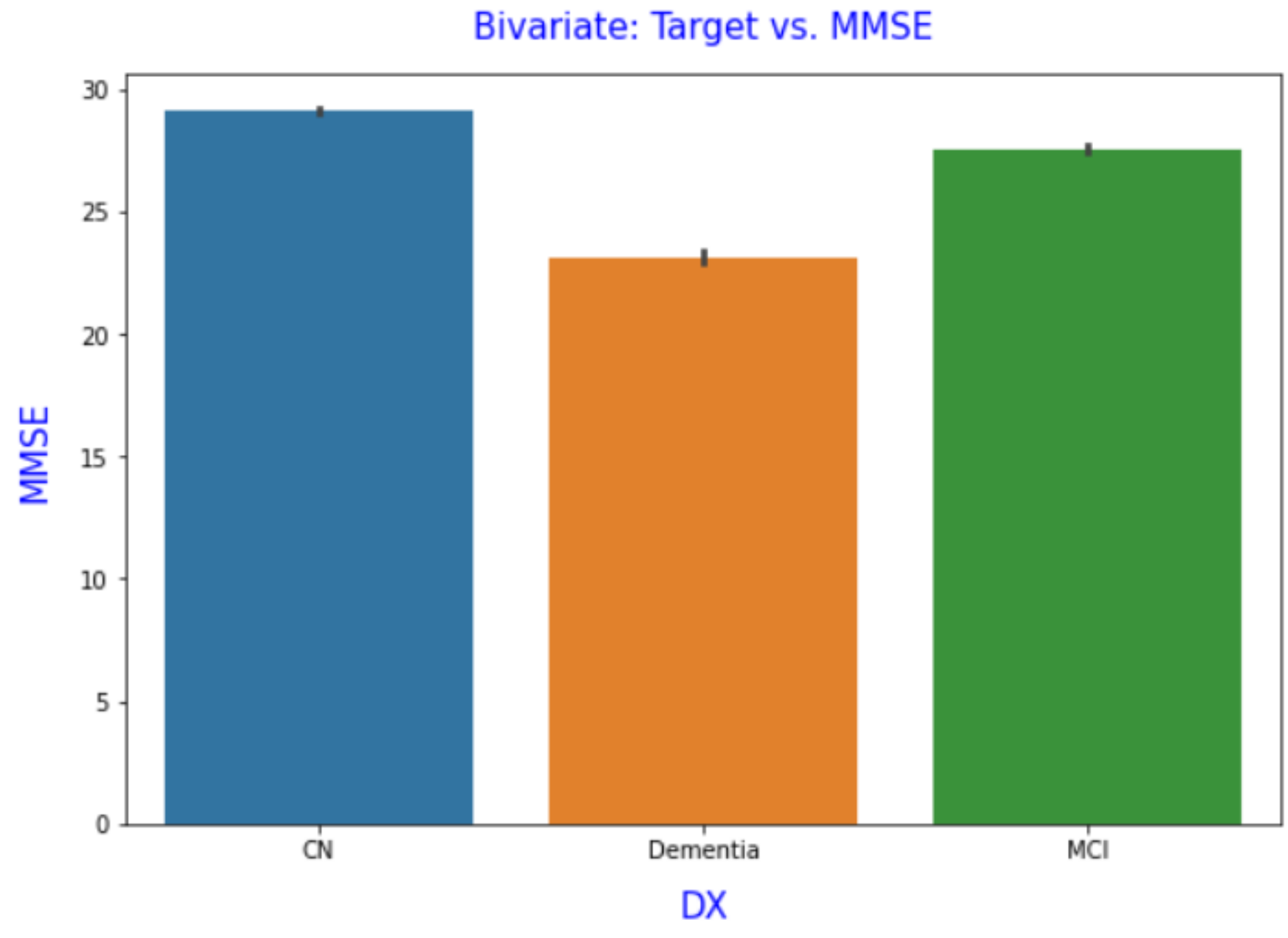

*Figure 4.53 Bar Plot between DX (Target) vs MMSE*

From below figure 4.54, it is clearly observed that MMSE rating for Dementia affected people are mostly ranges between 16 to 26. While 24 to 26 is an overlapping between AD and MCI. For healthy people this score mostly spans near 30. Hence, this variable can give a clear identification of target groups.

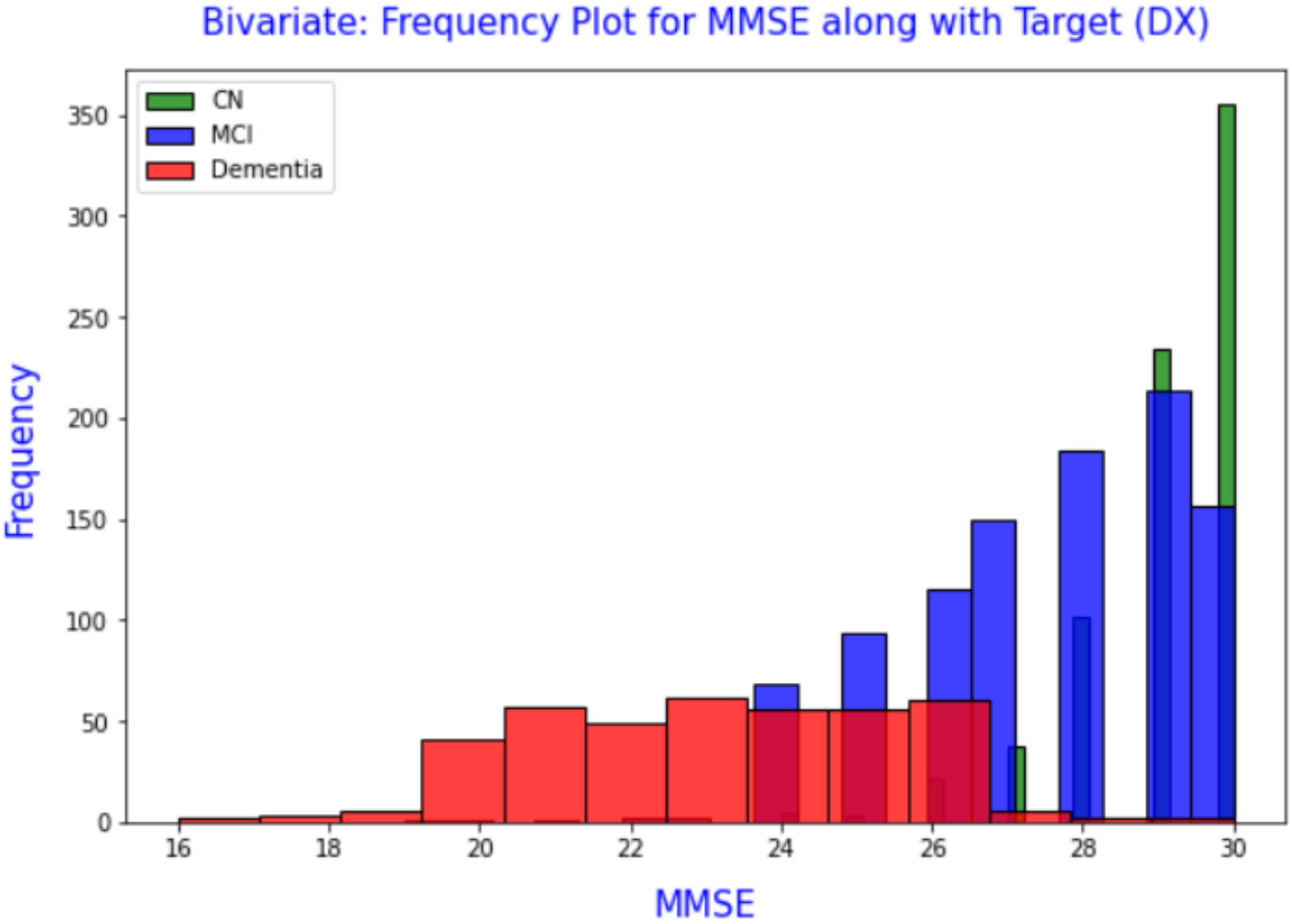

*Figure 4.54 Comparative Hist Plot between DX (Target) vs MMSE*

#### 4.4.2.2 Categorical vs Categorical Analysis

This section will go over various bivariate visualisations, such as bar plots and count plots, for some significant independent categorical attributes. Among all the predictive indicators, this section focuses on few most important biomarkers. This analysis focuses on the target variable (DX) as a categorical feature versus some other vital categorical features.

**a) PTGENDER (Participant's Gender)**

Figure 4.55 shows that among dementia sufferers, the percentage of male participants is slightly higher than that of females. Males are affected by dementia at a rate of 19.67%, while females are affected at a rate of 17.52%. On the other hand, 28.85% of male candidates are classified as normal, while 42.84% of female candidates are diagnosed as healthy.

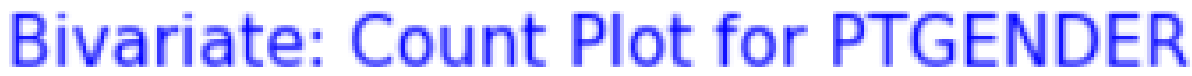

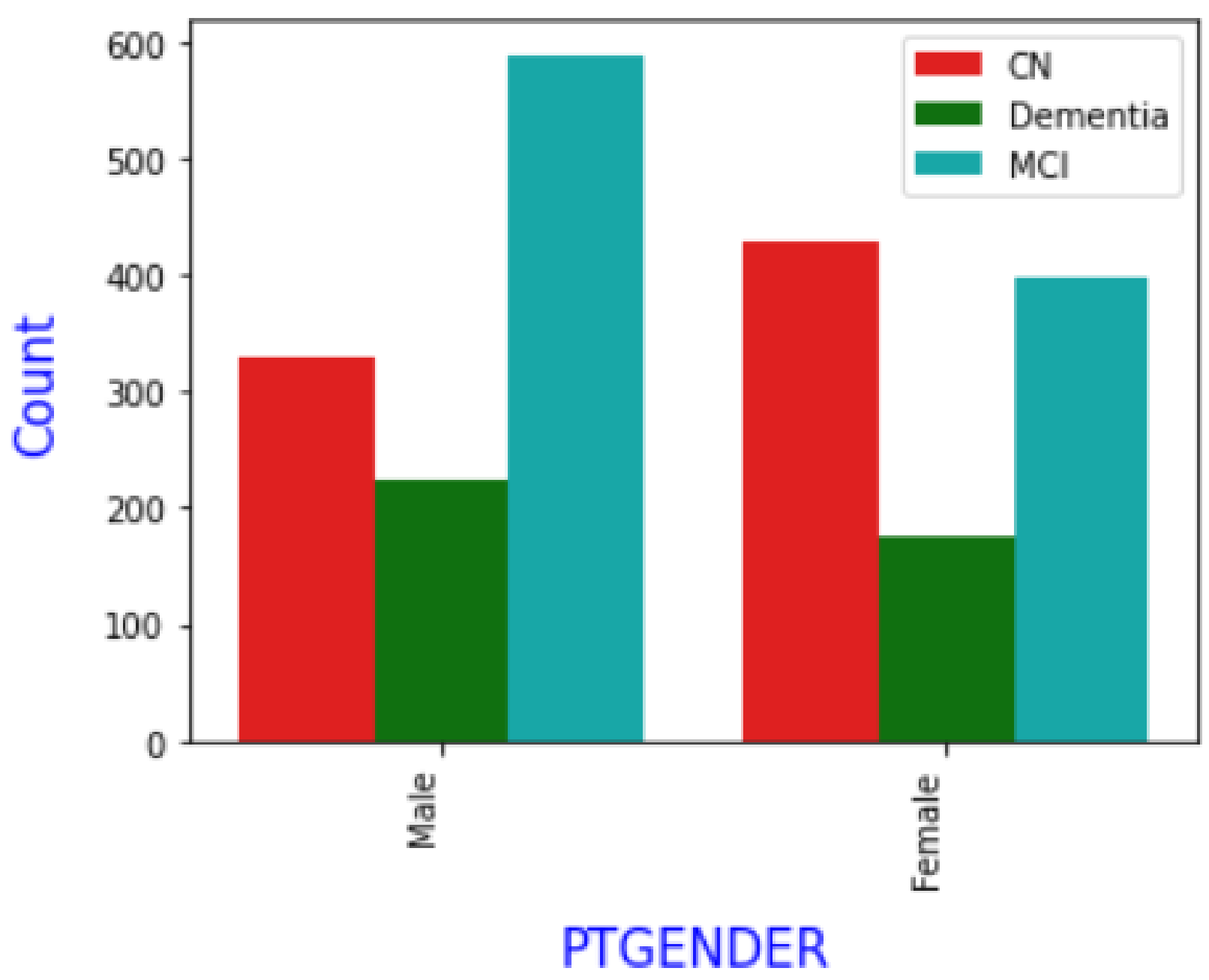

```
PTGENDER   DX
Female     CN          0.428428
           MCI         0.396396
           Dementia    0.175175
Male       MCI         0.514860
           CN          0.288462
           Dementia    0.196678
```

*Figure 4.55 Comparative Frequency between DX (Target) vs PTGENDER*

**b) PTETHCAT (Participant’s Ethnicity)**

Figure 4.56 shows that the dataset contains majority of observations with ethnicity ‘Not Hisp/Latino’. People with ‘Not Hisp/Latino’ ethnic group are having highest percentage (46.18%) of mild cognitive impairment. Also, the percentage of Alzheimer's sufferers are more (around 18.91%) in this ‘Not Hisp/Latino’ group. Most of the participant’s from ‘Hisp/Latino’ group has been diagnosed as normal (45.00%).

| PTETHCAT | DX | |
|---|---|---|
| Hisp/Latino | CN | 0.450000 |
| | MCI | 0.420000 |
| | Dementia | 0.130000 |
| Not Hisp/Latino | MCI | 0.461841 |
| | CN | 0.349089 |
| | Dementia | 0.189069 |
| Unknown | MCI | 0.416667 |
| | CN | 0.333333 |
| | Dementia | 0.250000 |

*Figure 4.56 Comparative Frequency between DX (Target) vs PTETHCAT*

**c) PTRACCAT (Participant's Race)**

Figure 4.57 shows that that a majority of our population is white. Among white and Asian people, the percentage of dementia sufferers are slightly higher than others (19.09% and 19.14% respectively). Most of the people, around 60% individuals, are suffering from mild impairment among the Am Indian/Alaskan race. On the other hand, 50.75% candidates are classified as normal among the black members.

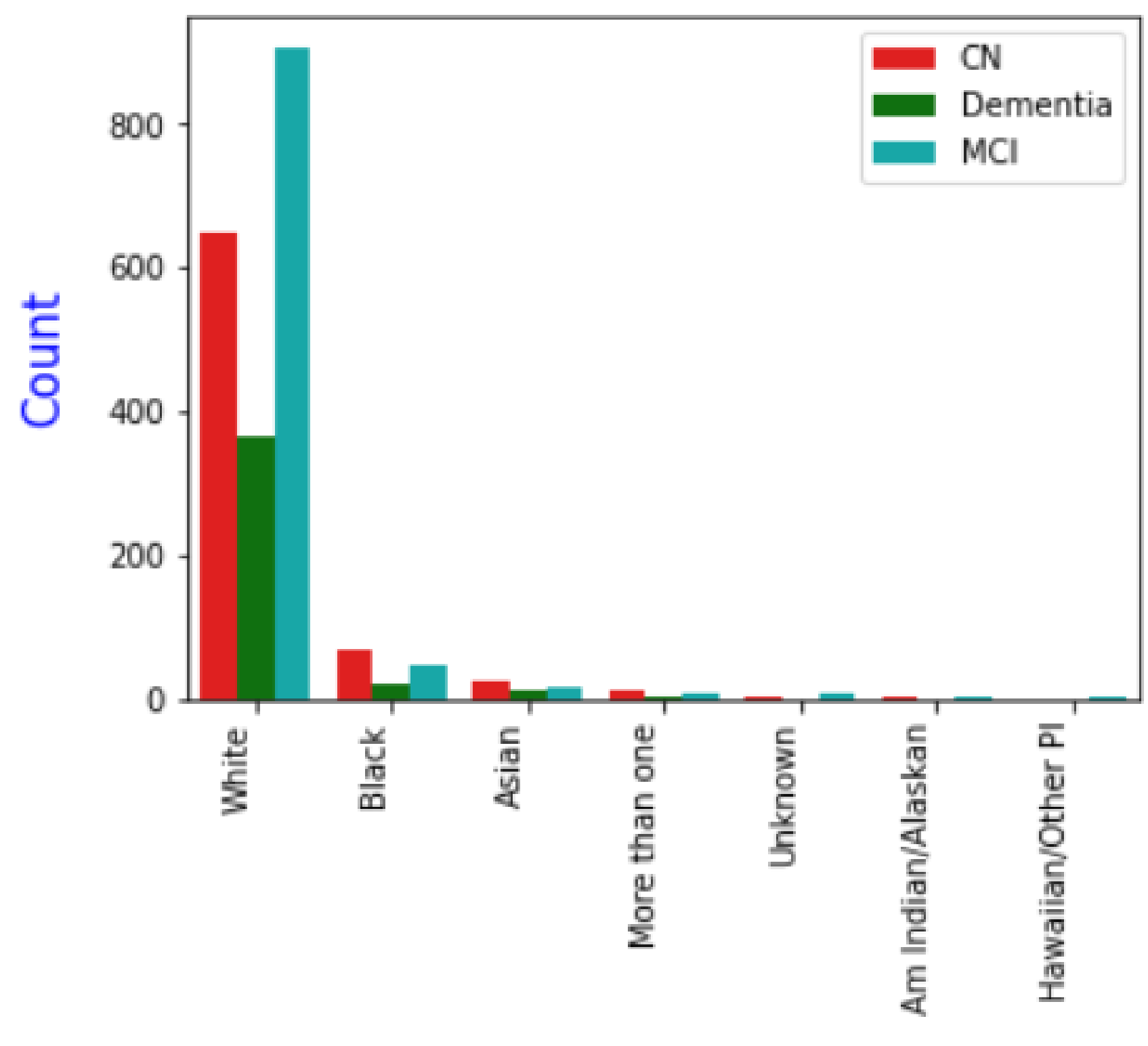

| PTRACCAT | DX | |
|---|---|---|
| Am Indian/Alaskan | MCI | 0.600000 |
| | CN | 0.400000 |
| Asian | CN | 0.489362 |
| | MCI | 0.319149 |
| | Dementia | 0.191489 |
| Black | CN | 0.507576 |
| | MCI | 0.340909 |
| | Dementia | 0.151515 |
| Hawaiian/Other PI | MCI | 1.000000 |
| More than one | CN | 0.500000 |
| | MCI | 0.333333 |
| | Dementia | 0.166667 |
| Unknown | MCI | 0.636364 |
| | CN | 0.363636 |
| White | MCI | 0.470864 |
| | CN | 0.338189 |
| | Dementia | 0.190947 |

*Figure 4.57 Comparative Frequency between DX (Target) vs PTRACCAT*

**d) PTMARRY (Participant's Marital Status)**

Figure 4.58 shows that that a majority of our population is married. Among married people, the percentage of dementia sufferers are slightly higher than others (around 20.43%). Percentage of mild dementia cases are high among married and divorced people (46.95% and 47.56% respectively). On the other hand, 51.89% candidates are classified as healthy among the never married category.

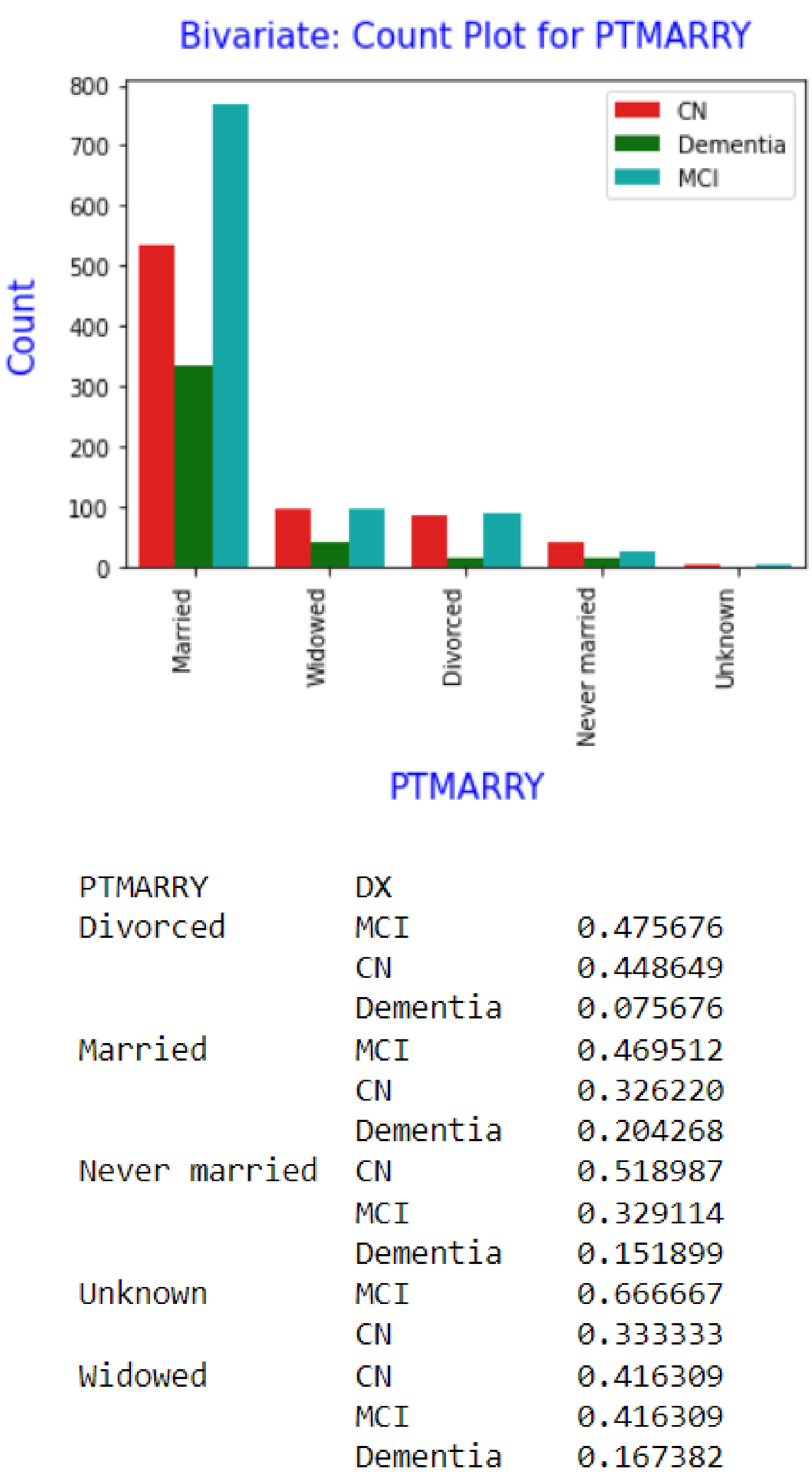

| PTMARRY | DX | |
|---|---|---|
| Divorced | MCI | 0.475676 |
| | CN | 0.448649 |
| | Dementia | 0.075676 |
| Married | MCI | 0.469512 |
| | CN | 0.326220 |
| | Dementia | 0.204268 |
| Never married | CN | 0.518987 |
| | MCI | 0.329114 |
| | Dementia | 0.151899 |
| Unknown | MCI | 0.666667 |
| | CN | 0.333333 |
| Widowed | CN | 0.416309 |
| | MCI | 0.416309 |
| | Dementia | 0.167382 |

*Figure 4.58 Comparative Frequency between DX (Target) vs PTMARRY*

### e) FLDSTRNG (MRI Field Strength)

Figure 4.59 shows that that a majority of our population is diagnosed with MRI Field Strength 1.5 Tesla MRI. People examined with 1.5 Tesla MRI field strength are more detected with Alzheimer's disease or dementia. On the other hand, 61.23% candidates are classified as mild dementia when detected using 3 Tesla MRI.

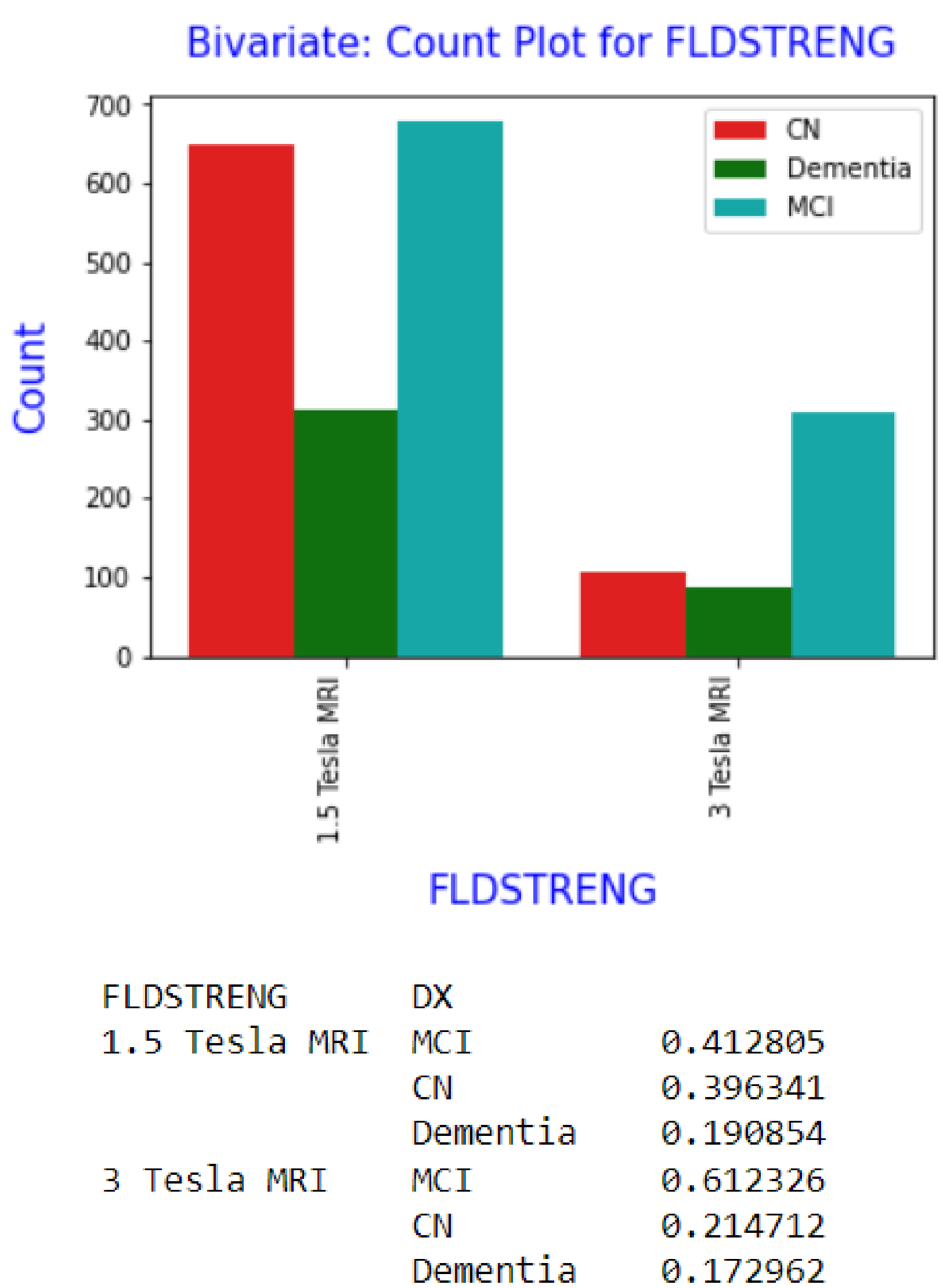

| FLDSTRENG | DX | |
|---|---|---|
| 1.5 Tesla MRI | MCI | 0.412805 |
| | CN | 0.396341 |
| | Dementia | 0.190854 |
| 3 Tesla MRI | MCI | 0.612326 |
| | CN | 0.214712 |
| | Dementia | 0.172962 |

*Figure 4.59 Comparative Frequency between DX (Target) vs FLDSTRENG*

#### 4.4.2.3 Numerical vs Numerical Analysis

This section will go over bivariate visualisations mainly scatter plots, for some significant independent numerical attributes. Among all the predictive indicators, this section focuses on few most important biomarkers. This analysis focuses on finding relationship between two numerical features and checks whether two continuous variables are corelated to each other or holding any kind of association between each other.

**a) CDRSB (Clinical Dementia Rating Sum of Boxes)**

From below scatterplot figure 4.60, it is shown that Clinical dementia rating sum of boxes test result of the patients participated in ADNI study does have a positively increasing relationships with other numerical features like with ADAS13 or with EcogSPTotal. It follows a good negative linear relationship with mPACCdigit, MMSE and MOCA. Rest other relationships are merely random, and no patterns are observed.

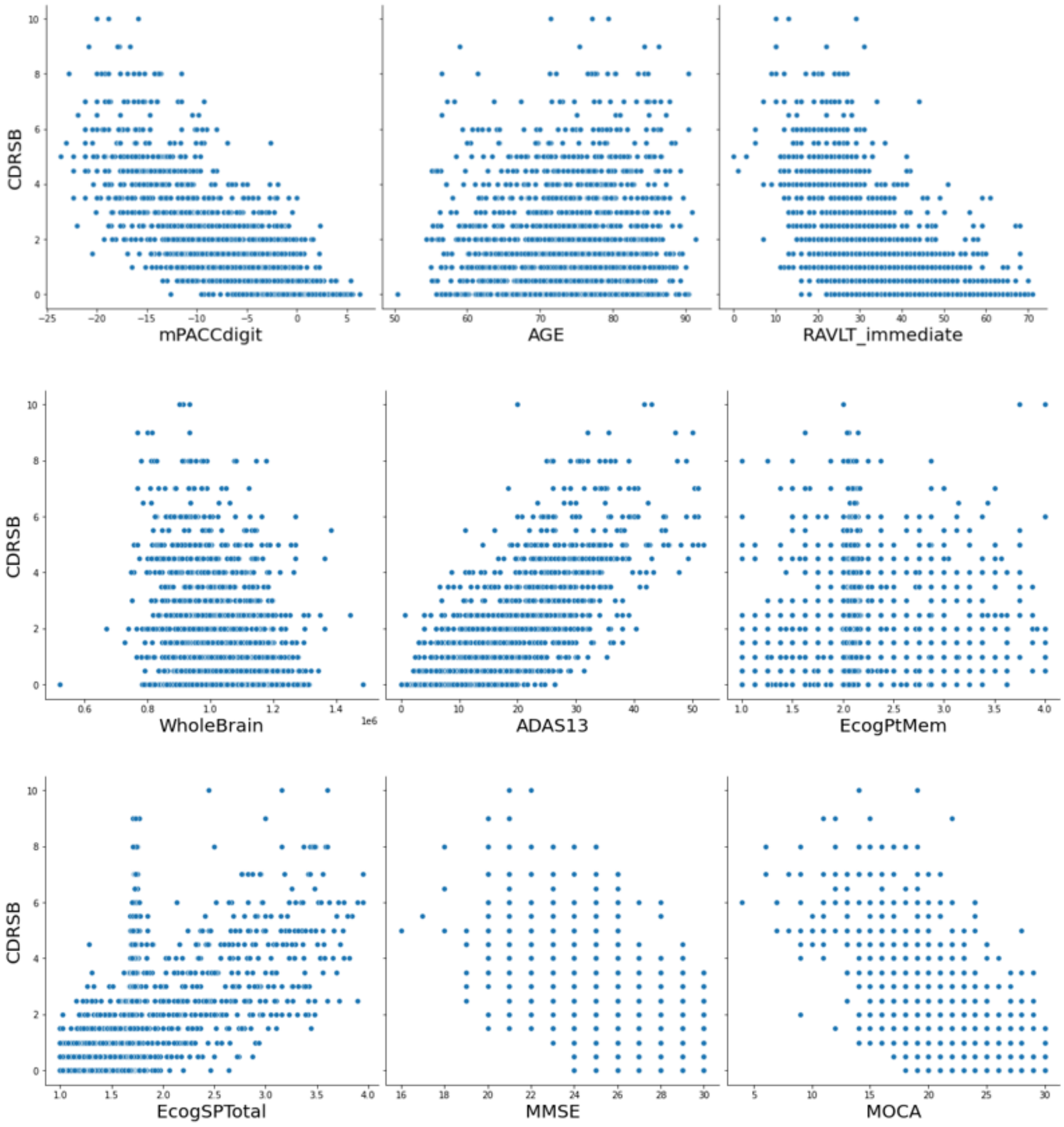

*Figure 4.60 Scatter Plot between CDRSB vs Other Numerical Variables*

**b) mPACCdigit (modified Preclinical Alzheimer's Cognitive Composite with Digit)**

From below scatterplot figure 4.61, it is shown that PACC digit test result of the patients participated in ADNI study does have a positively increasing relationships with other numerical features like with RAVLT_immediate or with MMSE or with MOCA. It follows a good negative linear relationship with CDRSB, ADAS13 and EcogSPTotal. Rest other relationships are merely random, and no patterns are observed.

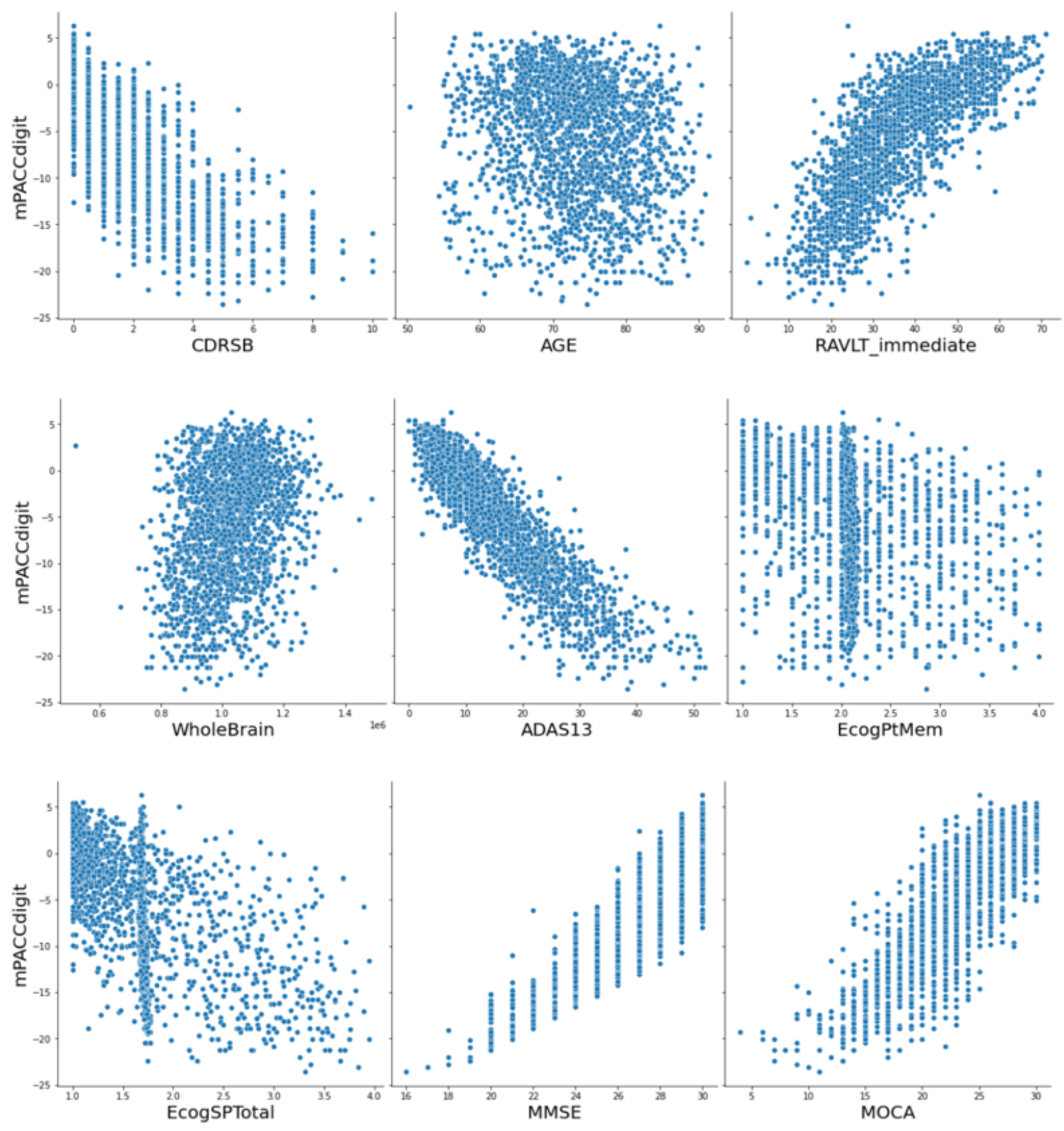

*Figure 4.61 Scatter Plot between mPACCdigit vs Other Numerical Variables*

**c) AGE (Participant's Age)**

From below scatterplot figure 4.62, it is shown that age of the patients participated in ADNI study does not have any type of relationships with any other numerical features like with RAVLT_immediate or with MMSE or with MOCA or with CDRSB or with ADAS13. All relationships are merely random, and no patterns are observed.

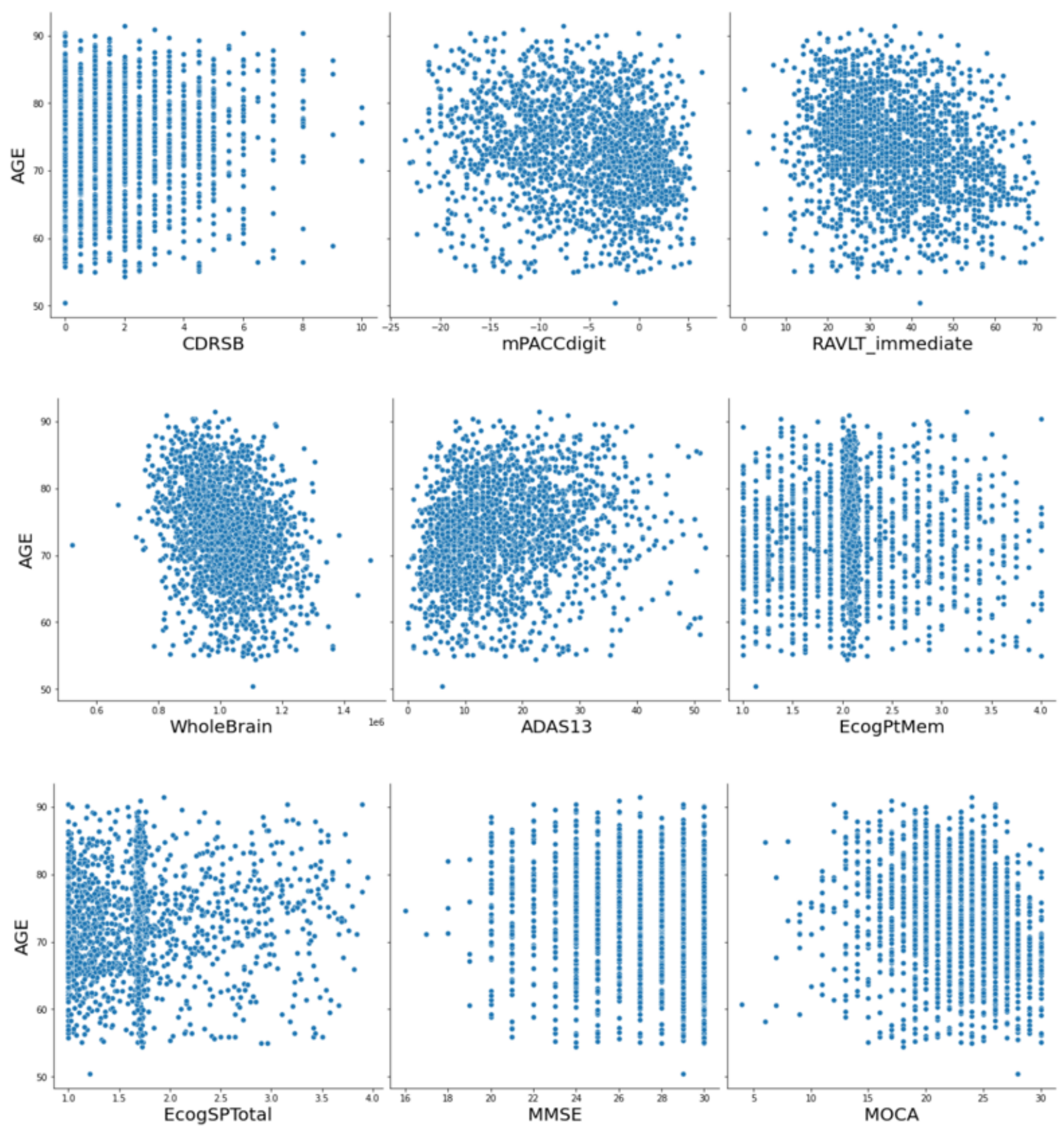

*Figure 4.62 Scatter Plot between AGE vs Other Numerical Variables*

**d) RAVLT_immediate (Rey Auditory Verbal Learning Test for immediate response)**

From below scatterplot figure 4.63, it is shown that verbal learning test result of the patients participated in ADNI study does have a positively increasing relationships with other numerical features like with mPACCdigit or with MMSE or with MOCA variables. It follows a good negative linear relationship with CDRSB, ADAS13 and EcogSPTotal. Rest other relationships are merely random, and no patterns are observed.

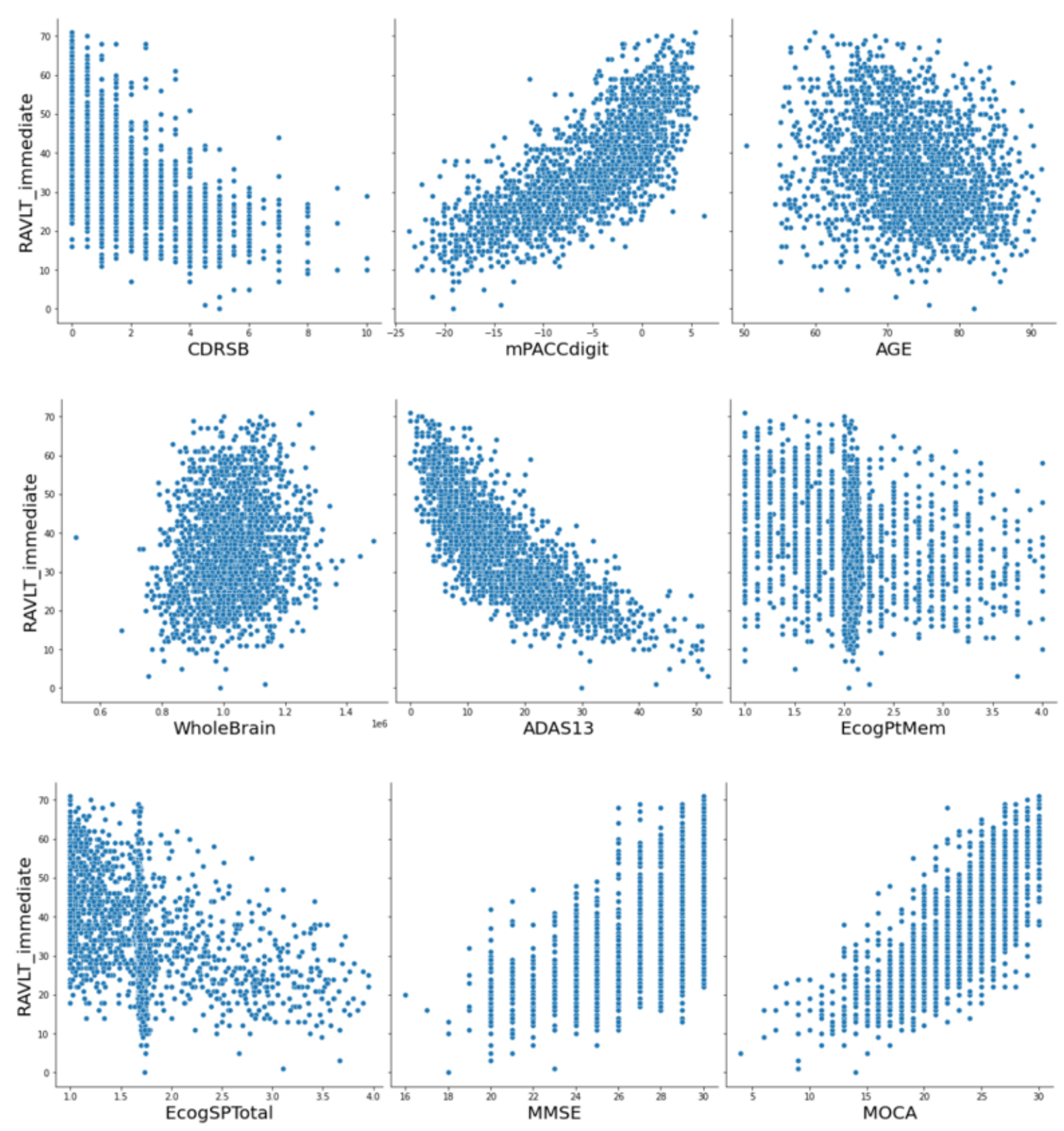

*Figure 4.63 Scatter Plot between RAVLT_immediate vs Other Numerical Variables*

**e) WholeBrain (Entire Brain Volume)**

From below scatterplot figure 4.64, it is shown that entire brain volume of the patients participated in ADNI study does not have any strong relationships with any other numerical features like with mPACCdigit or with MMSE or with MOCA variables. It follows a weak negative linear relationship with CDRSB. Rest other relationships are merely random, and no patterns are observed.

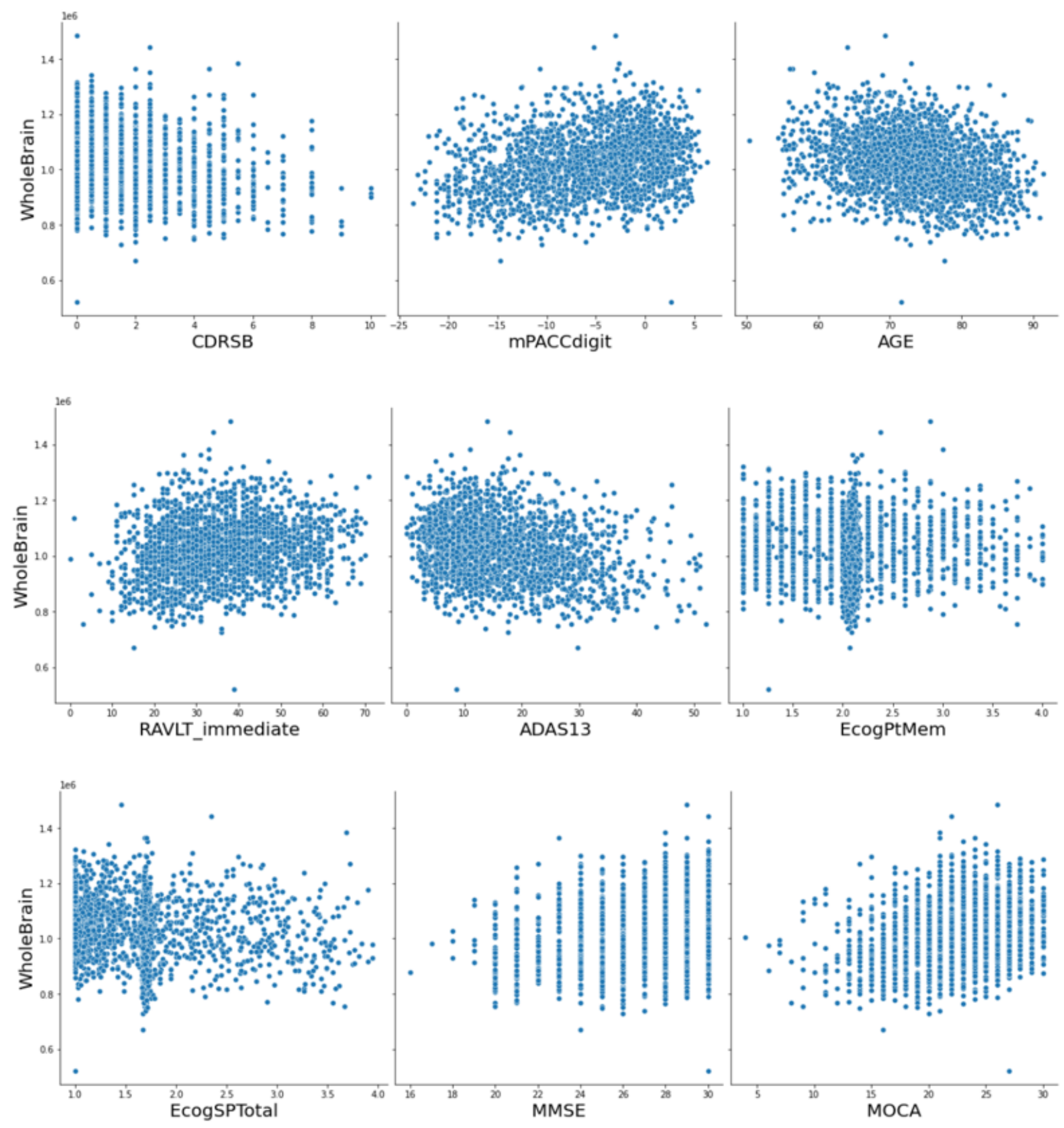

*Figure 4.64 Scatter Plot between WholeBrain vs Other Numerical Variables*

**f) ADAS13 (Alzheimer's Disease Assessment Scale - 13 items)**

From below scatterplot figure 4.65, it is shown that alzheimer's disease assessment scale test with 13 items of the patients participated in ADNI study does have a strong positive linear relationship with other numerical features like with CDRSB or with EcogSPTotal variables. It follows a good negative linear relationship with mPACCdigit or with RAVLT_immediate or with MMSE and MOCA. Rest other relationships are merely random, and no patterns are observed.

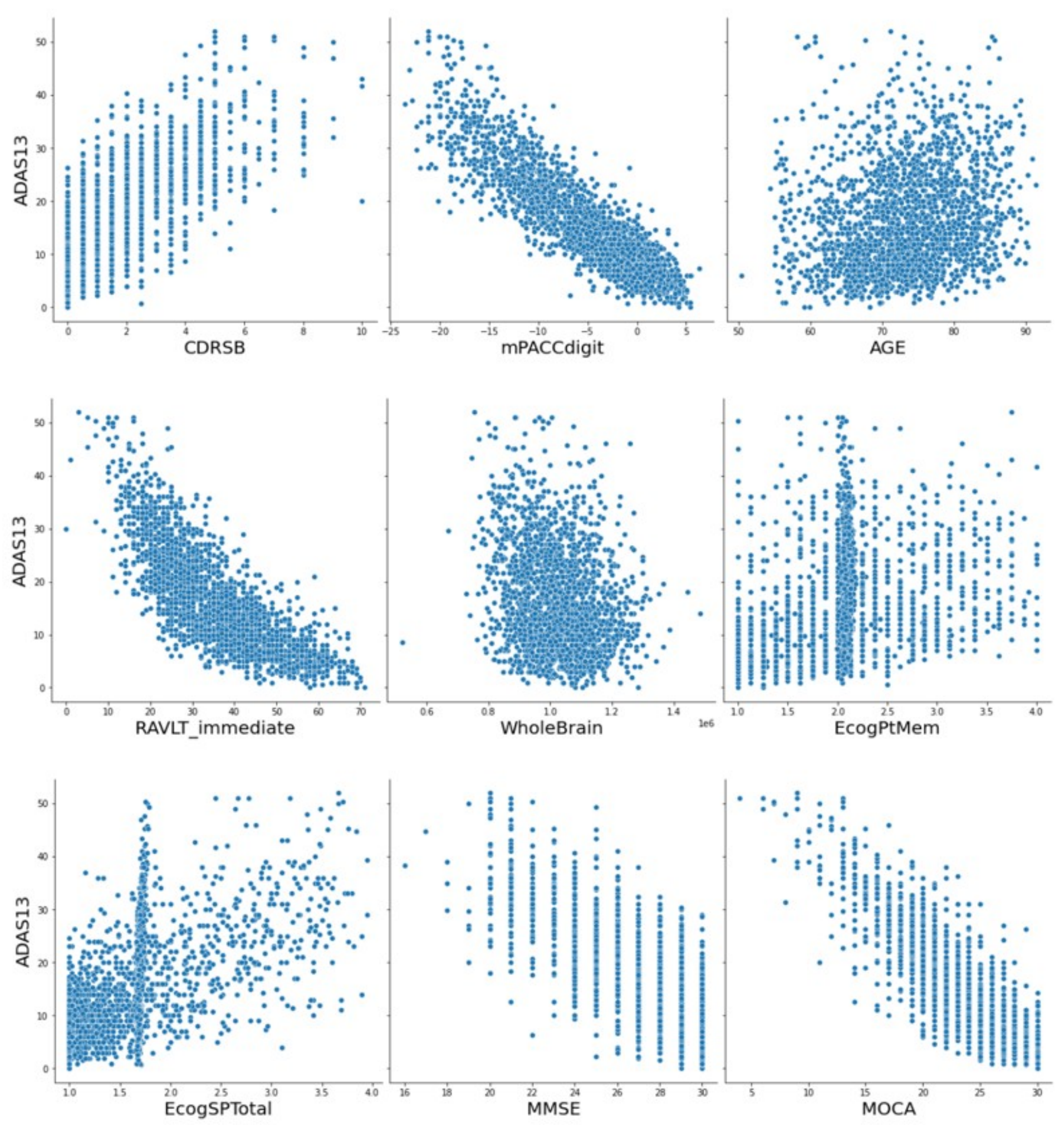

*Figure 4.65 Scatter Plot between ADAS13 vs Other Numerical Variables*

**g) EcogPtMem (Measurement of Everyday Cognition completed out by Participant)**

From below scatterplot figure 4.66, it is shown that everyday cognitive assessment test completed by patients participated in ADNI study does not have any relationships with any other numerical features like with CDRSB or with EcogSPTotal or AGE or MMSE etc. variables. These relationships are merely random, and no patterns are observed.

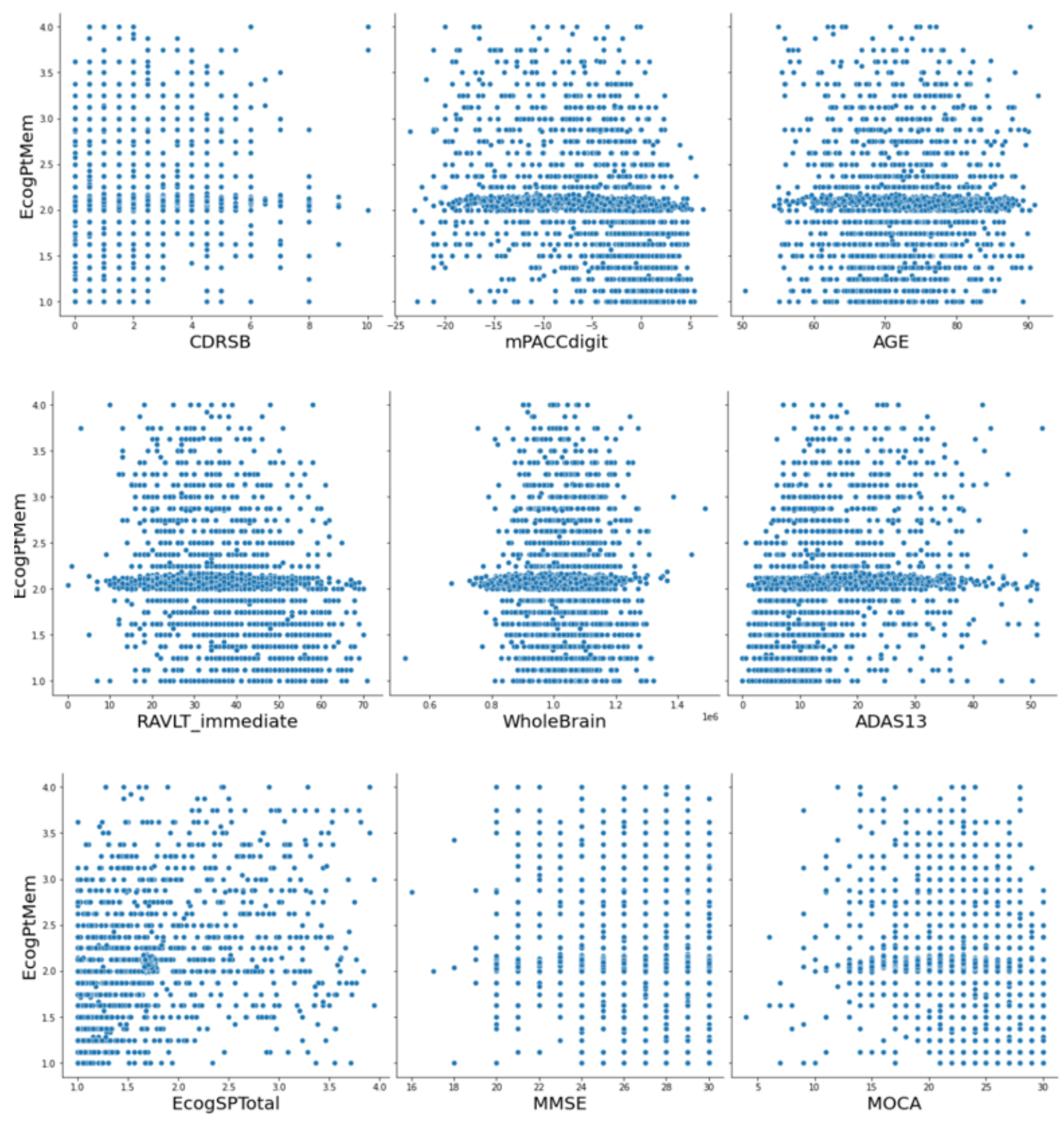

*Figure 4.66 Scatter Plot between EcogPtMem vs Other Numerical Variables*

**h) EcogSPTotal (Everyday Cognition Study Partner – Total)**

From below scatterplot figure 4.67, it is shown that everyday cognitive test completed by study partners of the patients participated in ADNI study does have a poor positive linear relationship with other numerical features like with CDRSB or with ADAS13 variables. It follows a slight negative linear relationship with mPACCdigit or with RAVLT_immediate. Rest other relationships are merely random, and no patterns are observed.

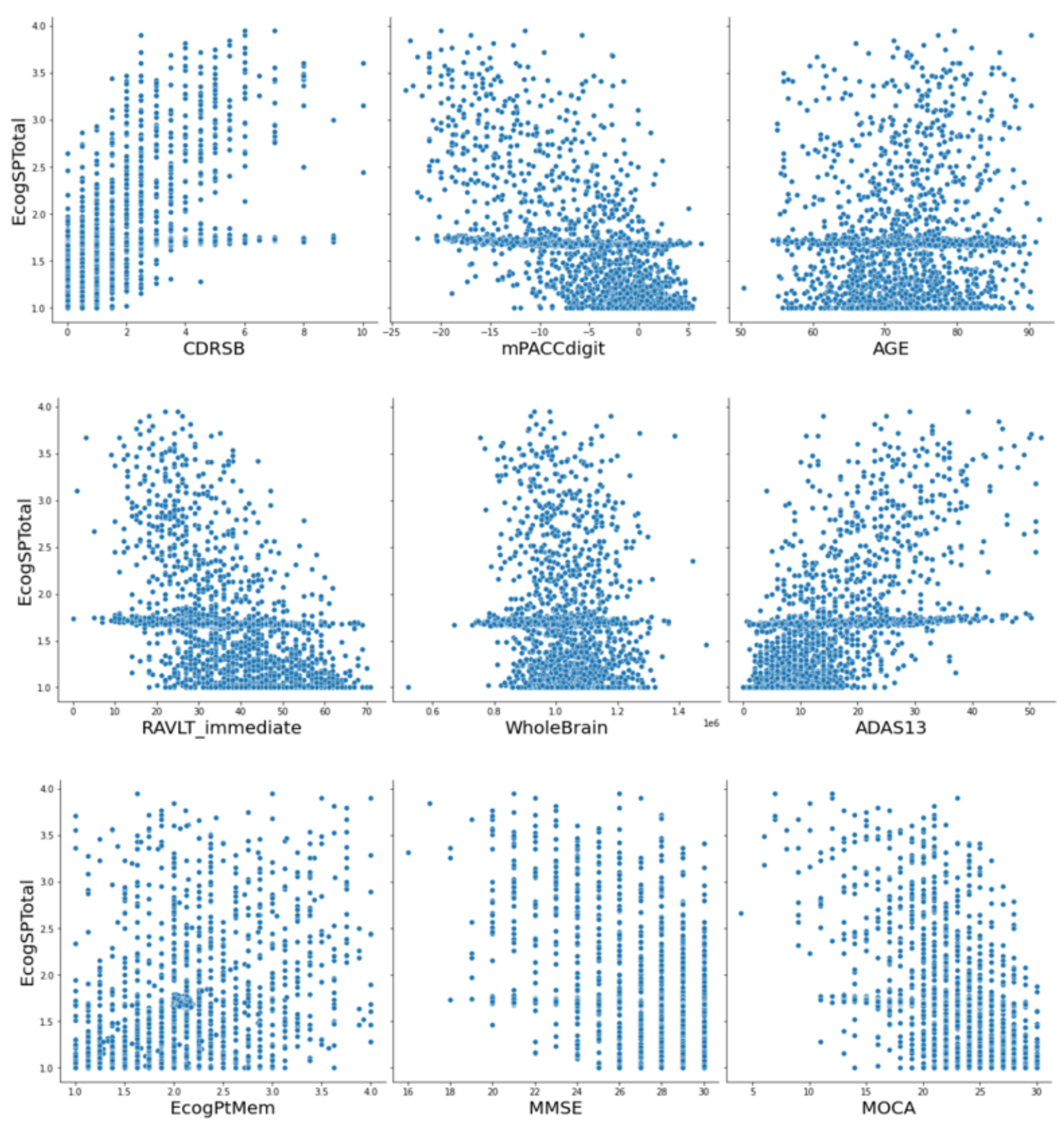

*Figure 4.67 Scatter Plot between EcogSPTotal vs Other Numerical Variables*

### i) MMSE (Mini-Mental State Examination)

From below scatterplot figure 4.68, it is shown that mini-mental state examination score of the patients participated in ADNI study does have a strong positive linear relationship with other numerical features like with mPACCdigit or with RAVLT_immediate or with MOCA variables. It follows a good negative linear relationship with CDRSB or with ADAS13. Rest other relationships are merely random, and no specific patterns are observed.

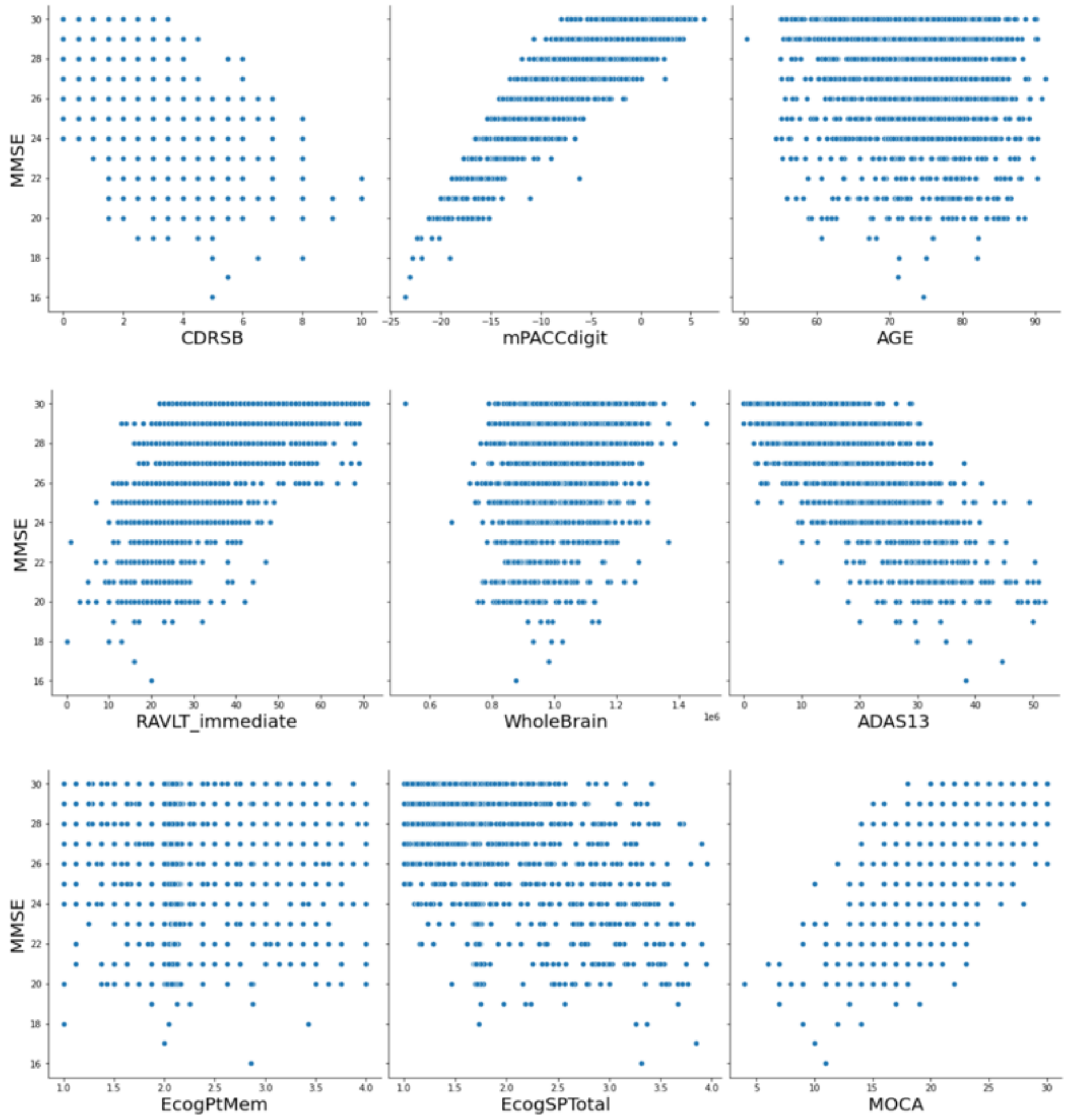

*Figure 4.68 Scatter Plot between MMSE vs Other Numerical Variables*

### j) MOCA (Montreal Cognitive Assessment)

From below scatterplot figure 4.69, it is shown that Montreal cognitive assessment test score of the patients participated in ADNI study does have a good positive linear relationship with other numerical features like with mPACCdigit or with PAVLT_immediate or with MMSE variables. It follows a good negative linear relationship with CDRSB or with ADAS13. Rest other relationships are merely random, and no specific patterns are observed.

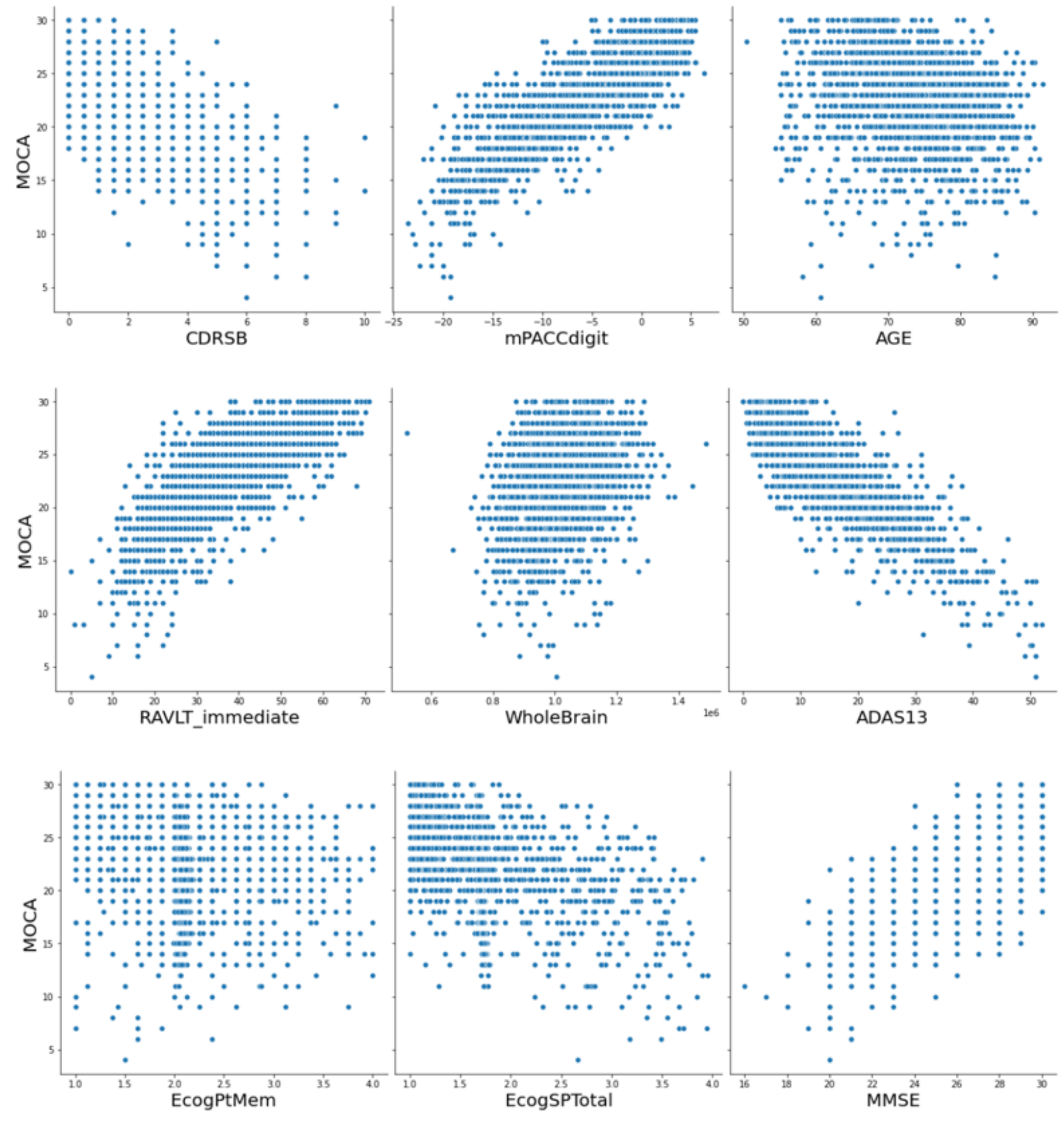

*Figure 4.69 Scatter Plot between MOCA vs Other Numerical Variables*

### 4.4.3 Multivariate and Correlation Analysis

Multivariate analysis examines multiple features (more than two) to see whether there is a possible relationship among them. In this study, Pearson correlations among numerical features were examined as part of multivariate analysis, and the top 10 positive and negative correlations were reported correspondingly.

Below table 4.31 shows the top 10 positive correlations among the numerical features.

*Table 4.31 Top 10 Positive Correlations among Numerical Variables*

| Feature1 | Feature2 | Correlation |
|---|---|---|
| mPACCtrailsB | mPACCdigit | 0.97936 |
| ADAS13 | ADAS11 | 0.976009 |
| EcogSPTotal | EcogSPPlan | 0.912253 |

| EcogSPTotal | EcogSPMem | 0.909163 |
|---|---|---|
| EcogSPTotal | EcogSPOrgan | 0.899913 |
| mPACCdigit | MMSE | 0.898784 |
| mPACCtrailsB | MMSE | 0.892887 |
| EcogSPTotal | EcogSPDivatt | 0.890344 |
| EcogSPTotal | EcogSPLang | 0.888824 |
| EcogPtTotal | EcogPtLang | 0.885935 |

The above table indicates that there is a strong positive correlation between mPACCtrailsB and mPACCdigit. All the ECog related variables are also showing positive correlations among them. That means increasing one variable will also linearly increase the other one in positive direction.

Below table 4.32 shows the top 10 negative correlations among the numerical features.

*Table 4.32 Top 10 Negative Correlations among Numerical Variables*

| **Feature1** | **Feature2** | **Correlation** |
|---|---|---|
| mPACCtrailsB | ADAS13 | -0.888201 |
| mPACCdigit | ADAS13 | -0.885473 |
| mPACCdigit | ADASQ4 | -0.871039 |
| mPACCtrailsB | ADASQ4 | -0.861363 |
| MOCA | ADAS13 | -0.842525 |
| mPACCtrailsB | ADAS11 | -0.828055 |
| mPACCdigit | ADAS11 | -0.826164 |
| MOCA | ADAS11 | -0.812793 |
| RAVLT_immediate | ADAS13 | -0.782312 |
| mPACCdigit | CDRSB | -0.778004 |

From above table, it is noticed that there is a strong negative correlation exists between mPACCtrailsB and ADAS13; mPACCdigit and ADASQ4 etc. That means increasing one variable say mPACCtrailsB will linearly decrease the other one ADAS13.

Below figure 4.70 visualizes the heatmap, which represents the correlation between different numerical features.

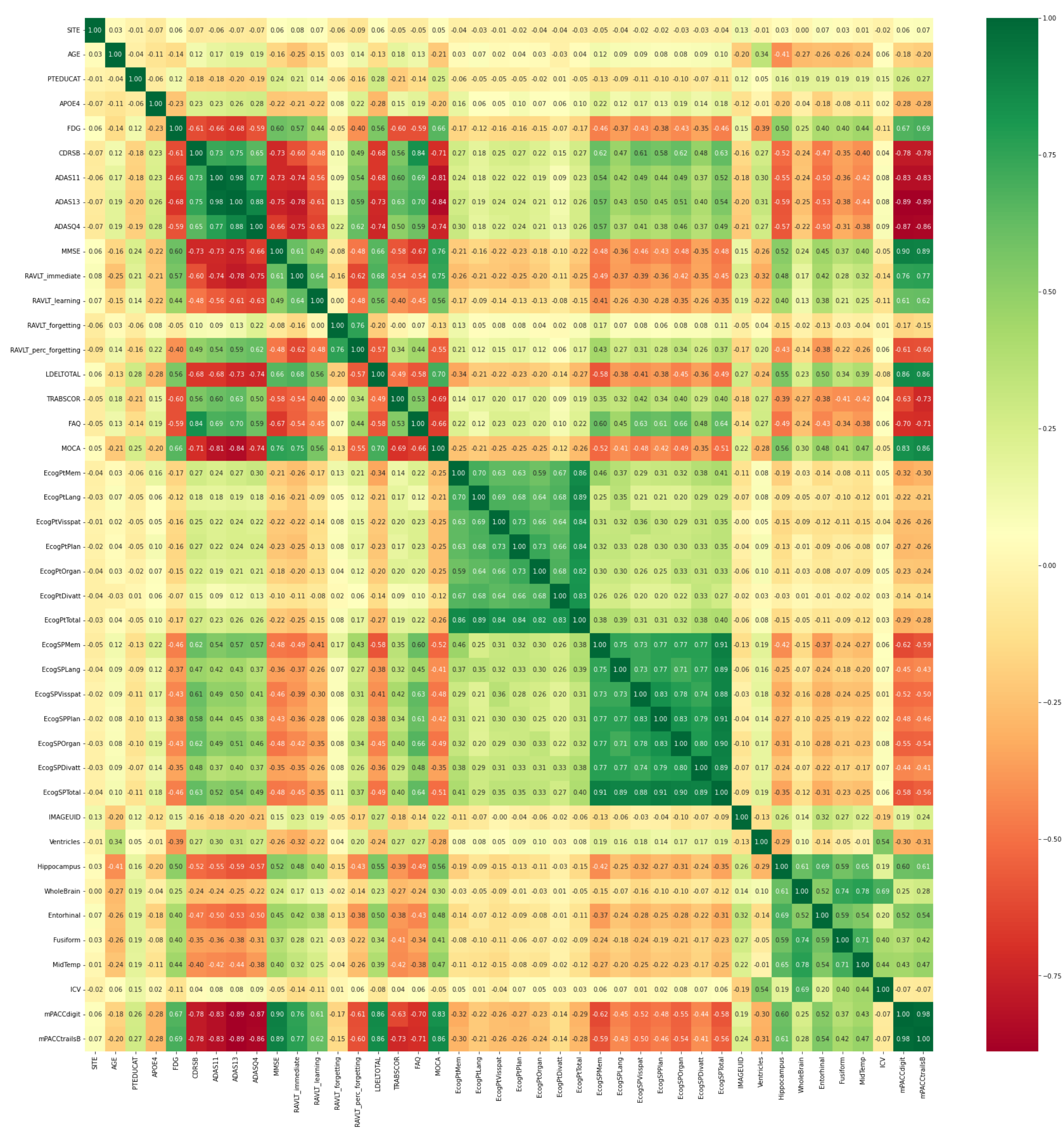

*Figure 4.70 Heat Map representing Correlation among Multiple Numerical Variables*

The pair-wise relationship between numerical features is depicted in figure 4.71 below. It can be shown that various features, such as RAVLT immediate and MOCA; RAVLT immediate and mPACCdigit; ADAS13 and CDRSB; ADAS13 and RAVLT immediate etc., have a linear relationship. Combination of these variables are also able to distinctly identify target groups, i.e., CN, MCI, Dementia or AD. The plots below showing the colour codes for Class 0 (CN) is green, Class 1 (MCI) is blue, and Class 2 (AD) is red. Fusion of these variables can show a potential discriminative power between the three classes.

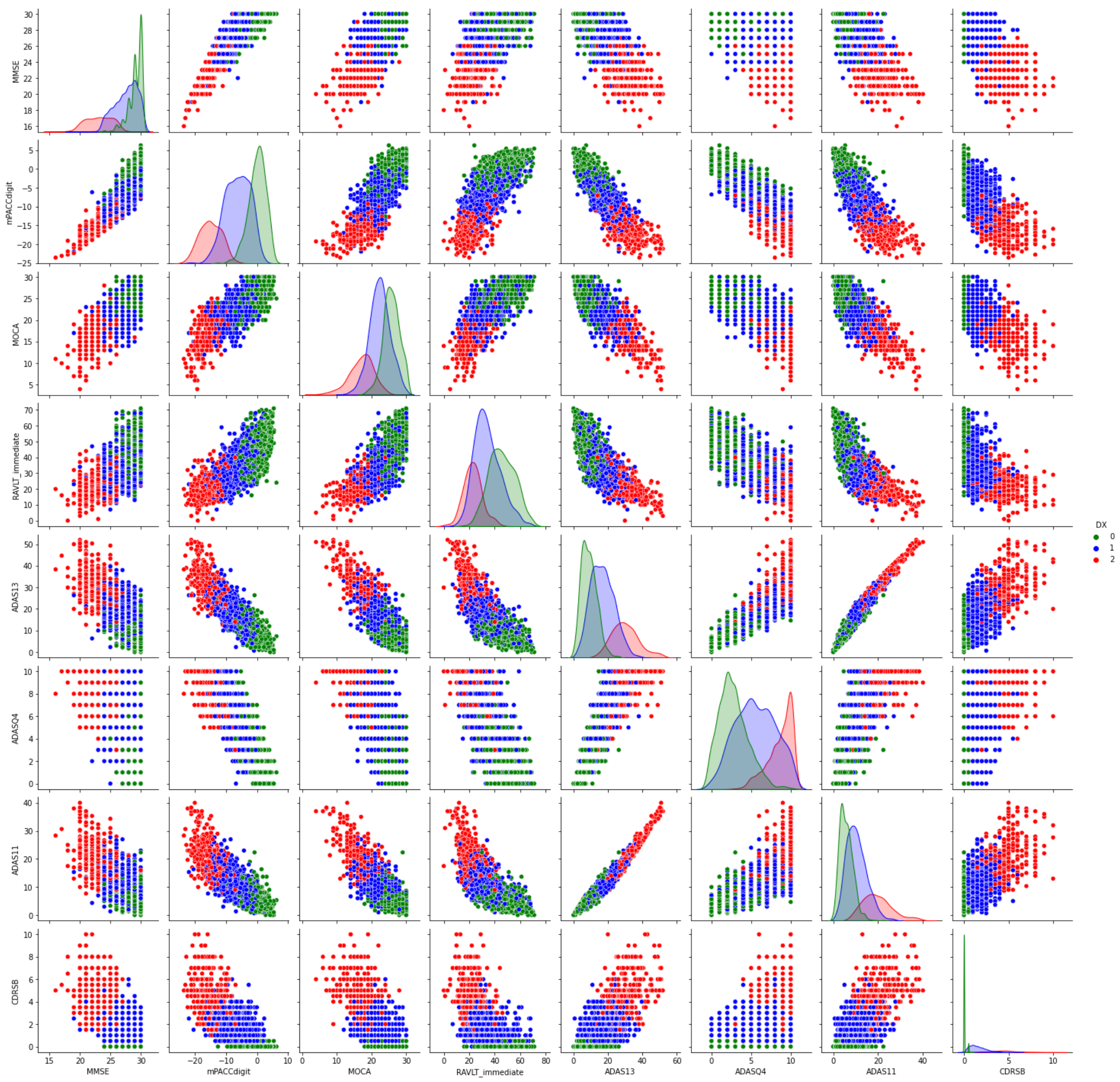

*Figure 4.71 Pair Plot representing Correlation among Multiple Numerical Variables*

## 4.5 Model Implementation

Figure 4.72 depicts the steps involved in implementing classification algorithms to develop a prediction model for identifying early stages of Alzheimer's disease using the ADNI dataset used in this study. Six classification algorithms, namely Logistic regression, Light GBM, Extra Tree, Bagging KNN, Stacking Blending and Artificial neural network were employed in the study. The process flow of these classifiers is shown in the diagram below.

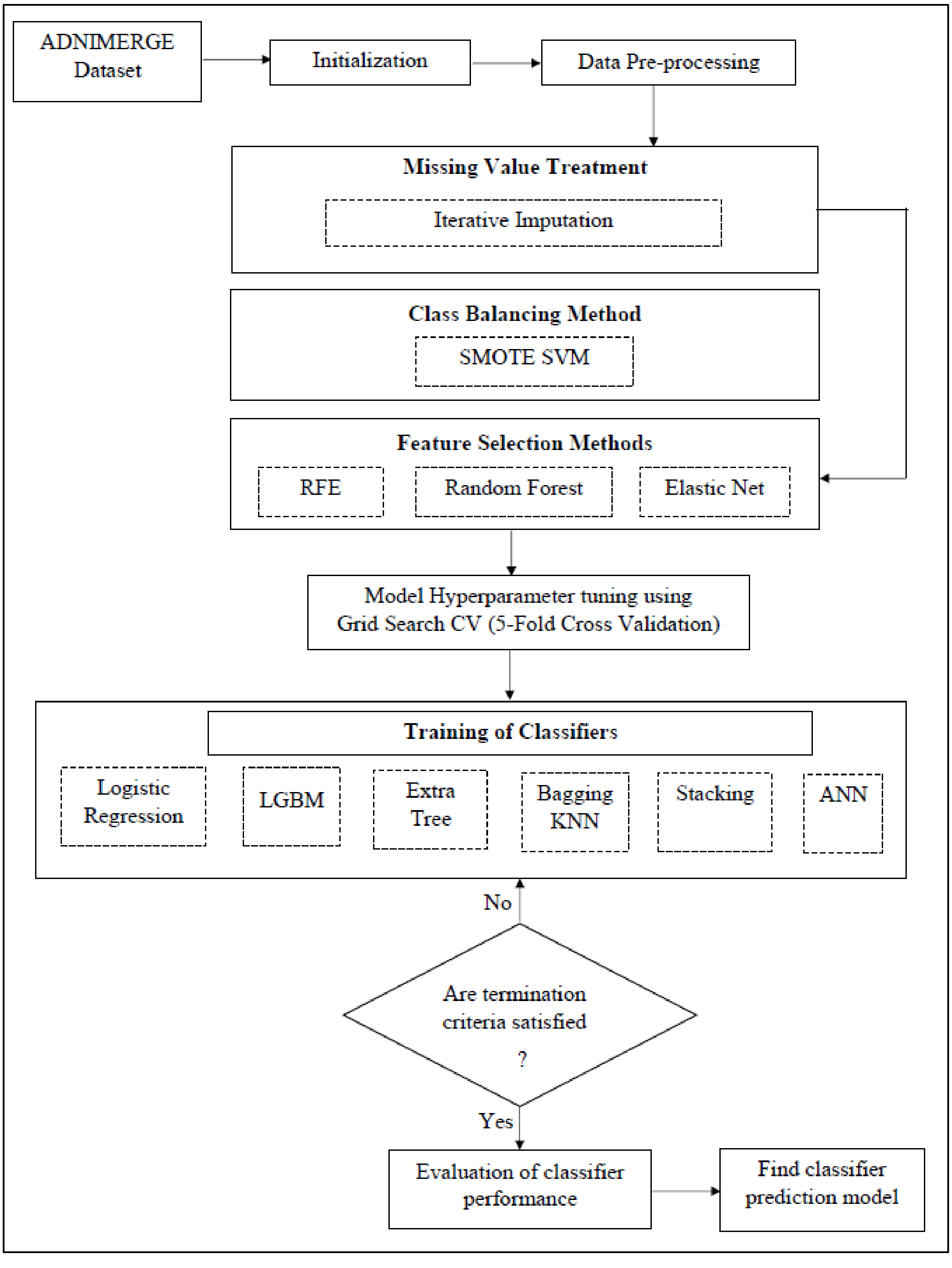

*Figure 4.72 Process flow of Classifiers*

### 4.5.1 Initialization of ADNI Dataset

The Alzheimer's Disease Neuroimaging Initiative Study (ADNI) dataset, which was in.csv format, was imported and loaded into a Jupiter notebook using Python version 3.8.5. There were 116 attributes with 15,762 instances when the data was loaded. The type for each selected characteristic was shown, and the parameters in this dataset were of diverse data types, including continuous, categorical, and datetime. The target class variable was assigned to the categorical variable DX.

### 4.5.2 Data Pre-processing

The "NaN" values, or missing values, were analysed and replaced. Missing values were imputed using the iterative imputation technique for all numerical attributes and simple imputation (most frequent value) for categorical attributes, except for a few (16 attributes) features which were not null. Iterative Imputer was implemented with hyperparameters max_iter as 100, random_state as 42 and rest all default parameter settings. After importing the entire dataset as a.csv file into the local jupyter notebook, the missing values were replaced. Also, some of the unrelated features such as RID, PTID, COLPROT, VISCODE etc. and attributes with more than 45% missing values like PIB, FBB, DIGITSCOR, AV45, ABETA, TAU, and PTAU were discarded from the dataset. Using the default split percentage, the entire dataset was then partitioned into training and test datasets.

### 4.5.3 Class Balancing

The class rebalancing algorithms were employed to a training dataset of 1,500 instances. These balancing procedures were used to address the target variable's class imbalance, DX (Diagnosis). An advance version of Synthetic Minority Oversampling Technique, namely Borderline SVM-SMOTE balancing technique was only used for class imbalance handling. For SVM-SMOTE, the default hyperparameters were applied along with random state of 42 for imbalance management. The use of these approach modifies the number of instances in the training dataset. The model is also built using the imbalanced dataset as well for comparative analysis. The number of instances for each of these techniques are listed below.

1) Training dataset before imbalance handling – 1,500 records (for model training)
2) Training dataset after imbalance handling using SVM-SMOTE – 2,073 records (for model training)

### 4.5.4 Feature Selection Techniques

Feature selection is a technique for limiting the input variable before feeding to any machine learning model by only using useful data and eliminating noise. The number of predictors or features used to train the machine learning model is directly proportional to its prediction power. Feature selection can be done both manually and automatically by limiting the number of features based on any statistical relationship. In this study, three automated feature selection techniques have been implemented two times separately, once before applying class balancing technique and once after handling the class balancing problem.

Below tables 4.33 listed down the hyper parameters used during the wrapper-based feature selection process using Recursive feature elimination technique.

*Table 4.33 Hyperparameters used for RFE Feature Selection Method*

| Hyperparameter Name | Hyperparameter Value |
|---|---|
| estimator | LogisticRegression |
| n_features_to_select | 45 |

Table 4.34 shows the hyperparameters of embedded feature selection technique, called RandomForest classifier (Tree based) used in this study.

*Table 4.34 Hyperparameters used for RandomForest Feature Selection Method*

| Hyperparameter Name | Hyperparameter Value |
|---|---|
| random_state | 42 |
| bootstrap | True |
| max_samples | 15 |
| n_estimators | 500 |

Below table 4.35 displays Elastic Net feature selection method, implemented via logistic regression, along with its hyper parameter.

*Table 4.35 Hyperparameters used for Elastic Net Feature Selection Method*

| Hyperparameter Name | Hyperparameter Value |
|---|---|
| penalty | elasticnet |
| l1_ratio | 0.1 |
| solver | saga |
| random_state | 42 |
| multi_class | multinomial |

Finally, as shown in figure 4.7 (class imbalanced set) and figure 4.8 (class balanced set), the top 40 independent characteristics were chosen from a total of 57 independent features for further processing and classifiers training/modelling.

### 4.5.5 Training of Classification Models

The classification model was trained using the techniques Logistic regression, Light GBM, Extra Tree, KNN bagging classifier, stacking blending, and Artificial neural network on the training dataset, one with class balancing methods used and the other without. The models were validated using Grid search CV with 5-fold cross validation, with one-fold used for testing and the other four for training. This method was performed five times until each fold had been used as a test set. Each of the modelling strategies, as well as the best tuned hyperparameters utilised to train them, are presented in detail in the sub-sections below on both the imbalance and class balanced training sets. The hyper parameters have been chosen based on grid search cross validation and monitoring the f1-weighted score. For the combination of hyperparameter values, the monitored metric value came as highest, that parameter values were considered as best set of hyper parameters.

#### 4.5.5.1 Modelling with Logistic Regression

A classification model was implemented using a basic multiclass logistic regression; this classifier was trained twice, once without the imbalance-handled dataset and once after the class imbalance treatment (with the SMOTE SVM method).

#### 4.5.5.1.1 Training without Class Imbalance Handling

The training dataset, which was used to train the model before imbalance handling, has 1,500 entries. Hyperparameters utilised for logistic regression model training without class imbalance handling are listed below in table 4.36. The rest of the hyperparameters were set to their default values.

*Table 4.36 Hyperparameters used for Logistic Regression on Imbalanced dataset*

| Hyperparameter Name | Hyperparameter Value |
|---|---|
| multi_class | multinomial |
| penalty | l1 |
| C | 0.6 |
| random_state | 42 |
| solver | saga |

#### 4.5.5.1.2 Training after Class imbalance Handling using SVM-SMOTE

The training dataset, which was used to train the model after applying imbalance handling strategy using SMOTE SVM, has 2,073 entries. Hyperparameters utilised for logistic regression model training after class imbalance handling are listed below in table 4.37. The rest of the hyperparameters were set to their default values.

*Table 4.37 Hyperparameters used for Logistic Regression on Balanced dataset*

| Hyperparameter Name | Hyperparameter Value |
|---|---|
| multi_class | multinomial |
| penalty | l2 |
| C | 3.6 |
| random_state | 42 |
| solver | newton-cg |

### 4.5.5.2 Modelling with Extra Tree Classifier

A classification model was implemented using a tree-based bagging classifier model Extra Tree; this classifier was trained twice, once without the imbalance-handled dataset and once after the class imbalance treatment (with the SMOTE SVM method).

#### 4.5.5.2.1 Training without Class Imbalance Handling

The training dataset, which was used to train the model before imbalance handling, has 1,500 entries. Hyperparameters utilised for extra tree model training without class imbalance handling are listed below in table 4.38. The rest of the hyperparameters were set to their default values.

*Table 4.38 Hyperparameters used for Extra Tree on Imbalanced dataset*

| Hyperparameter Name | Hyperparameter Value |
|---|---|
| criterion | entropy |
| max_depth | 8 |
| max_features | 22 |
| min_samples_leaf | 3 |
| min_samples_split | 50 |
| n_estimators | 50 |
| random_state | 42 |

#### 4.5.5.2.2 Training after Class Imbalance Handling using SVM-SMOTE

The training dataset, which was used to train the model after applying imbalance handling strategy using SMOTE SVM, has 2,073 entries. Hyperparameters utilised for extra tree model training after class imbalance handling are listed below in table 4.39. The rest of the hyperparameters were set to their default values.

*Table 4.39 Hyperparameters used for Extra Tree on Balanced dataset*

| Hyperparameter Name | Hyperparameter Value |
|---|---|
| criterion | entropy |
| max_depth | 8 |
| max_features | 22 |
| min_samples_leaf | 3 |
| min_samples_split | 100 |
| n_estimators | 100 |
| random_state | 42 |

#### 4.5.5.3 Modelling with Light Gradient Boosting Machine

A classification model was implemented using a tree-based boosting classifier model Light Gradient Boosting Machine (LGBM); this classifier was trained twice, once without the imbalance-handled dataset and once after the class imbalance treatment (with the SMOTE SVM method).

##### 4.5.5.3.1 Training without Class Imbalance Handling

The training dataset, which was used to train the model before imbalance handling, has 1,500 entries. Hyperparameters utilised for LGBM model training without class imbalance handling are listed below in table 4.40. The rest of the hyperparameters were set to their default values.

*Table 4.40 Hyperparameters used for LGBM on Imbalanced dataset*

| Hyperparameter Name | Hyperparameter Value |
|---|---|
| colsample_bytree | 1 |
| learning_rate | 0.006 |
| max_depth | 4 |
| min_child_samples | 150 |
| n_estimators | 500 |
| num_leaves | 50 |
| objective | multiclass |
| subsample | 0.1 |
| random_state | 42 |

##### 4.5.5.3.2 Training after Class Imbalance Handling using SVM-SMOTE

The training dataset, which was used to train the model after applying imbalance handling strategy using SMOTE SVM, has 2,073 entries. Hyperparameters utilised for Light GBM model training after class imbalance handling are listed below in table 4.41. The rest of the hyperparameters were set to their default values.

*Table 4.41 Hyperparameters used for LGBM on Balanced dataset*

| Hyperparameter Name | Hyperparameter Value |
|---|---|
| colsample_bytree | 0.5 |
| learning_rate | 0.5 |
| max_depth | 6 |
| min_child_samples | 100 |
| n_estimators | 150 |
| num_leaves | 50 |
| objective | multiclass |
| subsample | 0.1 |
| reg_alpha | 10 |
| random_state | 42 |

#### 4.5.5.4 Modelling with Bagging KNN Classifier

A classification model was implemented using a KNN-based bagging classifier model Bagging KNN classifier; this classifier was trained twice, once without the imbalance-handled dataset and once after the class imbalance treatment (with the SMOTE SVM method).

##### 4.5.5.4.1 Training without Class Imbalance Handling

The training dataset, which was used to train the model before imbalance handling, has 1,500 entries. Hyperparameters utilised for Bagging KNN model training without class imbalance handling are listed below in table 4.42. The rest of the hyperparameters were set to their default values.

*Table 4.42 Hyperparameters used for Bagging KNN on Imbalanced dataset*

| Classifier Name | Hyperparameter Name | Hyperparameter Value |
|---|---|---|
| BaggingClassifier | base_estimator | KNeighborsClassifier |
| | max_features | 20 |
| | max_samples | 500 |
| | n_estimators | 20 |
| | random_state | 42 |

| Hyperparameter’s for Base Estimator: | | |
|---|---|---|
| KNeighborsClassifier | n_neighbors | 5 |
| | p | 1 |
| | weights | distance |

#### 4.5.5.4.2 Training after Class Imbalance Handling using SVM-SMOTE

The training dataset, which was used to train the model after applying imbalance handling strategy using SMOTE SVM, has 2,073 entries. Hyperparameters utilised for Bagging KNN model training after class imbalance handling are listed below in table 4.43. The rest of the hyperparameters were set to their default values.

*Table 4.43 Hyperparameters used for Bagging KNN on Balanced dataset*

| Classifier Name | Hyperparameter Name | Hyperparameter Value |
|---|---|---|
| BaggingClassifier | base_estimator | KNeighborsClassifier |
| | max_features | 15 |
| | max_samples | 500 |
| | n_estimators | 100 |
| | random_state | 42 |
| **Hyperparameter’s for Base Estimator:** | | |
| KNeighborsClassifier | n_neighbors | 5 |
| | p | 1 |
| | weights | distance |

#### 4.5.5.5 Modelling with Stacking-Based Ensemble Learning

A classification model was implemented using a stacking classifier model Stacking-Based Ensemble Learning classifier; this classifier was trained twice, once without the imbalance-handled dataset and once after the class imbalance treatment (with the SMOTE SVM method).

#### 4.5.5.5.1 Training without Class Imbalance Handling

The training dataset, which was used to train the model before imbalance handling, has 1,500 entries. Hyperparameters utilised for Stacking-Based Ensemble Learning model training without class imbalance handling are listed below in table 4.44. The rest of the hyperparameters were set to their default values.

*Table 4.44 Hyperparameters used for Stacking-Ensemble Model on Imbalanced dataset*

| Classifier Name | Hyperparameter Name | Hyperparameter Value |
|---|---|---|
| StackingClassifier | estimators | LogisticRegression |
| | | LGBMClassifier |
| | | ExtraTreesClassifier |
| | | BaggingClassifier |
| | final_estimator | LogisticRegression(multi_class='multinomial') |
| | cv | StratifiedKFold(n_splits=5) |
| **Hyperparameter's for Estimators:** | | |
| LogisticRegression | c | 0.6 |
| | multi_class | multinomial |
| | penalty | l1 |
| | random_state | 42 |
| | solver | saga |
| | | |
| LGBMClassifier | colsample_bytree | 1 |
| | learning_rate | 0.006 |
| | max_depth | 4 |
| | min_child_samples | 150 |
| | n_estimators | 500 |
| | num_leaves | 50 |
| | objective | multiclass |
| | subsample | 0.1 |
| | random_state | 42 |
| | | |
| ExtraTreesClassifier | criterion | entropy |
| | max_depth | 8 |
| | max_features | 22 |
| | min_samples_leaf | 3 |

<table>
<tr><td rowspan="3"></td><td>min_samples_split</td><td>50</td></tr>
<tr><td>n_estimators</td><td>50</td></tr>
<tr><td>random_state</td><td>42</td></tr>
<tr><td colspan="3"></td></tr>
<tr><td rowspan="5">BaggingClassifier</td><td>base_estimator</td><td>KNeighborsClassifier<table><tr><td>n_neighbors</td><td>5</td></tr><tr><td>p</td><td>1</td></tr><tr><td>weights</td><td>distance</td></tr></table></td></tr>
<tr><td>max_features</td><td>20</td></tr>
<tr><td>max_samples</td><td>500</td></tr>
<tr><td>n_estimators</td><td>20</td></tr>
<tr><td>random_state</td><td>42</td></tr>
</table>

#### 4.5.5.5.2 Training after Class Imbalance Handling using SVM-SMOTE

The training dataset, which was used to train the model after applying imbalance handling strategy using SMOTE SVM, has 2,073 entries. Hyperparameters utilised Stacking-Based Ensemble Learning model training after class imbalance handling are listed below in table 4.45. The rest of the hyperparameters were set to their default values.

*Table 4.45 Hyperparameters used for Stacking-Ensemble Model on Balanced dataset*

| **Classifier Name** | **Hyperparameter Name** | **Hyperparameter Value** |
|---|---|---|
| StackingClassifier | estimators | LogisticRegression |
| | | LGBMClassifier |
| | | ExtraTreesClassifier |
| | | BaggingClassifier |
| | final_estimator | LogisticRegression(multi_class='multinomial') |
| | cv | StratifiedKFold(n_splits=5) |
| **Hyperparameter's for Estimators:** | | |
| LogisticRegression | multi_class | multinomial |
| | penalty | l2 |
| | C | 3.6 |

<table>
<tr><td rowspan="2"></td><td>random_state</td><td>42</td></tr>
<tr><td>solver</td><td>newton-cg</td></tr>
<tr><td colspan="3"></td></tr>
<tr><td rowspan="10">LGBMClassifier</td><td>colsample_bytree</td><td>0.5</td></tr>
<tr><td>learning_rate</td><td>0.5</td></tr>
<tr><td>max_depth</td><td>6</td></tr>
<tr><td>min_child_samples</td><td>100</td></tr>
<tr><td>n_estimators</td><td>150</td></tr>
<tr><td>num_leaves</td><td>50</td></tr>
<tr><td>objective</td><td>multiclass</td></tr>
<tr><td>subsample</td><td>0.1</td></tr>
<tr><td>reg_alpha</td><td>10</td></tr>
<tr><td>random_state</td><td>42</td></tr>
<tr><td colspan="3"></td></tr>
<tr><td rowspan="7">ExtraTreesClassifier</td><td>criterion</td><td>entropy</td></tr>
<tr><td>max_depth</td><td>8</td></tr>
<tr><td>max_features</td><td>22</td></tr>
<tr><td>min_samples_leaf</td><td>3</td></tr>
<tr><td>min_samples_split</td><td>100</td></tr>
<tr><td>n_estimators</td><td>100</td></tr>
<tr><td>random_state</td><td>42</td></tr>
<tr><td colspan="3"></td></tr>
<tr><td rowspan="5">BaggingClassifier</td><td>base_estimator</td><td>KNeighborsClassifier<table><tr><td>n_neighbors</td><td>5</td></tr><tr><td>p</td><td>1</td></tr><tr><td>weights</td><td>distance</td></tr></table></td></tr>
<tr><td>max_features</td><td>15</td></tr>
<tr><td>max_samples</td><td>500</td></tr>
<tr><td>n_estimators</td><td>100</td></tr>
<tr><td>random_state</td><td>42</td></tr>
</table>

#### 4.5.5.6 Modelling with Artificial Neural Network

An artificial neural network (ANN) is a complex model made up of many processing layers that accept inputs and outputs based on their activation functions. In this study, the results of the classical supervised algorithms prediction results will be compared against this neural architecture. A classification model was implemented using an Artificial Neural Network (ANN); this classifier was trained twice, once without the imbalance-handled dataset and once after the class imbalance treatment (with the SMOTE SVM method).

##### 4.5.5.6.1 Training without Class Imbalance Handling

The training dataset, which was used to train the model before imbalance handling, has 1,500 entries. Hyperparameters utilised for artificial neural network model training without class imbalance handling are listed below in table 4.46. The rest of the hyperparameters were set to their default values.

*Table 4.46 Hyperparameters used for ANN on Imbalanced dataset*

| Layers | Hyperparameter Name | Hyperparameter Value |
|---|---|---|
| Sequential () | layers | none |
| | name | none |
| Dense | units | 64 |
| | input_dim | 40 |
| | activation | relu |
| Dense | units | 128 |
| | activation | relu |
| Dropout | rate | 0.33 |
| Dense | units | 3 |
| | activation | softmax |

##### 4.5.5.6.2 Training after Class Imbalance Handling using SVM-SMOTE

The training dataset, which was used to train the model after applying imbalance handling strategy using SMOTE SVM, has 2,073 entries. Hyperparameters utilised for ANN model training after class imbalance handling are listed below in table 4.47. The rest of the hyperparameters were set to their default values.

*Table 4.47 Hyperparameters used for ANN on Balanced dataset*

| Layers | Hyperparameter Name | Hyperparameter Value |
|---|---|---|
| Sequential () | layers | none |
| | name | none |
| Dense | units | 64 |
| | input_dim | 40 |
| | activation | relu |
| BatchNormalization | momentum | 0.99 |
| | epsilon | 0.001 |
| | beta_initializer | zeros |
| | gamma_initializer | ones |
| Dropout | rate | 0.33 |
| Dense | units | 128 |
| | activation | relu |
| BatchNormalization | momentum | 0.99 |
| | epsilon | 0.001 |
| | beta_initializer | zeros |
| | gamma_initializer | ones |
| Dropout | rate | 0.4 |
| Dense | units | 3 |
| | activation | softmax |

### 4.5.6 Evaluation of Classifier Performance

The performance evaluation criteria were used to evaluate the classifiers that were built using the training datasets. Each of the classifiers was used twice in the modelling process. Once all classifiers have been fitted with the unbalanced training dataset. After that, the class balanced training dataset is utilised to develop the model, and the results are compared to the performance of the unbalanced model. The performance measurements will be used to evaluate all the classifiers developed from the chosen class unbalanced and balanced dataset to determine the best predictive model. For multiclass class labels or multiclass classification, the confusion matrix for each classifier was utilized to extract various measures such as accuracy, weighted average F1 score, weighted average precision, weighted average recall, and the ROCAUC curve.

The classifiers were contrasted after evaluating all the performance parameters produced from the confusion matrix. The classifier with the overall best effectiveness was chosen as the final classification or prediction model for early-stage detection of Alzheimer's disease.

If class label No is considered as the negative case and class label Yes is considered as the positive case, the TP, FP, FN, and TN are presented in the sequence stated in below table 4.48 confusion matrix.

*Table 4.48 General confusion matrix for binary classification*

| | | **Predicted** | |
|---|---|---|---|
| | | **No** | **Yes** |
| **Actual** | **No** | TN | FP |
| | **Yes** | FN | TP |

In the above table, TP means the number of cases correctly defined as positive case, TN means the number of cases correctly defined as negative case, FP means the number of cases incorrectly defined as positive case, FN means the number of cases incorrectly defined as negative case. After analysing all the confusion matrix's performance parameters, the classifiers were compared. The final classification or prediction model for early stage of dementia was chosen as the classifier with the highest overall effectiveness.

## 4.6 Summary

Overall, this chapter described the dataset used in this study as well as how the data was used to do analysis. The data for this study obtained from the Alzheimer's Disease Neuroimaging Initiative database, which contained 115 Alzheimer's-related risk variables. The target variable, DX (Diagnosis), indicates the stage of Alzheimer's disease that a patient is experiencing. This dataset had a total of 15,762 records when it was first imported. The baseline investigation was extracted from a subset of the dataset containing 2,379 records and 65 characteristics, as this study focused on early-stage diagnosis. To make the data easier to analyse, unrelated variables were removed, and categorical variables were turned into numerical variables using dummification to make the analysis and modelling process easier. Using the iterative/ mode imputation technique, missing data were detected and imputed.

The exploratory data analysis was conducted using the ADNI dataset with imputed missing values. To further understand the distribution of values in each variable, a univariate analysis was used. The summary of descriptive statistics shown in frequency tables, as well as the count plot or percentage plot for a few key variables, were analysed in depth. The percentage analysis for each target class, as well as the variable relationships, were checked and presented as part of bivariate and correlation analysis. The data visualisation improved understanding of the dataset, and each visualisation report indicated if the predictor variable has any influence on the development of dementia or Alzheimer's disease. The report also includes graphs depicting the association between demographic and cognitive test-related characteristics in order to see if the combined relationship has any effect on the outcome variable, Alzheimer's Diagnosis.

The original dataset with imputed missing values was then randomly divided into two sets, with the training set accounting for 70% and the test set for 30%. The analysis revealed a variety of patterns across the dementia factors; however, the variable DX had more classes labelled as normal or mild dementia than those labelled as Dementia. As a result, the findings favoured the class without dementia. On the training dataset, a class balancing method, namely SMOTE SVM, was used given this insight. Grid Search cross validation technique with 5 folds was used to tune the six classifiers. These classifiers were trained on both the imbalanced and balanced training datasets, and then tested against the test dataset. At the end of the process, 12 prediction models were created. Finally, the confusion matrix was shown for each of the three classes to derive the model's performance measures. The algorithm that produced the best prediction model on the class-balanced data was chosen based on a comparison of the performances.

# CHAPTER 5

# RESULTS AND DISCUSSION

## 5.1 Introduction

This chapter will contain all the findings from the analysis as well as the output results from the Python libraries version 3.8.5 that were used in this study. The results of the assessment of several modelling strategies with balanced and unbalanced data will be discussed in detail. The confusion matrices obtained from the development of the six classifiers (Logistic regression, LGBM, Bagging KNN, Extra Tree, Stacking Blending, and ANN) using both class balanced and class imbalanced training datasets and examined with a test dataset using SMOTE SVM class balancing technique will be displayed and interpreted in more detail. As a result, each classifier will have two confusion matrices derived from testing the models on the test set, which was previously trained using the following datasets:

i) Dataset before class imbalance treatment,
ii) Dataset after class imbalance treatment with SMOTE SVM method.

This sub-chapter will interpret twelve confusion matrices resulting from the use of six classifiers. A table comparing the balancing approach applied on the training set and evaluated on the test set will be shown with the classifiers' performance measurements for each classification model. Because this is a multiclass problem, these measurements for target class DX (Alzheimer's Diagnosis) comprise accuracy, weighted recall, weighted precision, weighted F-measure, and ROC curve for the three class labels. The classifier with the best overall performance will be chosen as the final prediction model for Alzheimer's diagnosis in its early stages. Based on the balancing method and classification model, this study will highlight the best approach for predicting Alzheimer's illness.

## 5.2 Significant Biomarkers from Visualization and Feature Selection

The varied types of visualizations were explained in Chapter 4. It showed how the demographic and clinical variables (cognitive test results) interacted with the target variable, DX (Alzheimer's diagnosis). The correlations between the demographic and clinical variables were also portrayed to identify any possible patterns or trends between these factors associated with Alzheimer's disease. The interactions and visualizations that provide an essential and deeper insight into the ADNI dataset were analysed, which provides an overview of all the variables in the dataset as well as a summary of the relationships between the variables.

The important biological indicators or variables that were found via feature selection (through RFE and Random Forest) were then visualised, providing a critical and deeper insight into the Alzheimer's dataset as well as a synopsis of their relationships. The details of such attributes are discussed in this sub-section. Patients suffering from dementia from various age groups participated in the ADNI trial. The bulk of those taking part in this study are between the ages of 70 and 80 years old. 60 to 70 and 80 to 90 years old are the second and third most populous age groups, respectively. Only a few of the participants in this study are between the ages of 50 and 60. According to this facts, Alzheimer's disease primarily affects the elderly citizens.

Variable CDRSB, a cognitive test, has been proven to clearly represent a good separability among the three target classes (CN, MCI, and Dementia). Dementia-affected people's CDRSB ratings are often in the 4 plus range. While a perfect indicator of cognitive normalcy is a score of zero. This score usually ranges from 1 to 3 for mildly afflicted people. Another psychological test-related variable, the PACC digit test rating (mPACCdigit), for dementia patients ranges from -25 to -5. While -15 to -5 represents an overlap between AD and MCI. This score is usually greater than 1 for healthy people. Therefore, this is also a clear identifier among the phases of dementia.

Similarly, RAVLT_immediate, an auditory verbal learning test, showed that the ratings for dementia affected people mostly ranged between 0 to 40. While value 15 to 40 is an overlapping between AD and MCI. This score is usually higher than 60 in healthy people. So, from this rating, one can distinguish between AD and non-AD patients. The attribute Alzheimer's assessment scale with 13 items (ADAS13) also showed a difference in test scores between normal and dementia patients. This rating ranges from 0 to 20 for the control group, and from 15 to 50 for Alzheimer's patients. The range observed for the variable everyday cognitive test completed by the study partner (EcogSPTotal) among the dementia patients is between 1.5 and 4. But for the majority of normal people, this score is 1.0. It can also be a good indicator between normal and affected people.

Furthermore, another mental test, namely the MMSE examination rating for dementia patients, typically ranges from 16 to 26. While the range of 24 to 26 overlaps between AD and MCI subjects. This score is usually around 30 for healthy people. As a result, this variable can provide an extremely precise identification of target groups.

From the visualization of demographic variables, different patterns are noted to classify the target groups. From participant's gender (PTGENDER) it is observed that mostly male people are part of this Alzheimer's Disease Neuroimaging Initiative (ADNI) study. Also, males are the biggest sufferers compared to female participants. Males are affected by dementia at a rate of 19.67%, while females are affected at a rate of 17.52%. On the other hand, variable PTRACCAT (patient's race) showed that among white and Asian people, the percentage of dementia sufferers is slightly higher than others (19.09% and 19.14%, respectively). Another demographic variable PTETHCAT (patient's ethnicity) showed that most of the participants from the 'Hisp/Latino' group had been diagnosed as normal (45.00%), whereas people from the ‘Not Hisp/Latino’ ethnic group had the highest percentage (46.18%) of mild cognitive impairment as well as dementia (around 18.91%). Furthermore, it was also seen that married people are slightly more susceptible to dementia (around 20.43%) than never married or divorced people.

Multivariate correlation analysis was also used to examine the predictors' interdependence and the combined effect of these features on the target class variable, DX. CDRSB and ADAS13, as well as CDRSB and EcogSPTotal, have a strong positive association. It has a good negative linear association with mPACCdigit, MMSE, and MOCA. It is clear from the graphic that a combination of these variables can effectively separate the target classes, DX (Alzheimer's Diagnosis). MMSE and RAVLT immediate have a good linearly increasing association; MMSE and mPACCdigit have a favourable relationship. When combined, these factors show a high level of separability among target classes. A strong positive link was found between mPACCdigit and RAVLT immediate, while a strong negative relationship was found between mPACCdigit and ADAS13. Besides this, the MOVCA and mPACCdigit clinical variables have a strong positive association, and when used together, they can be used to determine the condition of a patient's sickness. Together, these variables can distinguish patients who are suffering from dementia from mildly affected people.

Apart from the above key variables, there are some more important features like Apolipoprotein E4, Entire Brain Volume, Fusiform, size of Hippocampus, complex brain structure embedded deep in the temporal lobe, Functional Activities Questionnaire, participant’s education level, and Logical Memory-Delayed Recall, etc., based on which doctors or physicians can get a quick overview of the factors related to the fatal consequences of acute Alzheimer's patients, and these features can act as a guideline for dementia patients in understanding the impact of certain risk variables when predicting the stages of this disorder.

### 5.3 Evaluation of Balancing Methods and Results on Test Dataset

The performance of each classifier was evaluated using their balancing approaches, and the classifier with the best overall performance was chosen as the best prediction model for Alzheimer's disease early-stage diagnosis. The performance of the classifiers was evaluated using many assessment metrics, including the properly classified instances percentage or accuracy, weighted average F1 score, weighted average precision, weighted average recall, and the ROCAUC curve for this multiclass classification. Because the Alzheimer's dataset used in this study involves medical data, specific evaluation measures are critical in assessing the prediction model constructed using an algorithm. The accuracy, sensitivity or recall, precision, and AUCROC area are the four metrics.

A higher true positive rate for an ideal disease prediction model in the medical domain shows that diseased patients are properly predicted, which is the study's purpose. A greater true negative rate is likewise preferable, but this metric is not quite as important as the true positive rate. It is critical for a prediction model to appropriately detect the existence of any disease and avoid misdiagnosis. Patients may die because of misleading results caused by a false negative rate or a false positive rate, as severe dementia is a fatal disease, and the earlier it is diagnosed, the greater the possibilities of survival.

Usually, the number of appropriately and erroneously classified instances described in the confusion matrix table illustrates the four components, which are TP, FP, TN, and FN value. Since this study is dealing with a multi-class classification problem, these TP, TN, FP, and FN values are not directly obtained as in the binary classification problem. In a multi-class problem, with respect to each class, these TP, TN, FP, and FN values need to be calculated. Below, figure 5.1 represents a sample confusion matrix for multi-class classification problems, created from an early-stage dementia detection predictive model, which displays that the True Positive (TP) value is the condition at which the actual and anticipated values are equal. The True Negative (TN) value for a class will be the sum of the values of all columns and rows except the values of that class that are being calculated. The False Negative (FP) value for a class will be the sum of the values of analogous rows excluding the TP value. The False Positive (FP) value for a class will be the sum of the values of the related columns aside from the TP value. Figure 5.1 depicts the TP, TN, FP, and FN in relation to the Dementia class (Class label 2). Similarly, the confusion matrix can be used to calculate these four components for each class. This will be applied to explain the confusion matrices in below sub-sections.

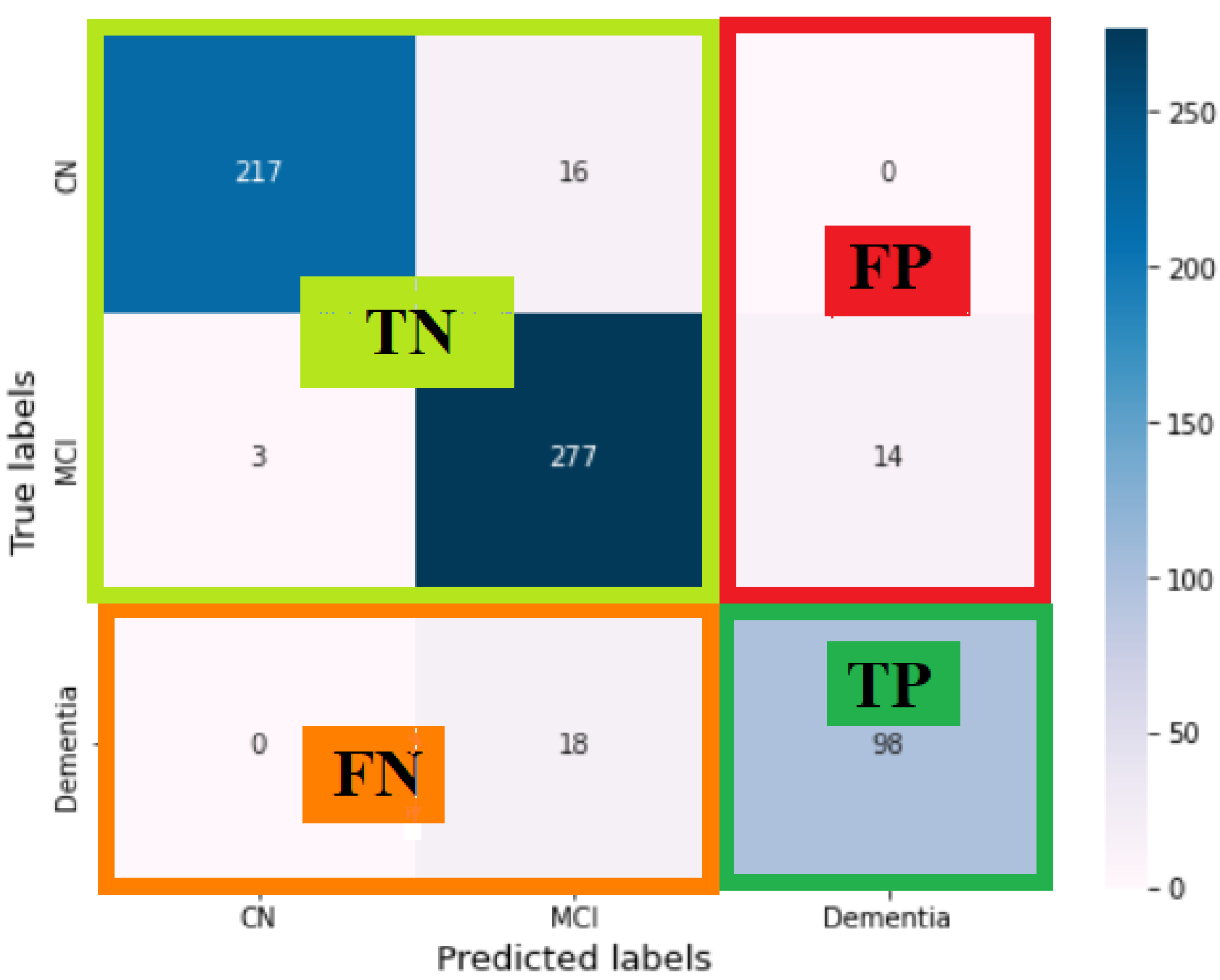

*Figure 5.1 Sample Confusion Matrix for Dementia Detection from ADNI Dataset*

The main goal of this study is to correctly predict the early phases of dementia among elderly population, and the fact that there are three target classes or diagnosis outcomes that the model must correctly predict makes it a multiclass classification problem, therefore the important assessment metrics for this multiclass classification task would be weighted average F1 score, weighted average precision, weighted average recall, and the ROCAUC curve. Also, the individual target class's recall or precision from each modelling approach, as well as the results of each balancing method, will be compared and represented in the performance tables.

### 5.3.1 Logistic Regression Classifier Evaluation and Results

For building a classification model, a basic multiclass logistic regression was utilised; this classifier was trained twice, once without imbalance handling and once after class imbalance correction (with SMOTE SVM minority class oversampling method). The evaluations/results of the classifier on the test set, as well as performance metrics such as the confusion matrix, weighted average F1 score, weighted average precision, weighted average recall, and the ROCAUC curve, will be shown in the subsections below.

#### 5.3.1.1 Evaluation and Results without Class Balancing

The training dataset, which was used to train the model before imbalance handling, has 1,500 entries. The model's performance was evaluated using a test dataset of 643 records. Below, figure 5.2 refers to the without class balancing confusion matrix for the Logistic Regression model, where rows represent actual or true class labels and columns represent predicted class labels. The TP, TN, FP, and FN for this multi-class problem have been calculated following the logic depicted in figure 5.1. For the cognitive normal (CN) class, there are 227 instances classified as CN that were predicted correctly (TP). The actual or true class values of non-CN classes, which have been predicted correctly, are 404 instances (TN). There are 6 instances which have been observed (true labels) as non-CN but are predicted as CN class (FP). At the same time, 6 instances that were predicted as non-CN where the true class value is given as CN class (FN). For the MCI class, there are 276 instances classified as MCI that were predicted correctly (TP). The actual or true class values of non-MCI classes, which have been predicted correctly, are 328 instances (TN). There are 21 instances which have been observed (true labels) as non-MCI but predicted as MCI class (FP). At the same time, 18 instances that were predicted as non-MCI where the true class value are given as MCI class (FN). For the dementia class, there are 101 instances classified as dementia that were predicted correctly (TP). The actual or true class values of non-dementia classes, which have been predicted correctly, are 515 instances (TN). There are 12 instances which have been observed (true labels) as non-dementia but predicted as dementia class (FP). At the same time, 15 instances that were predicted as non-dementia where the true class value is given as dementia class (FN).

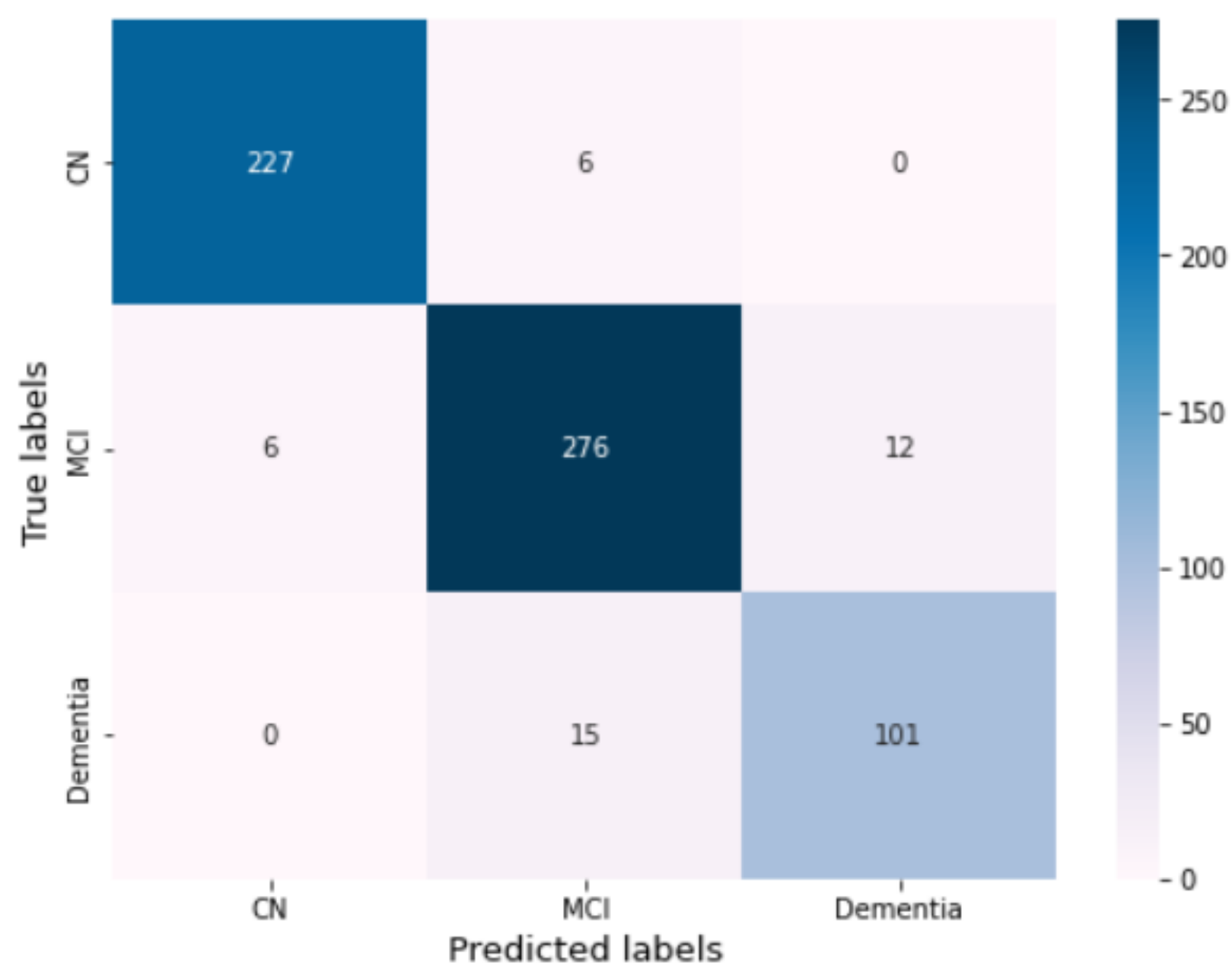

*Figure 5.2 Confusion Matrix for Logistic Regression without Class Balancing*

The comprehensive performance metrics for this multiclass classification task, as well as the precision, recall, and f1-score for each class, are provided in table 5.1. Additionally, for measuring overall model performance, the weighted average precision, weighted average recall, weighted average f1-score, and accuracy are all 94%.

*Table 5.1 Classification Report for Logistic Regression without Class Balancing*

| Target Class | Precision | Recall | f1-score | weighted avg Precision | weighted avg Recall | weighted avg f1-score | Accuracy |
|---|---|---|---|---|---|---|---|
| CN | 0.97 | 0.97 | 0.97 | 0.94 | 0.94 | 0.94 | 0.94 |
| MCI | 0.93 | 0.94 | 0.93 | | | | |
| Dementia | 0.89 | 0.87 | 0.88 | | | | |

The ROC curve presented in figure 5.3 shows how well the model predicts each class label. In addition, the whole model's weighted average AUCROC score is 99%.

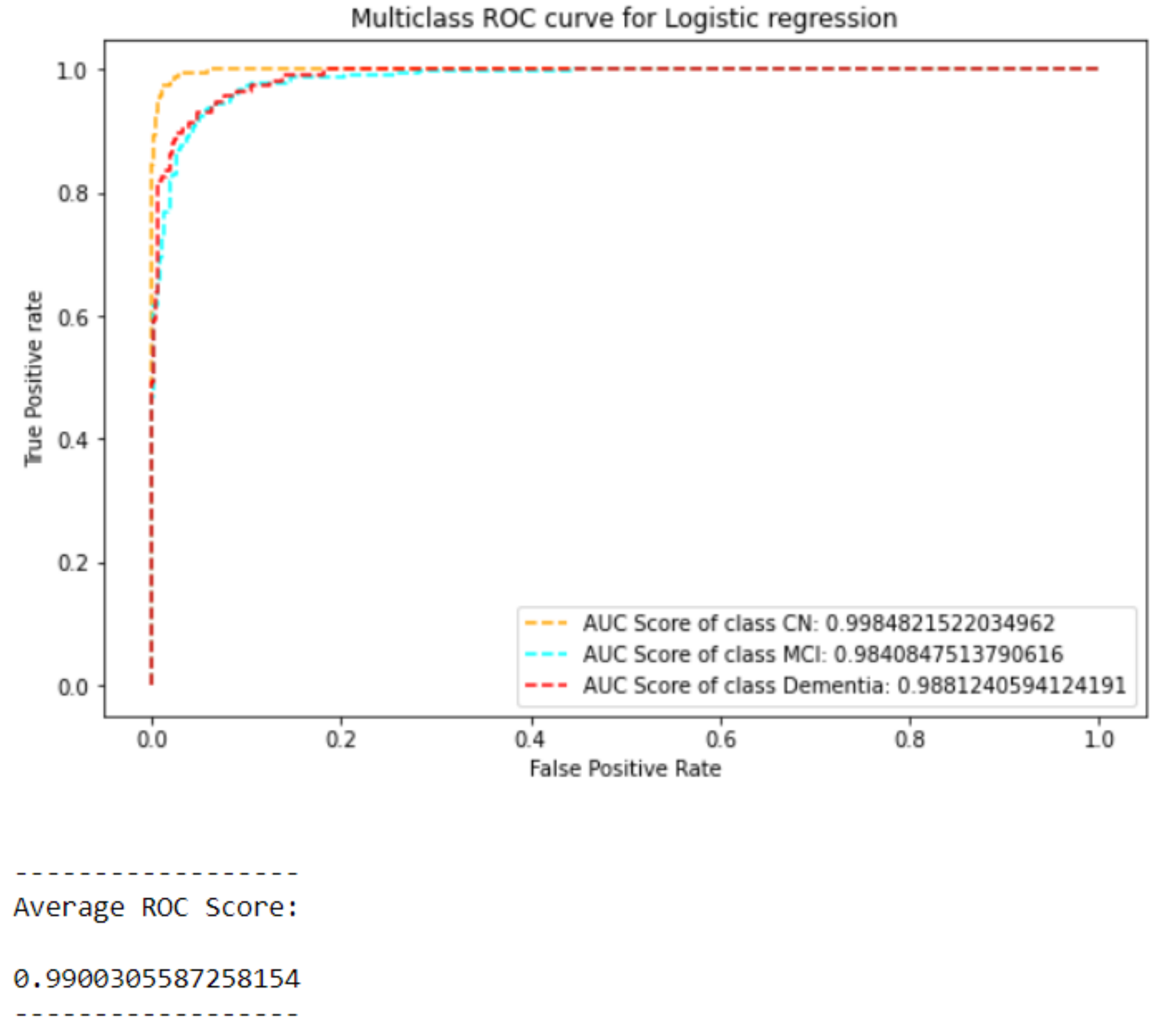

*Figure 5.3 AUCROC Curve for Logistic Regression without Class Balancing*

#### 5.3.1.2 Evaluation and Results with Class Balancing using SVM-SMOTE

The training dataset, which was used to train the model after imbalance handling using SVM-SMOTE method, has 2,073 entries. The model's performance was evaluated using a test dataset of 643 records. Below, figure 5.4 refers to the with class balancing confusion matrix for the Logistic Regression model, where rows represent actual or true class labels and columns represent predicted class labels. The TP, TN, FP, and FN for this multi-class problem have been calculated following the logic depicted in figure 5.1. For the cognitive normal (CN) class, there are 227 instances classified as CN that were predicted correctly (TP). The actual or true class values of non-CN classes, which have been predicted correctly, are 401 instances (TN). There are 9 instances which have been observed (true labels) as non-CN but are predicted as CN class (FP). At the same time, 6 instances that were predicted as non-CN where the true class value is given as CN class (FN). For the MCI class, there are 258 instances classified as MCI that were predicted correctly (TP). The actual or true class values of non-MCI classes, which have been predicted correctly, are 331 instances (TN). There are 18 instances which have been observed (true labels) as non-MCI but predicted as MCI class (FP). At the same time, 36 instances that were predicted as non-MCI where the true class value are given as MCI class (FN). For the dementia class, there are 104 instances classified as dementia that were predicted correctly (TP).

The actual or true class values of non-dementia classes, which have been predicted correctly, are 500 instances (TN). There are 27 instances which have been observed (true labels) as non-dementia but predicted as dementia class (FP). At the same time, 12 instances that were predicted as non-dementia where the true class value is given as dementia class (FN).

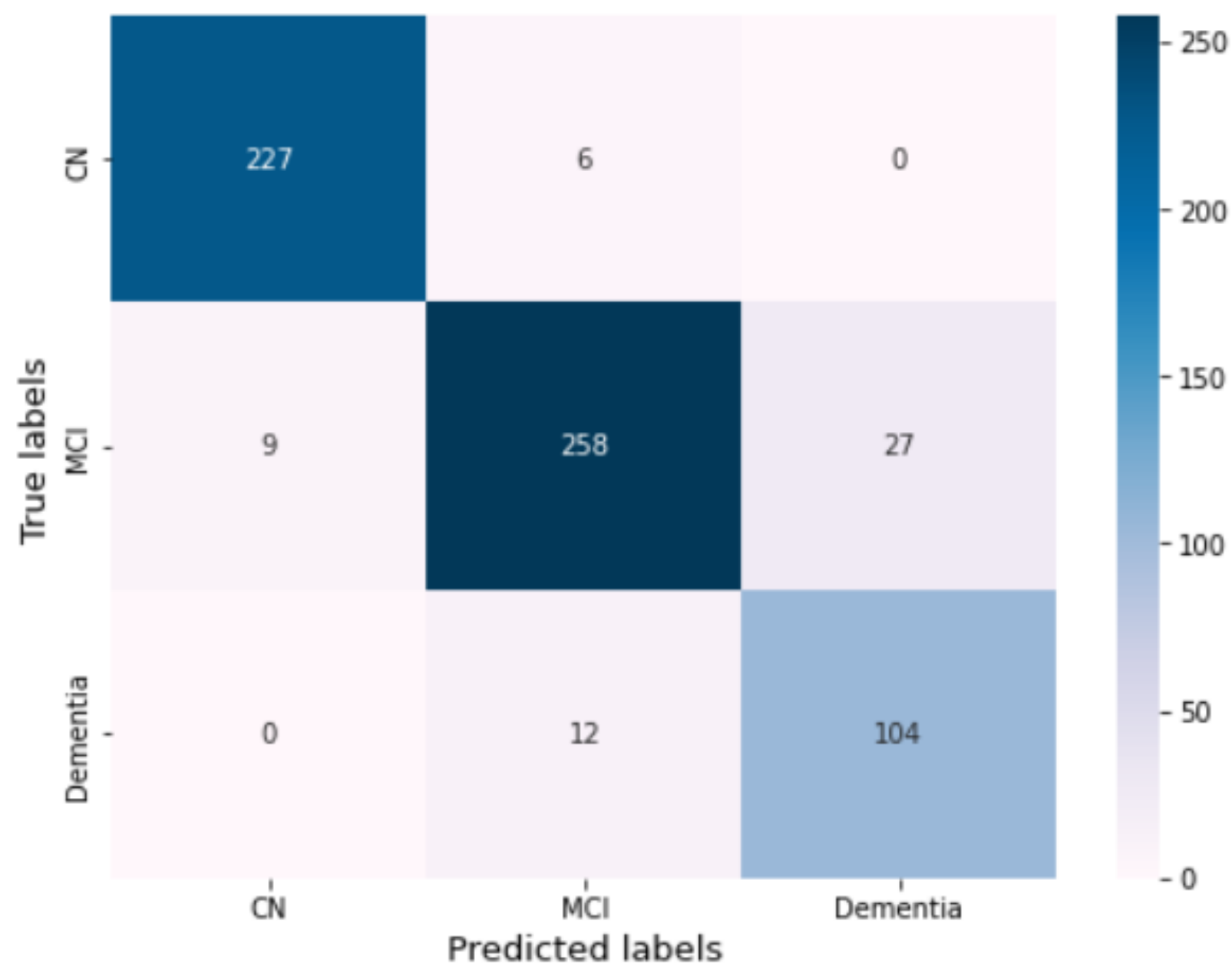

*Figure 5.4 Confusion Matrix for Logistic Regression after Class Balancing*

The comprehensive performance metrics for this multiclass classification task after class imbalance handling, as well as the precision, recall, and f1-score for each class, are provided in table 5.2. Additionally, for measuring overall model performance, the weighted average precision, weighted average recall, weighted average f1-score, and accuracy are all 92%.

*Table 5.2 Classification Report for Logistic Regression after Class Balancing*

| **Target Class** | **Precision** | **Recall** | **f1-score** | **weighted avg Precision** | **weighted avg Recall** | **weighted avg f1-score** | **Accuracy** |
|---|---|---|---|---|---|---|---|
| CN | 0.96 | 0.97 | 0.97 | 0.92 | 0.92 | 0.92 | 0.92 |
| MCI | 0.93 | 0.88 | 0.91 | | | | |
| Dementia | 0.79 | 0.90 | 0.84 | | | | |

The ROC curve presented in figure 5.5 after class imbalance handling shows how well the model predicts each class label. In addition, the whole model's weighted average AUCROC score is 98.77%.

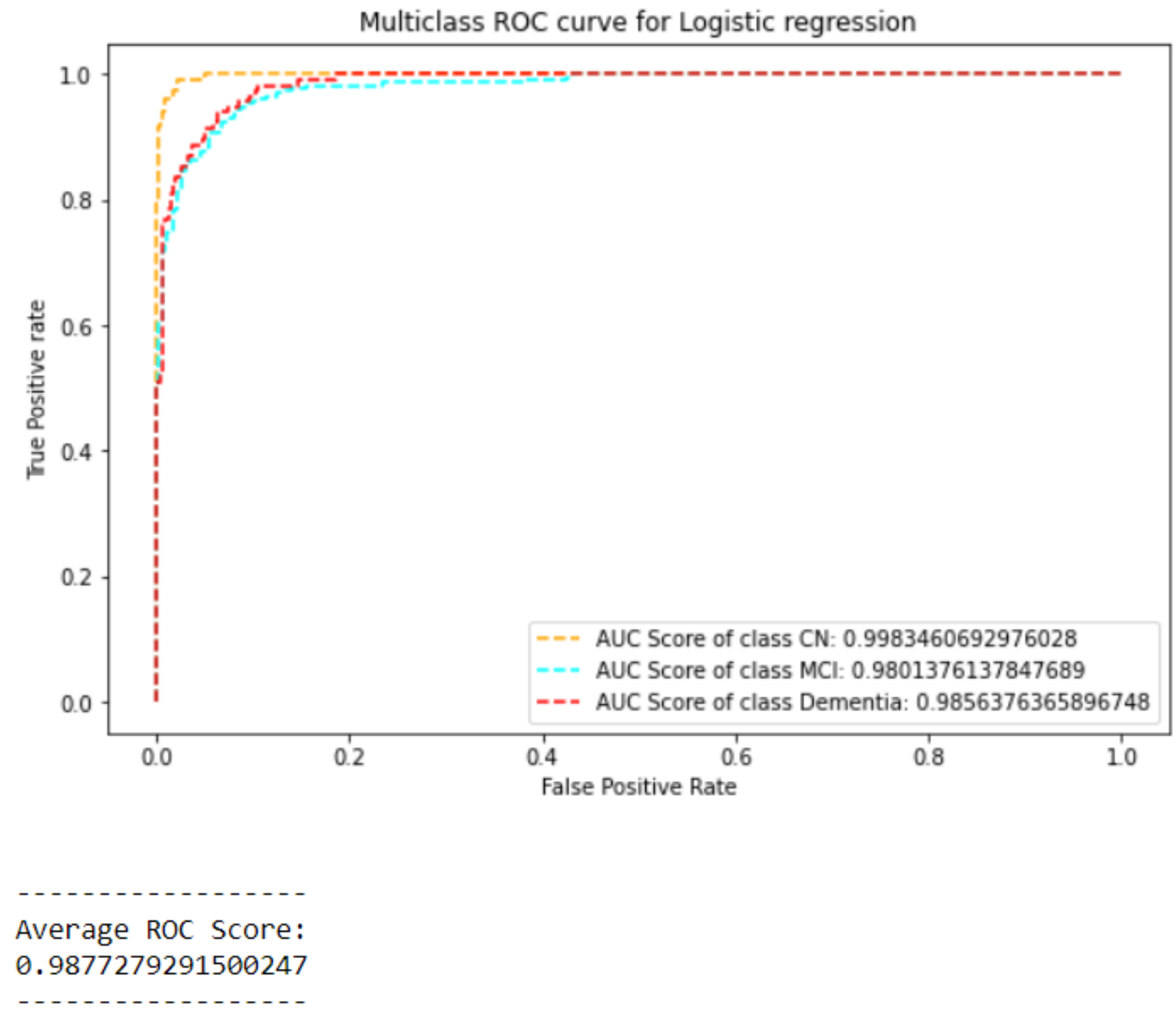

*Figure 5.5 AUCROC Curve for Logistic Regression after Class Balancing*

### 5.3.2 Light Gradient Boosting Machine Classifier Evaluation and Results

For building a classification model, a Light Gradient Boosting Machine (LGBM) model was utilised; this classifier was trained twice, once without imbalance handling and once after class imbalance correction (with SMOTE SVM minority class oversampling method). The evaluations/results of the classifier on the test set, as well as performance metrics such as the confusion matrix, weighted average F1 score, weighted average precision, weighted average recall, and the ROCAUC curve, will be shown in the subsections below.

#### 5.3.2.1 Evaluation and Results without Class Balancing

The training dataset, which was used to train the model before imbalance handling, has 1,500 entries. The model's performance was evaluated using a test dataset of 643 records. Below, figure 5.6 refers to the without class balancing confusion matrix for the LGBM model, where rows represent actual or true class labels and columns represent predicted class labels. The TP,

TN, FP, and FN for this multi-class problem have been calculated following the logic depicted in figure 5.1. For the cognitive normal (CN) class, there are 222 instances classified as CN that were predicted correctly (TP). The actual or true class values of non-CN classes, which have been predicted correctly, are 406 instances (TN). There are 4 instances which have been observed (true labels) as non-CN but are predicted as CN class (FP). At the same time, 11 instances that were predicted as non-CN where the true class value is given as CN class (FN). For the MCI class, there are 275 instances classified as MCI that were predicted correctly (TP). The actual or true class values of non-MCI classes, which have been predicted correctly, are 323 instances (TN). There are 26 instances which have been observed (true labels) as non-MCI but predicted as MCI class (FP). At the same time, 19 instances that were predicted as non-MCI where the true class value are given as MCI class (FN). For the dementia class, there are 101 instances classified as dementia that were predicted correctly (TP). The actual or true class values of non-dementia classes, which have been predicted correctly, are 512 instances (TN). There are 15 instances which have been observed (true labels) as non-dementia but predicted as dementia class (FP). At the same time, 15 instances that were predicted as non-dementia where the true class value is given as dementia class (FN).

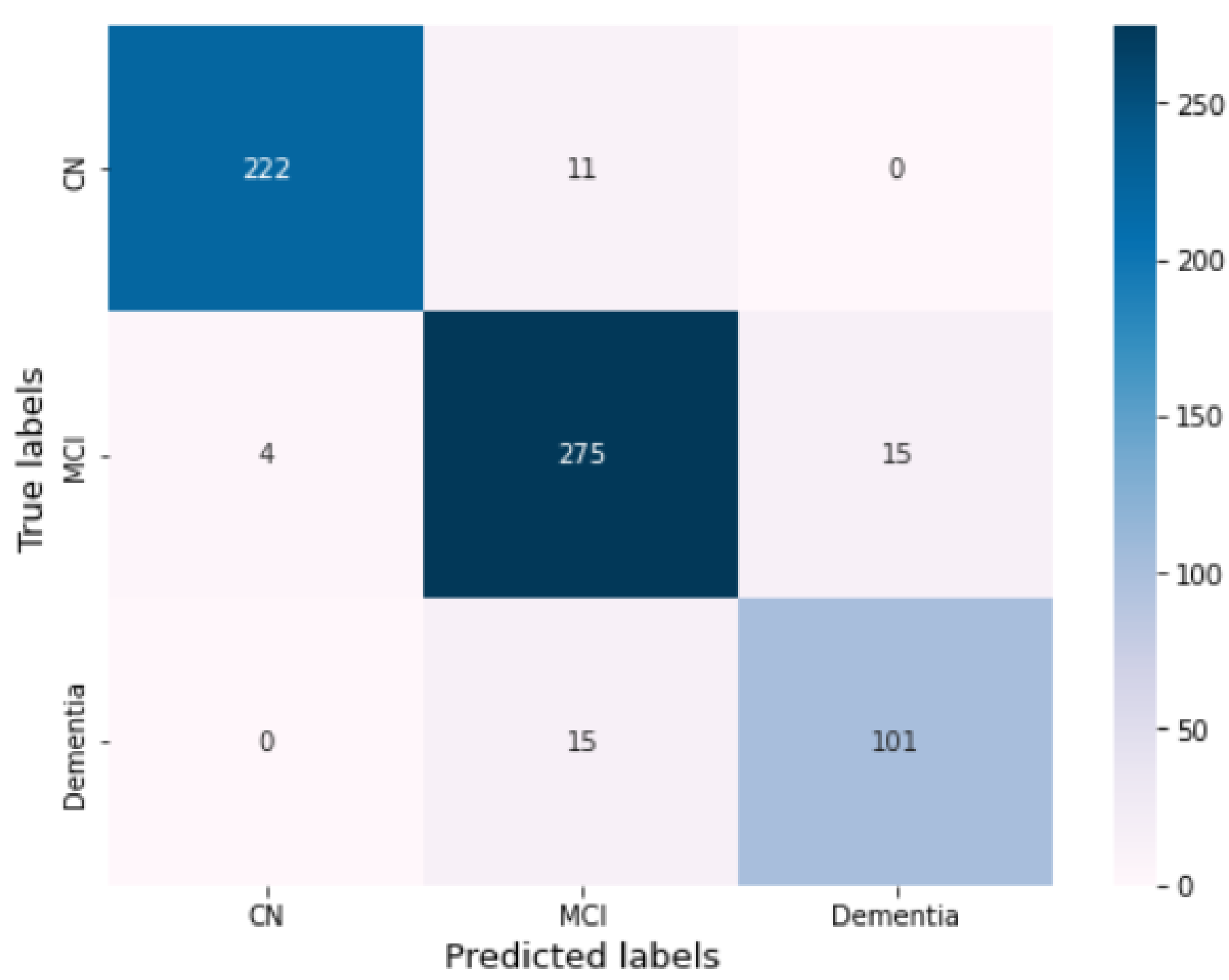

*Figure 5.6 Confusion Matrix for LGBM without Class Balancing*

The comprehensive performance metrics for this multiclass classification task, as well as the precision, recall, and f1-score for each class, are provided in table 5.3. Additionally, for measuring overall model performance, the weighted average precision, weighted average recall, weighted average f1-score, and accuracy are all 93%.

*Table 5.3 Classification Report for LGBM without Class Balancing*

| Target Class | Precision | Recall | f1-score | weighted avg Precision | weighted avg Recall | weighted avg f1-score | Accuracy |
|---|---|---|---|---|---|---|---|
| CN | 0.98 | 0.95 | 0.97 | 0.93 | 0.93 | 0.93 | 0.93 |
| MCI | 0.91 | 0.94 | 0.92 | | | | |
| Dementia | 0.87 | 0.87 | 0.87 | | | | |

The ROC curve presented in figure 5.7 shows how well the model predicts each class label. In addition, the whole model's weighted average AUCROC score is 98.86%.

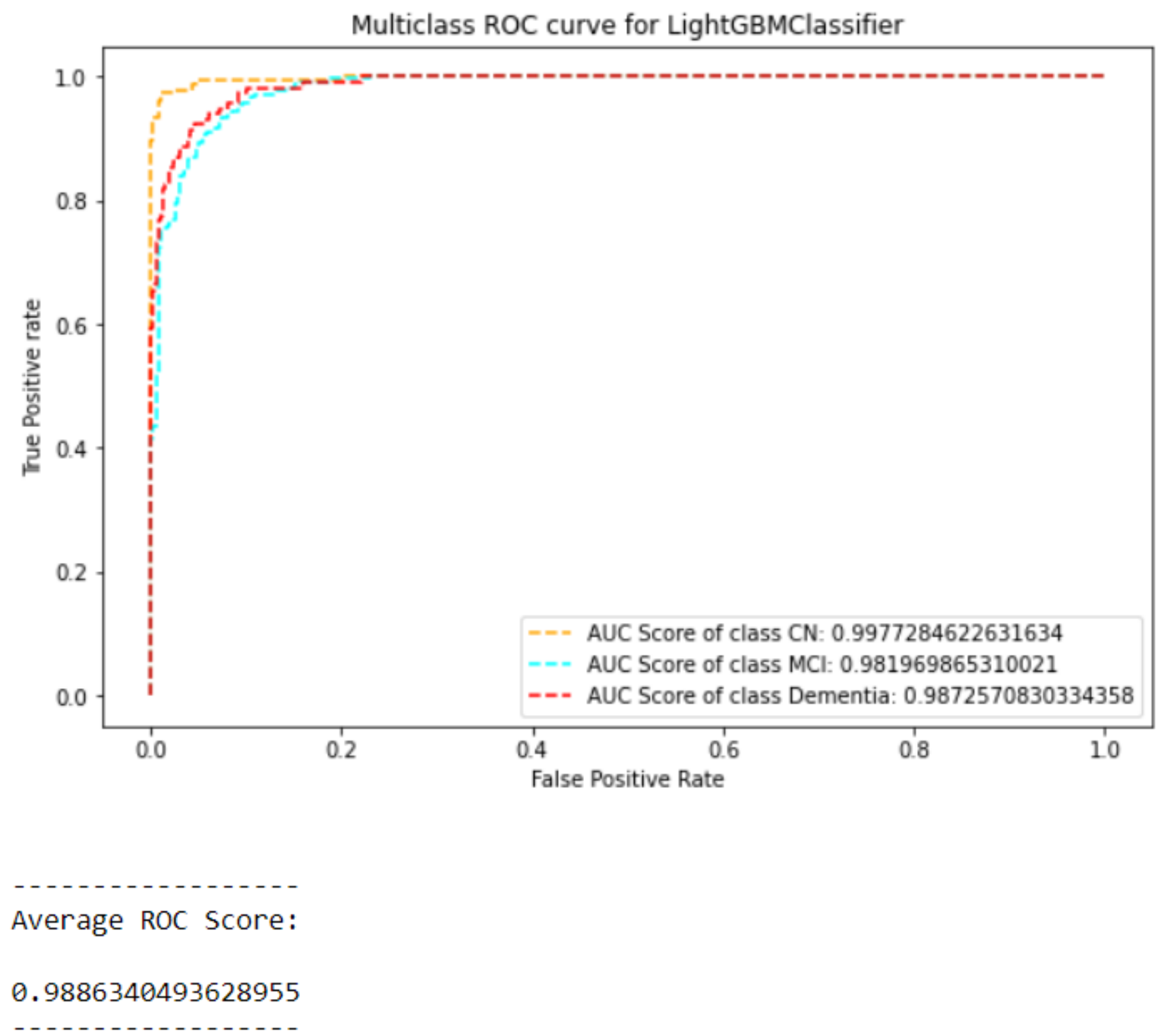

*Figure 5.7 AUCROC Curve for LGBM without Class Balancing*

#### 5.3.2.2 Evaluation and Results with Class Balancing using SVM-SMOTE

The training dataset, which was used to train the model after imbalance handling using SVM-SMOTE method, has 2,073 entries. The model's performance was evaluated using a test dataset of 643 records. Below, figure 5.8 refers to the with class balancing confusion matrix for the LGBM model, where rows represent actual or true class labels and columns represent predicted class labels. The TP, TN, FP, and FN for this multi-class problem have been calculated following the logic depicted in figure 5.1. For the cognitive normal (CN) class, there are 222 instances classified as CN that were predicted correctly (TP). The actual or true class values of non-CN classes, which have been predicted correctly, are 406 instances (TN). There are 4 instances which have been observed (true labels) as non-CN but are predicted as CN class (FP). At the same time, 11 instances that were predicted as non-CN where the true class value is given as CN class (FN). For the MCI class, there are 263 instances classified as MCI that were predicted correctly (TP). The actual or true class values of non-MCI classes, which have been predicted correctly, are 328 instances (TN). There are 21 instances which have been observed (true labels) as non-MCI but predicted as MCI class (FP). At the same time, 31 instances that were predicted as non-MCI where the true class value are given as MCI class (FN). For the dementia class, there are 106 instances classified as dementia that were predicted correctly (TP). The actual or true class values of non-dementia classes, which have been predicted correctly, are 500 instances (TN). There are 27 instances which have been observed (true labels) as non-dementia but predicted as dementia class (FP). At the same time, 10 instances that were predicted as non-dementia where the true class value is given as dementia class (FN).

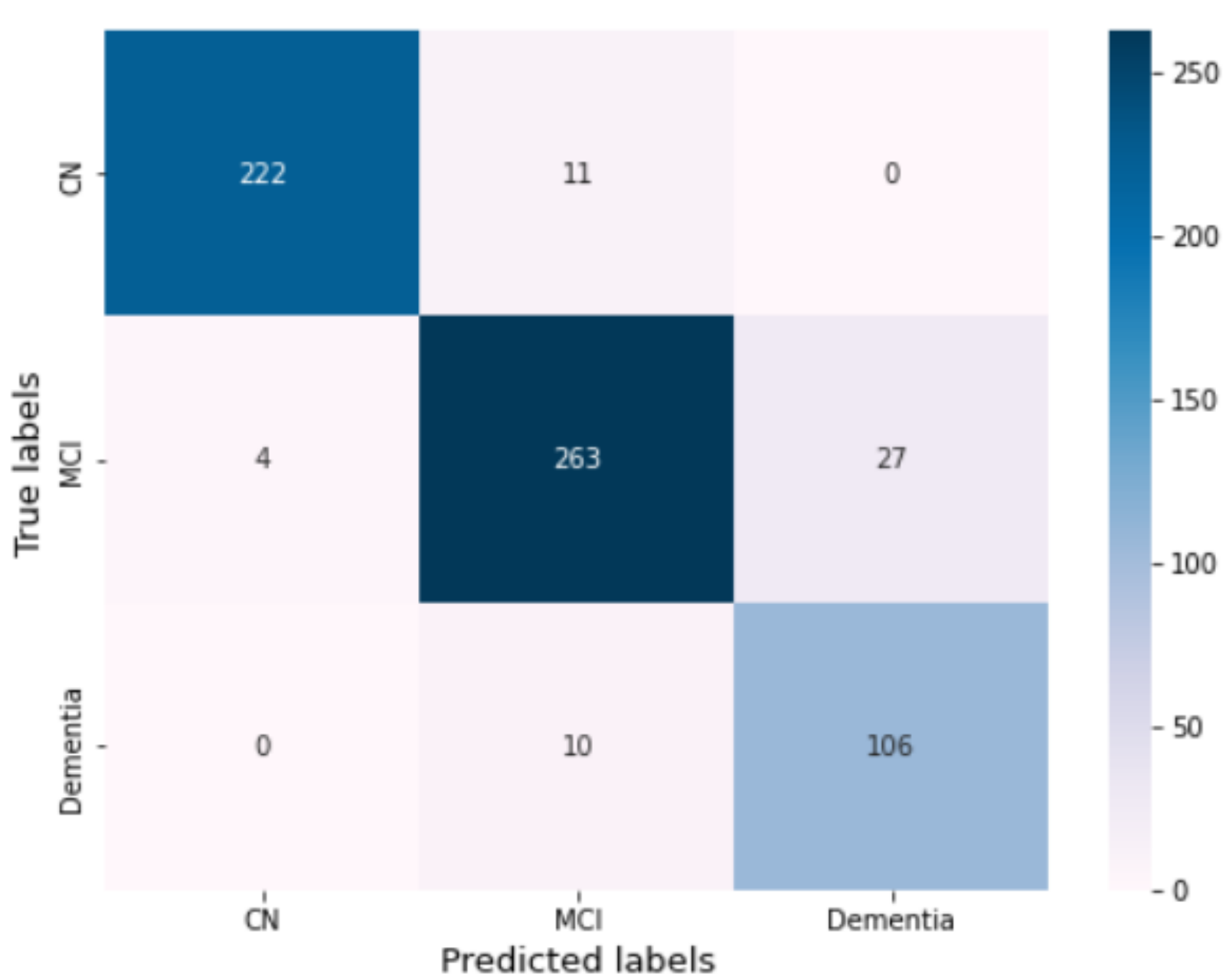

*Figure 5.8 Confusion Matrix for LGBM after Class Balancing*

The comprehensive performance metrics for this multiclass classification task after class imbalance handling, as well as the precision, recall, and f1-score for each class, are provided in table 5.4. Additionally, for measuring overall model performance, the weighted average precision, weighted average recall, weighted average f1-score, and accuracy are all 92%.

*Table 5.4 Classification Report for LGBM after Class Balancing*

| **Target Class** | **Precision** | **Recall** | **f1-score** | **weighted avg Precision** | **weighted avg Recall** | **weighted avg f1-score** | **Accuracy** |
|---|---|---|---|---|---|---|---|
| CN | 0.98 | 0.95 | 0.97 | 0.92 | 0.92 | 0.92 | 0.92 |
| MCI | 0.93 | 0.89 | 0.91 | | | | |
| Dementia | 0.80 | 0.91 | 0.85 | | | | |

The ROC curve presented in figure 5.9 after class imbalance handling shows how well the model predicts each class label. In addition, the whole model's weighted average AUCROC score is 98.73%.

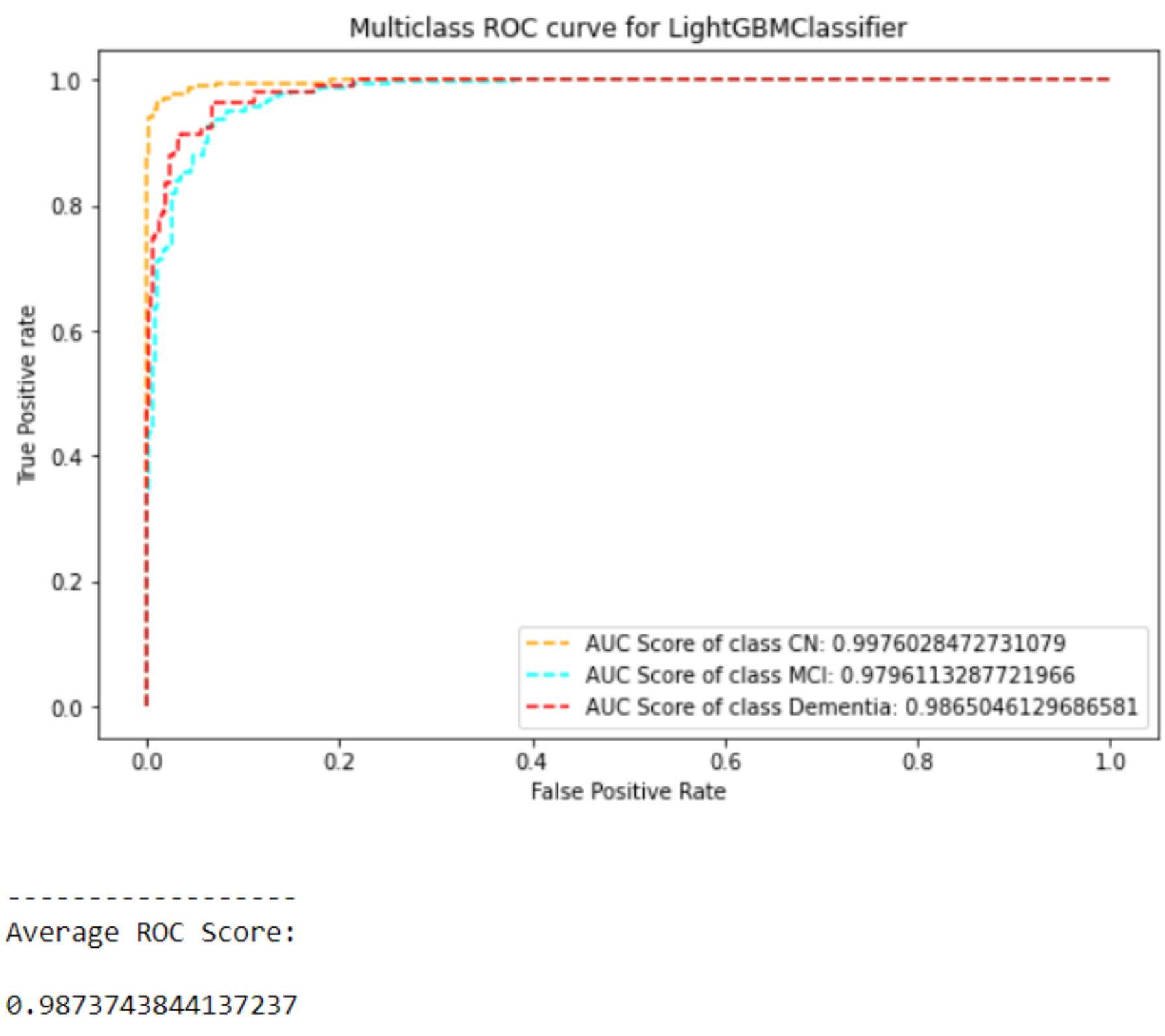

```
-----------------
Average ROC Score:

0.9873743844137237
-----------------
```

*Figure 5.9 AUCROC Curve for LGBM after Class Balancing*

### 5.3.3 Extra Tree Classifier Evaluation and Results

For building a classification model, an Extremely Randomized Tree model was utilised; this classifier was trained twice, once without imbalance handling and once after class imbalance correction (with SMOTE SVM minority class oversampling method). The evaluations/results of the classifier on the test set, as well as performance metrics such as the confusion matrix, weighted average F1 score, weighted average precision, weighted average recall, and the ROCAUC curve, will be shown in the subsections below.

#### 5.3.3.1 Evaluation and Results without Class Balancing

The training dataset, which was used to train the model before imbalance handling, has 1,500 entries. The model's performance was evaluated using a test dataset of 643 records. Below, figure 5.10 refers to the without class balancing confusion matrix for the Extra Tree classifier model, where rows represent actual or true class labels and columns represent predicted class labels. The TP, TN, FP, and FN for this multi-class problem have been calculated following the logic depicted in figure 5.1. For the cognitive normal (CN) class, there are 217 instances classified as CN that were predicted correctly (TP). The actual or true class values of non-CN classes, which have been predicted correctly, are 407 instances (TN). There are 3 instances which have been observed (true labels) as non-CN but are predicted as CN class (FP). At the same time, 16 instances that were predicted as non-CN where the true class value is given as CN class (FN). For the MCI class, there are 277 instances classified as MCI that were predicted correctly (TP). The actual or true class values of non-MCI classes, which have been predicted correctly, are 315 instances (TN). There are 34 instances which have been observed (true labels) as non-MCI but predicted as MCI class (FP). At the same time, 17 instances that were predicted as non-MCI where the true class value are given as MCI class (FN). For the dementia class, there are 98 instances classified as dementia that were predicted correctly (TP). The actual or true class values of non-dementia classes, which have been predicted correctly, are 513 instances (TN). There are 14 instances which have been observed (true labels) as non-dementia but predicted as dementia class (FP). At the same time, 18 instances that were predicted as non-dementia where the true class value is given as dementia class (FN).

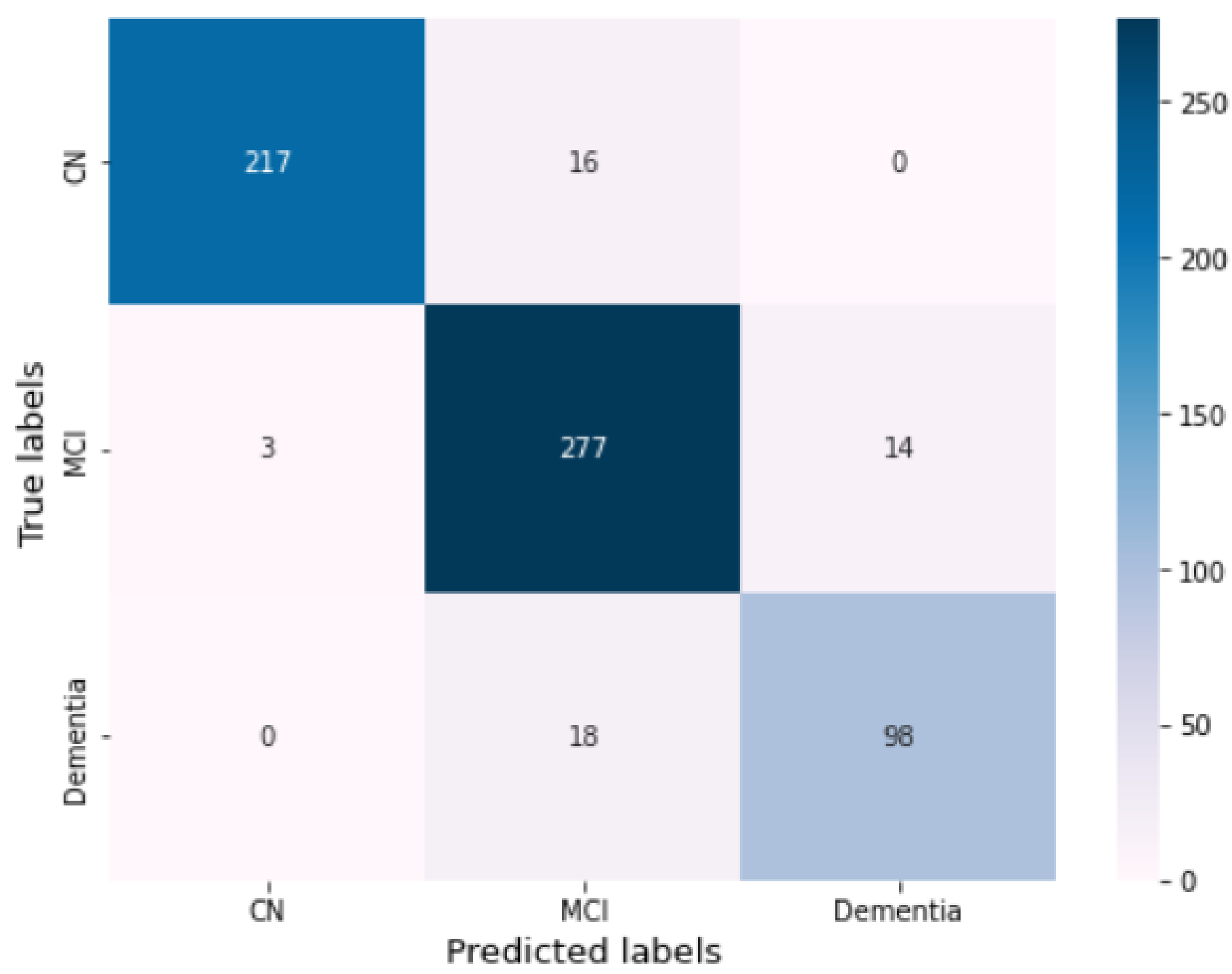

*Figure 5.10 Confusion Matrix for Extra Tree Classifier without Class Balancing*

The comprehensive performance metrics for this multiclass classification task, as well as the precision, recall, and f1-score for each class, are provided in table 5.5. Additionally, for measuring overall model performance, the weighted average precision, weighted average recall, weighted average f1-score, and accuracy are all 92%.

*Table 5.5 Classification Report for Extra Tree Classifier without Class Balancing*

| Target Class | Precision | Recall | f1-score | weighted avg Precision | weighted avg Recall | weighted avg f1-score | Accuracy |
|---|---|---|---|---|---|---|---|
| CN | 0.99 | 0.93 | 0.96 | 0.92 | 0.92 | 0.92 | 0.92 |
| MCI | 0.89 | 0.94 | 0.92 | | | | |
| Dementia | 0.88 | 0.84 | 0.86 | | | | |

The ROC curve presented in figure 5.11 shows how well the model predicts each class label. In addition, the whole model's weighted average AUCROC score is 98.94%.

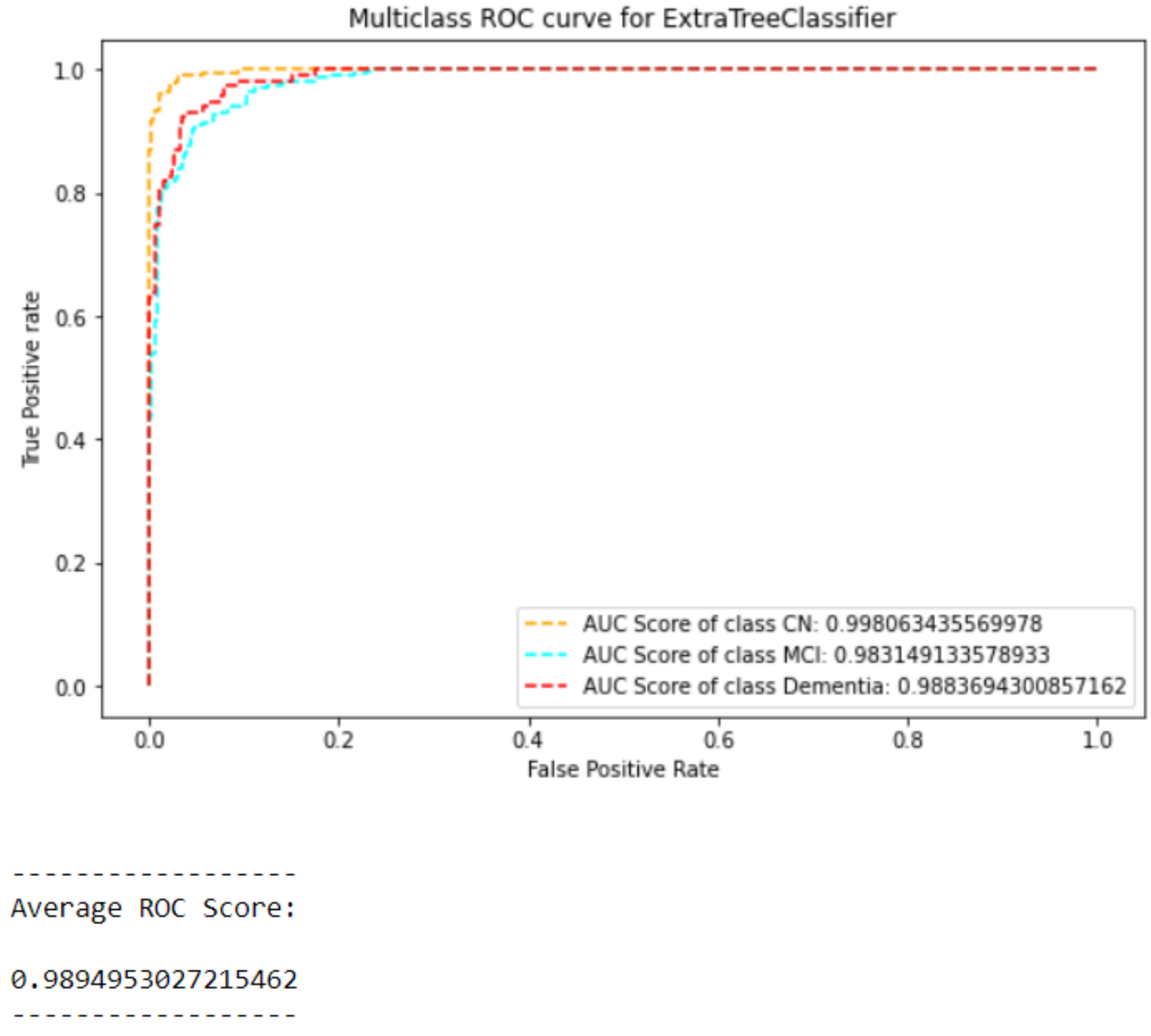

*Figure 5.11 AUCROC Curve for Extra Tree Classifier without Class Balancing*

#### 5.3.3.2 Evaluation and Results with Class Balancing using SVM-SMOTE

The training dataset, which was used to train the model after imbalance handling using SVM-SMOTE method, has 2,073 entries. The model's performance was evaluated using a test dataset of 643 records. Below, figure 5.12 refers to the with class balancing confusion matrix for the Extra Tree classifier model, where rows represent actual or true class labels and columns represent predicted class labels. The TP, TN, FP, and FN for this multi-class problem have been calculated following the logic depicted in figure 5.1. For the cognitive normal (CN) class, there are 222 instances classified as CN that were predicted correctly (TP). The actual or true class values of non-CN classes, which have been predicted correctly, are 403 instances (TN). There are 7 instances which have been observed (true labels) as non-CN but are predicted as CN class (FP). At the same time, 11 instances that were predicted as non-CN where the true class value is given as CN class (FN). For the MCI class, there are 259 instances classified as MCI that were predicted correctly (TP). The actual or true class values of non-MCI classes, which have been predicted correctly, are 328 instances (TN). There are 21 instances which have been observed (true labels) as non-MCI but predicted as MCI class (FP). At the same time, 35 instances that were predicted as non-MCI where the true class value are given as MCI class (FN). For the dementia class, there are 106 instances classified as dementia that were predicted correctly (TP). The actual or true class values of non-dementia classes, which have been predicted correctly, are 499 instances (TN). There are 28 instances which have been observed (true labels) as non-dementia but predicted as dementia class (FP). At the same time, 10 instances that were predicted as non-dementia where the true class value is given as dementia class (FN).

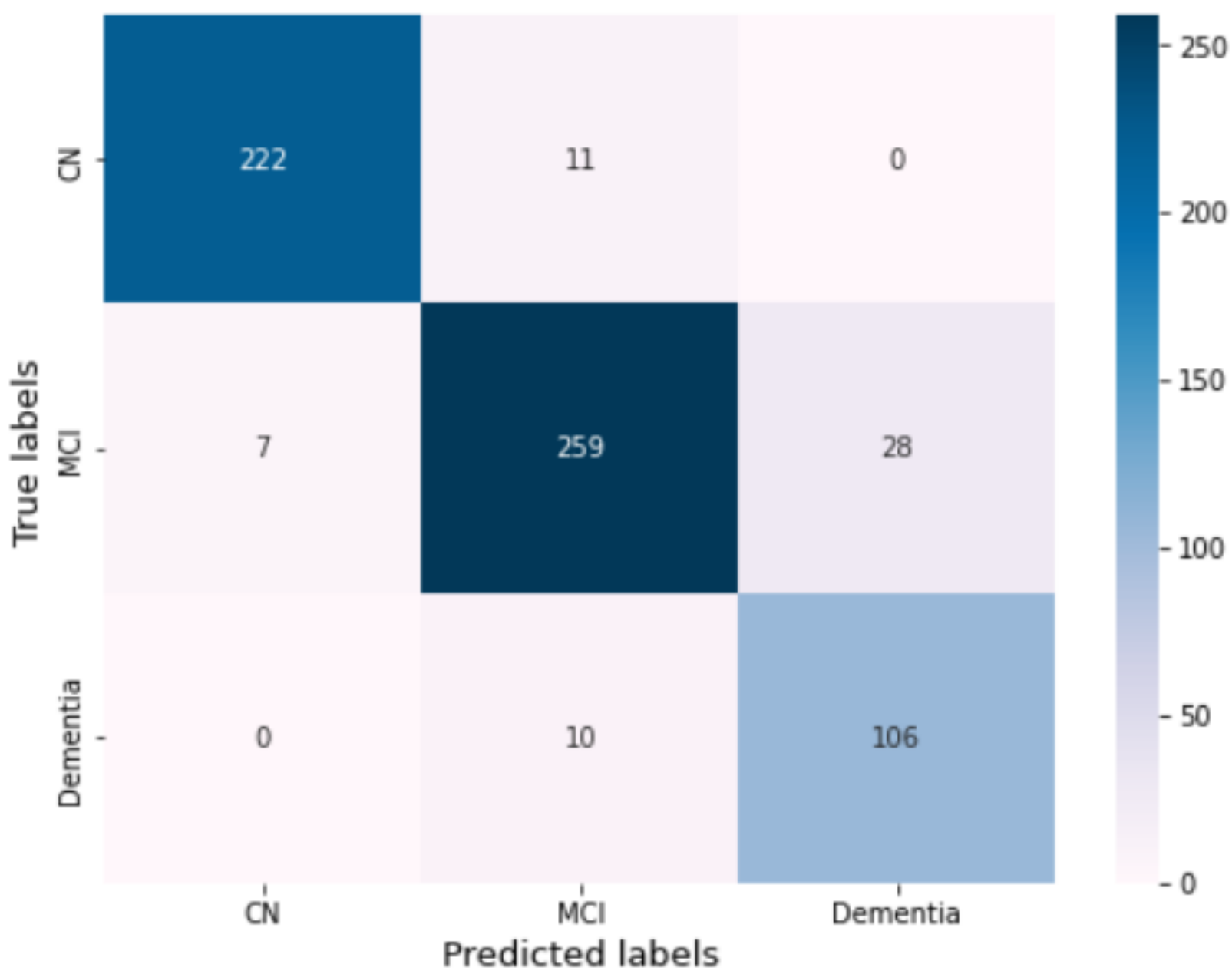

*Figure 5.12 Confusion Matrix for Extra Tree Classifier after Class Balancing*

The comprehensive performance metrics for this multiclass classification task after class imbalance handling, as well as the precision, recall, and f1-score for each class, are provided in table 5.6. Additionally, for measuring overall model performance, the weighted average precision, weighted average recall, weighted average f1-score, and accuracy are 92%, 91%, 91%, and 91%, respectively.

*Table 5.6 Classification Report for Extra Tree Classifier after Class Balancing*

| Target Class | Precision | Recall | f1-score | weighted avg Precision | weighted avg Recall | weighted avg f1-score | Accuracy |
|---|---|---|---|---|---|---|---|
| CN | 0.97 | 0.95 | 0.96 | 0.92 | 0.91 | 0.91 | 0.91 |
| MCI | 0.93 | 0.88 | 0.90 | | | | |
| Dementia | 0.79 | 0.91 | 0.85 | | | | |

The ROC curve presented in figure 5.13 after class imbalance handling shows how well the model predicts each class label. In addition, the whole model's weighted average AUCROC score is 98.55%.

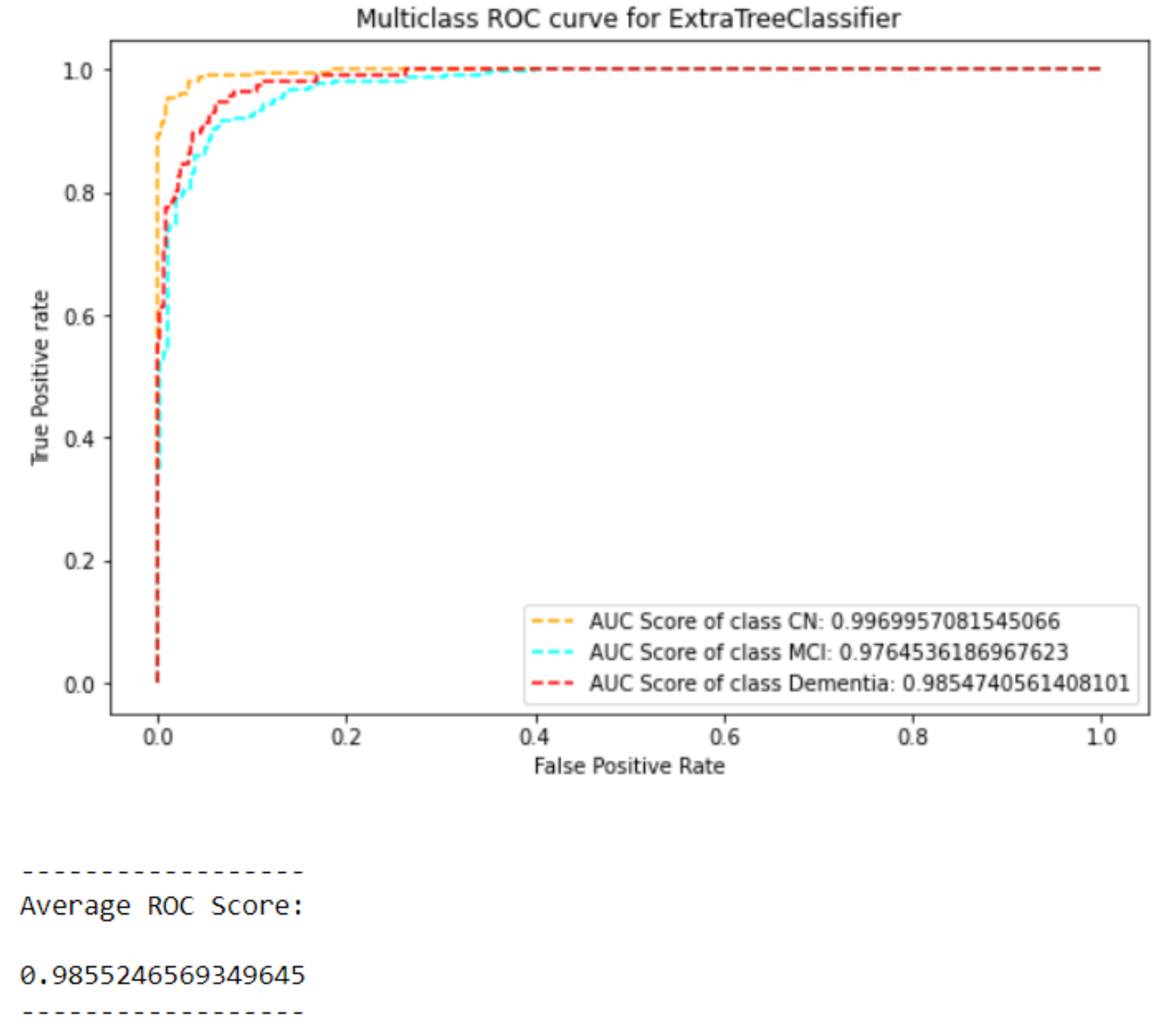

*Figure 5.13 AUCROC Curve for Extra Tree Classifier after Class Balancing*

### 5.3.4 Bagging KNN Classifier Evaluation and Results

For building a classification model, a bagging classifier with KNN estimator was utilised; this classifier was trained twice, once without imbalance handling and once after class imbalance correction (with SMOTE SVM minority class oversampling method). The evaluations/results of the classifier on the test set, as well as performance metrics such as the confusion matrix, weighted average F1 score, weighted average precision, weighted average recall, and the ROCAUC curve, will be shown in the subsections below.

#### 5.3.4.1 Evaluation and Results without Class Balancing

The training dataset, which was used to train the model before imbalance handling, has 1,500 entries. The model's performance was evaluated using a test dataset of 643 records. Below, figure 5.14 refers to the without class balancing confusion matrix for the bagging KNN classifier model, where rows represent actual or true class labels and columns represent predicted class labels. The TP, TN, FP, and FN for this multi-class problem have been calculated following the logic depicted in figure 5.1. For the cognitive normal (CN) class, there are 226 instances classified as CN that were predicted correctly (TP). The actual or true class values of non-CN classes, which have been predicted correctly, are 368 instances (TN). There are 42 instances which have been observed (true labels) as non-CN but are predicted as CN class (FP). At the same time, 7 instances that were predicted as non-CN where the true class value is given as CN class (FN). For the MCI class, there are 237 instances classified as MCI that were predicted correctly (TP). The actual or true class values of non-MCI classes, which have been predicted correctly, are 322 instances (TN). There are 27 instances which have been observed (true labels) as non-MCI but predicted as MCI class (FP). At the same time, 57 instances that were predicted as non-MCI where the true class value are given as MCI class (FN). For the dementia class, there are 96 instances classified as dementia that were predicted correctly (TP). The actual or true class values of non-dementia classes, which have been predicted correctly, are 512 instances (TN). There are 15 instances which have been observed (true labels) as non-dementia but predicted as dementia class (FP). At the same time, 20 instances that were predicted as non-dementia where the true class value is given as dementia class (FN).

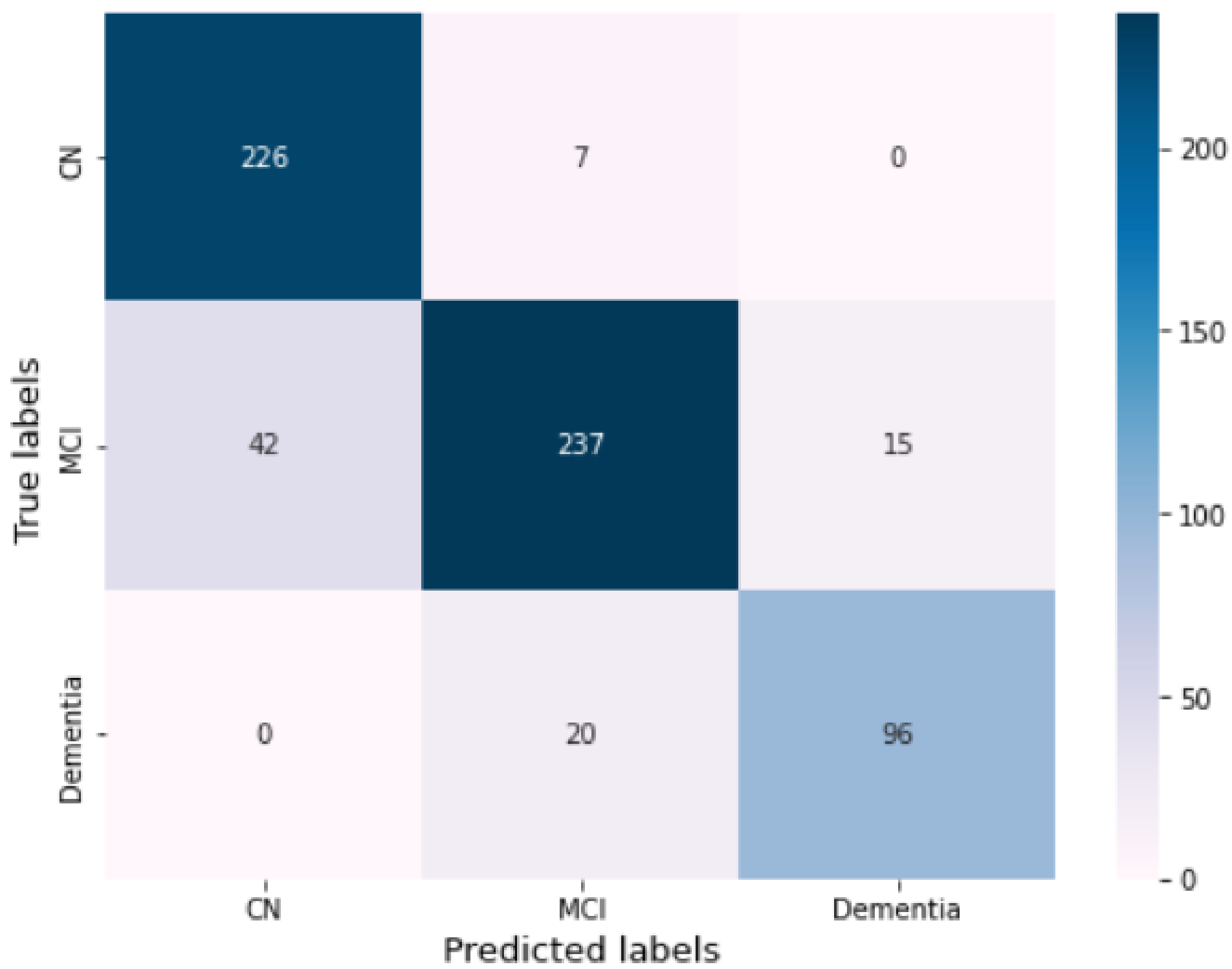

*Figure 5.14 Confusion Matrix for Bagging KNN Classifier without Class Balancing*

The comprehensive performance metrics for this multiclass classification task, as well as the precision, recall, and f1-score for each class, are provided in table 5.7. Additionally, for measuring overall model performance, the weighted average precision, weighted average recall, weighted average f1-score, and accuracy are all 87%.

*Table 5.7 Classification Report for Bagging KNN Classifier without Class Balancing*

| Target Class | Precision | Recall | f1-score | weighted avg Precision | weighted avg Recall | weighted avg f1-score | Accuracy |
|---|---|---|---|---|---|---|---|
| CN | 0.84 | 0.97 | 0.90 | 0.87 | 0.87 | 0.87 | 0.87 |
| MCI | 0.90 | 0.81 | 0.85 | | | | |
| Dementia | 0.86 | 0.83 | 0.85 | | | | |

The ROC curve presented in figure 5.15 shows how well the model predicts each class label. In addition, the whole model's weighted average AUCROC score is 96.48%.

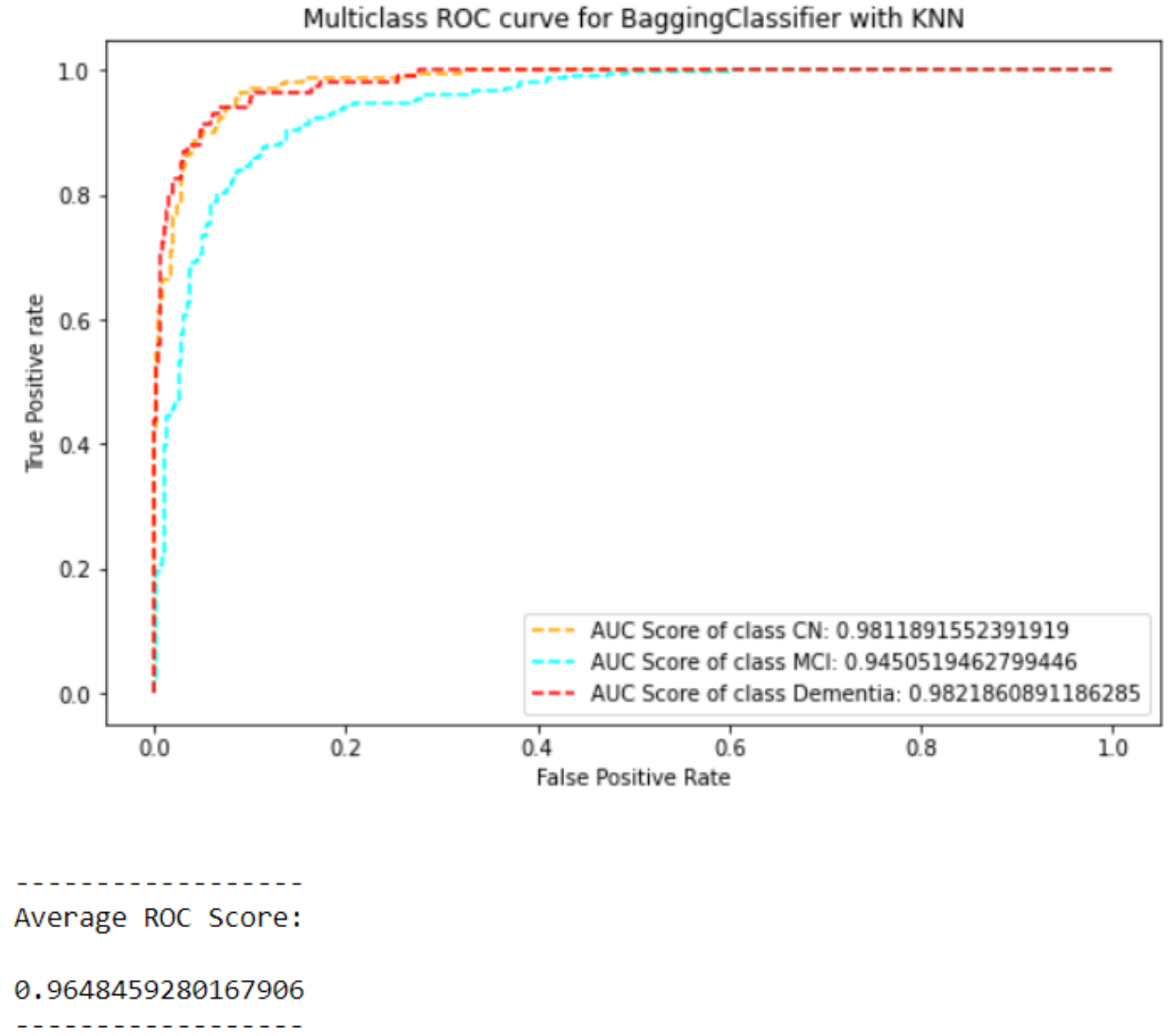

*Figure 5.15 AUCROC Curve for Bagging KNN Classifier without Class Balancing*

#### 5.3.4.2 Evaluation and Results with Class Balancing using SVM-SMOTE

The training dataset, which was used to train the model after imbalance handling using SVM-SMOTE method, has 2,073 entries. The model's performance was evaluated using a test dataset of 643 records. Below, figure 5.16 refers to the with class balancing confusion matrix for the bagging KNN classifier model, where rows represent actual or true class labels and columns represent predicted class labels. The TP, TN, FP, and FN for this multi-class problem have been calculated following the logic depicted in figure 5.1. For the cognitive normal (CN) class, there are 225 instances classified as CN that were predicted correctly (TP). The actual or true class values of non-CN classes, which have been predicted correctly, are 351 instances (TN). There are 59 instances which have been observed (true labels) as non-CN but are predicted as CN class (FP). At the same time, 8 instances that were predicted as non-CN where the true class value is given as CN class (FN). For the MCI class, there are 203 instances classified as MCI that were predicted correctly (TP). The actual or true class values of non-MCI classes, which have been predicted correctly, are 333 instances (TN). There are 16 instances which have been observed (true labels) as non-MCI but predicted as MCI class (FP). At the same time, 91 instances that were predicted as non-MCI where the true class value are given as MCI class (FN). For the dementia class, there are 108 instances classified as dementia that were predicted correctly (TP). The actual or true class values of non-dementia classes, which have been predicted correctly, are 495 instances (TN). There are 32 instances which have been observed (true labels) as non-dementia but predicted as dementia class (FP). At the same time, 8 instances that were predicted as non-dementia where the true class value is given as dementia class (FN).

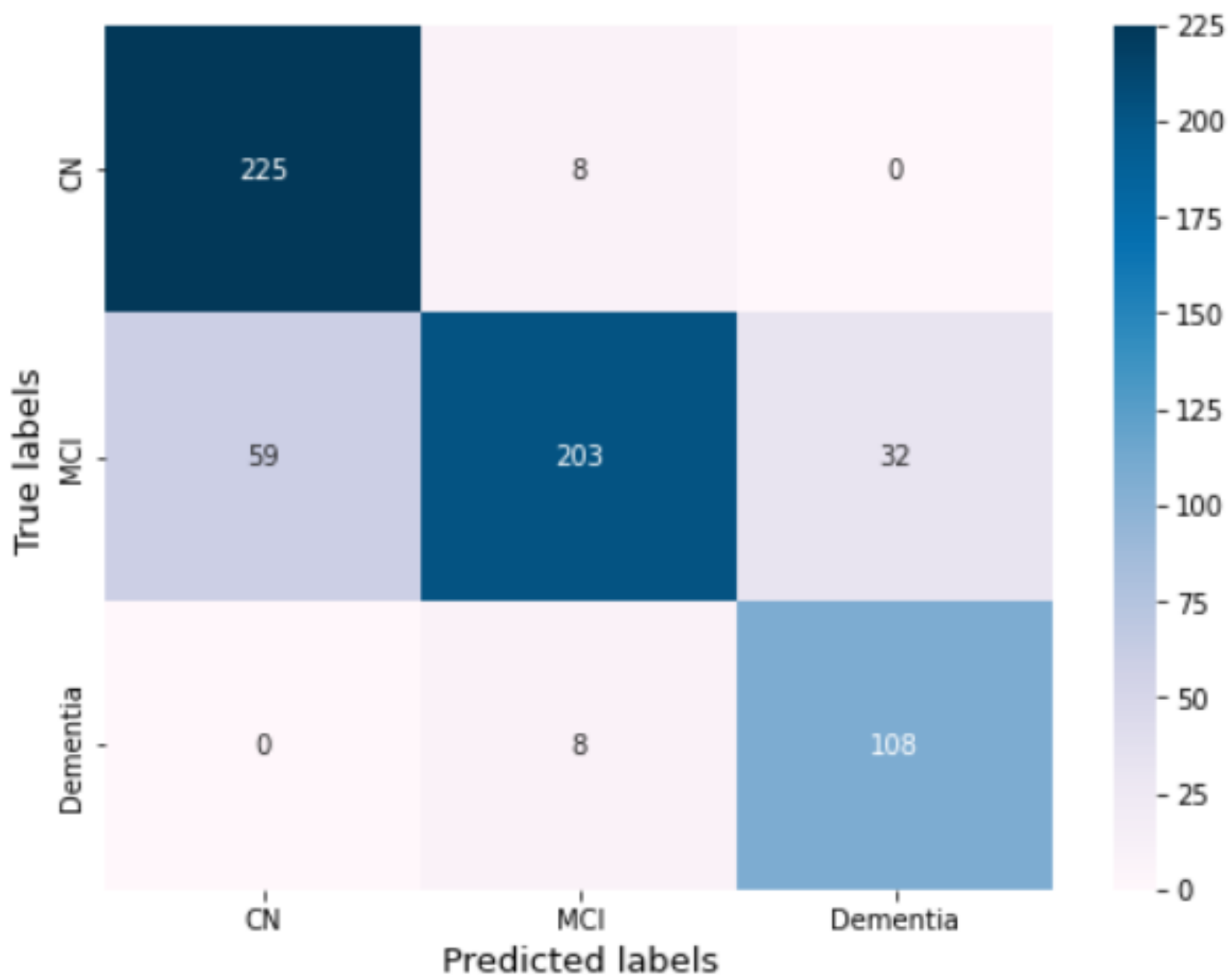

*Figure 5.16 Confusion Matrix for Bagging KNN Classifier after Class Balancing*

The comprehensive performance metrics for this multiclass classification task after class imbalance handling, as well as the precision, recall, and f1-score for each class, are provided in table 5.8. Additionally, for measuring overall model performance, the weighted average precision, weighted average recall, weighted average f1-score, and accuracy are 85%, 83%, 83%, and 83%, respectively.

*Table 5.8 Classification Report for Bagging KNN Classifier after Class Balancing*

| **Target Class** | **Precision** | **Recall** | **f1-score** | **weighted avg Precision** | **weighted avg Recall** | **weighted avg f1-score** | **Accuracy** |
|---|---|---|---|---|---|---|---|
| CN | 0.79 | 0.97 | 0.87 | 0.85 | 0.83 | 0.83 | 0.83 |
| MCI | 0.93 | 0.69 | 0.79 | | | | |
| Dementia | 0.77 | 0.93 | 0.84 | | | | |

The ROC curve presented in figure 5.17 after class imbalance handling shows how well the model predicts each class label. In addition, the whole model's weighted average AUCROC score is 96.65%.

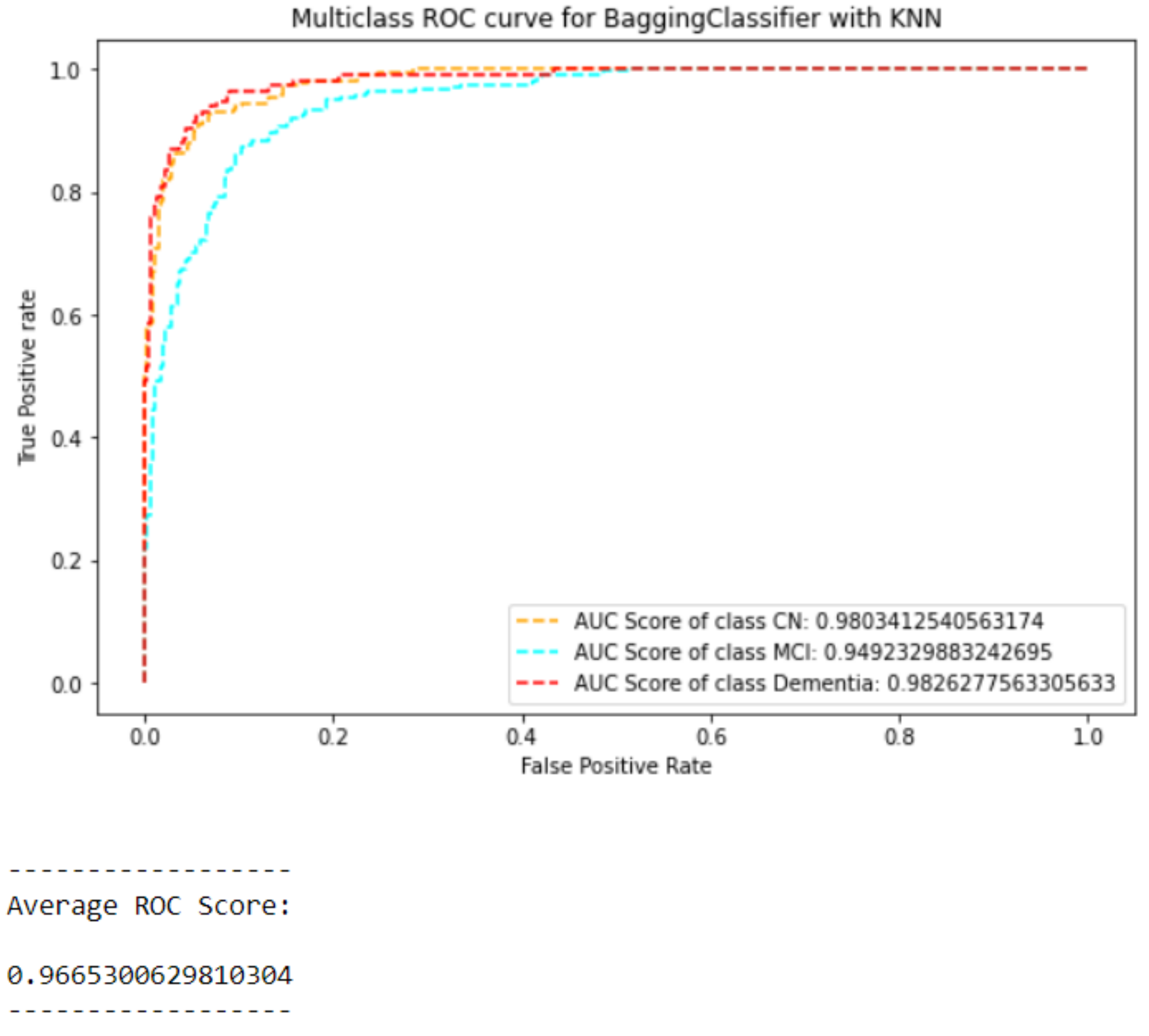

*Figure 5.17 AUCROC Curve for Bagging KNN Classifier after Class Balancing*

### 5.3.5 Stacking-Based Ensemble Learning Classifier Evaluation and Results

For building a classification model, a stacking blending (ensemble) classifier was utilised; this classifier was trained twice, once without imbalance handling and once after class imbalance correction (with SMOTE SVM minority class oversampling method). The evaluations/results of the classifier on the test set, as well as performance metrics such as the confusion matrix, weighted average F1 score, weighted average precision, weighted average recall, and the ROCAUC curve, will be shown in the subsections below.

#### 5.3.5.1 Evaluation and Results without Class Balancing

The training dataset, which was used to train the model before imbalance handling, has 1,500 entries. The model's performance was evaluated using a test dataset of 643 records. Below, figure 5.18 refers to the without class balancing confusion matrix for the stacking-ensemble classifier model, where rows represent actual or true class labels and columns represent predicted class labels. The TP, TN, FP, and FN for this multi-class problem have been calculated following the logic depicted in figure 5.1. For the cognitive normal (CN) class, there are 224 instances classified as CN that were predicted correctly (TP). The actual or true class values of non-CN classes, which have been predicted correctly, are 406 instances (TN). There are 4 instances which have been observed (true labels) as non-CN but are predicted as CN class (FP). At the same time, 9 instances that were predicted as non-CN where the true class value is given as CN class (FN). For the MCI class, there are 274 instances classified as MCI that were predicted correctly (TP). The actual or true class values of non-MCI classes, which have been predicted correctly, are 327 instances (TN). There are 22 instances which have been observed (true labels) as non-MCI but predicted as MCI class (FP). At the same time, 20 instances that were predicted as non-MCI where the true class value are given as MCI class (FN). For the dementia class, there are 103 instances classified as dementia that were predicted correctly (TP). The actual or true class values of non-dementia classes, which have been predicted correctly, are 511 instances (TN). There are 16 instances which have been observed (true labels) as non-dementia but predicted as dementia class (FP). At the same time, 13 instances that were predicted as non-dementia where the true class value is given as dementia class (FN).

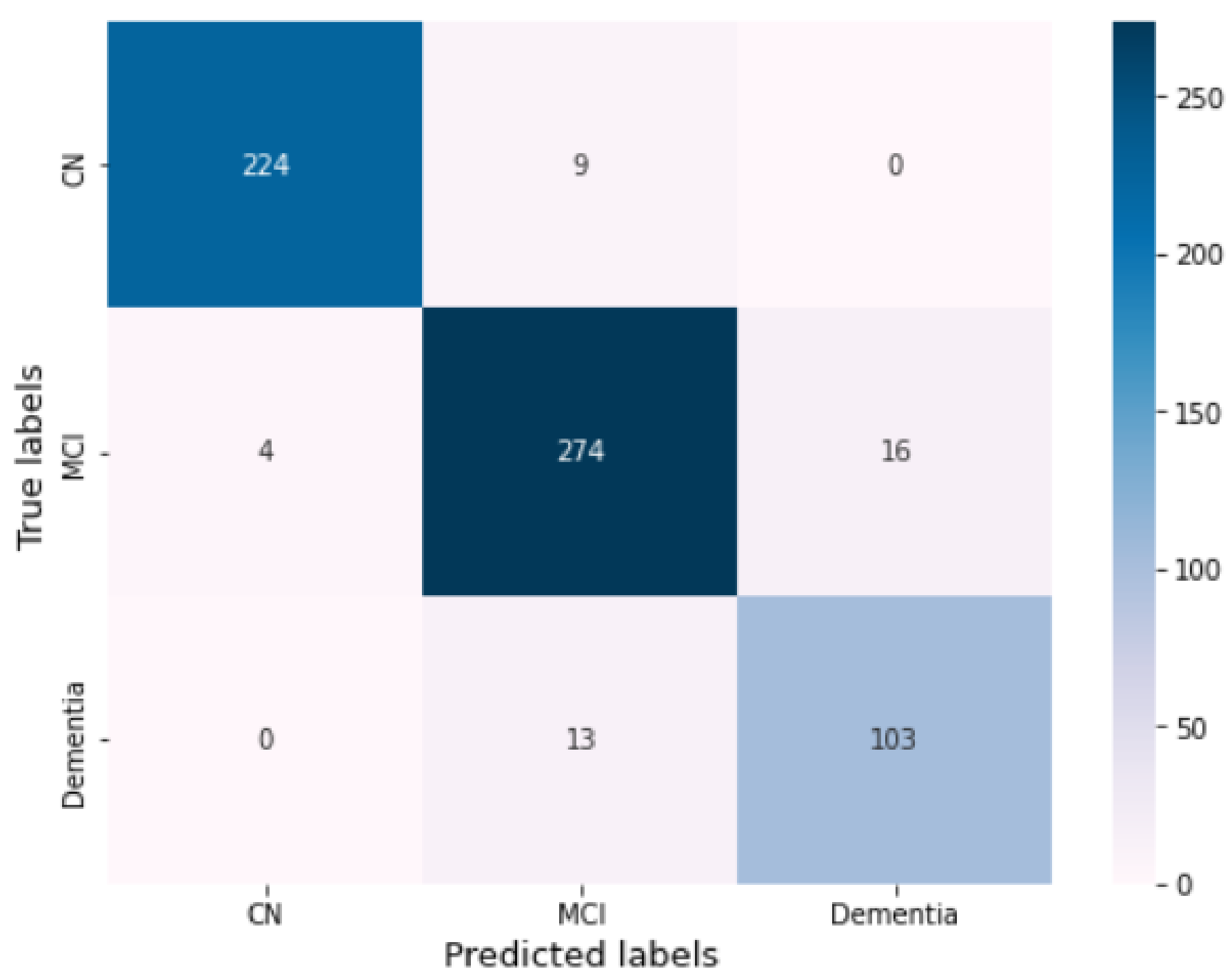

*Figure 5.18 Confusion Matrix for Stacking-Ensemble Classifier without Class Balancing*

The comprehensive performance metrics for this multiclass classification task, as well as the precision, recall, and f1-score for each class, are provided in table 5.9. Additionally, for measuring overall model performance, the weighted average precision, weighted average recall, weighted average f1-score, and accuracy are 94%, 93%, 93%, and 93%, respectively.

*Table 5.9 Classification Report for Stacking-Ensemble Classifier without Class Balancing*

| Target Class | Precision | Recall | f1-score | weighted avg Precision | weighted avg Recall | weighted avg f1-score | Accuracy |
|---|---|---|---|---|---|---|---|
| CN | 0.98 | 0.96 | 0.97 | 0.94 | 0.93 | 0.93 | 0.93 |
| MCI | 0.93 | 0.93 | 0.93 | | | | |
| Dementia | 0.87 | 0.89 | 0.88 | | | | |

The ROC curve presented in figure 5.19 shows how well the model predicts each class label. In addition, the whole model's weighted average AUCROC score is 99.05%.

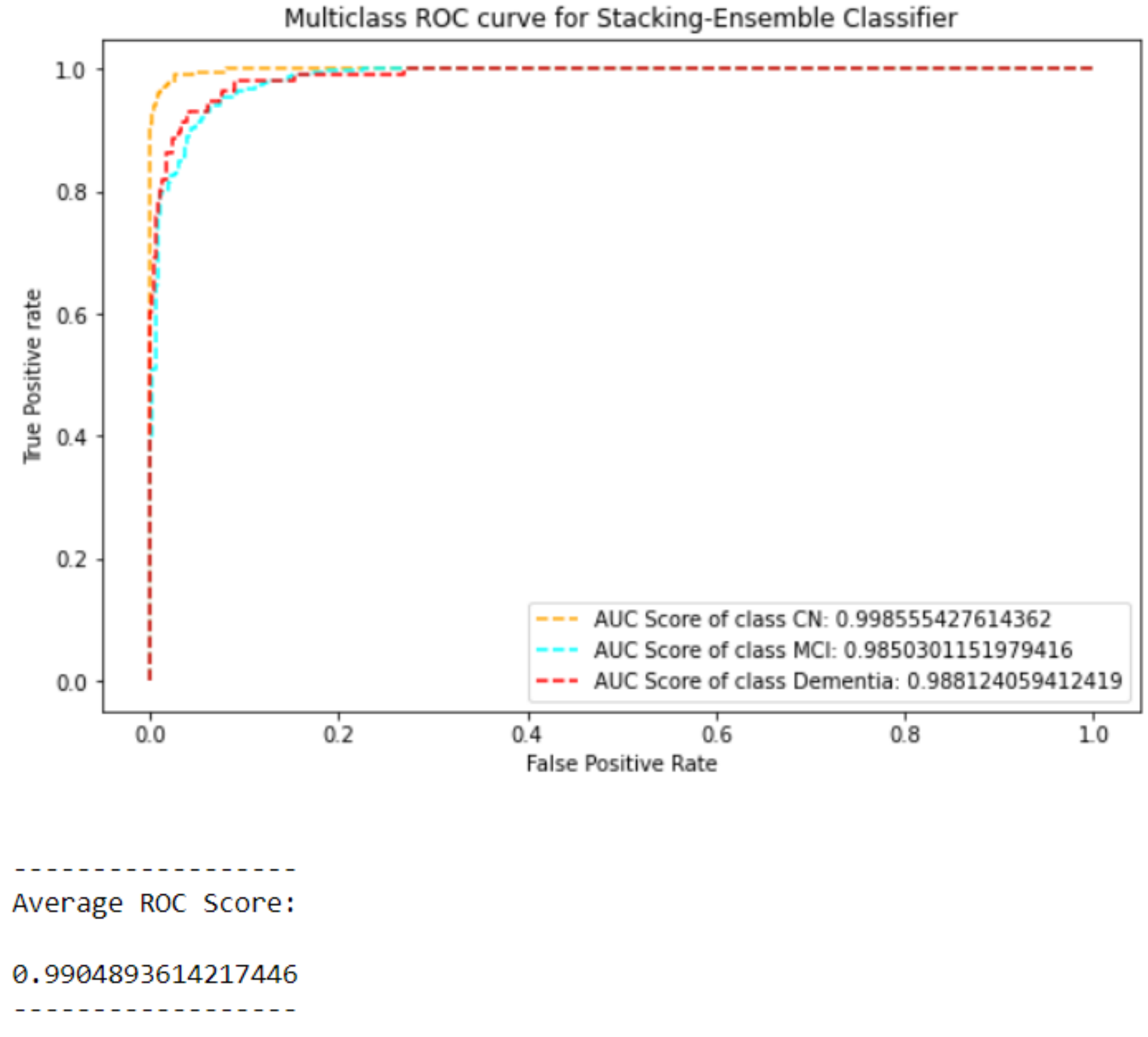

*Figure 5.19 AUCROC Curve for Stacking-Ensemble Classifier without Class Balancing*

#### 5.3.5.2 Evaluation and Results with Class Balancing using SVM-SMOTE

The training dataset, which was used to train the model after imbalance handling using SVM-SMOTE method, has 2,073 entries. The model's performance was evaluated using a test dataset of 643 records. Below, figure 5.20 refers to the with class balancing confusion matrix for the stacking-ensemble classifier model, where rows represent actual or true class labels and columns represent predicted class labels. The TP, TN, FP, and FN for this multi-class problem have been calculated following the logic depicted in figure 5.1. For the cognitive normal (CN) class, there are 226 instances classified as CN that were predicted correctly (TP). The actual or true class values of non-CN classes, which have been predicted correctly, are 405 instances (TN). There are 5 instances which have been observed (true labels) as non-CN but are predicted as CN class (FP). At the same time, 7 instances that were predicted as non-CN where the true class value is given as CN class (FN). For the MCI class, there are 265 instances classified as MCI that were predicted correctly (TP). The actual or true class values of non-MCI classes, which have been predicted correctly, are 332 instances (TN). There are 17 instances which have been observed (true labels) as non-MCI but predicted as MCI class (FP). At the same time, 29 instances that were predicted as non-MCI where the true class value are given as MCI class (FN). For the dementia class, there are 106 instances classified as dementia that were predicted correctly (TP). The actual or true class values of non-dementia classes, which have been predicted correctly, are 503 instances (TN). There are 24 instances which have been observed (true labels) as non-dementia but predicted as dementia class (FP). At the same time, 10 instances that were predicted as non-dementia where the true class value is given as dementia class (FN).

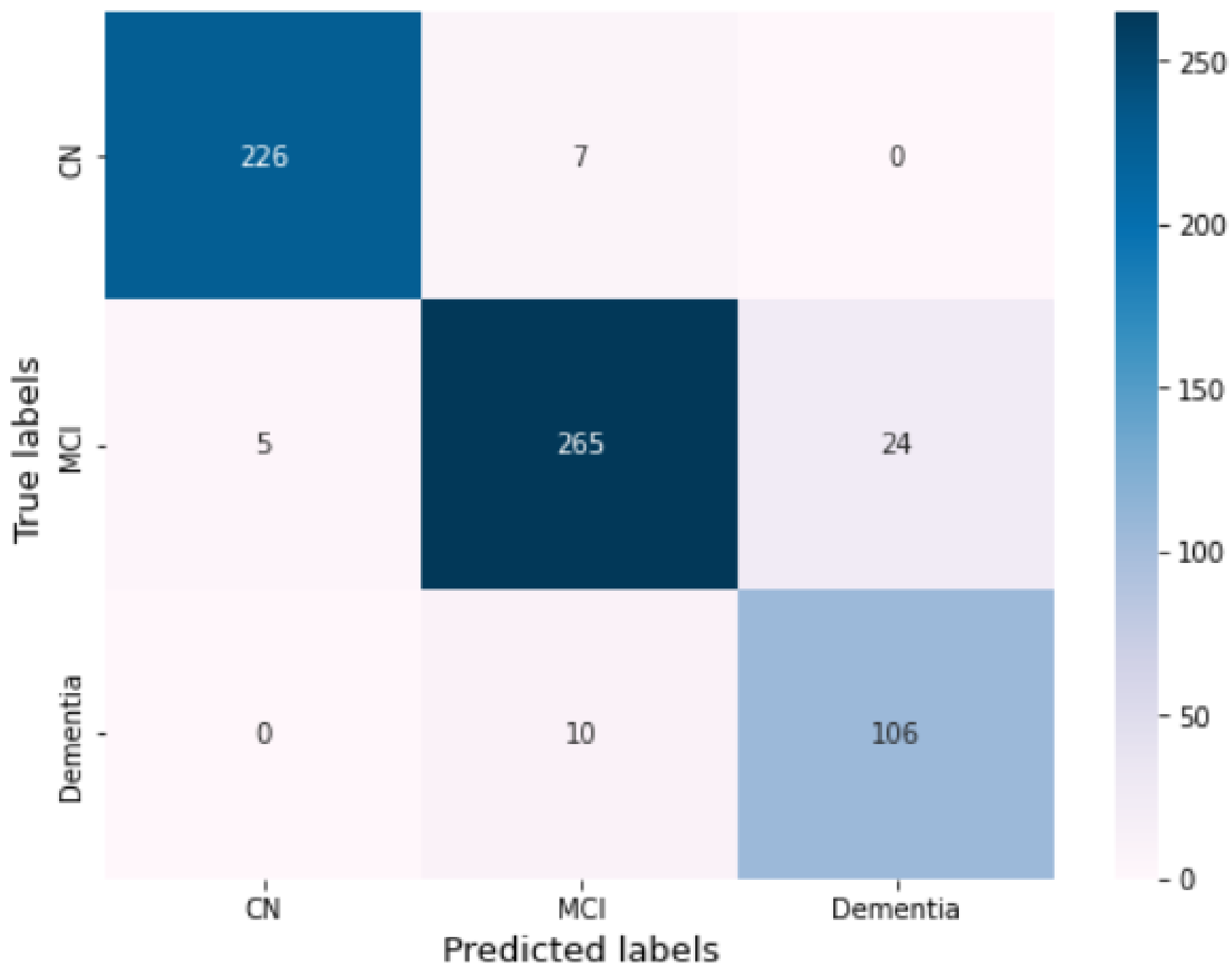

*Figure 5.20 Confusion Matrix for Stacking-Ensemble Classifier after Class Balancing*

The comprehensive performance metrics for this multiclass classification task after class imbalance handling, as well as the precision, recall, and f1-score for each class, are provided in table 5.10. Additionally, for measuring overall model performance, the weighted average precision, weighted average recall, weighted average f1-score, and accuracy are all 93%.

*Table 5.10 Classification Report for Stacking-Ensemble Classifier after Class Balancing*

| Target Class | Precision | Recall | f1-score | weighted avg Precision | weighted avg Recall | weighted avg f1-score | Accuracy |
|---|---|---|---|---|---|---|---|
| CN | 0.98 | 0.97 | 0.97 | 0.93 | 0.93 | 0.93 | 0.93 |
| MCI | 0.94 | 0.90 | 0.92 | | | | |
| Dementia | 0.82 | 0.91 | 0.86 | | | | |

The ROC curve presented in figure 5.21 after class imbalance handling shows how well the model predicts each class label. In addition, the whole model's weighted average AUCROC score is 98.84%.

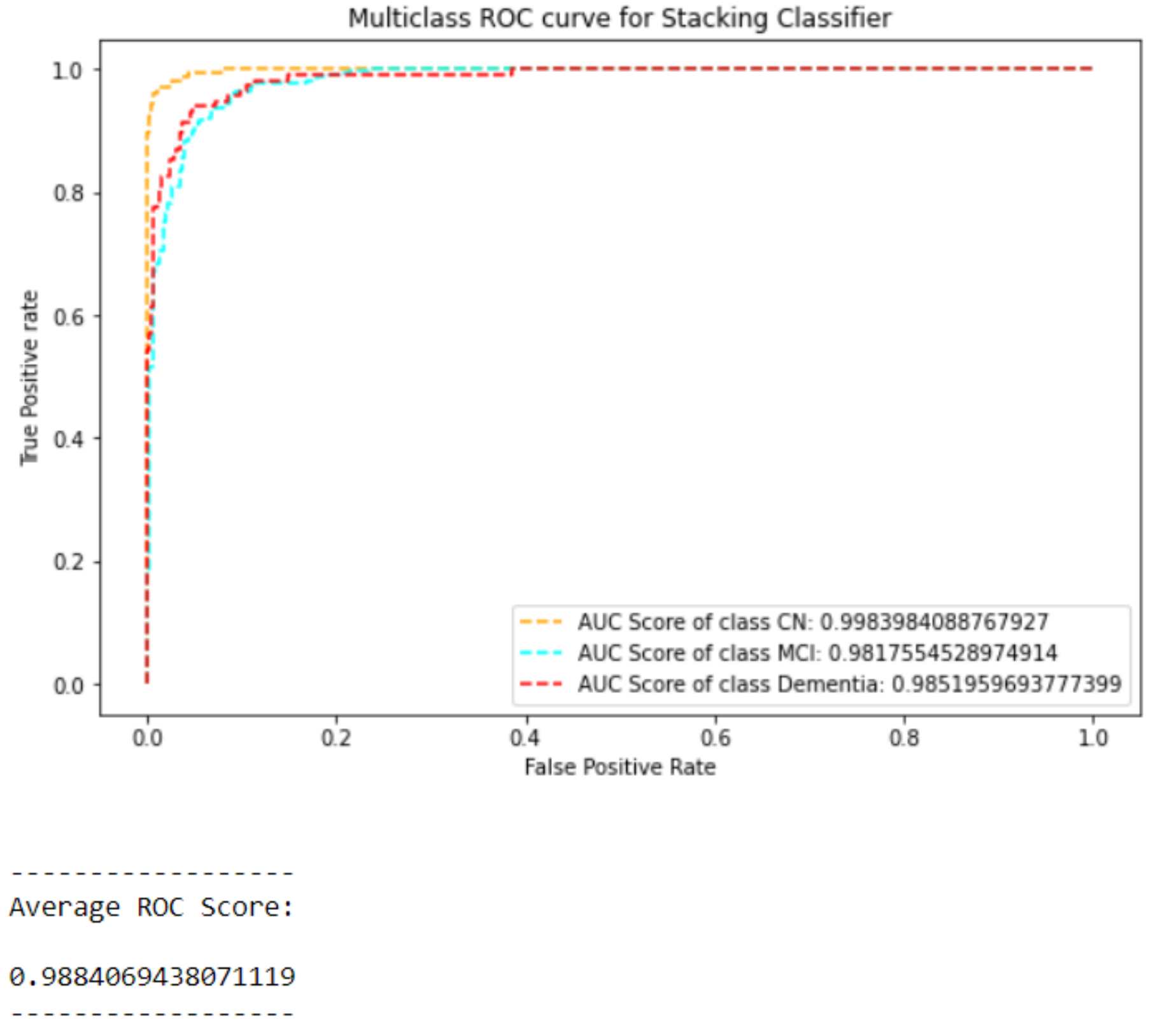

*Figure 5.21 AUCROC Curve for Stacking-Ensemble Classifier after Class Balancing*

### 5.3.6 Artificial Neural Network Evaluation and Results

For building a classification model, a classifier with Artificial Neural Network (ANN) was utilised; this classifier was trained twice, once without imbalance handling and once after class imbalance correction (with SMOTE SVM minority class oversampling method). The evaluations/results of the classifier on the test set, as well as performance metrics such as the confusion matrix, weighted average F1 score, weighted average precision, weighted average recall, and the ROCAUC curve, will be shown in the subsections below.

#### 5.3.6.1 Evaluation and Results without Class Balancing

The training dataset, which was used to train the model before imbalance handling, has 1,500 entries. The model's performance was evaluated using a test dataset of 643 records. Below, figure 5.22 refers to the without class balancing confusion matrix for the ANN classifier model, where rows represent actual or true class labels and columns represent predicted class labels. The TP, TN, FP, and FN for this multi-class problem have been calculated following the logic depicted in figure 5.1. For the cognitive normal (CN) class, there are 221 instances classified as CN that were predicted correctly (TP). The actual or true class values of non-CN classes, which have been predicted correctly, are 399 instances (TN). There are 11 instances which have been observed (true labels) as non-CN but are predicted as CN class (FP). At the same time, 12 instances that were predicted as non-CN where the true class value is given as CN class (FN). For the MCI class, there are 269 instances classified as MCI that were predicted correctly (TP). The actual or true class values of non-MCI classes, which have been predicted correctly, are 323 instances (TN). There are 26 instances which have been observed (true labels) as non-MCI but predicted as MCI class (FP). At the same time, 25 instances that were predicted as non-MCI where the true class value are given as MCI class (FN). For the dementia class, there are 102 instances classified as dementia that were predicted correctly (TP). The actual or true class values of non-dementia classes, which have been predicted correctly, are 513 instances (TN). There are 14 instances which have been observed (true labels) as non-dementia but predicted as dementia class (FP). At the same time, 14 instances that were predicted as non-dementia where the true class value is given as dementia class (FN).

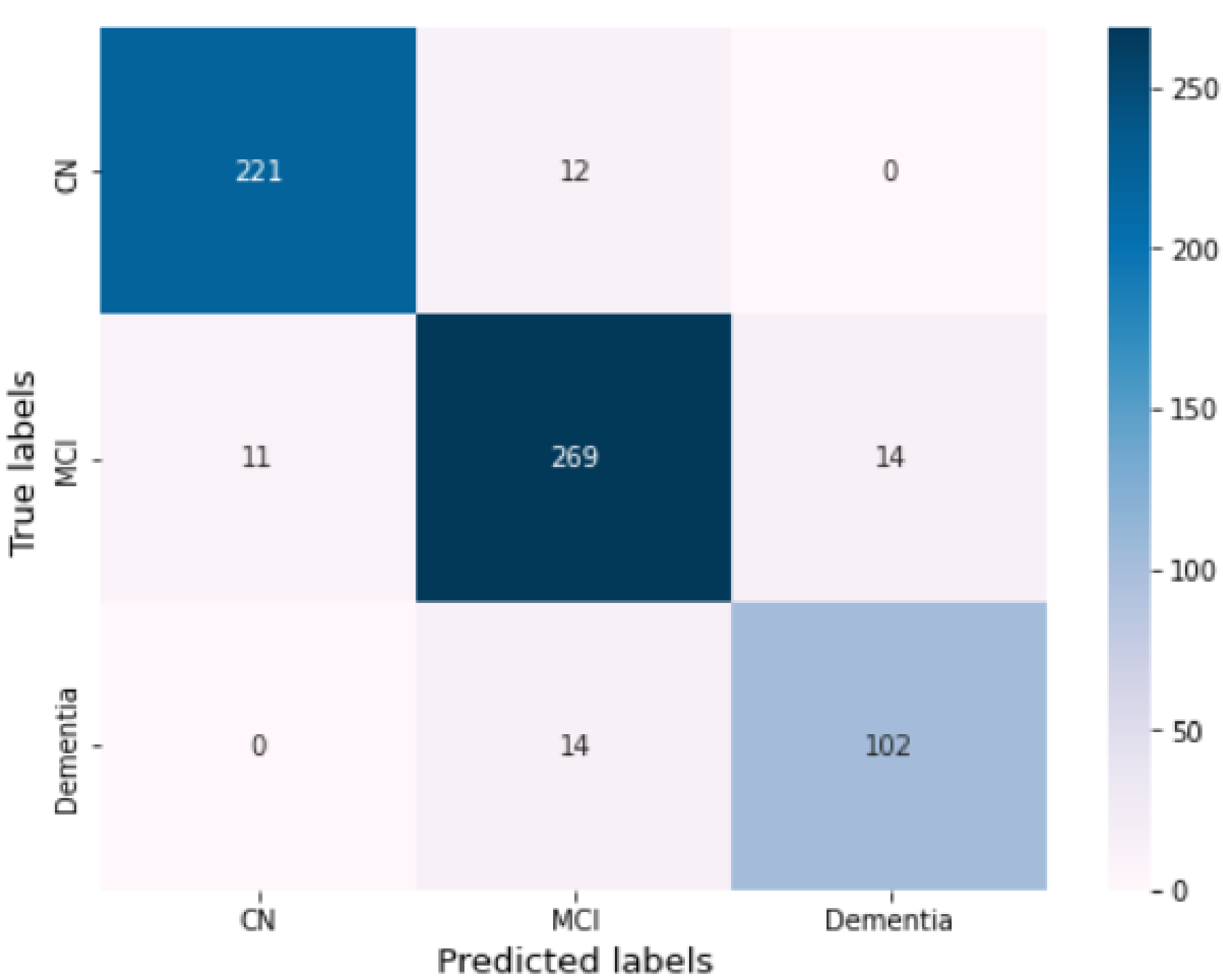

*Figure 5.22 Confusion Matrix for Artificial Neural Network without Class Balancing*

The comprehensive performance metrics for this multiclass classification task, as well as the precision, recall, and f1-score for each class, are provided in table 5.11. Additionally, for measuring overall model performance, the weighted average precision, weighted average recall, weighted average f1-score, and accuracy are all 92%.

*Table 5.11 Classification Report for Artificial Neural Network without Class Balancing*

| Target Class | Precision | Recall | f1-score | weighted avg Precision | weighted avg Recall | weighted avg f1-score | Accuracy |
|---|---|---|---|---|---|---|---|
| CN | 0.95 | 0.95 | 0.95 | 0.92 | 0.92 | 0.92 | 0.92 |
| MCI | 0.91 | 0.91 | 0.91 | | | | |
| Dementia | 0.88 | 0.88 | 0.88 | | | | |

The ROC curve presented in figure 5.23 shows how well the model predicts each class label. In addition, the whole model's weighted average AUCROC score is 98.18%.

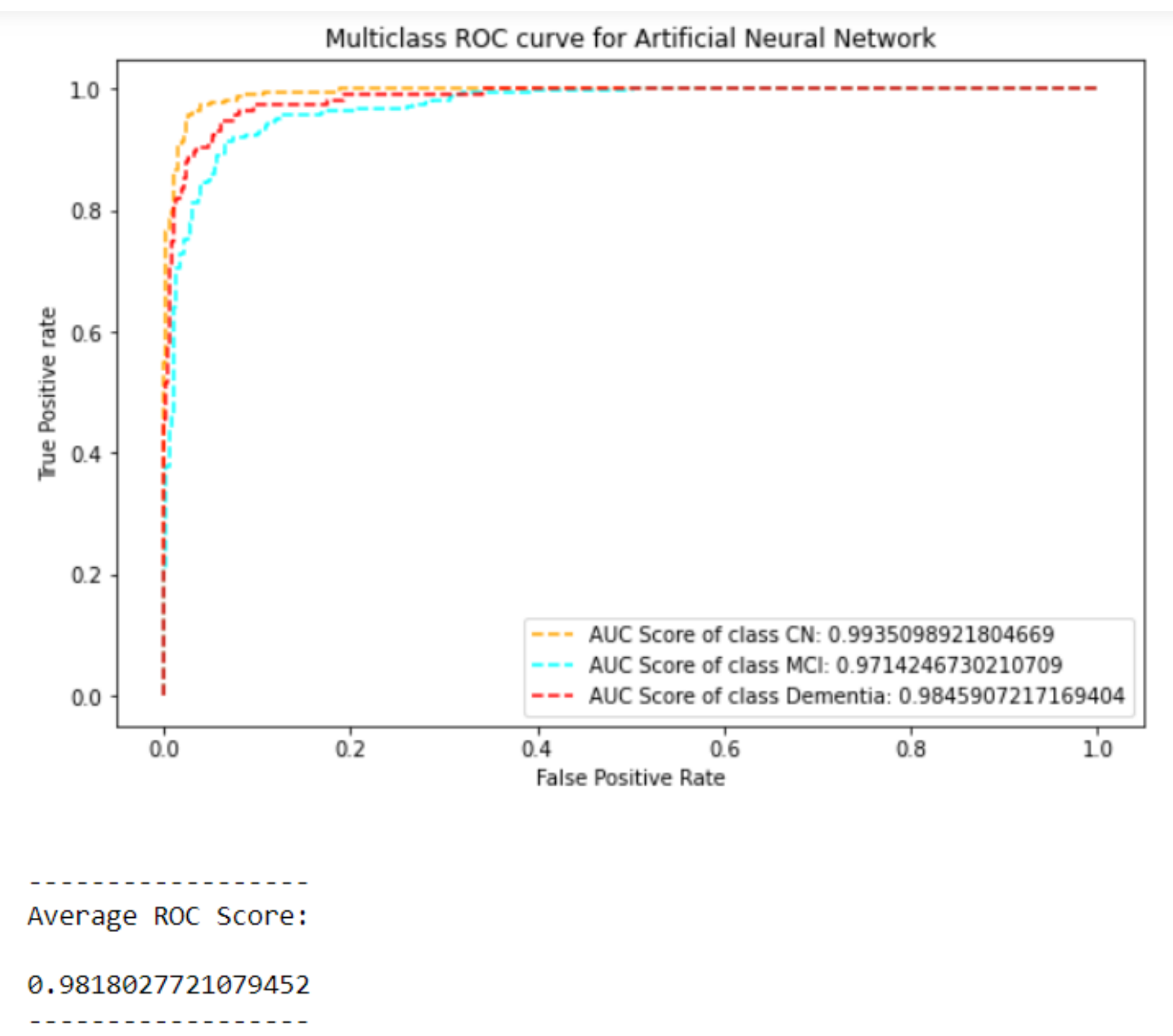

*Figure 5.23 AUCROC Curve for Artificial Neural Network without Class Balancing*

#### 5.3.6.2 Evaluation and Results with Class Balancing using SVM-SMOTE

The training dataset, which was used to train the model after imbalance handling using SVM-SMOTE method, has 2,073 entries. The model's performance was evaluated using a test dataset of 643 records. Below, figure 5.24 refers to the with class balancing confusion matrix for the ANN classifier model, where rows represent actual or true class labels and columns represent predicted class labels. The TP, TN, FP, and FN for this multi-class problem have been calculated following the logic depicted in figure 5.1. For the cognitive normal (CN) class, there are 224 instances classified as CN that were predicted correctly (TP). The actual or true class values of non-CN classes, which have been predicted correctly, are 393 instances (TN). There are 17 instances which have been observed (true labels) as non-CN but are predicted as CN class (FP). At the same time, 9 instances that were predicted as non-CN where the true class value is given as CN class (FN). For the MCI class, there are 256 instances classified as MCI that were predicted correctly (TP). The actual or true class values of non-MCI classes, which have been predicted correctly, are 328 instances (TN). There are 21 instances which have been observed (true labels) as non-MCI but predicted as MCI class (FP). At the same time, 38 instances that were predicted as non-MCI where the true class value are given as MCI class (FN). For the dementia class, there are 104 instances classified as dementia that were predicted correctly (TP). The actual or true class values of non-dementia classes, which have been predicted correctly, are 506 instances (TN). There are 21 instances which have been observed (true labels) as non-dementia but predicted as dementia class (FP). At the same time, 12 instances that were predicted as non-dementia where the true class value is given as dementia class (FN).

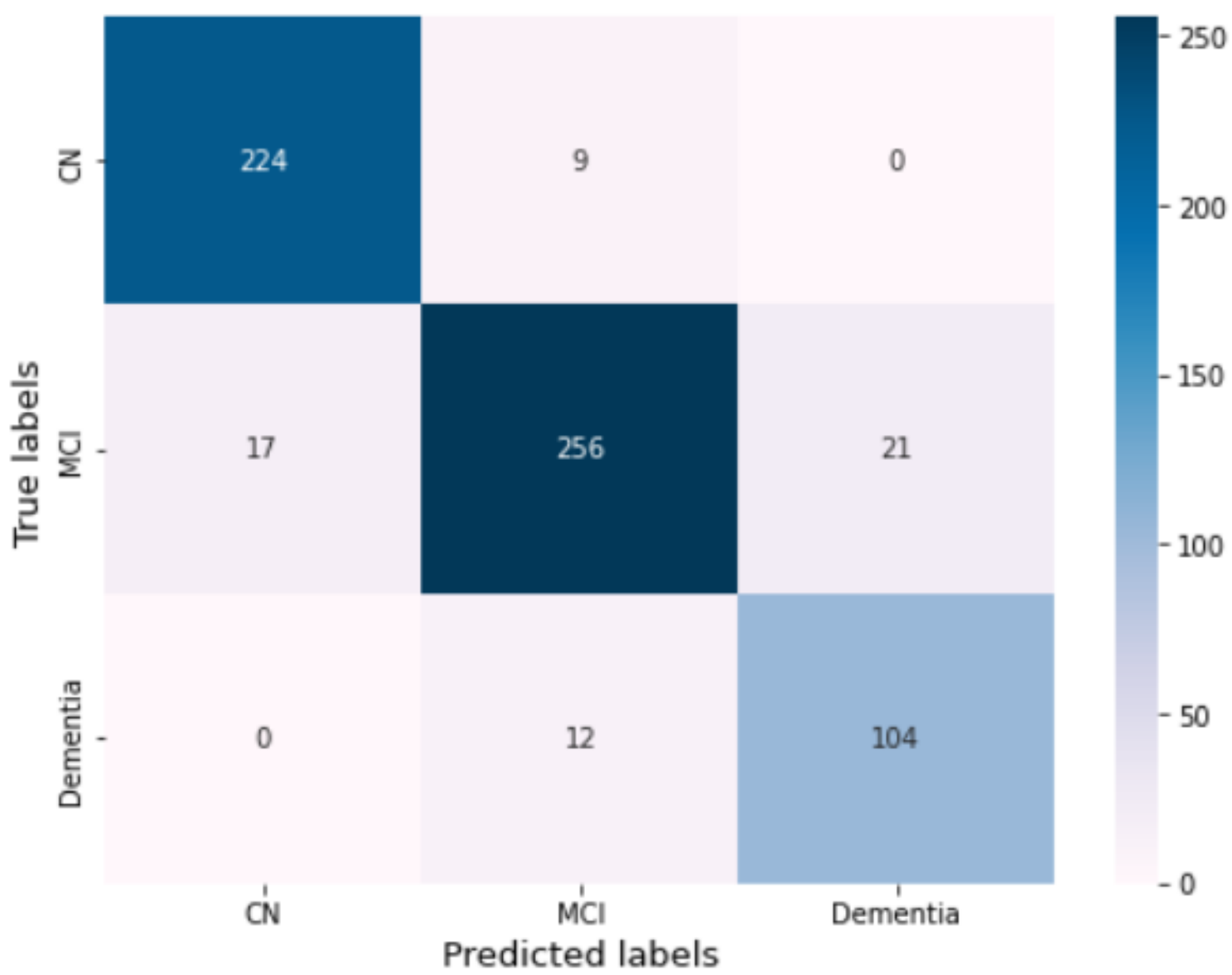

*Figure 5.24 Confusion Matrix for Artificial Neural Network after Class Balancing*

The comprehensive performance metrics for this multiclass classification task after class imbalance handling, as well as the precision, recall, and f1-score for each class, are provided in table 5.12. Additionally, for measuring overall model performance, the weighted average precision, weighted average recall, weighted average f1-score, and accuracy are all 91%.

*Table 5.12 Classification Report for Artificial Neural Network after Class Balancing*

| **Target Class** | **Precision** | **Recall** | **f1-score** | **weighted avg Precision** | **weighted avg Recall** | **weighted avg f1-score** | **Accuracy** |
|---|---|---|---|---|---|---|---|
| CN | 0.93 | 0.96 | 0.95 | 0.91 | 0.91 | 0.91 | 0.91 |
| MCI | 0.92 | 0.87 | 0.90 | | | | |
| Dementia | 0.83 | 0.90 | 0.86 | | | | |

The ROC curve presented in figure 5.25 after class imbalance handling shows how well the model predicts each class label. In addition, the whole model's weighted average AUCROC score is 98.06%.

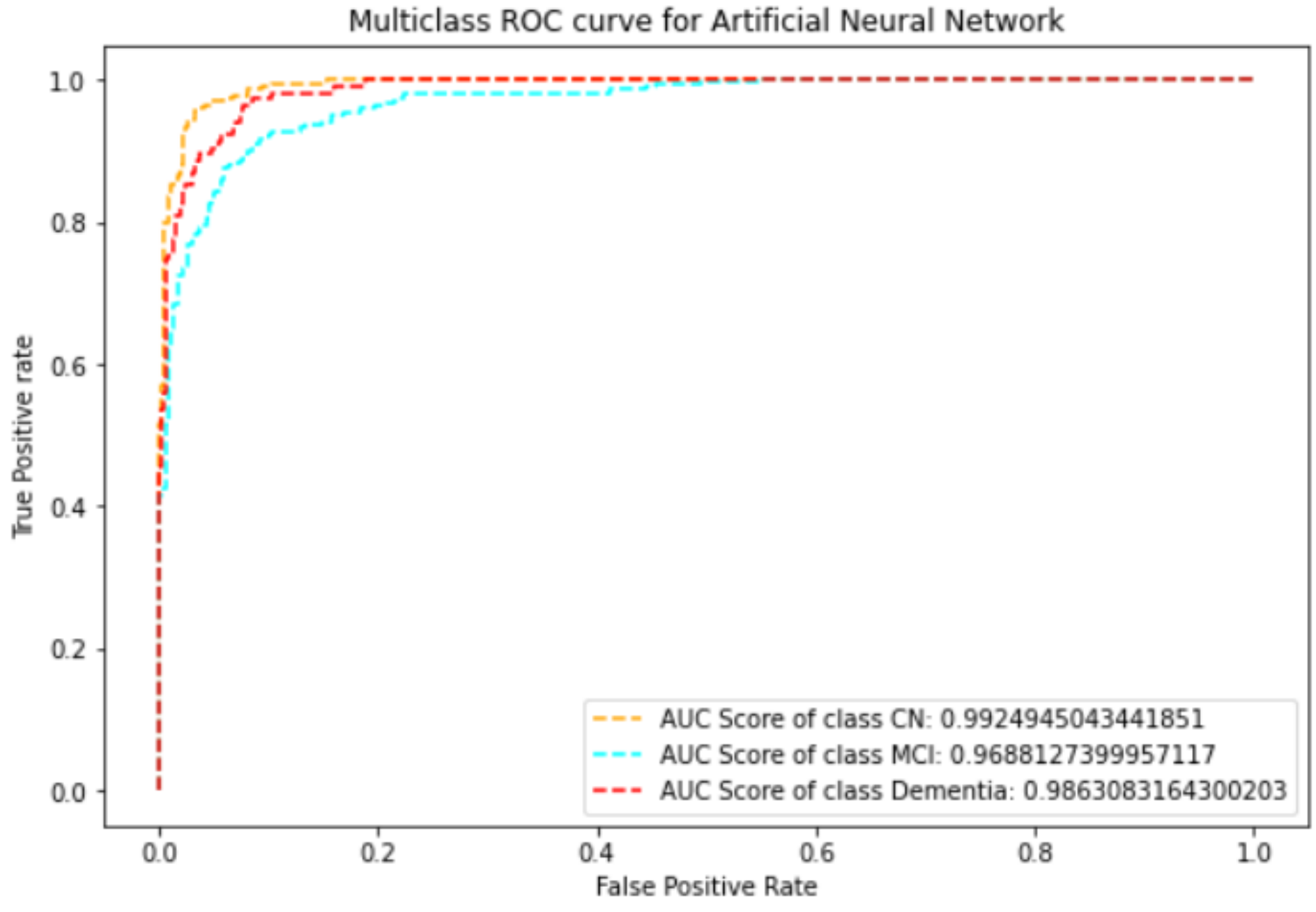

```
Average ROC Score:
0.98055043511169
```

*Figure 5.25 AUCROC Curve for Artificial Neural Network after Class Balancing*

## 5.4 Comparison between Proposed Classifiers and Results

The top classification model for the prediction of Alzheimer's disease using the ADNI study dataset was determined by analysing various performance measures used to evaluate the classifiers. The test dataset used to validate the classifiers built on the training set resulted in the production of classifiers with nearly identical values of evaluation metrics and no significant differences between them. This demonstrates that all of the classifiers performed admirably in the test set evaluation. However, some of the most popular evaluation measures used in medical diagnosis were compared amongst the six proposed classifiers in order to select the best or most robust one. Both the class imbalance and class balanced datasets were used to determine the optimal classification model.

As shown in below table 5.13, in terms of total weighted average metrices as well as individual class prediction performance, the Logistic Regression and Stacking-Ensemble classification model has delivered the best results among all other classifiers. The weighted average precision, weighted average recall, weighted average fi-score, and accuracy for Logistic Regression are all 94%. For the same model, the area under the curve (AUC) was 99%. Individual class performance metrics are also calculated and listed below. Precision, recall, and f1-score were all 97% for the Cognitive Normal (CN) class. For the Mild Cognitive Impairment (MCI) class, the precision, recall, and f1-score were 93%, 94%, and 93%, respectively. For the Dementia class, the precision, recall, and f1-score were 87%, 89%, and 88%, respectively.

Furthermore, for the Stacking-Ensemble model the weighted average precision, weighted average recall, weighted average fi-score, and accuracy are 94%, 93%, 93%, 93% respectively. The AUC score achieved for the same predictor was 99.05%. Individual class performance metrics are also calculated and listed below. Precision, recall, and f1-score were 98%, 96%, and 97%, respectively for the Cognitive Normal (CN) class. For the Mild Cognitive Impairment (MCI) class, the precision, recall, and f1-score were all 93%. For the Dementia class, the precision, recall, and f1-score were 89%, 87%, and 88%, respectively.

*Table 5.13 Performance Evaluation Metrics of Classifiers without Imbalance Handling*

| **Classifier Name** | **Target Class Labels** | **Performance Evaluation Metrics (Class Imbalanced Dataset)** | | | | | | | |
|---|---|---|---|---|---|---|---|---|---|
| | | | | | **Weighted Average** | | | | |
| | | **Precision (%)** | **Recall (%)** | **f1-score (%)** | **Precision (%)** | **Recall (%)** | **f1-score (%)** | **AUC (%)** | **Accuracy (%)** |
| **Logistic Regression** | CN | **97** | **97** | **97** | **94** | **94** | **94** | **99** | **94** |
| | MCI | **93** | **94** | **93** | | | | | |
| | Dementia | **89** | **87** | **88** | | | | | |
| **Light GBM** | CN | 98 | 95 | 97 | 93 | 93 | 93 | 98.86 | 93 |
| | MCI | 91 | 94 | 92 | | | | | |
| | Dementia | 87 | 87 | 87 | | | | | |
| **Extra Tree** | CN | 99 | 93 | 96 | 92 | 92 | 92 | 98.95 | 92 |
| | MCI | 89 | 94 | 92 | | | | | |
| | Dementia | 88 | 84 | 86 | | | | | |
| **Bagging-KNN** | CN | 84 | 97 | 90 | 87 | 87 | 87 | 96.48 | 87 |
| | MCI | 90 | 81 | 85 | | | | | |
| | Dementia | 86 | 83 | 85 | | | | | |
| **Stacking-Ensemble** | CN | **98** | **96** | **97** | **94** | **93** | **93** | **99.05** | **93** |
| | MCI | **93** | **93** | **93** | | | | | |
| | Dementia | **87** | **89** | **88** | | | | | |
| **Artificial Neural Network** | CN | 95 | 95 | 95 | 92 | 92 | 92 | 98.18 | 92 |
| | MCI | 91 | 91 | 91 | | | | | |
| | Dementia | 88 | 88 | 88 | | | | | |

The comparison between the predictors' performance after applying the class balancing strategy has been reported in below table 5.14. The Stacking-Ensemble classifier surpasses all other classification models in terms of weighted average metrics. The weighted average precision, weighted average recall, weighted average f1-score, and accuracy are all 93% for the stacking model, which is the highest among the others. Also, the AUC score, which indicates how well a classification model fits, was 98.84%.

Individual class performance metrics are computed and mentioned below. The Stacking-Ensemble model outperforms others in terms of individual class prediction outcomes as well. Precision, recall, and f1-score for the Cognitive Normal (CN) class were 98%, 96%, and 97%, respectively. Precision, recall, and f1-score were all 93% for the Mild Cognitive Impairment (MCI) group. The precision, recall, and f1-score for the Dementia class were 89%, 87%, and 88%, respectively.

*Table 5.14 Performance Evaluation Metrics of Classifiers after Imbalance Handling*

| Classifier Name | Target Class Labels | Performance Evaluation Metrics (Class Balanced Dataset) | | | | | | | |
|---|---|---|---|---|---|---|---|---|---|
| | | Precision (%) | Recall (%) | f1-score (%) | Weighted Average Precision (%) | Weighted Average Recall (%) | Weighted Average f1-score (%) | Weighted Average AUC (%) | Weighted Average Accuracy (%) |
| **Logistic Regression** | CN | 96 | 97 | 97 | 92 | 92 | 92 | 98.11 | 92 |
| | MCI | 93 | 88 | 91 | | | | | |
| | Dementia | 79 | 90 | 84 | | | | | |
| **Light GBM** | CN | 98 | 95 | 97 | 92 | 92 | 92 | 98.74 | 92 |
| | MCI | 93 | 89 | 91 | | | | | |
| | Dementia | 80 | 91 | 85 | | | | | |
| **Extra Tree** | CN | 97 | 95 | 96 | 92 | 91 | 91 | 98.55 | 91 |
| | MCI | 93 | 88 | 90 | | | | | |
| | Dementia | 79 | 91 | 85 | | | | | |
| **Bagging-KNN** | CN | 79 | 97 | 87 | 85 | 83 | 83 | 96.65 | 83 |
| | MCI | 93 | 69 | 79 | | | | | |
| | Dementia | 77 | 93 | 84 | | | | | |
| **Stacking-Ensemble** | CN | **98** | **97** | **97** | **93** | **93** | **93** | **98.84** | **93** |
| | MCI | **94** | **90** | **92** | | | | | |
| | Dementia | **82** | **91** | **86** | | | | | |
| **Artificial Neural Network** | CN | 93 | 96 | 95 | 91 | 91 | 91 | 98.06 | 91 |
| | MCI | 92 | 87 | 90 | | | | | |
| | Dementia | 83 | 90 | 86 | | | | | |

As discussed in above performance comparison results, it has been observed that Stacking-Ensemble classifiers (with LR, LGBM, Extra Tree and Bagging-KNN as base estimators and LR as meta estimator) is the best predictive model for early-stage diagnosis of Alzheimer's disease on the ADNI study dataset. This model can identify mild dementia stage (MCI), which is the early phase before the severe cognitive impairment or acute dementia, with 93% precision, recall and fi-score before class imbalance handling and with 94% precision, 90% recall and 92% fi-score after class imbalance handling. Hence, the final model that can be chosen as best early-stage predictor is the stacking-ensemble classifier.

## 5.5 Summary

This chapter summarised the results and analysis of the whole study, and the findings of the classification methods were interpreted. Section 5.2 (Significant Biomarkers from visualization and Feature Selection) included only the essential risk variables that were identified to play a pivotal role in presenting deeper insights into the ADNI dataset. The interactions between the variables with the target Alzheimer's diagnosis (DX) and the interdependency among the independent variables were discussed in depth. These crucial biomarkers will provide clinicians an overview of the key parameters associated with dementia illness at a quick glance, and for elderly patients, they will act as a guide for understanding the impact of certain risk factors in predicting the stage of Alzheimer's disease.

The confusion matrices of the classification models, which were constructed using the test datasets without and with balancing techniques, were interpreted in terms of their recall, precision, f1-score, accuracy, and ROC score. The performance measure table was displayed for each classifier, and the results without and with balancing methods were evaluated and compared. The results showed that across all the six classifiers (Logistic regression, LGBM, Extra Tree, Bagging KNN, stacking blending, and ANN), Logistic Regression and Stacking blending or Stacking-Ensemble classifiers achieved the best performance when the model was generated from the dataset without class balancing. The Stacking-Ensemble classifiers achieved the best performance when the model was generated from the dataset after class balancing with SVM SMOTE.

To finalize the best predictor model, not only the weighted average metrics but also the individual class prediction precision, recall, f1-score, etc. metrices were considered. It is observed that the performance of the best-chosen models is very comparable between the class imbalance and class balanced datasets. Because the presented dataset is not highly imbalanced, the evaluation results from the models on the test dataset before and after imbalance handling with SVM SMOTE do not differ significantly.

# CHAPTER 6

# CONCLUSION AND FUTURE RECOMMENDATION

## 6.1 Introduction

The overall study will be addressed briefly in this chapter, and conclusions based on the study's findings will be drawn. A summary of the steps used, such as data selection, pre-processing, missing value imputation, feature selection, and class balancing technique, will be included in the discussion. The findings of various analyses, such as univariate and bivariate analysis, correlation studies, visualisation, model creation, and model validation using the test set, will also be discussed. Following that, a conclusion will be drawn to show if the objective of the study has been met or not. Extensions of this investigation, application of new methodologies to the data, or a different perspective on the study to offer ways to improve the suggested model or find a better model than the one in this study will be among the future recommendations.

## 6.2 Summary of the Study

This study was conducted using the Alzheimer's Disease Neuroimaging Initiative dataset, which consisted of 15,762 records when it was first imported, and the demographic and cognitive test factors who participated in the ADNI1, ADNI2, ADNIGO, and ADNI3 studies. The baseline investigation was extracted from a subset of the dataset containing 2,379 records and 65 characteristics, as this study focused on early-stage diagnosis. The target variable, DX (Alzheimer's Diagnosis), indicates the stage of Alzheimer's disease that a patient is experiencing. As a part of data pre-processing and transformation, unrelated variables like RID, PTID, COLPROT, and VISCODE were removed, 5 categorical variables were turned into numerical variables using dummy features, and 4 numerical features into categorical columns to make the analysis and modelling process efficient and effective. As a part of missing value analysis, there are around 48 features that consist of missing data that were detected and imputed using proper missing value imputation techniques, namely, iterative and mode imputation techniques for numerical and categorical attributes, respectively. Furthermore, it has also been identified that there are 7 features, namely, PIB, FBB, DIGITSCOR, AV45, ABETA, TAU, and PTAU, that consist of more than 45% null values. Imputing more than 45 percent of missing data can introduce bias into the model and reduce the classifier's efficiency. Dropping those features would be a better strategy.

Following that, univariate analysis was conducted to analyse the distribution of data in each variable as part of the exploratory data analysis on the ADNI dataset with imputed missing values. The count plot or percentage plot for a few significant variables, as well as the overall statistical analysis, were thoroughly examined. As part of the bivariate and correlation analysis, the percentage analysis for each target class, as well as the variable relationships, were checked and reported. Each visualization report demonstrated if the predictor variable had any impact on the dementia progression. If so, then whether the target classes are well separated by the predictor. The report and graphical representation also illustrate the link between the numerical elements that decide if the combined relationship has any effect on the targeted variable, DX (Alzheimer's diagnosis).

The prediction capacity of a machine learning model is directly proportional to the number of relevant characteristics or features used to train it. The performance of ML models will degrade if irrelevant features are used in their development. Feature selection is the process of reducing the number of unneeded attributes, also known as dimensionality reduction, based on any statistical relationship, either automatically or manually. This could improve the performance of machine learning models or their ability to anticipate outcomes. To pick the most acceptable and relevant features for classification model development, the wrapper-based and embedded feature selection methods RFE (recursive feature elimination) and tree-based classifier Random Forest were used on the presented dataset. The top 40 traits were chosen as the most important elements in predicting the target class, and they were later used to develop a classification model.

The interactions and visualisations, which are based on selected important features, provide a critical and deeper insight into the Alzheimer's dataset that was examined and analysed, providing a clear image of all the variables in the dataset as well as a summary of their associations. Some of the important biomarkers that are identified from this study are the psychometric exam related variables such as Clinical Dementia Rating Sum of Boxes (CDRSB), Rey Auditory Verbal Learning Test for immediate response (RAVLT_immediate), Everyday Cognition Study Partner - Total (EcogSPTotal), Alzheimer's Disease Assessment Scale - 13 items (ADAS13), Mini-Mental State Examination (MMSE), modified Preclinical Alzheimer's Cognitive Composite with Digit (mPACCdigit). These important biomarkers or variables will give clinicians a fast summary of the key characteristics linked with dementia and will serve as a guide for elderly patients in understanding the influence of various risk factors when anticipating disease stage. Using the default split, the original dataset with imputed missing values was divided into two sets, with the training set accounting for 70% and the test set for 30%. Because the supplied dataset is unbalanced, a class balancing method, particularly the SVM SMOTE method, was used to resolve the issue of class imbalance on the training dataset. The six classifiers were tuned using the Grid Search cross validation technique with five folds. These six classifiers (Logistic regression, LGBM, Bagging KNN, Extra Tree, Stacking Blending, and ANN) were trained on both the imbalanced and balanced training datasets, and then validated or evaluated on the test dataset.

Following that, the classification algorithms' confusion matrices were explained using test datasets and balance procedures, and they were produced using recall, precision, f1-score, accuracy, and ROC score. The results showed that logistic regression and stacking ensemble classifiers performed best when the model was generated from the dataset without class balance handling. For Logistic Regression, the weighted average values of precision, recall, fi-score, area under ROC, accuracy were found to be 94%, 94%, 94%, 99%, 94%, respectively. For Stacking-Ensemble classifier, the weighted average values of precision, recall, fi-score, area under ROC, accuracy were found to be 94%, 93%, 93%, 99.05%, 93%, respectively. In contrast, the stacking blending classifiers performed best when the model was generated from the dataset after class balance treatment with the SVM SMOTE minority class oversampling method with the weighted average values of precision, recall, fi-score, area under ROC, accuracy of 93%, 93%, 93%, 98.84%, 93%, respectively.

Before and after using class balancing procedures, the weighted average values, as well as each class's recall, precision, and f1-score for each classifier, were analysed to arrive at a single method. Individual class prediction precision, recall, f1-score, and other metrics were examined in addition to the weighted metrics for determining the optimal predictor model. The performance of the best-chosen models is shown to be relatively similar in both the class imbalance and class balanced datasets. The evaluation results from the models on the test dataset before and after imbalance treatment with SVM SMOTE do not change considerably because the presented dataset is not severely imbalanced. From the estimated results of the classification models, it is evident that the performance of the stacking ensemble technique is superior when compared to the individual predictor's performance.

### 6.3 Future Recommendations

There is potential for improvement in this study, which can be addressed as part of future suggestions, most notably, improved class imbalance managing strategies. As an improvement to the SVM SMOTE method of class imbalance handling, hybrid imbalance handling methods, such as a combination of oversampling and under sampling like the SMOTE Tomek method or Spread Sample method, along with SMOTE, can be used for better class imbalance treatment, rather than only generating synthetic data to balance the minority class with the majority.

The study can be further improved by performing more robust feature selection techniques using information value and statistical testing such as ANOVA and Chi-Square along with the wrapper-based and embedded selection techniques used in this study. Information value, or IV, will determine feature importance based on weight of evidence and rank each feature independently of weight or cover, comparable to the feature importance of any tree model. Statistical testing, on the other hand, will determine the influence of individual predictors on the dependent variable. The most significant features that can produce more accurate and stable model performance include whether the independent feature is a good separator of the target class, whether the variance or mean value of the predictor attributes is equal or not among the classes, and so on.

This ADNI dataset contains only 2,379 data points or observations for aged patients, which is inadequate to train a classification model or learn more diverse information about this condition. More observations or data from more senior patients suffering from Alzheimer's illness from other health institutions or studies needs to be collected.

The missing value imputation approach used in this thesis was iterative imputation; however, the KNN imputation method of missing value imputation can be used instead, and the classification model performance can then be compared.

As part of the missing value treatment, a few features were dropped. For variables related to cerebrospinal fluid (CSF Biomarkers), there are more than 45% missing values in the supplied dataset. As a result, such variables have been removed from the dataset without being imputed. The molecular abnormalities in the AD brain are reflected in the cerebrospinal fluid. Amyloid-beta peptide level (ABETA), total tau (T-tau), and phosphorylated tau (P-tau) are three primary possible CSF biomarkers that have been shown to have high diagnostic performance in diagnosing Alzheimer's disease even in its early stages. If data for these variables can be collected and used for model training, predictor performance can be improved.